%% file: main.tex
\documentclass{ecai} 

\usepackage{tikz}
\usetikzlibrary{shapes}
\usetikzlibrary{arrows}
\usetikzlibrary{automata}
\usetikzlibrary{fit}
\usetikzlibrary{shadows}
\usetikzlibrary{shapes}
\usetikzlibrary{trees}
\usetikzlibrary{positioning}
\usetikzlibrary{patterns}
\usetikzlibrary{backgrounds}
\usetikzlibrary{matrix}
\usetikzlibrary{external}

\usepackage{pgfplots}
\pgfplotsset{compat=1.18}

\tikzstyle{every state}=[draw=black,text=black,inner color= white,outer color= white,draw= black,text=black]
\tikzstyle{place}=[thick,draw=sthlmBlue,fill=blue!20,minimum size=8mm, opacity=.5]
\tikzstyle{red place}=[square,place, draw=sthlmRed, fill=sthlmLightRed]
\tikzstyle{green place}=[diamond, place, draw=sthlmGreen, fill=sthlmLightGreen]
\tikzset{chance/.style={state,place}}
\tikzset{maxnod/.style={state,red place}}
\tikzset{minnod/.style={state,green place}}
\tikzset{termin/.style={align=left}}

	\definecolor{sthlmLightBlue}{RGB}{214,237,252} 
		
	\definecolor{sthlmBlue}{RGB}{0,110,191} 
		
	\definecolor{sthlmLightGreen}{RGB}{213,247,244} 
		
	\definecolor{sthlmGreen}{RGB}{0,134,127} 

	\definecolor{sthlmLightGrey}{RGB}{213,217,225} 
		
	\definecolor{sthlmGrey}{RGB}{245,243,238} 
		
	\definecolor{sthlmDarkGrey}{RGB}{51,51,51} 

	\definecolor{sthlmLightOrange}{RGB}{255,215,210} 
		
	\definecolor{sthlmOrange}{RGB}{221,74,44} 

	\definecolor{sthlmLightPurple}{RGB}{241,230,252} 
		
	\definecolor{sthlmPurple}{RGB}{93,35,125} 

	\definecolor{sthlmLightRed}{RGB}{254,222,237} 
		
	\definecolor{sthlmRed}{RGB}{196,0,100} 
		
	\definecolor{sthlmYellow}{RGB}{252,191,10} 

\usepackage[framemethod=TikZ]{mdframed}

\usepackage{caption,subcaption}

\usepackage{amsmath}
\usepackage{amssymb}
\usepackage{xspace}
\def\ie{{\em i.e.}\xspace}

\newcommand{\maxx}[1]{\underset{#1}{max}}
\newcommand{\minn}[1]{\underset{#1}{min}}
\newcommand{\Ex}[2]{\underset{#1}{\mathbb{E}}\left[#2\right]}

\usepackage{thmtools}
\usepackage{thm-restate}
\usepackage{cleveref}

\usepackage{array, multirow,makecell}
\usepackage{paralist}
\usepackage{cancel}
\usepackage{diagbox}
\usepackage{graphicx}
\usepackage{booktabs}       
\usepackage{rotating}
\usepackage{algorithm2e}
\usepackage{bbm}

\newcommand{\neuralnetwork}{\theta}

\newcommand{\distance}[4]{\mathcal{F}_{#1}^{#2} (#3,#4)}

\newcommand{\timestep}{t}
\newcommand{\agent}{i}

\newcommand{\Time}[2]{#1^{#2}}
\newcommand{\AgentPov}[2]{#1_{#2}}
\newcommand{\TimeAgentPov}[3]{#1^{#2}_{#3}}

\newcommand{\worldstateTime}[1]{\Time{s}{#1}}
\newcommand{\worldstate}{\worldstateTime{}}
\newcommand{\worldstateSet}{\mathcal{S}}

\newcommand{\actionTimePov}[2]{\TimeAgentPov{a}{#1}{#2}}
\newcommand{\actionTime}[1]{\actionTimePov{#1}{}}
\newcommand{\action}{\actionTimePov{}{}}
\newcommand{\actionSetPov}[1]{\AgentPov{\mathcal{A}}{#1}}
\newcommand{\actionSet}{\actionSetPov{}}

\newcommand{\rewardTimePov}[2]{\TimeAgentPov{r}{#1}{#2}}
\newcommand{\reward}{\rewardTimePov{}{}}
\newcommand{\rewardPovCond}[4]{\rewardTimePov{#1}{#2} (#3,#4)}

\newcommand{\transition}{\mathcal{T}}
\newcommand{\transitionCond}[3]{\transition{} (#1|#2,#3)}

\newcommand{\discount}{\gamma}
\newcommand{\horizon}{\mathrm{T}}
\newcommand{\trajectory}{\tau}

\newcommand{\budget}{budget}
\newcommand{\Count}[1]{\mathrm{N}_{#1}}
\newcommand{\CountTuple}[3]{\Count{#1} (#2,#3)}
\newcommand{\CountOne}[2]{\Count{#1} (#2)}
\newcommand{\qvalueTime}[1]{\Time{\mathrm{Q}}{#1}}
\newcommand{\qvalue}{\qvalueTime{}}
\newcommand{\qvalueTimeCond}[3]{\qvalueTime{#1} (#2,#3)}
\newcommand{\qvaluemechanism}[2]{\widehat{\qvalue{#1}} (#2)}

\newcommand{\uctconstante}{c_1}
\newcommand{\puctconstante}{c_2}
\newcommand{\dirichetnoiseproba}{\rho}
\newcommand{\dirichetnoise}{\epsilon}
\newcommand{\dirichetnoisedistribution}{\mathcal{N}_D (\dirichetnoise)}
\newcommand{\temperature}{\mathrm{T}}

\newcommand{\continue}{c}

\newcommand{\dataset}[1]{\mathrm{D}^{#1}}

\newcommand{\policyTimePov}[2]{\TimeAgentPov{\pi}{#1}{#2}}

\newcommand{\policyTimePovCond}[4]{\policyTimePov{#1}{#2} (#3|#4)}

\newcommand{\distanceActor}[3]{\distance{#1}{A}{#2}{#3}}
\newcommand{\distanceCritic}[3]{\distance{#1}{C}{#2}{#3}}

\newcommand{\weightcritic}[1]{\lambda^{C}_{#1}}
\newcommand{\weightactor}[1]{\lambda^{A}_{#1}}
\newcommand{\weightfullactor}[1]{\alpha^{A}_{#1}}
\newcommand{\maxlambda}[1]{\lambda^{Max}_{#1}}
\newcommand{\expoTerm}[1]{\tau_{#1}}

\newcommand{\lossCriticSub}[1]{\loss^{C,Sub}(#1)}
\newcommand{\lossActorSub}[1]{\loss^{A,Sub}(#1)}
\newcommand{\lossCritic}{\loss^{C}}
\newcommand{\lossActor}{\loss^{A}}
\newcommand{\loss}{\mathcal{L}_{\neuralnetwork}}

\newcommand{\expert}{\mathrm{E}}

\newcommand{\weightdiscrete}[1]{\AgentPov{p}{#1}}

\newcommand{\bucketSize}{B}

\newcommand{\valuefunctionPov}[1]{\AgentPov{V}{#1}}
\newcommand{\valuefunctionPovCond}[2]{\valuefunctionPov{#1} (#2)}
\newcommand{\valuefunctionTimePov}[2]{\TimeAgentPov{V}{#1}{#2}}
\newcommand{\valuefunctionlambdaTimePovCond}[3]{\valuefunctionTimePov{\lambda,#1}{#2} (#3)}

\newcommand{\discountlambda}{\lambda}
\newcommand{\horizonbootstrap}{\mathrm{N}}

\newcommand{\valuetargetPov}[1]{\AgentPov{\bar{V}}{#1}}
\newcommand{\valuetargetPovCond}[2]{\valuetargetPov{#1} (#2)}

\newcommand{\normalizationTerm}{S}

\begin{document}


\begin{frontmatter}


\paperid{719} 

\title{Enhancing Reinforcement Learning Through Guided Search}
\author[A]{\fnms{Jérôme}~\snm{Arjonilla}\orcid{0000-0002-0082-1939}\thanks{Corresponding Author. Email: jerome.arjonilla@hotmail.fr}}
\author[A]{\fnms{Abdallah}~\snm{Saffidine}\orcid{0000-0001-9805-8291}}
\author[B]{\fnms{Tristan}~\snm{Cazenave}\orcid{0000-0003-4669-9374}} 

\address[A]{PSL University - Dauphine, Paris, France }
\address[B]{Potassco Solutions, Potsdam, Germany}

\begin{abstract}
    With the aim of improving performance in Markov Decision Problem in an Off-Policy setting, we suggest taking inspiration from what is done in Offline Reinforcement Learning (RL). 
    In Offline RL, it is a common practice during policy learning to maintain proximity to a reference policy to mitigate uncertainty, reduce potential policy errors, and help improve performance. 
    We find ourselves in a different setting, yet it raises questions about whether a similar concept can be applied to enhance performance \ie, whether it is possible to find a guiding policy capable of contributing to performance improvement, and how to incorporate it into our RL agent. 
    Our attention is particularly focused on algorithms based on Monte Carlo Tree Search (MCTS) as a guide. 
    MCTS renowned for its state-of-the-art capabilities across various domains, catches our interest due to its ability to converge to equilibrium in single-player and two-player contexts. 
    By harnessing the power of MCTS as a guide for our RL agent, we observed a significant performance improvement, surpassing the outcomes achieved by utilizing each method in isolation. 
    Our experiments were carried out on the Atari 100k benchmark.    
\end{abstract}

\end{frontmatter}


\footnotetext{Paper accepted at the 27th European Conference on Artificial Intelligence (ECAI 2024).}

\section{Introduction}

Reinforcement Learning (RL) is a leading field in artificial intelligence, advancing our grasp of intelligent decision-making in complex environments~\cite{arulkumaran2017brief,sutton2018reinforcement}. 
Despite the remarkable progress, the pursuit of optimizing RL algorithms remains a central focus. 
In this pursuit, we turn our attention to a foundational concept within the realm of RL.
In Offline RL~\cite{levine2020offline,prudencio2023survey}, the primary objective is to derive the best possible policy solely from a dataset originating from an auxiliary policy, without interacting with the environment. 
The prevalent idea is to align the new policy closely with the auxiliary policy to enhance performance.
This strategy derives from the principle that deviating from the limits of the auxiliary policy often leads to uncertainty which leads to erroneous judgments about the policy's efficacy. 

Our scenario diverges from this framework and pivots back to a more classical approach where the constraints of an auxiliary policy fade away and we once again interact with the environment. 
Despite this paradigm shift, we question whether it is possible to preserve the concept of Offline RL \ie, staying as close as possible to an auxiliary policy to enhance performance. 
Considering our lack of auxiliary policy, we inquire whether it is plausible to use an online algorithm proficient enough to act as our guiding reference, and how to integrate such a guiding agent into our RL agent. 

In our investigation, we initially explore various online algorithms that can potentially serve as a guide. 
The existing literature presents algorithms that already exploit guide knowledge to improve performance.
For instance, algorithms such as Soft Actor-Critic (SAC) and Asynchronous Advantage Actor Critic (A3C)~\cite{haarnoja2018soft,haarnoja2018softb,mnih2016asynchronous} integrate an entropy term into the reinforcement learning (RL) agent. 
This entropy, in another formulation, is a measure of the distance between the current policy and the policy of a guide, of which this guide happens to be a random agent. 

In our research, we turn our attention to search algorithms, specifically focusing on Monte Carlo Tree Search (MCTS) as a guiding policy for reinforcement learning (RL) agents. 
MCTS-based approaches, well-established in game theory literature~\cite{browne_survey_2012,swiechowski2023monte}, obtain state-of-the-art performance across a spectrum of games, converging towards equilibrium even in complex scenarios involving one or two players.

Integrating MCTS as a guide yields significant performance improvements. Our analysis reveals that, in the majority of cases, this integration leads to enhanced performance. Even in instances where performance does not improve, the algorithms achieve optimal outcomes when compared individually.
By combining an RL algorithm with MCTS as a guide, we harness the generalization and learning capabilities inherent to RL, while also capitalizing on MCTS's optimal online decision-making prowess.
Furthermore, we extend our investigation by exploring various hyperparameters, with a keen focus on the degree of integration of the guide's policy. 
Through experimentation, we are demonstrating that it is possible to reduce the frequency of use of the guide, thereby mitigating associated overhead while retaining performance enhancements.

In Section~\ref{sec:notation}, we establish the formalism and notation employed throughout the paper.
Section~\ref{sec:main} presents multiple online guides, discussing their respective strengths and weaknesses, and elucidates the process of integrating a guide into the RL agent. 
Particularly incorporating MCTS-based algorithms as a guide offers valuable guidance to the RL agent in several key points: the actor and the critic components. 
In Section~\ref{sec:experiment}, we conduct experiments using various guides on the Atari100k benchmark.
Section~\ref{sec:related} provides an overview of related work in the field.
Lastly, Section~\ref{sec:conclusion} offers a summary of our findings and outlines avenues for future research.

\section{Formalism and Notation}~\label{sec:notation}

\subsection{Markov Decision Process}

A dynamic system is typically characterized by a Markov Decision Process (MDP), which is represented as $\mathcal{M} = (\worldstateSet, \actionSet, \transition, \reward, \discount)$. 
Here, $\worldstateSet$ denotes the state space where $\worldstate \in \worldstateSet$, $\actionSet$ represents the action space with $\action \in \actionSet$, $\transitionCond{\worldstateTime{t+1}}{\worldstateTime{\timestep}}{\actionTime{\timestep}}$ signifies the transition probability distribution governing the system dynamics, $\rewardPovCond{}{}{\worldstate}{\action}$ stands for the reward function, and $\discount \in (0, 1]$ serves as a discount factor. 

Dealing with an exact MDP can impose considerable computational burdens. 
Utilizing an approximation of $\mathcal{M}$, known as a world model~\cite{hafner2019dream,schrittwieser2020mastering, hafner2020mastering,hafner2023mastering}, can offer significant advantages. 
Employing the world model for information retrieval not only expedites computations compared to exact methods but also facilitates parallel processing of state batches, particularly when computing complex tools such as N-step bootstrapped $\lambda$-returns or employing MCTS.
This parallel processing is often performed on GPUs rather than CPUs, further enhancing computational efficiency.

\subsection{Reinforcement Learning}

Reinforcement learning confronts the problem of learning to control the MDP, where the agent tries to acquire a policy $\pi$, which is defined as a distribution over actions conditioned on state $\policyTimePovCond{}{}{\action}{\worldstate}$ that maximizes the long-term discounted cumulative reward defined as follow:
\begin{align}
    \policyTimePov{*}{} = \max_{\policyTimePov{}{}} \Ex{\trajectory \sim \policyTimePov{}{}}{\sum_{t=0}^\horizon \discount^t \rewardTimePov{\timestep}{}}
\end{align}
where $\trajectory=(\worldstateTime{0},\actionTime{0}, \rewardTimePov{0}{},\dots)$ is a sequence of states, actions, and rewards generated from the current policy.
To maximize the policy $\policyTimePov{}{}$, one of the primary methods utilized is the \emph{actor-critic} approach involves learning a critic and an actor-network. 
The learning can be conducted online by generating new trajectories or by leveraging a data buffer $\dataset{}$, which comprises past trajectories ${\trajectory_0, \trajectory_1, \dots, \trajectory_{k-1}}$. 

\subsubsection{Critic}

The critic aims to estimate the value functions, \ie the expected cumulative rewards an agent can obtain at a state:
\begin{align}
\label{equa:stateValueFunction}
\valuefunctionPovCond{\policyTimePov{}{}}{\worldstateTime{\timestep}} &= \Ex{\actionTime{\timestep} \sim \policyTimePovCond{}{}{\cdot}{\worldstateTime{\timestep}}}{\rewardTimePov{\timestep}{}+ \discount \Ex{\worldstateTime{t+1}\sim \transitionCond{\cdot}{\worldstateTime{\timestep}}{\actionTime{\timestep}}}{\valuefunctionPovCond{\policyTimePov{}{}}{\worldstateTime{t+1}}}} 
\end{align}

The loss function of the critic $\lossCritic$ is formulated to minimize the disparity between the value target $\valuetargetPovCond{\neuralnetwork}{\worldstate}$ and the predicted value $\valuefunctionPovCond{\neuralnetwork}{\worldstate}$ over a batch of state.
\begin{align}
    \label{equa:lossCritic}
    \lossCritic = \Ex{\worldstate \sim \dataset{}}{\lossCriticSub{\worldstate} }
\end{align}

Previous studies have emphasized the benefits of employing cross-entropy over a discrete representation in reinforcement learning \cite{bellemare2017distributional, schrittwieser2020mastering, hafner2023mastering, bellemare2023distributional, farebrother2024stop}.
This method involves the critic to learn a discrete weight distribution $\weightdiscrete{\neuralnetwork} = \{\weightdiscrete{1}, . . . , \weightdiscrete{\bucketSize} \} \in \mathbb{R}^\bucketSize$ instead of learning the mean of the distribution/ A function $y()$ is used to convert a target value into a corresponding weight distribution of size $\bucketSize$.
This leads to the following sub-loss for the critic:

\begin{align}
    \lossCriticSub{\worldstate} = y(\valuetargetPovCond{\neuralnetwork}{\worldstate})^T \log \weightdiscrete{\neuralnetwork}
\end{align}

The value target often corresponds to the Q-Value, yet, to enhance stability, an alternative approach involves using the $\horizonbootstrap$-step bootstrapped $\lambda$-returns~\cite{sutton2018reinforcement,hafner2023mastering}.
These returns incorporate predicted rewards and values~\cite{schulman2015high, sutton2018reinforcement} over a depth of $\horizonbootstrap$:

\begin{align}
\begin{cases}
     \valuefunctionPovCond{\neuralnetwork}{\worldstateTime{\timestep}} & \text{if } \horizonbootstrap = 0\\
     \rewardTimePov{\timestep} + \discount \left( (1-\discountlambda)\valuefunctionPovCond{\neuralnetwork}{\worldstateTime{\timestep+1}} + \discountlambda \valuefunctionlambdaTimePovCond{\horizonbootstrap-1}{\neuralnetwork}{\worldstateTime{\timestep+1}} \right) & \text{if } \horizonbootstrap>0
\end{cases}
\end{align}

\subsubsection{Actor}

The actor’s loss function, denoted as \(\lossActor\), is designed to maximize the expected reward by optimizing the actions that lead to states with the highest predicted values from the critic.
\begin{align}
    \label{equa:lossActor}
    \lossActor = \Ex{\worldstate \sim \dataset{}}{\lossActorSub{\worldstate}}
\end{align}

In the context of Atari Benchmarks, as observed in~\cite{hafner2020mastering}, authors have found it more advantageous to employ the Reinforce~\cite{williams1992simple} algorithm, which is the approach adopted in our work as well. 
Reinforce maximizes the actor's probability of its own sampled actions weighted by the values of those actions. 
One can reduce the variance of this estimator by subtracting the state value as a baseline. 
Therefore, we obtain the following loss for the actor:

\begin{align}
    \lossActorSub{\worldstate} = \Ex{\substack{\action \sim \policyTimePovCond{}{\neuralnetwork}{\cdot}{\worldstate}}}{-\ln \policyTimePovCond{}{\neuralnetwork}{\action}{\worldstate} \left( \frac{\valuetargetPovCond{\neuralnetwork}{\worldstate} - \valuefunctionPovCond{\neuralnetwork}{\worldstate}}{\normalizationTerm_\neuralnetwork } \right)}
\end{align}
where the term ${\normalizationTerm_\neuralnetwork }$ refers to the normalization factor used to stabilize the scale of returns. The normalization is carried out using an exponentially decaying average, is robust to outliers by taking the returns from the $5th$ to the $95th$ batch percentile, and reduces large returns without increasing small returns. 

\begin{align}
    \normalizationTerm_{\neuralnetwork} = \max\left(1,\text{Per}_{95}\left(\valuetargetPovCond{\neuralnetwork}{\cdot}\right) - \text{Per}_{5}\left(\valuetargetPovCond{\neuralnetwork}{\cdot}\right)\right)
\end{align}

\subsection{Behavior Cloning}

Behavior Cloning (BC)~\cite{hussein2017imitation} is a method employed in RL where the objective is to develop an agent capable of executing tasks closely resembling those of the demonstrator.  In this approach, the agent's policy, denoted as $\policyTimePov{}{BC}$, undergoes a supervised learning process to closely replicate the actions present in the dataset.

\begin{align}
    \policyTimePov{}{BC} = \max_{\policyTimePov{}{}} \Ex{(\action,\worldstate)\sim \dataset{}} {\log \policyTimePovCond{}{}{\action}{\worldstate}}
\end{align}

\subsection{Search Algorithm}

Search algorithms are algorithms that aim to explore the game tree efficiently to make informed decisions that maximize the chances of winning. 
To do this, search algorithms are given a larger budget in the given state that they wish to solve, and during the budget they efficiently explore the different possible paths of action, thus obtaining a better estimate of the value function and a better policy in the given state.

Search algorithms encompass a diverse range of techniques tailored to handle various game scenarios, from single-player to multi-player, and from perfect to imperfect information settings. 
In perfect information games like Chess or Go, where players have complete knowledge of the game state, algorithms like Minimax with Alpha-Beta Pruning \cite{knuth_analysis_1975, cohen2020minimax} or MCTS~\cite{browne_survey_2012,swiechowski2023monte}  are widely employed. 
Conversely, imperfect information games like Poker or Skat pose additional challenges due to hidden information. In such cases, techniques like Perfect Information Monte Carlo \cite{long_understanding_2010}, Information Set Monte Carlo Tree Search \cite{cowling_information_2012}, or Counterfactual Regret Minimization based method \cite{neller2013introduction} are utilized.

\subsubsection{Monte Carlo Tree Search}
MCTS is a tree search algorithm, for perfect information game that converges towards equilibrium with one and two players.
At each time step of the budget, MCTS (\romannumeral 1) selects the best path of node, (\romannumeral 2) expands the tree by adding a child node, (\romannumeral 3) estimates the child node, (\romannumeral 4) backpropagates the result obtained through the nodes chosen.
At the end of the budget, the algorithms return the distribution of actions $\policyTimePov{}{MCTS}$ that has been visited, and the value $\valuefunctionPov{MCTS}$ obtained when running MCTS. 

Starting from AlphaGo/AlphaZero (AZ) series~\cite{Silver2016MasteringTG,silver_general_2018,silver_mastering_2017}, MCTS has been combined with neural networks to enhance performance where an actor-network is used to help the search and a critic network is used to give a better estimate of the new state.
We denote $\policyTimePov{}{AZ}$/$\valuefunctionPov{AZ}$ the information returned when running MCTS with AlphaZero. 
This information is then utilized to compute the sub-actor loss $\lossActorSub{\worldstate}$ and the sub-critic loss $\lossCriticSub{\worldstate}$.
\begin{align}~\label{equa:lossActorMCTS}
    \lossCriticSub{\worldstate} &= y(\valuetargetPovCond{AZ}{\worldstate})^T \log \weightdiscrete{\neuralnetwork} \\
    \lossActorSub{\worldstate} &= \policyTimePovCond{}{AZ}{\cdot}{\worldstate}^T \log \policyTimePovCond{}{\neuralnetwork}{\cdot}{\worldstate}
\end{align}

\section{Guide}
~\label{sec:main}
As mentioned in the introduction, we aim to find an online algorithm that can guide our RL agent to improve its performance. 
In this objective, we will first investigate the advantages and disadvantages of different guides, and then we will explain how to integrate the guide into the RL agent.

\begin{table*}[!htbp]
\small
\bigskip
\caption{\centering Advantage and Inconvenient of using each guide. $\text{\textcolor{teal}{Yes}}^*$ implies that the algorithm is relevant to improve exploration, but only if the action produced is also relevant to improve performance.}
\bigskip
\centering
\begin{tabular}{l *{7}c} 
   \toprule
   \diagbox{Criteria}{Guide} & $\policyTimePov{}{Random}$ & $\policyTimePov{}{BC}$ & $\policyTimePov{\neuralnetwork}{BC}$ &$\policyTimePov{}{MCTS}$ & $\policyTimePov{}{AlphaZero}$ & $\policyTimePov{}{Human}$ \\
   \midrule

   Available at each ($\worldstate$,$\action$) & \textcolor{teal}{Yes} & \textcolor{red}{No} & \textcolor{teal}{Yes} & \textcolor{teal}{Yes} & \textcolor{teal}{Yes} & \textcolor{red}{No}\\

    Relevant for exploration & \textcolor{teal}{Yes} & \textcolor{red}{No} & \textcolor{red}{No} & $\text{\textcolor{teal}{Yes}}^*$ & $\text{\textcolor{teal}{Yes}}^*$ & $\text{\textcolor{teal}{Yes}}^*$ \\

    Relevant for performance & \textcolor{red}{No} & \textcolor{red}{No} & \textcolor{red}{No} & \textcolor{teal}{Yes} & \textcolor{teal}{Yes} & \textcolor{teal}{Yes} \\
    Reduce extrapolation error & \textcolor{red}{No} & \textcolor{teal}{Yes} & \textcolor{teal}{Yes} & \textcolor{red}{No} & \textcolor{red}{No} & \textcolor{red}{No} \\

   Performance is not time-dependent & \textcolor{teal}{Yes} & \textcolor{red}{No} & \textcolor{red}{No} & \textcolor{teal}{Yes} & \textcolor{red}{No} & \textcolor{teal}{Yes}\\

    Online Cost & \textcolor{teal}{Low} & \textcolor{teal}{Low}  & \textcolor{teal}{Low} & \textcolor{orange}{Medium} & \textcolor{orange}{Medium} & \textcolor{red}{High} \\
    
    Offline Cost & \textcolor{teal}{Low} & \textcolor{teal}{Low}  & \textcolor{orange}{Medium} & \textcolor{teal}{Low} & \textcolor{orange}{Medium} & \textcolor{red}{High}\\
   
   \bottomrule

\end{tabular} 
~\label{tab:procons}
\end{table*}

\subsection{Analysis of the various guides}

To thoroughly assess the efficacy of different guides and determine their suitability for guiding the reinforcement learning algorithm, we conducted a comprehensive evaluation based on multiple criteria. The gathered information is summarized in Table~\ref{tab:procons}. 
The guides discussed are detailed below and are identified as follows `Human', `Random', `BC', and `MCTS'.%

The criteria take into account their capacity to be available in each state-action, if they are relevant for exploring/performance, their online and offline cost, if they can reduce the extrapolation error, and if they are time dependent. 
Time-dependent algorithms are those that require learning before they are operational, for example, learning a neural network. 
Extrapolation error~\cite{fujimoto2019off} is an error present in Off-Policy and Offline problems that arise when the target selects actions rarely present in the dataset, affecting the accuracy of the value estimate.

\subsubsection{Human}

The use of guides is often associated with the use of human guides, whether for learning to drive~\cite{huang2022efficient,wu2023toward}, for conversing with other humans~\cite{jaques2019way} or even for trying to play as much as a human~\cite{bakhtin2022mastering}. 
It is a necessity in scenarios where real-time interaction is either infeasible or the risk is too significant.
The initial stages of a game present a valuable opportunity for the incorporation of human policies. 
During this phase, RL policies may prove ineffective, whereas human policies are directly applicable and advantageous. 
Unfortunately, the data are available in a restricted subset of all state-action, are expensive and complex to obtain.

\subsubsection{Random}

In algorithms like SAC~\cite{haarnoja2018soft} and several state-of-the-art counterparts~\cite{hafner2023mastering}, the RL agent is coupled with an entropy term to enhance exploration. 
In an alternative perspective, this entropy is a measure of the distance between the current policy $\policyTimePov{}{}$ and the policy of a guide, of which this guide happens to be a random agent. 
The choice of a random agent as a guide holds distinct advantages, particularly when exploration of the state space is desired, its minimal computational cost and immediate availability make it an ideal choice in many scenarios.
However, reservations emerged when considering the utility of such a guide in enhancing overall performance.

\subsubsection{Behavior Cloning}

In Offline RL, a common strategy involves approximating closeness to the behavioral policy that underlies the $\dataset{}$ dataset. Achieving this requires an initial step of estimating the behavioral policy by behavioral cloning. This estimate of the behavior policy is then used as a guide for RL agents. This method yields a significant advantage by minimizing extrapolation errors. By aligning the new policy closely with the behavior policy, the algorithm performs actions for which accurate approximations exist, reducing uncertainties of the new policy.
However, several considerations come into play. Firstly, the guide is not inherently well-suited for exploration or enhancing performance. Secondly, the data is confined to a subspace of the state space and depends on the amount of interaction.


\subsubsection{Search Algorithm}

Leveraging a search algorithm as a guide stands as a reasonable choice given its constant availability in each state and its relatively low cost compared to human guidance. 
Particularly, in contrast to employing either a random guide or a guide relying solely on past data, search-based algorithms hold greater potential for performance enhancement due to their abilities to explore and converge toward the optimal solution. 
It is noteworthy, however, that while search algorithms are less expensive than human guidance, they may incur higher costs than alternative methods. 
Additionally, under constrained resource budgets or insufficient training of neural networks, search algorithms may encounter challenges in converging toward the optimal solution.

\subsection{How to integrate a guide into the RL agent}

Offline RL~\cite{levine2020offline,prudencio2023survey} domain offers diverse methods for aligning one policy with another, contingent on the degree of closeness desired between them. Possible methods include value penalty where the penalty term is incorporated into the reward function or policy regularization where the penalty term is incorporated after the calculation of the loss. 
In our work, we have chosen to implement regularization techniques. 

Our approach to incorporating the guide is largely inspired by the work of~\citet{shi2023offline}, which, to our knowledge, stands as the sole study employing both policy and critic information. 
This choice is based on our ability to leverage the information provided by the search algorithms, especially $\policyTimePov{}{AZ}$ help to influence the actor policy and $\valuefunctionPov{AZ}$ help to shape the critic. 

In the subsequent discussion, we adopt general notations that consider the possibility of multiple guides.
$\expert = \{\expert_\agent\}_{\agent \in \mathcal{N}}$ denotes the set of guide algorithms, each exerting varying degrees of influence on the decision-making process. 

\subsubsection{Critic Incorporation}

By integrating the guide into the critic, our objective is to refine the estimation of the value function by considering the insights provided by the guide. 
Incorporating a penalty into the critic using value regularization amounts to change from Equation~\eqref{equa:lossCritic} to equation the following new loss function $\lossCritic$:
\begin{align}
\label{equa:NewlossCritic}
    \Ex{\worldstate \sim \dataset{}}{\lossCriticSub{\worldstate}  \boldsymbol{+ \sum_{\expert_\agent \in \expert} \weightcritic{\expert_\agent}(\worldstate) \distanceCritic{\expert_\agent}{\valuefunctionPovCond{\neuralnetwork}{\worldstate}}{\valuetargetPovCond{\expert_\agent}{\worldstate}}} }
\end{align}
where $\distanceCritic{\expert_\agent}{}{}$ is the penalty term between the guide target $\valuetargetPovCond{\expert_\agent}{\worldstate}$ and the predicted value, and $\weightcritic{\expert_\agent}(\worldstate)$ is the function weight used for regularizing the penalty term. 
The penalty term can be any function that evaluates the disparity, and in particular, the same function as the critic's sub-loss.
Similarly, to enhance stability, one can compute the $\horizonbootstrap$-step bootstrapped $\lambda$-returns on the target value. 

\subsubsection{Actor Incorporation}

To incorporate the guide on the actor, we used information from the guide on the actor and the critic. The use of the critic allows us to increase guidance when states are promising or have high potential. Incorporating a penalty into the actor using regularization amounts to change from Equation~\eqref{equa:lossActor} to the following loss function of the actor $\lossActor$:
\begin{align}
\label{equa:actor_final}
    \Ex{\worldstate \sim \dataset{}} {\frac{\lossActorSub{\worldstate}}{\boldsymbol{\Ex{}{|\lossActorSub{\cdot}|}}} \boldsymbol{ + \sum_{\expert_\agent \in \expert} \weightfullactor{\expert_\agent}(\worldstate) \distanceActor{\expert_\agent}{{\policyTimePovCond{}{\neuralnetwork}{\cdot}{\worldstate}}}{{\policyTimePovCond{}{\expert_\agent}{\cdot}{\worldstate}}}}}
\end{align}
where $\distanceActor{\expert_\agent}{}{}$ represents the penalty term between the actor-network and the target policy, and $\weightfullactor{\expert_\agent}(\worldstate)$ is a function determining the penalty weight based on the current state. 
The penalty term can be any function that evaluates the disparity, yet, in Offline RL, the penalty term is often the KL divergence \cite{wu2019behavior}.

The loss of the actor-network significantly depends on the scale of the internal loss values. To address this, we normalize $\lossActorSub{\worldstate}$ by the average absolute value of $\lossActorSub{\cdot}$. This mean term is estimated over mini-batches and is solely used for scaling purposes.
The weight $\weightfullactor{\expert_\agent}(\worldstate) \in [\weightactor{\expert_\agent}(\worldstate), \weightactor{\expert_\agent}(\worldstate) \cdot \maxlambda{\expert_\agent}]$ is a function that serves to emphasize the increased penalty on high-quality state \ie, more weight is given to states that perform better than the target, which results in more attention toward the policy given by the guide. 
\begin{align}
    \weightactor{\expert_\agent}(\worldstate) \cdot \text{Clip}\left[ \text{exp} \left(\expoTerm{\expert_\agent} \frac{ \valuetargetPovCond{\expert_\agent}{\worldstate}-\valuetargetPovCond{\neuralnetwork}{\worldstate}}{\normalizationTerm_{\expert_\agent}} \right), (1,\maxlambda{\expert_\agent}) \right]
\end{align}

In this equation, the state's quality is assessed through the term $\valuefunctionPovCond{\expert_\agent}{\worldstate}-\valuefunctionPovCond{\neuralnetwork}{\worldstate}$ normalized by $\normalizationTerm_{\expert_\agent}$. 
The normalization is carried out using an exponentially decaying average, robust to outliers by taking the returns from the $5th$ to the $95th$ batch percentile, and reduces large returns without increasing small returns. 
\begin{align}
    \normalizationTerm_{\expert_\agent} = \max\left(1,\text{Per}_{95}\left(\valuefunctionPovCond{\expert_\agent}{\cdot}\right) - \text{Per}_{5}\left(\valuefunctionPovCond{\expert_\agent}{\cdot}\right)\right)
\end{align}

\section{Experimentation}
~\label{sec:experiment}
\subsection{Experimental Information}

\subsubsection{Benchmarks}

Atari 100k~\cite{kaiser2019model} serves as a comprehensive benchmark comprising $26$ Atari games, providing a diverse range of challenges to assess various algorithms' performance. 
In this benchmark, agents train for $100$k steps, equivalent to $400$k frames (considering a frameskip of $4$). 
Each block of $100$k steps approximately aligns with $2$ hours of real-time gameplay per environment.

\subsubsection{Algorithms}

The algorithms used are namely (\romannumeral1) AlphaZero (AZ)~\cite{silver_mastering_2017}; (\romannumeral2) A2C (Advantage Actor-Critic)~\cite{mnih2016asynchronous}; (\romannumeral3) A2C with random agent as a guide, noted as A2C-Rand (similar to SAC); (\romannumeral4) A2C with behavior cloning as an guide, noted as A2C-BC; (\romannumeral5) A2C agent with AlphaZero as an guide, noted as A2C-AZ or A2C-AZ* where A2C-AZ uses a fixed hyperparameter $\weightactor{}$ for all games and A2C-AZ* uses a fine-tuned $\weightactor{}$ for each game.

Given the novelty of our approach, we conducted experiments with a single guide ($|\expert|=1$) and uniform weights assigned across all states ($\forall \worldstate, \weightcritic{\expert_\agent}(\worldstate) = \weightcritic{\expert_\agent}$ and $\forall \weightactor{\expert_\agent}(\worldstate) = \weightactor{\expert_\agent}$).
Furthermore, in our study, we used A2C as the reinforcement learning algorithm and AlphaZero as the guide. However, our implementation is not limited to these specific algorithms. 
Various other RL algorithms and search algorithms could have been explored as alternative options for experimentation.

\subsubsection{Actor/Critic}

All the algorithms use a critic and an actor network, composed of a two-layered MLP network of 512 hidden units.
As defined in the introduction, the critic loss sub $\lossCriticSub{}$ uses a cross-entropy based on a discrete representation~\cite{schrittwieser2020mastering,hafner2023mastering} and the actor loss sub $\lossActorSub{}$ uses reinforce with an advantage baseline to reduce the variance. 

The distance function $\distanceActor{}{}{}$ used for the actor is a KL-divergence function and the distance function $\distanceCritic{}{}{}$ used for the critic is a cross-entropy. 
The weight of the guide penalty $\weightactor{}$ is fixed at $0.08$ for behavior cloning and at $0.03$ for random (both where chosen between [0.03, 0.08, 0.3]), and unless otherwise stated, set at $0.7$ for MCTS. 
For A2C-AZ, the weight for the critic is fixed at $0.05$. For enhancing stability, the guide value target $\valuetargetPovCond{\expert_\agent}{}$ and the value target $\valuetargetPovCond{\neuralnetwork}{}$ use the $\horizonbootstrap$-step bootstrapped $\discountlambda$-returns.

\subsubsection{Monte Carlo Tree Search}

A2C-AZ utilizes the actor and critic networks of the A2C agent, ensuring that it does not deviate significantly from it.
Our implementation of MCTS in A2C-AZ and AlphaZero is built on previous famous MCTS implementations~\cite{Silver2016MasteringTG, silver_mastering_2017,schrittwieser2020mastering,ye2021mastering}.
It uses a search budget of $50$, PUCT in the selection and Dirichlet noise distribution to help explore.
However, three differences should be noted (\romannumeral1) we do not use Re-Analyse; (\romannumeral2) we do not use prioritized experience replay~\cite{schaul2015prioritized}; (\romannumeral3) we do not use the search algorithm in the test phase. 
These differences were made to effectively compare the different algorithms.

\subsubsection{Metrics}

We report the raw performance on each game, the human normalized score, as well as the Interquartile Mean (IQM) and Optimality Gap. 
The IQM and the Optimality Gap are metrics recommended for Atari100K benchmarks~\cite{agarwal2021deep} where the authors recommend using IQM instead of the Median, and Optimality Gap instead of Mean, as both methods are more robust. 
IQM calculates the average over the data, removing the top and bottom 25\%. 
Optimality Gap computes the amount by which the algorithm fails to meet a minimum score. 
A higher score is better for the IQM and a lower score is better for the optimality gap.

\subsubsection{Other}

Each agent uses a single environment instance with a single \emph{NVIDIA V100 GPU}. 
Each algorithm is run using $5$ seeds, we evaluated performance every $10$k training step with $10$ independent run of the game. 
To mitigate training expenses, we conducted our experiments by using a world model. 
We employed the fixed-trained weights from the Dreamer algorithm~\cite{hafner2023mastering}, a state-of-the-art model-based technique trained over $50,000$k steps. 
The world model is used to compute the $\horizonbootstrap$-step bootstrapped $\lambda$-returns for A2C algorithms and facilitating MCTS in A2C-AZ and AlphaZero. 
Additionally, to enhance cost-effectiveness and stability, we restricted our experimentation to $21$ out of $26$ games, excluding those where the world model demonstrated poorer performance in terms of mean human-normalized scores.

\subsection{Experiments}

Initially, we will examine the overall impact of the various algorithms and guides. 
Subsequently, we narrow our focus on MCTS as a guide, analyzing the experiments in greater detail.
Finally, we analyze the impact of the guide's weight, by testing several weights and trying to observe the impact when the guide is called less often. 

\subsubsection{Overall analysis}
 
\begin{figure}[!htbp]
    \centering
    \subcaptionbox{IQM}[\columnwidth]{\includegraphics[scale=0.37]{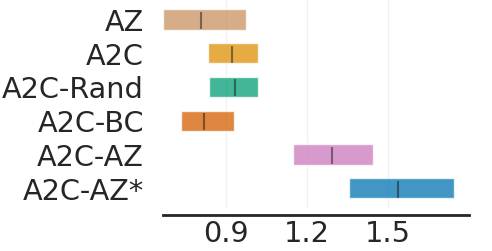}}
    
    \bigskip 

    \subcaptionbox{Optimality Gap}[\columnwidth]{\includegraphics[scale=0.37]{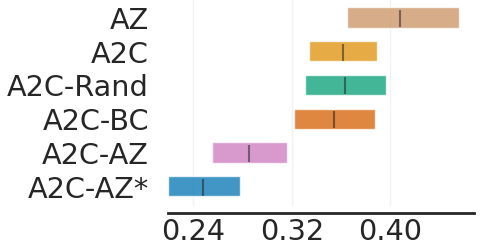}}

    \bigskip 

    \caption{\centering Aggregate performance. Shaded area shows $95\%$ stratified bootstrap confidence interval. The x-axis represents the human normalized score.} 
    \bigskip

~\label{fig:aggregatateData}
\end{figure}

Figure~\ref{fig:aggregatateData} analyzes the overall performance using the IQM and Optimality Gap metrics for all the different guides considered.
We notice a significant enhancement in performance when utilizing AZ as a guide across both metrics. Specifically, A2C-AZ with a fixed weight surpasses A2C by over $0.4$ on the IQM and $0.1$ on the Optimality Gap.
Fine-tuning the weight further improves performance, with IQM increasing from $1.3$ to $1.5$ and the Optimality Gap decreasing from $0.28$ to $0.24$.

\subsubsection{MCTS as a guide}

\begin{figure}[!htbp]
    \centering
        \subcaptionbox{Asterix \label{fig:asterix}}[\columnwidth]{\scalebox{0.65}{\input{Image/asterix_reduced.tex}}}
        
        \bigskip

        \subcaptionbox{Assault\label{fig:breakout}}[\columnwidth]{\scalebox{0.65}{\input{Image/assault_reduced.tex}}}
        
        \bigskip
        
        \caption{\centering Learning curves on $2$ different game of Atari100k benchmarks with $3$ algorithms presented. The shaded area shows $95\%$ confidence interval.}
        \bigskip

    ~\label{fig:MctsOrRl}
\end{figure}
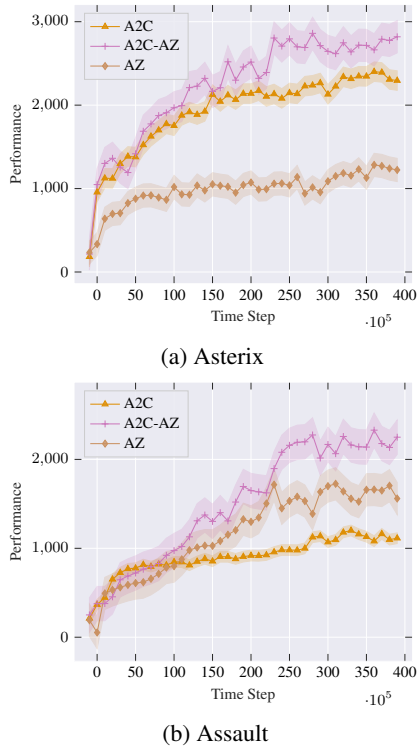

\begin{figure*}[!htbp]
\centering
    \subcaptionbox{X is A2C-AZ*, Y is A2C agent}[0.3\textwidth]{\includegraphics[width=\linewidth]{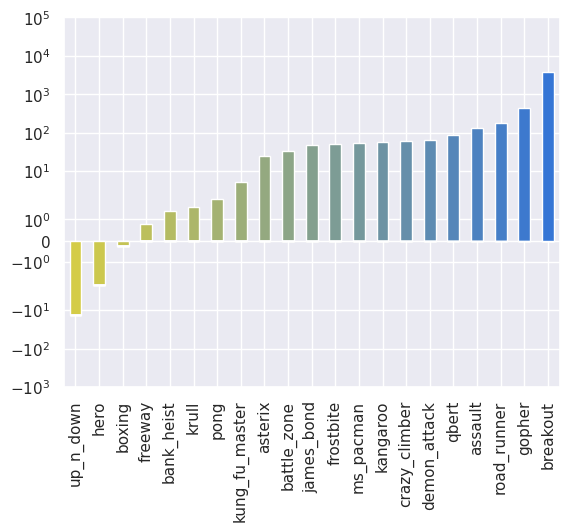}}
    \hfill
    \subcaptionbox{X is A2C-AZ, Y is A2C agent\label{fig:RLMCTS_vs_RL}}[0.3\textwidth]{\includegraphics[width=\linewidth]{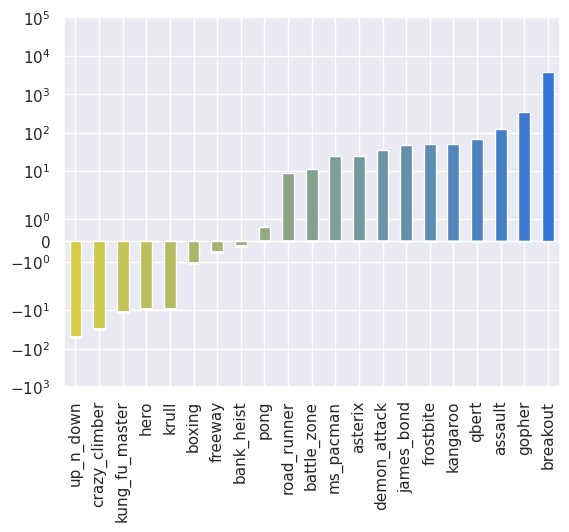}}
    \hfill
    \subcaptionbox{X is AlphaZero, Y is A2C agent \label{fig:MCTS_vs_RL}}[0.3\textwidth]{\includegraphics[width=\linewidth]{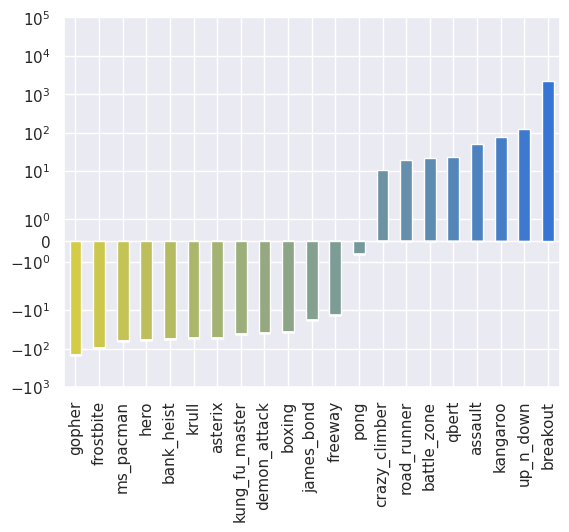}}
    
    \bigskip
    \caption{\centering Percentage improvement of algorithm X compared to algorithm Y on Atari100k Benchmarks. Improvement is measured as a percentage of mean human-normalized return. }
    \bigskip

    ~\label{fig:increase_performance_mcts}
\end{figure*}

In Figure~\ref{fig:increase_performance_mcts}, we observe the percentage improvement of A2C-AZ*/A2C-AZ/AlphaZero over A2C.
Furthermore, Figure~\ref{fig:MctsOrRl} displays a series of learning curves for A2C-AZ, A2C, and AlphaZero, forming the foundation for our subsequent analysis.

We begin our analysis by comparing the performance of AZ and A2C agents independently. Each figure represents distinct scenarios: one where AZ outperforms A2C (Figure \ref{fig:MctsOrRl}.a) and another where A2C outperforms AZ (Figure \ref{fig:MctsOrRl}.b). These figures provide an initial glimpse into the broader performance trends.

Overall, A2C demonstrates superior performance in 12 games, while AZ surpasses A2C in 8 games, with 1 game showing equivalent performance. Despite the general advantage of A2C, it is essential to highlight instances where A2C falls short, indicating the potential benefits of integrating AZ as a guide.

When considering the incorporation of AZ as a guide, several critical questions arise: can this integration elevate the agent's performance to at least match the best of the two individual agents? Is it possible to create an agent superior to the best individual performer, or might utilizing the guide lead to a weakened agent?

Our analysis across games shows that, compared to A2C, $12$/$17$ games exhibit performance improvements, $4$/$2$ show equivalent performance, and $5$/$2$ show lower performance when using the combined approaches A2C-AZ and A2C-AZ*. 
Interestingly, in the subset of 8 games where AZ outperformed A2C in isolation, integrating AZ as a guide resulted in superior performance in $6$/$7$ of those instances.

Although not visible in the figure, but observable in supplementary material. Our observations indicate that the combined algorithm outperforms both individual algorithms in $9$/$11$ instances, achieves the performance of the better of the two methods in $7$/$9$ instances, is lower than the best but bounded by the two algorithms in $4$/$1$ instances, and shows lower performance than both in only $1$/$0$ instance.

\subsubsection{Weight of the guide}

\begin{figure}[!htbp]
    \centering
    \subcaptionbox{IQM}[\columnwidth]{\includegraphics[scale=0.35]{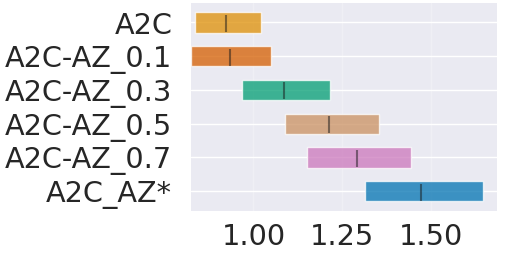}}
    
    \bigskip

    \subcaptionbox{Optimality Gap}[\columnwidth]{\includegraphics[scale=0.35]{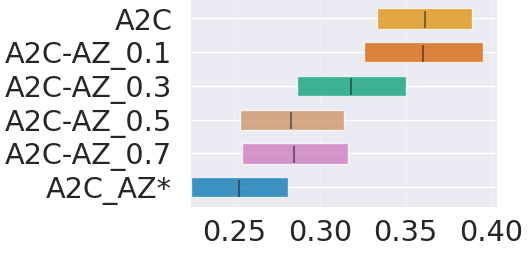}}
    
    \bigskip

    \caption{\centering Aggregate performance metrics according to the weight. The shaded area shows $95\%$ stratified bootstrap confidence interval.  The x-axis represents the human normalized score.}
    \bigskip

    ~\label{fig:aggregatateData3}
\end{figure}

In Figure \ref{fig:aggregatateData3}, we explore the impact of the weight parameter, $\weightactor{}$, on performance. We compare several fixed values of $\weightactor{}$ ranging from $0.1$ to $0.7$, alongside the optimal weight selected for each game.
Each variant of A2C-AZ is denoted by A2C-AZ-X, where X represents the specific weight used. For instance, A2C-AZ-0.3 employs AZ as a guide with a weight of $0.3$.

Upon examination, we find that the optimal fixed weight is $0.7$, resulting in an IQM of $1.29$ and an Optimality Gap of $0.28$. Notably, reducing the weight significantly leads to performance outcomes closely resembling those of A2C alone. 


\subsubsection{Cost of using a guide}

Throughout our previous experiments, we have observed a significant advantage in using AZ as a guide. However, as indicated in Table~\ref{tab:procons}, there is an overhead cost associated with employing AZ.

In practice, several MCTS methods can significantly reduce this cost, such as batch MCTS \cite{cazenave_batch_2021} and various parallelization techniques (leaf \cite{cazenave2007parallelization}, root \cite{chaslot2008parallel}, and tree \cite{cazenave2008parallel}). Additionally, many implementations utilize extensive computational resources to better distribute the workload. For instance, the basic version of AlphaGo uses 40 search threads, 48 CPUs, and 8 GPUs.

In the following experiment, we demonstrate another way to reduce the cost of incorporating a guide, which is quite natural in our context. Currently, the guide is executed at every iteration. However, our goal is to avoid deploying the guide in every situation. We aim to activate the guide only when necessary—specifically, in scenarios where our RL agent faces challenges or when the guide is known to excel.

\begin{figure}[!htbp]
    \centering
    \subcaptionbox{IQM}[\columnwidth]{\includegraphics[scale=0.35]{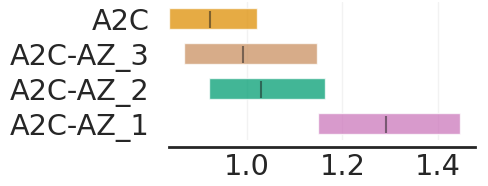}}
    
    \bigskip

    \subcaptionbox{Optimality Gap}[\columnwidth]{\includegraphics[scale=0.35]{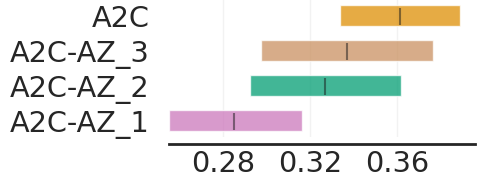}}

    \bigskip

    \caption{\centering Aggregate performance according to the number of calls made to the guide. The shaded area shows $95\%$ stratified bootstrap confidence interval. The x-axis represents the human normalised score. }
    \bigskip

    \label{fig:aggregatateData4}
\end{figure}

Figure~\ref{fig:aggregatateData4} shows the impact of using the guide less frequently. Instead of employing the guide at each iteration, we use it at every N iterations. We introduce the notation A2C-AZ-X, where X indicates how often the guide is called. For example, A2C-AZ-3 uses MCTS every three steps. Additionally, Table~\ref{tab:time_algo} presents the overhead cost of using AZ as a guide according to different values of N.

\begin{table}[!htbp]
    \small
    \medskip

    \caption{Runtime on Atari100k Benchmarks.}
    \medskip

    \centering
    \begin{tabular}{lcccc} 
       \toprule
       \textbf{Algorithm} & A2C & A2C-AZ-1 & A2C-AZ-2 & A2C-AZ-3 \\
        \midrule
        \textbf{Time} & 4:00& 18:00& 11:00 & 8:30 \\
        \bottomrule
    \end{tabular} 
    \label{tab:time_algo}
\end{table}

As observed, when running AZ at every iteration, the algorithm will achieve the best performance but take $18$ hours to complete instead of the $4$ hours required by A2C alone. However, by reducing the frequency to every two iterations, the runtime is reduced by half while still achieving a performance close to the best.



\section{Related Work}~\label{sec:related}
\subsection{Offline Reinforcement Learning}

Our work is strongly linked to the field of Offline RL as inspired by one of the key methods in the field. 
In our case, we have chosen to use regularization methods to align the policy and the value function with the guide. 
Yet, within the realm of regularization methods, there exist many methods, these include penalties applied within the reward function~\cite{wu2019behavior} or regularization penalties applied after its computation of the loss~\cite{kumar2020conservative,kostrikov2021offline}. 
Additionally, the calculation of the penalty can be accomplished by using various functions including KL divergence, Maximum Mean Discrepancy~\cite{wu2019behavior}, or even Fisher information~\cite{kostrikov2021offline}. 

\subsection{Monte Carlo Tree Search }

MCTS~\cite{browne_survey_2012,swiechowski2023monte} stands as a state-of-the-art algorithm that has significantly enhanced performance and tackled complex problems. In recent years, MCTS has been integrated with neural networks to boost its performance~\cite{Silver2016MasteringTG,silver_general_2018,silver_mastering_2017}, however, in most scenarios, neural networks are employed to predict the outcomes generated by MCTS. 

To our knowledge, no prior work has attempted to utilize MCTS as a guide while retaining the RL module. The most closely related study we encountered is~\cite{kartal2019action}, where the authors incorporated A3C with K workers, among which MCTS was one, resulting in notable performance enhancements. However, unlike their method, our approach utilizes MCTS as a guide in every state. Furthermore, we consider not only the policy distribution but also the value returned by MCTS to enrich learning. Additionally, we introduce an adaptive weight for actor learning and conduct a more comprehensive set of experiments.
\section{Conclusion}
~\label{sec:conclusion}

In this paper, we investigate the influence of leveraging online algorithms as a guide to enhance the learning process of RL algorithms. 
Inspired by techniques in Offline RL, we adapt these methodologies to the context of using an online algorithm, as a guide. 
Our approach involves regularizing the loss functions for both the actor and the critic to incorporate the information provided by the guide effectively.

Among the array of online algorithms explored from existing literature, our focus lies on Monte Carlo Tree Search (MCTS), a cutting-edge planning algorithm renowned for its convergence capabilities in both single-player and two-player scenarios. 
Notably, employing MCTS as a guide yields superior results compared to employing either of the two methods in isolation. 
Furthermore, fine-tuning just one hyperparameter can extend performance gains. 
Additionally, reducing the frequency of the guide calls can mitigate the cost associated with it while still resulting in enhanced performance. 

In the future, there exist promising avenues for further exploration. 
Experimenting with diverse hyperparameters, such as alternative distance functions, different search algorithms or different reinforcement learning algorithms, could illuminate nuanced insights. 
Additionally, exploring the integration of multiple guides could broaden the range of possibilities, incorporating different perspectives from various guides. Finally, investigating the utilization of an automatic weight, potentially based on neural networks, could provide a more adaptive, efficient, and general approach.

\bibliography{Doctorat}

\input{Appendix.tex}


\end{document}

%% file: Image/asterix_reduced.tex
\begin{tikzpicture}

\definecolor{chocolate213940}{RGB}{213,94,0}
\definecolor{darkcyan1115178}{RGB}{1,115,178}
\definecolor{darkcyan2158115}{RGB}{2,158,115}
\definecolor{darkorange2221435}{RGB}{222,143,5}
\definecolor{darkslategray38}{RGB}{38,38,38}
\definecolor{lavender234234242}{RGB}{234,234,242}
\definecolor{lightgray204}{RGB}{204,204,204}
\definecolor{orchid204120188}{RGB}{204,120,188}
\definecolor{peru20214597}{RGB}{202,145,97}

\begin{axis}[
axis background/.style={fill=lavender234234242},
axis line style={white},
legend cell align={left},
legend style={
  fill opacity=0.8,
  draw opacity=1,
  text opacity=1,
  at={(0.03,0.97)},
  anchor=north west,
  draw=lightgray204,
  fill=lavender234234242
},
x grid style={white},
xlabel=\textcolor{darkslategray38}{Time Step},
xmajorgrids,
xmajorticks=true,
xmin=-30000, xmax=410000,
xtick style={color=darkslategray38},
xtick={0,50000,100000,150000,200000,250000,300000,350000,400000},
xticklabels={0,50,100,150,200,250,300,350,400},
y grid style={white},
ylabel=\textcolor{darkslategray38}{Performance},
ymajorgrids,
ymajorticks=true,
ymin=-140.097425583967, ymax=3205.53861093376,
ytick style={color=darkslategray38}
]

\path [draw=darkorange2221435, fill=darkorange2221435, opacity=0.2]
(axis cs:-10000,299.41567007528)
--(axis cs:-10000,65.5843299247201)
--(axis cs:0,839.08432992472)
--(axis cs:10000,1007.08432992472)
--(axis cs:20000,1004.58432992472)
--(axis cs:30000,1180.58432992472)
--(axis cs:40000,1268.58432992472)
--(axis cs:50000,1260.58432992472)
--(axis cs:60000,1406.08432992472)
--(axis cs:70000,1509.08432992472)
--(axis cs:80000,1583.08432992472)
--(axis cs:90000,1658.08432992472)
--(axis cs:100000,1636.08432992472)
--(axis cs:110000,1760.08432992472)
--(axis cs:120000,1800.58432992472)
--(axis cs:130000,1769.58432992472)
--(axis cs:140000,1807.58432992472)
--(axis cs:150000,2009.58432992472)
--(axis cs:160000,1926.58432992472)
--(axis cs:170000,2003.08432992472)
--(axis cs:180000,1949.08432992472)
--(axis cs:190000,2021.58432992472)
--(axis cs:200000,2022.08432992472)
--(axis cs:210000,2056.08432992472)
--(axis cs:220000,1987.08432992472)
--(axis cs:230000,2017.08432992472)
--(axis cs:240000,1963.58432992472)
--(axis cs:250000,2029.58432992472)
--(axis cs:260000,2018.08432992472)
--(axis cs:270000,2110.58432992472)
--(axis cs:280000,2121.08432992472)
--(axis cs:290000,2148.58432992472)
--(axis cs:300000,2012.58432992472)
--(axis cs:310000,2108.58432992472)
--(axis cs:320000,2222.58432992472)
--(axis cs:330000,2198.58432992472)
--(axis cs:340000,2229.58432992472)
--(axis cs:350000,2228.08432992472)
--(axis cs:360000,2284.08432992472)
--(axis cs:370000,2275.58432992472)
--(axis cs:380000,2190.58432992472)
--(axis cs:390000,2177.58432992472)
--(axis cs:390000,2411.41567007528)
--(axis cs:390000,2411.41567007528)
--(axis cs:380000,2424.41567007528)
--(axis cs:370000,2509.41567007528)
--(axis cs:360000,2517.91567007528)
--(axis cs:350000,2461.91567007528)
--(axis cs:340000,2463.41567007528)
--(axis cs:330000,2432.41567007528)
--(axis cs:320000,2456.41567007528)
--(axis cs:310000,2342.41567007528)
--(axis cs:300000,2246.41567007528)
--(axis cs:290000,2382.41567007528)
--(axis cs:280000,2354.91567007528)
--(axis cs:270000,2344.41567007528)
--(axis cs:260000,2251.91567007528)
--(axis cs:250000,2263.41567007528)
--(axis cs:240000,2197.41567007528)
--(axis cs:230000,2250.91567007528)
--(axis cs:220000,2220.91567007528)
--(axis cs:210000,2289.91567007528)
--(axis cs:200000,2255.91567007528)
--(axis cs:190000,2255.41567007528)
--(axis cs:180000,2182.91567007528)
--(axis cs:170000,2236.91567007528)
--(axis cs:160000,2160.41567007528)
--(axis cs:150000,2243.41567007528)
--(axis cs:140000,2041.41567007528)
--(axis cs:130000,2003.41567007528)
--(axis cs:120000,2034.41567007528)
--(axis cs:110000,1993.91567007528)
--(axis cs:100000,1869.91567007528)
--(axis cs:90000,1891.91567007528)
--(axis cs:80000,1816.91567007528)
--(axis cs:70000,1742.91567007528)
--(axis cs:60000,1639.91567007528)
--(axis cs:50000,1494.41567007528)
--(axis cs:40000,1502.41567007528)
--(axis cs:30000,1414.41567007528)
--(axis cs:20000,1238.41567007528)
--(axis cs:10000,1240.91567007528)
--(axis cs:0,1072.91567007528)
--(axis cs:-10000,299.41567007528)
--cycle;

\path [draw=orchid204120188, fill=orchid204120188, opacity=0.2]
(axis cs:-10000,408.464245637496)
--(axis cs:-10000,22.9357543625044)
--(axis cs:0,855.135754362504)
--(axis cs:10000,1108.6357543625)
--(axis cs:20000,1170.1357543625)
--(axis cs:30000,1064.3357543625)
--(axis cs:40000,999.335754362504)
--(axis cs:50000,1224.3357543625)
--(axis cs:60000,1494.3357543625)
--(axis cs:70000,1582.2357543625)
--(axis cs:80000,1682.2357543625)
--(axis cs:90000,1715.8357543625)
--(axis cs:100000,1775.8357543625)
--(axis cs:110000,1801.5357543625)
--(axis cs:120000,2016.5357543625)
--(axis cs:130000,2034.3357543625)
--(axis cs:140000,2127.9357543625)
--(axis cs:150000,1971.5357543625)
--(axis cs:160000,2015.8357543625)
--(axis cs:170000,2326.5357543625)
--(axis cs:180000,2106.5357543625)
--(axis cs:190000,2264.3357543625)
--(axis cs:200000,2322.9357543625)
--(axis cs:210000,2128.6357543625)
--(axis cs:220000,2198.6357543625)
--(axis cs:230000,2611.5357543625)
--(axis cs:240000,2515.1357543625)
--(axis cs:250000,2601.5357543625)
--(axis cs:260000,2506.5357543625)
--(axis cs:270000,2497.9357543625)
--(axis cs:280000,2667.9357543625)
--(axis cs:290000,2520.8357543625)
--(axis cs:300000,2451.5357543625)
--(axis cs:310000,2424.3357543625)
--(axis cs:320000,2554.3357543625)
--(axis cs:330000,2446.5357543625)
--(axis cs:340000,2525.8357543625)
--(axis cs:350000,2522.9357543625)
--(axis cs:360000,2470.1357543625)
--(axis cs:370000,2597.2357543625)
--(axis cs:380000,2582.9357543625)
--(axis cs:390000,2626.5357543625)
--(axis cs:390000,3012.0642456375)
--(axis cs:390000,3012.0642456375)
--(axis cs:380000,2968.4642456375)
--(axis cs:370000,2982.7642456375)
--(axis cs:360000,2855.6642456375)
--(axis cs:350000,2908.4642456375)
--(axis cs:340000,2911.3642456375)
--(axis cs:330000,2832.0642456375)
--(axis cs:320000,2939.8642456375)
--(axis cs:310000,2809.8642456375)
--(axis cs:300000,2837.0642456375)
--(axis cs:290000,2906.3642456375)
--(axis cs:280000,3053.4642456375)
--(axis cs:270000,2883.4642456375)
--(axis cs:260000,2892.0642456375)
--(axis cs:250000,2987.0642456375)
--(axis cs:240000,2900.6642456375)
--(axis cs:230000,2997.0642456375)
--(axis cs:220000,2584.1642456375)
--(axis cs:210000,2514.1642456375)
--(axis cs:200000,2708.4642456375)
--(axis cs:190000,2649.8642456375)
--(axis cs:180000,2492.0642456375)
--(axis cs:170000,2712.0642456375)
--(axis cs:160000,2401.3642456375)
--(axis cs:150000,2357.0642456375)
--(axis cs:140000,2513.4642456375)
--(axis cs:130000,2419.8642456375)
--(axis cs:120000,2402.0642456375)
--(axis cs:110000,2187.0642456375)
--(axis cs:100000,2161.3642456375)
--(axis cs:90000,2101.3642456375)
--(axis cs:80000,2067.7642456375)
--(axis cs:70000,1967.7642456375)
--(axis cs:60000,1879.8642456375)
--(axis cs:50000,1609.8642456375)
--(axis cs:40000,1384.8642456375)
--(axis cs:30000,1449.8642456375)
--(axis cs:20000,1555.6642456375)
--(axis cs:10000,1494.1642456375)
--(axis cs:0,1240.6642456375)
--(axis cs:-10000,408.464245637496)
--cycle;

\path [draw=peru20214597, fill=peru20214597, opacity=0.2]
(axis cs:-10000,373.563177557596)
--(axis cs:-10000,95.0368224424043)
--(axis cs:0,195.036822442404)
--(axis cs:10000,498.636822442404)
--(axis cs:20000,557.136822442404)
--(axis cs:30000,566.436822442404)
--(axis cs:40000,686.436822442404)
--(axis cs:50000,738.636822442404)
--(axis cs:60000,777.136822442404)
--(axis cs:70000,780.036822442404)
--(axis cs:80000,753.636822442404)
--(axis cs:90000,725.736822442404)
--(axis cs:100000,879.336822442404)
--(axis cs:110000,791.436822442404)
--(axis cs:120000,787.136822442404)
--(axis cs:130000,897.136822442404)
--(axis cs:140000,838.636822442404)
--(axis cs:150000,912.136822442404)
--(axis cs:160000,895.036822442404)
--(axis cs:170000,882.136822442404)
--(axis cs:180000,811.436822442404)
--(axis cs:190000,903.636822442404)
--(axis cs:200000,933.636822442404)
--(axis cs:210000,849.336822442404)
--(axis cs:220000,853.636822442404)
--(axis cs:230000,919.336822442404)
--(axis cs:240000,922.836822442404)
--(axis cs:250000,899.336822442404)
--(axis cs:260000,998.636822442404)
--(axis cs:270000,801.436822442404)
--(axis cs:280000,875.036822442404)
--(axis cs:290000,817.836822442404)
--(axis cs:300000,948.636822442404)
--(axis cs:310000,1010.0368224424)
--(axis cs:320000,1045.7368224424)
--(axis cs:330000,1016.4368224424)
--(axis cs:340000,1091.4368224424)
--(axis cs:350000,990.036822442404)
--(axis cs:360000,1144.3368224424)
--(axis cs:370000,1131.4368224424)
--(axis cs:380000,1101.4368224424)
--(axis cs:390000,1083.6368224424)
--(axis cs:390000,1362.1631775576)
--(axis cs:390000,1362.1631775576)
--(axis cs:380000,1379.9631775576)
--(axis cs:370000,1409.9631775576)
--(axis cs:360000,1422.8631775576)
--(axis cs:350000,1268.5631775576)
--(axis cs:340000,1369.9631775576)
--(axis cs:330000,1294.9631775576)
--(axis cs:320000,1324.2631775576)
--(axis cs:310000,1288.5631775576)
--(axis cs:300000,1227.1631775576)
--(axis cs:290000,1096.3631775576)
--(axis cs:280000,1153.5631775576)
--(axis cs:270000,1079.9631775576)
--(axis cs:260000,1277.1631775576)
--(axis cs:250000,1177.8631775576)
--(axis cs:240000,1201.3631775576)
--(axis cs:230000,1197.8631775576)
--(axis cs:220000,1132.1631775576)
--(axis cs:210000,1127.8631775576)
--(axis cs:200000,1212.1631775576)
--(axis cs:190000,1182.1631775576)
--(axis cs:180000,1089.9631775576)
--(axis cs:170000,1160.6631775576)
--(axis cs:160000,1173.5631775576)
--(axis cs:150000,1190.6631775576)
--(axis cs:140000,1117.1631775576)
--(axis cs:130000,1175.6631775576)
--(axis cs:120000,1065.6631775576)
--(axis cs:110000,1069.9631775576)
--(axis cs:100000,1157.8631775576)
--(axis cs:90000,1004.2631775576)
--(axis cs:80000,1032.1631775576)
--(axis cs:70000,1058.5631775576)
--(axis cs:60000,1055.6631775576)
--(axis cs:50000,1017.1631775576)
--(axis cs:40000,964.963177557596)
--(axis cs:30000,844.963177557596)
--(axis cs:20000,835.663177557596)
--(axis cs:10000,777.163177557596)
--(axis cs:0,473.563177557596)
--(axis cs:-10000,373.563177557596)
--cycle;

\addplot [semithick, darkorange2221435, mark=triangle*, mark size=2, mark options={solid}]
table {%
-10000 182.5
0 956
10000 1124
20000 1121.5
30000 1297.5
40000 1385.5
50000 1377.5
60000 1523
70000 1626
80000 1700
90000 1775
100000 1753
110000 1877
120000 1917.5
130000 1886.5
140000 1924.5
150000 2126.5
160000 2043.5
170000 2120
180000 2066
190000 2138.5
200000 2139
210000 2173
220000 2104
230000 2134
240000 2080.5
250000 2146.5
260000 2135
270000 2227.5
280000 2238
290000 2265.5
300000 2129.5
310000 2225.5
320000 2339.5
330000 2315.5
340000 2346.5
350000 2345
360000 2401
370000 2392.5
380000 2307.5
390000 2294.5
};
\addlegendentry{A2C}
\addplot [semithick, orchid204120188, mark=+, mark size=2, mark options={solid}]
table {%
-10000 215.7
0 1047.9
10000 1301.4
20000 1362.9
30000 1257.1
40000 1192.1
50000 1417.1
60000 1687.1
70000 1775
80000 1875
90000 1908.6
100000 1968.6
110000 1994.3
120000 2209.3
130000 2227.1
140000 2320.7
150000 2164.3
160000 2208.6
170000 2519.3
180000 2299.3
190000 2457.1
200000 2515.7
210000 2321.4
220000 2391.4
230000 2804.3
240000 2707.9
250000 2794.3
260000 2699.3
270000 2690.7
280000 2860.7
290000 2713.6
300000 2644.3
310000 2617.1
320000 2747.1
330000 2639.3
340000 2718.6
350000 2715.7
360000 2662.9
370000 2790
380000 2775.7
390000 2819.3
};
\addlegendentry{A2C-AZ}
\addplot [semithick, peru20214597, mark=diamond*, mark size=2, mark options={solid}]
table {%
-10000 234.3
0 334.3
10000 637.9
20000 696.4
30000 705.7
40000 825.7
50000 877.9
60000 916.4
70000 919.3
80000 892.9
90000 865
100000 1018.6
110000 930.7
120000 926.4
130000 1036.4
140000 977.9
150000 1051.4
160000 1034.3
170000 1021.4
180000 950.7
190000 1042.9
200000 1072.9
210000 988.6
220000 992.9
230000 1058.6
240000 1062.1
250000 1038.6
260000 1137.9
270000 940.7
280000 1014.3
290000 957.1
300000 1087.9
310000 1149.3
320000 1185
330000 1155.7
340000 1230.7
350000 1129.3
360000 1283.6
370000 1270.7
380000 1240.7
390000 1222.9
};
\addlegendentry{AZ}

\end{axis}

\end{tikzpicture}

%% file: Image/assault_reduced.tex
\begin{tikzpicture}

\definecolor{chocolate213940}{RGB}{213,94,0}
\definecolor{darkcyan1115178}{RGB}{1,115,178}
\definecolor{darkcyan2158115}{RGB}{2,158,115}
\definecolor{darkorange2221435}{RGB}{222,143,5}
\definecolor{darkslategray38}{RGB}{38,38,38}
\definecolor{lavender234234242}{RGB}{234,234,242}
\definecolor{lightgray204}{RGB}{204,204,204}
\definecolor{orchid204120188}{RGB}{204,120,188}
\definecolor{peru20214597}{RGB}{202,145,97}

\begin{axis}[
axis background/.style={fill=lavender234234242},
axis line style={white},
legend cell align={left},
legend style={
  fill opacity=0.8,
  draw opacity=1,
  text opacity=1,
  at={(0.03,0.97)},
  anchor=north west,
  draw=lightgray204,
  fill=lavender234234242
},
x grid style={white},
xlabel=\textcolor{darkslategray38}{Time Step},
xmajorgrids,
xmajorticks=true,
xmin=-30000, xmax=410000,
xtick style={color=darkslategray38},
xtick={0,50000,100000,150000,200000,250000,300000,350000,400000},
xticklabels={0,50,100,150,200,250,300,350,400},
y grid style={white},
ylabel=\textcolor{darkslategray38}{Performance},
ymajorgrids,
ymajorticks=true,
ymin=-270.599651940714, ymax=2868.88604675176,
ytick style={color=darkslategray38}
]

\path [draw=darkorange2221435, fill=darkorange2221435, opacity=0.2]
(axis cs:-10000,272.776488129916)
--(axis cs:-10000,152.223511870084)
--(axis cs:0,305.723511870084)
--(axis cs:10000,385.723511870084)
--(axis cs:20000,590.923511870084)
--(axis cs:30000,665.023511870084)
--(axis cs:40000,708.523511870084)
--(axis cs:50000,715.923511870084)
--(axis cs:60000,755.623511870084)
--(axis cs:70000,733.723511870084)
--(axis cs:80000,760.223511870084)
--(axis cs:90000,750.523511870084)
--(axis cs:100000,786.823511870084)
--(axis cs:110000,784.723511870084)
--(axis cs:120000,749.523511870084)
--(axis cs:130000,791.323511870084)
--(axis cs:140000,821.923511870084)
--(axis cs:150000,794.023511870084)
--(axis cs:160000,844.823511870084)
--(axis cs:170000,844.623511870084)
--(axis cs:180000,820.423511870084)
--(axis cs:190000,845.923511870084)
--(axis cs:200000,858.223511870084)
--(axis cs:210000,855.723511870084)
--(axis cs:220000,864.123511870084)
--(axis cs:230000,897.123511870084)
--(axis cs:240000,921.023511870084)
--(axis cs:250000,919.823511870084)
--(axis cs:260000,918.323511870084)
--(axis cs:270000,938.923511870084)
--(axis cs:280000,1064.42351187008)
--(axis cs:290000,1076.92351187008)
--(axis cs:300000,1009.42351187008)
--(axis cs:310000,1035.12351187008)
--(axis cs:320000,1119.42351187008)
--(axis cs:330000,1140.22351187008)
--(axis cs:340000,1098.72351187008)
--(axis cs:350000,1069.82351187008)
--(axis cs:360000,1020.22351187008)
--(axis cs:370000,1101.72351187008)
--(axis cs:380000,1037.32351187008)
--(axis cs:390000,1053.32351187008)
--(axis cs:390000,1173.87648812992)
--(axis cs:390000,1173.87648812992)
--(axis cs:380000,1157.87648812992)
--(axis cs:370000,1222.27648812992)
--(axis cs:360000,1140.77648812992)
--(axis cs:350000,1190.37648812992)
--(axis cs:340000,1219.27648812992)
--(axis cs:330000,1260.77648812992)
--(axis cs:320000,1239.97648812992)
--(axis cs:310000,1155.67648812992)
--(axis cs:300000,1129.97648812992)
--(axis cs:290000,1197.47648812992)
--(axis cs:280000,1184.97648812992)
--(axis cs:270000,1059.47648812992)
--(axis cs:260000,1038.87648812992)
--(axis cs:250000,1040.37648812992)
--(axis cs:240000,1041.57648812992)
--(axis cs:230000,1017.67648812992)
--(axis cs:220000,984.676488129916)
--(axis cs:210000,976.276488129916)
--(axis cs:200000,978.776488129916)
--(axis cs:190000,966.476488129916)
--(axis cs:180000,940.976488129916)
--(axis cs:170000,965.176488129916)
--(axis cs:160000,965.376488129916)
--(axis cs:150000,914.576488129916)
--(axis cs:140000,942.476488129916)
--(axis cs:130000,911.876488129916)
--(axis cs:120000,870.076488129916)
--(axis cs:110000,905.276488129916)
--(axis cs:100000,907.376488129916)
--(axis cs:90000,871.076488129916)
--(axis cs:80000,880.776488129916)
--(axis cs:70000,854.276488129916)
--(axis cs:60000,876.176488129916)
--(axis cs:50000,836.476488129916)
--(axis cs:40000,829.076488129916)
--(axis cs:30000,785.576488129916)
--(axis cs:20000,711.476488129916)
--(axis cs:10000,506.276488129916)
--(axis cs:0,426.276488129916)
--(axis cs:-10000,272.776488129916)
--cycle;

\path [draw=orchid204120188, fill=orchid204120188, opacity=0.2]
(axis cs:-10000,443.989181096645)
--(axis cs:-10000,57.6108189033548)
--(axis cs:0,184.510818903355)
--(axis cs:10000,185.410818903355)
--(axis cs:20000,261.310818903355)
--(axis cs:30000,452.410818903355)
--(axis cs:40000,495.010818903355)
--(axis cs:50000,528.910818903355)
--(axis cs:60000,571.610818903355)
--(axis cs:70000,598.810818903355)
--(axis cs:80000,635.510818903355)
--(axis cs:90000,727.610818903355)
--(axis cs:100000,781.310818903355)
--(axis cs:110000,827.110818903355)
--(axis cs:120000,936.010818903355)
--(axis cs:130000,1116.11081890335)
--(axis cs:140000,1181.21081890335)
--(axis cs:150000,1109.11081890335)
--(axis cs:160000,1207.81081890335)
--(axis cs:170000,1115.31081890335)
--(axis cs:180000,1327.01081890335)
--(axis cs:190000,1501.81081890335)
--(axis cs:200000,1456.31081890335)
--(axis cs:210000,1441.51081890335)
--(axis cs:220000,1430.31081890335)
--(axis cs:230000,1703.41081890335)
--(axis cs:240000,1886.81081890335)
--(axis cs:250000,1965.11081890336)
--(axis cs:260000,1998.81081890335)
--(axis cs:270000,2004.81081890335)
--(axis cs:280000,2081.51081890335)
--(axis cs:290000,1822.41081890335)
--(axis cs:300000,1974.81081890335)
--(axis cs:310000,1871.11081890336)
--(axis cs:320000,2064.61081890335)
--(axis cs:330000,1969.21081890335)
--(axis cs:340000,1948.41081890335)
--(axis cs:350000,1945.41081890335)
--(axis cs:360000,2135.61081890335)
--(axis cs:370000,1985.21081890335)
--(axis cs:380000,1941.11081890336)
--(axis cs:390000,2056.51081890335)
--(axis cs:390000,2442.88918109665)
--(axis cs:390000,2442.88918109665)
--(axis cs:380000,2327.48918109665)
--(axis cs:370000,2371.58918109665)
--(axis cs:360000,2521.98918109665)
--(axis cs:350000,2331.78918109665)
--(axis cs:340000,2334.78918109665)
--(axis cs:330000,2355.58918109665)
--(axis cs:320000,2450.98918109665)
--(axis cs:310000,2257.48918109665)
--(axis cs:300000,2361.18918109665)
--(axis cs:290000,2208.78918109665)
--(axis cs:280000,2467.88918109665)
--(axis cs:270000,2391.18918109665)
--(axis cs:260000,2385.18918109665)
--(axis cs:250000,2351.48918109665)
--(axis cs:240000,2273.18918109665)
--(axis cs:230000,2089.78918109665)
--(axis cs:220000,1816.68918109665)
--(axis cs:210000,1827.88918109665)
--(axis cs:200000,1842.68918109665)
--(axis cs:190000,1888.18918109665)
--(axis cs:180000,1713.38918109665)
--(axis cs:170000,1501.68918109665)
--(axis cs:160000,1594.18918109665)
--(axis cs:150000,1495.48918109665)
--(axis cs:140000,1567.58918109665)
--(axis cs:130000,1502.48918109665)
--(axis cs:120000,1322.38918109665)
--(axis cs:110000,1213.48918109665)
--(axis cs:100000,1167.68918109665)
--(axis cs:90000,1113.98918109665)
--(axis cs:80000,1021.88918109665)
--(axis cs:70000,985.189181096645)
--(axis cs:60000,957.989181096645)
--(axis cs:50000,915.289181096645)
--(axis cs:40000,881.389181096645)
--(axis cs:30000,838.789181096645)
--(axis cs:20000,647.689181096645)
--(axis cs:10000,571.789181096645)
--(axis cs:0,570.889181096645)
--(axis cs:-10000,443.989181096645)
--cycle;

\path [draw=peru20214597, fill=peru20214597, opacity=0.2]
(axis cs:-10000,372.395756545601)
--(axis cs:-10000,10.4042434543987)
--(axis cs:0,-127.895756545601)
--(axis cs:10000,311.604243454399)
--(axis cs:20000,350.904243454399)
--(axis cs:30000,380.304243454399)
--(axis cs:40000,407.304243454399)
--(axis cs:50000,428.004243454399)
--(axis cs:60000,438.204243454399)
--(axis cs:70000,475.704243454399)
--(axis cs:80000,531.904243454399)
--(axis cs:90000,602.504243454399)
--(axis cs:100000,618.004243454399)
--(axis cs:110000,695.004243454399)
--(axis cs:120000,797.704243454399)
--(axis cs:130000,828.904243454399)
--(axis cs:140000,849.004243454399)
--(axis cs:150000,846.804243454399)
--(axis cs:160000,901.304243454399)
--(axis cs:170000,971.804243454399)
--(axis cs:180000,1022.9042434544)
--(axis cs:190000,1145.7042434544)
--(axis cs:200000,1116.4042434544)
--(axis cs:210000,1164.5042434544)
--(axis cs:220000,1323.0042434544)
--(axis cs:230000,1534.8042434544)
--(axis cs:240000,1268.6042434544)
--(axis cs:250000,1351.1042434544)
--(axis cs:260000,1398.9042434544)
--(axis cs:270000,1350.4042434544)
--(axis cs:280000,1206.9042434544)
--(axis cs:290000,1453.5042434544)
--(axis cs:300000,1518.5042434544)
--(axis cs:310000,1544.5042434544)
--(axis cs:320000,1458.8042434544)
--(axis cs:330000,1377.0042434544)
--(axis cs:340000,1343.3042434544)
--(axis cs:350000,1477.2042434544)
--(axis cs:360000,1479.8042434544)
--(axis cs:370000,1468.8042434544)
--(axis cs:380000,1523.0042434544)
--(axis cs:390000,1378.7042434544)
--(axis cs:390000,1740.6957565456)
--(axis cs:390000,1740.6957565456)
--(axis cs:380000,1884.9957565456)
--(axis cs:370000,1830.7957565456)
--(axis cs:360000,1841.7957565456)
--(axis cs:350000,1839.1957565456)
--(axis cs:340000,1705.2957565456)
--(axis cs:330000,1738.9957565456)
--(axis cs:320000,1820.7957565456)
--(axis cs:310000,1906.4957565456)
--(axis cs:300000,1880.4957565456)
--(axis cs:290000,1815.4957565456)
--(axis cs:280000,1568.8957565456)
--(axis cs:270000,1712.3957565456)
--(axis cs:260000,1760.8957565456)
--(axis cs:250000,1713.0957565456)
--(axis cs:240000,1630.5957565456)
--(axis cs:230000,1896.7957565456)
--(axis cs:220000,1684.9957565456)
--(axis cs:210000,1526.4957565456)
--(axis cs:200000,1478.3957565456)
--(axis cs:190000,1507.6957565456)
--(axis cs:180000,1384.8957565456)
--(axis cs:170000,1333.7957565456)
--(axis cs:160000,1263.2957565456)
--(axis cs:150000,1208.7957565456)
--(axis cs:140000,1210.9957565456)
--(axis cs:130000,1190.8957565456)
--(axis cs:120000,1159.6957565456)
--(axis cs:110000,1056.9957565456)
--(axis cs:100000,979.995756545601)
--(axis cs:90000,964.495756545601)
--(axis cs:80000,893.895756545601)
--(axis cs:70000,837.695756545601)
--(axis cs:60000,800.195756545601)
--(axis cs:50000,789.995756545601)
--(axis cs:40000,769.295756545601)
--(axis cs:30000,742.295756545601)
--(axis cs:20000,712.895756545601)
--(axis cs:10000,673.595756545601)
--(axis cs:0,234.095756545601)
--(axis cs:-10000,372.395756545601)
--cycle;

\addplot [semithick, darkorange2221435, mark=triangle*, mark size=2, mark options={solid}]
table {%
-10000 212.5
0 366
10000 446
20000 651.2
30000 725.3
40000 768.8
50000 776.2
60000 815.9
70000 794
80000 820.5
90000 810.8
100000 847.1
110000 845
120000 809.8
130000 851.6
140000 882.2
150000 854.3
160000 905.1
170000 904.9
180000 880.7
190000 906.2
200000 918.5
210000 916
220000 924.4
230000 957.4
240000 981.3
250000 980.1
260000 978.6
270000 999.2
280000 1124.7
290000 1137.2
300000 1069.7
310000 1095.4
320000 1179.7
330000 1200.5
340000 1159
350000 1130.1
360000 1080.5
370000 1162
380000 1097.6
390000 1113.6
};
\addlegendentry{A2C}
\addplot [semithick, orchid204120188, mark=+, mark size=2, mark options={solid}]
table {%
-10000 250.8
0 377.7
10000 378.6
20000 454.5
30000 645.6
40000 688.2
50000 722.1
60000 764.8
70000 792
80000 828.7
90000 920.8
100000 974.5
110000 1020.3
120000 1129.2
130000 1309.3
140000 1374.4
150000 1302.3
160000 1401
170000 1308.5
180000 1520.2
190000 1695
200000 1649.5
210000 1634.7
220000 1623.5
230000 1896.6
240000 2080
250000 2158.3
260000 2192
270000 2198
280000 2274.7
290000 2015.6
300000 2168
310000 2064.3
320000 2257.8
330000 2162.4
340000 2141.6
350000 2138.6
360000 2328.8
370000 2178.4
380000 2134.3
390000 2249.7
};
\addlegendentry{A2C-AZ}
\addplot [semithick, peru20214597, mark=diamond*, mark size=2, mark options={solid}]
table {%
-10000 191.4
0 53.1
10000 492.6
20000 531.9
30000 561.3
40000 588.3
50000 609
60000 619.2
70000 656.7
80000 712.9
90000 783.5
100000 799
110000 876
120000 978.7
130000 1009.9
140000 1030
150000 1027.8
160000 1082.3
170000 1152.8
180000 1203.9
190000 1326.7
200000 1297.4
210000 1345.5
220000 1504
230000 1715.8
240000 1449.6
250000 1532.1
260000 1579.9
270000 1531.4
280000 1387.9
290000 1634.5
300000 1699.5
310000 1725.5
320000 1639.8
330000 1558
340000 1524.3
350000 1658.2
360000 1660.8
370000 1649.8
380000 1704
390000 1559.7
};
\addlegendentry{AZ}

\end{axis}

\end{tikzpicture}

%% file: Appendix.tex
\newpage
\appendix


\section{Implementation Details}


\subsection{Discrete representation for the critic}
The critic’s loss function, denoted as $\lossCriticSub{\worldstate}$, is formulated to minimize the disparity between the value target $\valuetargetPovCond{\neuralnetwork}{\worldstate}$ and the predicted value $\valuefunctionPovCond{\neuralnetwork}{\worldstate}$ at a specific state $\worldstate$.
Commonly, the disparity is computed with the mean squared error or the cross-entropy over a discrete representation.

Previous studies have emphasized the benefits of employing cross-entropy over a discrete representation in reinforcement learning \cite{bellemare2017distributional, schrittwieser2020mastering, hafner2023mastering, bellemare2023distributional, farebrother2024stop}.
This method involves the critic to learn a discrete weight distribution $\weightdiscrete{\neuralnetwork} = \{\weightdiscrete{1}, . . . , \weightdiscrete{\bucketSize} \} \in \mathbb{R}^\bucketSize$ instead of learning the mean of the distribution. A function $y()$ that converts a target value into a corresponding weight distribution of size $\bucketSize$.
This leads to the following sub-loss for the critic:

\begin{align}
    \lossCriticSub{\worldstate} = y(\valuetargetPovCond{\neuralnetwork}{\worldstate})^T \log \weightdiscrete{\neuralnetwork}
\end{align}

More specifically, transforming (function y) the reward/target into a discrete representation is done function by a method called two-hot encoding.
The two-hot encoding is a generalization of the one-hot encoding where all elements are 0 except for the two entries closest to y at positions m and m + 1. 
These two entries sum up to 1, with more weight given to the entry that is closer to y:

\begin{align*}
    y(x)=twohot(x)_i = 
    \begin{cases}
        |b_{m+1} - x| / |b_{m+1} - bm| & \text{ if }i = m \\
        |b_m - x| / |b_{m+1} - b_m| &\text{ if } i = m + 1 \\
        0 & \text{ else }
    \end{cases}
\end{align*}
Importantly, two-hot encoding can predict any continuous value in the interval because its expected bucket value can fall between the buckets.

\section{Monte Carlo Tree Search-Detailed}

Below, we provide a comprehensive overview of the Monte Carlo Tree Search (MCTS) algorithm, drawing from previous research on MCTS \cite{silver_general_2018, schrittwieser2020mastering, ye2021mastering}. 
Notably, in \citet{schrittwieser2020mastering}, it was demonstrated that a budget of $50$ is adequate for resolving the Atari100K benchmark, hence informing our decision in this regard.

Monte Carlo Tree Search (MCTS)~\cite{browne_survey_2012} is the state-of-the-art in the perfect information game. MCTS converges asymptotically to the optimal policy in single-agent domains and to the minimax value function in zero-sum games.
Starting from the AlphaGo~\cite{Silver2016MasteringTG,silver_general_2018,silver_mastering_2017}, MCTS has been combined with an offline neural network to enhance performance.

$MCTS(\worldstate,\budget)$ is an online tree search algorithm that runs at $\worldstate$ for a budget of $\budget$ and works as follows (\romannumeral1) \textbf{selection} --- selects a path of nodes until a leaf node; (\romannumeral2) \textbf{expansion} --- expands the tree by adding a new child node; (\romannumeral3) \textbf{backpropagation} --- backpropagates the result obtained through the nodes chosen during the selection phase; (\romannumeral4) repeats step (\romannumeral1) to (\romannumeral3) until the budget $\budget$ is finished; (\romannumeral5) returns the distribution of actions $\policyTimePov{}{MCTS}$ that has been visited, and the value $\valuefunctionPov{MCTS}$ obtained when running MCTS.
 
\vspace{5pt}
Every node of the search tree is associated with a state $\worldstate$ and each node stores the statistics estimating the value of the node $\valuefunctionPovCond{}{\worldstate}$. 
For every action $\action$ at $\worldstate$, there exists an associated edge represented as $(\worldstate, \action)$. 
These edges stores a set of statistics $\{$ $\CountTuple{\timestep}{\worldstate}{\action}$, $\qvalueTimeCond{t}{\worldstate}{\action}$, $\policyTimePovCond{}{}{\action}{\worldstate}$, $\rewardPovCond{}{}{\worldstate}{\action}$, $\transitionCond{\cdot}{\worldstate}{\action}$ $\}$, respectively representing visit counts $\Count{}$, mean $\qvalue$-Value observed, policy $\policyTimePov{}{}$, reward $\reward$, and state transition $\transition$.

\subsection{Selection}

In the selection, a simulation begins at the internal root state, denoted as $\worldstateTime{0}$, and progresses until it reaches a leaf node represented as $\worldstateTime{l}$. 
Throughout this selection, actions $\actionTime{k}$ for $k = {1, \cdots,l}$ are selected using the PUCT formula. This formula strikes a balance between exploration and exploitation, guiding the choice of actions during the simulation.

\begin{align*}
    \qvalueTimeCond{}{\worldstate}{\action} + 
    \policyTimePovCond{}{}{\action}{\worldstate}\frac{\sqrt{ \CountOne{}{\worldstate}}}{1+\CountTuple{}{\worldstate}{\action}} \left(\uctconstante + log (\frac{\CountOne{}{\worldstate} + \puctconstante + 1}{\puctconstante})\right)
\end{align*}
The best action is the one that maximizes where $\CountOne{}{\worldstate}$ represents the number of times that the state $\worldstate$ has been visited, $\uctconstante$ and $\puctconstante$ are variables that help to control the exploration. 

\subsection{Expansion}

Following the selection phase, at step $l$ of the simulation, the environment dynamics function $f(\worldstateTime{l-1},\actionTime{l})$ computes the reward $\rewardPovCond{}{}{\worldstateTime{l-1}}{\actionTime{l}}$ and the ensuing state $\worldstateTime{l}$. 
The policy network generates $\policyTimePovCond{}{\neuralnetwork}{\cdot}{\worldstateTime{l}}$ and the value network estimates $\valuefunctionPovCond{\neuralnetwork}{\worldstateTime{l}}$. 
For every action $\action$ within $\worldstateTime{l}$, a corresponding edge $(\worldstateTime{l},\action)$ is established. 
This edge is initialized with values: $\CountTuple{}{\worldstateTime{l}}{\action} = 0$, $\qvalueTimeCond{}{\worldstateTime{l}}{\action} = 0$, and $\policyTimePovCond{}{}{\action}{\worldstateTime{l}} = \policyTimePovCond{}{\neuralnetwork}{\action}{\worldstateTime{l}}$.

Initially, the default $\qvalue$ for unvisited nodes is set to $0$, indicating the worst possible state. 
To enhance the $\qvalue$ estimation for unvisited nodes, a mean $\qvalue$ mechanism is implemented during each simulation for tree nodes. 
This approach, similar to the one employed in Elf OpenGo~\cite{tian2019elf,ye2021mastering}, provides a more accurate estimation of $\qvalue$ for unvisited nodes. 
This evaluation is performed iteratively for $k = {0, \cdots,l}$.

\begin{align*}
    \qvaluemechanism{\worldstateTime{0}} &= 0 \\
    \qvaluemechanism{\worldstate} &= \frac{\qvaluemechanism{\worldstateTime{parent}}+\sum_b \mathbbm{1}_{\CountTuple{}{\worldstate}{\action}>0} \qvalueTimeCond{}{\worldstate}{b}}{1+\mathbbm{1}_{\CountTuple{}{\worldstate}{\action}>0}}  \\
    \qvalueTimeCond{}{\worldstate}{\action} &= \begin{cases}
        \qvalueTimeCond{}{\worldstate}{\action}& \CountTuple{}{\worldstate}{\action} >0 \\
        \qvaluemechanism{\worldstate}{}& \CountTuple{}{\worldstate}{\action} =0
    \end{cases}
\end{align*}
where $\qvaluemechanism{\worldstate}$ is the estimated $\qvalue$ value for unvisited nodes. 

In addition, when the root node is expanded, a Dirichlet noise to the policy prior is added. This technique is used for improving the exploration. $\mathcal{N}_D (\epsilon)$ is the Dirichlet noise distribution, $\rho$, $\epsilon$ is set to 0.25 and 0.3 respectively.

\begin{align*}
    \policyTimePovCond{}{}{\action}{\worldstate} = (1 - \dirichetnoiseproba )
    \policyTimePovCond{}{}{\action}{\worldstate} + \dirichetnoisedistribution
\end{align*}

\subsection{Backup}

For every step from $k = \{l-1, \dots,0\}$, we update the statistics associated with each edge ($\worldstateTime{k}$, $\actionTime{k}$) in the simulation path by using the boostrapped $\discountlambda$-returns.
\begin{align*}
    \qvalueTimeCond{}{\worldstateTime{k-1}}{\actionTime{k}} &= \frac{\CountTuple{}{\worldstateTime{k-1}}{\actionTime{k}} \qvalueTimeCond{}{\worldstateTime{k-1}}{\actionTime{k}} + \valuefunctionlambdaTimePovCond{l-k}{}{\worldstateTime{k}}}{\CountTuple{}{\worldstateTime{k-1}}{\actionTime{k}} +1} \\
    \CountTuple{}{\worldstateTime{k-1}}{\actionTime{k}} &= \CountTuple{}{\worldstateTime{k-1}}{\actionTime{k}} +1
\end{align*}

Ensuring the $\qvalue$-Value falls within the range of [0,1] is crucial for employing the PUCT formula, to achieve this, we normalize the $\qvalue$ value by referencing the minimum and maximum values observed in the search tree up to that particular point.

\begin{align*}
    \bar{Q}(\worldstateTime{k}, \actionTime{k} ) = \frac{\qvalueTimeCond{}{\worldstateTime{k}}{\actionTime{k}} - \minn{\worldstate,\action \in Tree} \qvalueTimeCond{}{\worldstate}{\action}}{\maxx{}(\maxx{\worldstate,\action \in Tree} \qvalueTimeCond{}{\worldstate}{\action} - \minn{\worldstate,\action \in Tree} \qvalueTimeCond{}{\worldstate}{\action},\epsilon)}
\end{align*}
, where $\epsilon$, the threshold to give a smooth range of the min-max bound.

\subsection{Additional Information}

After the budget $\budget$ is finished, we obtain the average value $\valuefunctionPov{AZ}$ and visit count distributions of the root node. To obtain the policy distribution $\policyTimePov{}{AZ}$, we use a temperature parameter of the visit count; \ie{}

\begin{align*}
    \policyTimePov{}{AZ}(\worldstate,\action) = \frac{\CountTuple{}{\worldstate}{\action}^{\frac{1}{\temperature}}}{\sum_b \CountTuple{}{\worldstate}{b}^{\frac{1}{\temperature}}}
\end{align*}
During the training process, we decay the temperature twice, at $50\%$ of training progress to $0.5$, and at the $75\%$ of training progress to $0.25$.

\section{Model-Based}

To reduce the cost of training, our experiments were carried out using a world model trained.
This means that the game (dynamics, reward, graphical representation, etc.) is approximated by a representation of the world and to obtain one of the game's pieces of information, a call is made to this representation. 
Using a world model that has already been trained allows us to reduce calculation time. 
Thanks to this, we already have an abstraction of the worlds and the dynamics, allowing us to concentrate our learning on the critic and the actor-network.

In our case, we used the world model of the Dreamer V3 algorithm~\cite{hafner2023mastering}. 
Dreamer is a state-of-the-art algorithm in a model-based setting and constitutes the first agent that achieves human-level performance on the Atari benchmark tasks by learning behaviors inside a separately trained world model.
Dreamer takes a world state $\worldstateTime{\timestep}$ as an input and returns an abstracted world state $\hat{\worldstate}_\timestep \sim \worldstateTime{\timestep}$, an expected reward $\hat{\reward}_\timestep \sim \rewardTimePov{\timestep}{}$ and a continue flag $\continue_\timestep$, which indicate if the game if finished or not. 
The world model uses a Recurrent State-Space Model (RSSM) which predicts future information ($\hat{\worldstate}_{t+1}$, $\hat{\reward}_{t+1}$  and $\continue_{t+1}$) when given $\actionTime{\timestep}$ to the current abstract world $\hat{\worldstate}_\timestep$. 
The world model is used to compute the $\horizonbootstrap$-step bootstrapped $\lambda$-returns for A2C algorithms and facilitating MCTS in A2C-AZ and AlphaZero. 

In our implementation, we use the weight of an already trained Dreamer algorithm during $50,000$k training steps.
To be precise, the world model is never modified (fixed), so our work is not related to MuZero/EfficientZero/Dreamer algorithms, which train the world model and at the same time learn the policy/critic, but much closer to algorithms such as AlphaZero/A2C that use a fixed world and learn the policy/critic.

\section{Neural Network}

The neural networks of the world model are the same as used in the paper in DreamerV3~\cite{hafner2023mastering} in which we used the Model size S. 
The encoder begins with stride $2$ convolutions neural networks (CNN)~\cite{lecun1989backpropagation} that progressively increase the depth of the representation until reaching a resolution of $4 \times 4$. 
Subsequently, the data is flattened for further processing. 
Conversely, the decoder initiates with a dense layer, which is followed by reshaping the data into a $4 \times 4 \times 32$ format. 
It then effectively reverses the architecture employed in the encoder to reconstruct the original data. 
The dynamics component is realized through a Recurrent State Space Model (RSSM)~\cite{hafner2019learning} utilizing vectors of categorical representations. 
This implementation comprises a Gated Recurrent Unit (GRU)~\cite{chung2014empirical} combined with MLP layers. 
The reward, critic, actor and continue predictors are also MLPs. 
Each MLP network is composed of $2$ linear network of $512$ hidden units. 

As clarified in the preceding section, our approach does not involve training the world model as in Dreamer. 
Instead, we utilize a pre-trained and fixed world model. Consequently, the only neural networks trained in our setup are the critic and actor networks which each are a MLP with $2$ linear layers of $512$ hidden units.

\section{Notation and Hyper-parameters}
All notations and hyperparameters utilized throughout the paper are summarized in Table~\ref{tab:hyper_param}. 
The hyperparameters in the World-Model/Actor-Critic section are derived from DreamerV3~\cite{hafner2023mastering}, while those in the Search section are adapted from EfficientZero~\cite{ye2021mastering}.
\begin{table*}
\small
\bigskip
\caption{Hyper parameters used.}
\bigskip
\centering
\begin{tabular}{lcc} 
   \toprule
   \textbf{Name} & \textbf{Notation} & \textbf{Value} \\
   
    \midrule

    \midrule
    \textbf{General} \\
    \midrule

        Training Step &  \_  & $100,000$\\

   \midrule
    \textbf{World Model} \\
    \midrule

        Batch size &  \_ & $16$ \\ 
        Batch length & \_ & $16$ \\
        Imagination horizon & \_& $15$\\
        Number of latents & \_ & $32$\\
        Classes per latent & \_ & $32$ \\
        Model Size & \_ & S\\

    \midrule
    \textbf{Actor Critic } \\
    \midrule
        
        Discount & $\discountlambda$ & $0.95$\\
        Critic EMA decay & \_ & $0.98$ \\
        Critic EMA regularizer & \_ & $1.0$ \\
        Normalization Ter & $\normalizationTerm$ & $Per(R,95) - Per(R,5)$ \\
        Learning rate & \_ & $3 \cdot 10^{-5}$\\
        Adam epsilon & $\_$ & $10^{-5}$\\
        Gradient clipping & \_ & 100 \\
        Guide Actor penalty & $\weightactor{}$ & $[ 0.1, 0.3 , 0.5, 0.7],$ \\
        Max Guide Penalty & $\maxlambda{}$ & $10.0$ \\
        Tau & $\expoTerm{}$ & 5.0 \\
        Guide Critic penalty & $\weightcritic{}$ & $0.05$ \\
        Distance function Actor & $\distanceActor{}{}{}$ & KL-divergence \\
        Distance function Critic & $\distanceCritic{}{}{}$ & Cross-Entropy \\
        Loss function Actor & $\lossActorSub{}$ & Reinforce/Cross-Entropy for RL/MCTS \\
        Loss function Critic & $\lossCriticSub{}$ & Cross-Entropy \\

    \midrule
    \textbf{Search} \\
    \midrule

        Exploration constant & $\uctconstante$ & $1.25$ \\
        Exploration constant 2 & $\puctconstante$ & $19652$ \\
        Search budget & $\budget$ & $50$ \\
        Dirichlet Noise & $ \dirichetnoise$ & $0.3$ \\
        Dirichlet Noise Proba & $ \dirichetnoiseproba$ & $0.25$\\
        Temperature & $\temperature$ & $[1.0,0.5,0.25]$\\
            
    \bottomrule
\end{tabular} 
~\label{tab:hyper_param}
\end{table*}

\section{Additional Experiment}

\begin{table*}[!htbp]
\small
\bigskip
\caption{Performance obtained on Atari100K benchmarks.}
\bigskip
\centering
\begin{tabular}{l|cc| *{5}c | c} 
   \toprule
   Game & Random & Human & $\text{A2C}$ & $\text{AZ}$ & $\text{A2C-Rand}$ & $\text{A2C-\_BC}$ & $\text{A2C-AZ}$ & $\text{A2C-AZ}^*$ \\
   \midrule

   Assault & 222.4 & 742.0 & 1113.6 & 1559.7 & 2138.62 & 1766.1 & 2249.6 & \textbf{2249.6}\\
   Asterix & 210.0 & 8503.3 & 2294.5& 1222.8 & 2225.0& 1924.0& \textbf{2819.2} & \textbf{2819.2}\\
   Bank Heist & 14.2 & 753.1 & 1328.0 & 595.1 & 1285.0 & 1132.6 & 1324.4 & \textbf{1346.0}\\
   BattleZone & 2360.0 & 37187.5 & 14300.0 &16985.7 & 19660.0 & \textbf{22660.0} & 15642.8 & 15642.8\\
   Boxing & 0.1 & 12.1 & \textbf{99.6} & 63.4 & \textbf{99.6} & 99.2& 98.6 & 99.2\\
   Breakout & 1.7 & 30.5 & 5.8 & 94.4 & 6.16&  8.7 & \textbf{162.4} & \textbf{162.4} \\
   Crazy Climber & 10780.5 & 35829.4 & 52137.0 & 62392.8 & 38264.0 & 26816.0 & 43157.1 & \textbf{75465.7} \\
   Demon Attack & 152.1 & 1971.0 & 5168.4 & 3197.7 & \textbf{11517.5} & 5822.4 & 6967.2 & 75465.7\\
   Freeway & 0.0 & 29.6 & 33.3 & 28.7 & \textbf{33.6} &  \textbf{33.6} & 33.1& 33.1\\
   Frostbite & 65.2 & 4334.7 & 2420.7 & 89.4 & 920.2&  2225.2 &\textbf{3588.1} & \textbf{3588.1} \\
   Gopher & 257.6 & 2412.5 & 289.6 & 242.8 & 356.4 & 287.6 & 398.5 & \textbf{434.5}\\
   Hero & 1027.0 & 30826.4 & 12598.4 & 5669.0 & 11093.9 & \textbf{14549.4} & 11531.3 & 12343.7\\
   Jamesbond & 29.0 & 302.8 & 671.0 & 557.8 & 816.0 & 947.0 & \textbf{981.4 }& \textbf{981.4 }\\
   Kangaroo & 52.0 & 3035.0 & 4084.0& \textbf{7288.5} & 792.0& 4084.0 & 6117.1 & 6117.1\\
   Krull & 1598.0 & 2665.5 & 10150.2 & 5593.5 & 9906.6 &  7699.4 & 9366.8 &\textbf{10257.7} \\
   Kung Fu Master & 258.5 & 22736.3 & 35207.0 & 20454.2 & 34592.0 & 20590.0& 31430.0 & \textbf{37002.8} \\
   Ms Pacman & 307.3 & 6951.6 & 1656.5 & 784.4 & \textbf{2904.0} & 2208.6 & 1985.8 & 2402.5\\
   Pong & -20.7 & 14.6 & 20.1 &19.8 & 20.6 & 18.5 & \textbf{20.4} & \textbf{20.9}\\
   Qbert & 163.9 & 13455.0 & 3780.2 & 4610.0 & 6569.5 & 5056.5 & 6268.9 & \textbf{6844.6}\\
   Road Runner & 11.5 & 7845.0 & 7827.0& 9341.4 & 9836.0&  8016.0 & 8521.4 & \textbf{12178.5}\\
   Up N Down & 533.4 & 11693.2 & 8622.3 &\textbf{18857.8} & 3922.8 & 4156.6 & 4688.7 &7570.2\\

   \midrule

    Mean ($\nearrow$)& 0.0& 1.0& 1.7 & 1.46 &1.89 & 1.62 &2.14 & 2.31 \\
    Median ($\nearrow$)& 0.0 & 1.0 & 1.63 & 1.48 & 1.87& 1.45& 2.18 & 2.23\\
    IQM ($\nearrow$)& 0.0& 1.0& 0.92& 0.81 & 0.93 & 0.82& 1.29& 1.47\\
    Optimality Gap ($\searrow$)& 1.0 & 0.0 & 0.36& 0.40& 0.36 & 0.35& 0.28 & 0.25\\

   \bottomrule

\end{tabular} 
\label{tab:all_data_mb}
\end{table*}

\begin{table*}
    \small
    \bigskip
    \caption{Performance obtained on Atari100K benchmarks. We denote A2C-AZ-X where x refers to the weight of the guide's penalty.}
    \bigskip

    \centering
    \begin{tabular}{l|cc| *{4}c } 
       \toprule
       Game & Random & Human & $\text{A2C-AZ\_0.1}$ & $\text{A2C-AZ\_0.3}$ & $\text{A2C-AZ\_0.5}$ & $\text{A2C-AZ\_0.7}$ \\
       \midrule
    
       Assault & 222.4 & 742.0 & 1753.5 & 1786.4 & 2154.8 & \textbf{2249.6}\\
       Asterix & 210.0 & 8503.3 & 2319.2& 2275.0 & 2500.7 & \textbf{2819.2}\\
       Bank Heist & 14.2 & 753.1 & 1342.2 & \textbf{1346.0} &  1303.7 & 1324.4\\
       BattleZone & 2360.0 & 37187.5 & 7457.1 & 10885.7 & 10328.5 & \textbf{15642.8} \\
       Boxing & 0.1 & 12.1 & 98.2 & \textbf{99.2} & 97.6 & 98.6\\
       Breakout & 1.7 & 30.5 & 6.9 & 86.3& 84.3 &\textbf{162.4} \\
       Crazy Climber & 10780.5 & 35829.4 & \textbf{75465.7} & 29448.5& 60411.4 & 43157.1 \\
       Demon Attack & 152.1 & 1971.0 & 5578.0 & \textbf{7361.7}& 6947.2 &  6967.2 \\
       Freeway & 0.0 & 29.6 & \textbf{33.1} & 33.0 & \textbf{33.1} &  33.1  \\
       Frostbite & 65.2 & 4334.7 & 2364.4 & 1639.8 &3113.1 &  \textbf{3588.1}  \\
       Gopher & 257.6 & 2412.5 &371.7& 336.0 & \textbf{434.5} & 398.5 \\
       Hero & 1027.0 & 30826.4 & 9145.7 & 10719.2& \textbf{12343.7}& 11531.3 \\
       Jamesbond & 29.0 & 302.8 & 546.4 & 855.7 & 793.5 & \textbf{981.4} \\
       Kangaroo & 52.0 & 3035.0 & 3248.0 & 3611.4 & 3945.7 & \textbf{6117.1} \\
       Krull & 1598.0 & 2665.5 & 9711.8 & \textbf{10257.7} &  8366.7&  9366.8 \\
       Kung Fu Master & 258.5 & 22736.3 & 25264.0 & 30807.1 & \textbf{37002.8} & 31430.0 \\
       Ms Pacman & 307.3 & 6951.6 & \textbf{2402.5} & 2261.4 &2143.7 & 1985.8 \\
       Pong & -20.7 & 14.6 & 20.3 & \textbf{20.9} & 20.3 & 20.4 \\
       Qbert & 163.9 & 13455.0 & 5411.7 &6021.7 &\textbf{ 6844.6 }& 6268.9 \\
       Road Runner & 11.5 & 7845.0 & 10070.0& 11460.0 & \textbf{12178.5} &  8521.4 \\
       Up N Down & 533.4 & 11693.2 & \textbf{7205.0}& 4708.7& 7178.7 & 4688.7 \\
    
       \midrule

       Mean ($\nearrow$)& 0.0&1.0 & 1.78 &1.95 & 1.95& 2.14\\
       Median ($\nearrow$)& 0.0& 1.0& 1.74 & 1.91 & 1.90 & 2.18 \\
       IQM ($\nearrow$)& 0.0 & 1.0& 0.93& 1.08&1.21 & 1.29 \\
       Optimality Gap ($\searrow$)&1.0 &0.0 & 0.36 &0.31 & 0.28 & 0.28\\

       \bottomrule

    \end{tabular} 
    \label{tab:all_data_mb2}
    \end{table*}

    \begin{table*}
        \small
        \bigskip
        \caption{ Performance obtained on Atari100K benchmarks. We denote A2C-AZ-X where X indicates how often the guide is called. }
        \bigskip
        \centering
        \begin{tabular}{l|cc| *{3}c} 
           \toprule
           Game & Random & Human & $\text{A2C-AZ\_1}$ & $\text{A2C-AZ\_2}$ & $\text{A2C-AZ\_3}$ \\
           \midrule
        
           Assault & 222.4 & 742.0 & \textbf{2249.6} & 1737.0 & 1819.3\\
           Asterix & 210.0 & 8503.3 & \textbf{2819.3} &2431.4&2587.9 \\
           Bank Heist & 14.2 & 753.1 &\textbf{1324.4} &1271.4 & 1264.7\\
           BattleZone & 2360.0 & 37187.5 & \textbf{15642.9} & 13585.7&12814.3\\
           Boxing & 0.1 & 12.1 & \textbf{99.2} &\textbf{99.2}& 98.8\\
           Breakout & 1.7 & 30.5 & \textbf{162.4} &74.3&47.3\\
           Crazy Climber & 10780.5 & 35829.4 & \textbf{43157.1}& 40022.9&32127.1\\
           Demon Attack & 152.1 & 1971.0 & 6967.2& \textbf{7476.4}& 6697.2\\
           Freeway & 0.0 & 29.6 & 33.0& 33.2& \textbf{33.1}\\
           Frostbite & 65.2 & 4334.7 & \textbf{3588.1}& 2722.0&3221.3\\
           Gopher & 257.6 & 2412.5 & \textbf{398.5} & 398.3&351.1\\
           Hero & 1027.0 & 30826.4 &  11531.3 &9462.1&\textbf{11860.3}\\
           Jamesbond & 29.0 & 302.8 & \textbf{981.4} &653.2&720.7\\
           Kangaroo & 52.0 & 3035.0 &  \textbf{6117.1} &2787.1&4262.9\\
           Krull & 1598.0 & 2665.5 & 9366.8 &8392.0&\textbf{9847.1}\\
           Kung Fu Master & 258.5 & 31430.0 & 26915.7 &31235.4&\textbf{32202.9}\\
           Ms Pacman & 307.3 & 6951.6 &1985.8 &\textbf{2312.6}&2176.1\\
           Pong & -20.7 & 14.6 & 20.4 &\textbf{21.0}&20.0\\
           Qbert & 163.9 & 13455.0 & \textbf{6268.9} &5517.5&6202.1\\
           Road Runner & 11.5 & 7845.0 & 8521.4 &\textbf{10144.3}&9850.0\\
           Up N Down & 533.4 & 11693.2  & 4688.7 &4895.3&\textbf{6031.6}\\

           \midrule

           Mean ($\nearrow$)& 0.0 &1.0 & 2.14 & 1.80& 1.83 \\
           Median ($\nearrow$)& 0.0& 1.0& 2.18 & 1.74 & 1.68 \\
           IQM ($\nearrow$)& 0.0&1.0 & 1.29 & 1.03&  0.99\\
           Optimality Gap ($\searrow$)&1.0 &0.0 & 0.28 & 0.32 &0.34\\

           \bottomrule

        \end{tabular} 
        \label{tab:all_data_mb3}
        \end{table*}

\begin{figure*}
\centering  
    \subcaptionbox{ Assault}[0.3\textwidth]{\scalebox{0.6}{\input{Image/assault}}}
 	\subcaptionbox{ Asterix}[0.3\textwidth]{\scalebox{0.6}{\input{Image/asterix}}}
	\subcaptionbox{ Bank Heist}[0.3\textwidth]{\scalebox{0.6}{\input{Image/bank_heist}}}

    \bigskip

	\subcaptionbox{ Battle Zone}[0.3\textwidth]{\scalebox{0.6}{\input{Image/battle_zone}}}
    \subcaptionbox{ Boxing}[0.3\textwidth]{\scalebox{0.6}{\input{Image/boxing}}}
   	\subcaptionbox{ Breakout}[0.3\textwidth]{\scalebox{0.6}{\input{Image/breakout}}}
    \bigskip

    \subcaptionbox{ Crazy Climber}[0.3\textwidth]{\scalebox{0.6}{\input{Image/crazy_climber}}}
	\subcaptionbox{ Demon Attack}[0.3\textwidth]{\scalebox{0.6}{\input{Image/demon_attack}}}
 	\subcaptionbox{ Freeway}[0.3\textwidth]{\scalebox{0.6}{\input{Image/freeway}}}
 	\bigskip

    \subcaptionbox{ Frostbite}[0.3\textwidth]{\scalebox{0.6}{\input{Image/frostbite}}}
    \subcaptionbox{ Gopher}[0.3\textwidth]{\scalebox{0.6}{\input{Image/gopher}}}
 	\subcaptionbox{ Hero}[0.3\textwidth]{\scalebox{0.6}{\input{Image/hero}}}
\end{figure*}
\begin{figure*}\ContinuedFloat
\centering
	\subcaptionbox{ James Bond}[0.3\textwidth]{\scalebox{0.6}{\input{Image/james_bond}}}
 	\subcaptionbox{ Kangaroo}[0.3\textwidth]{\scalebox{0.6}{\input{Image/kangaroo}}}
 	\subcaptionbox{ Krull}[0.3\textwidth]{\scalebox{0.6}{\input{Image/krull}}}
     \bigskip

    \subcaptionbox{ Kung Fu Master}[0.3\textwidth]{\scalebox{0.6}{\input{Image/kung_fu_master}}}
 	\subcaptionbox{ Ms Pacman}[0.3\textwidth]{\scalebox{0.6}{\input{Image/ms_pacman}}}
	\subcaptionbox{ Pong}[0.3\textwidth]{\scalebox{0.6}{\input{Image/pong}}}
    \bigskip

 	\subcaptionbox{ Qbert}[0.3\textwidth]{\scalebox{0.6}{\input{Image/qbert}}}
    \subcaptionbox{ Road Runner}[0.3\textwidth]{\scalebox{0.6}{\input{Image/road_runner}}}
	\subcaptionbox{ Up N Down}[0.3\textwidth]{\scalebox{0.6}{\input{Image/up_n_down}}}
    \bigskip

    \caption{\centering Learning curves on the $26$ game of Atari100k benchmarks with $5$ algorithms presented. The shaded area shows $95\%$ confidence interval (CI) over $5$ seeds.}
    \bigskip

    \label{fig:full_curves}
\end{figure*}
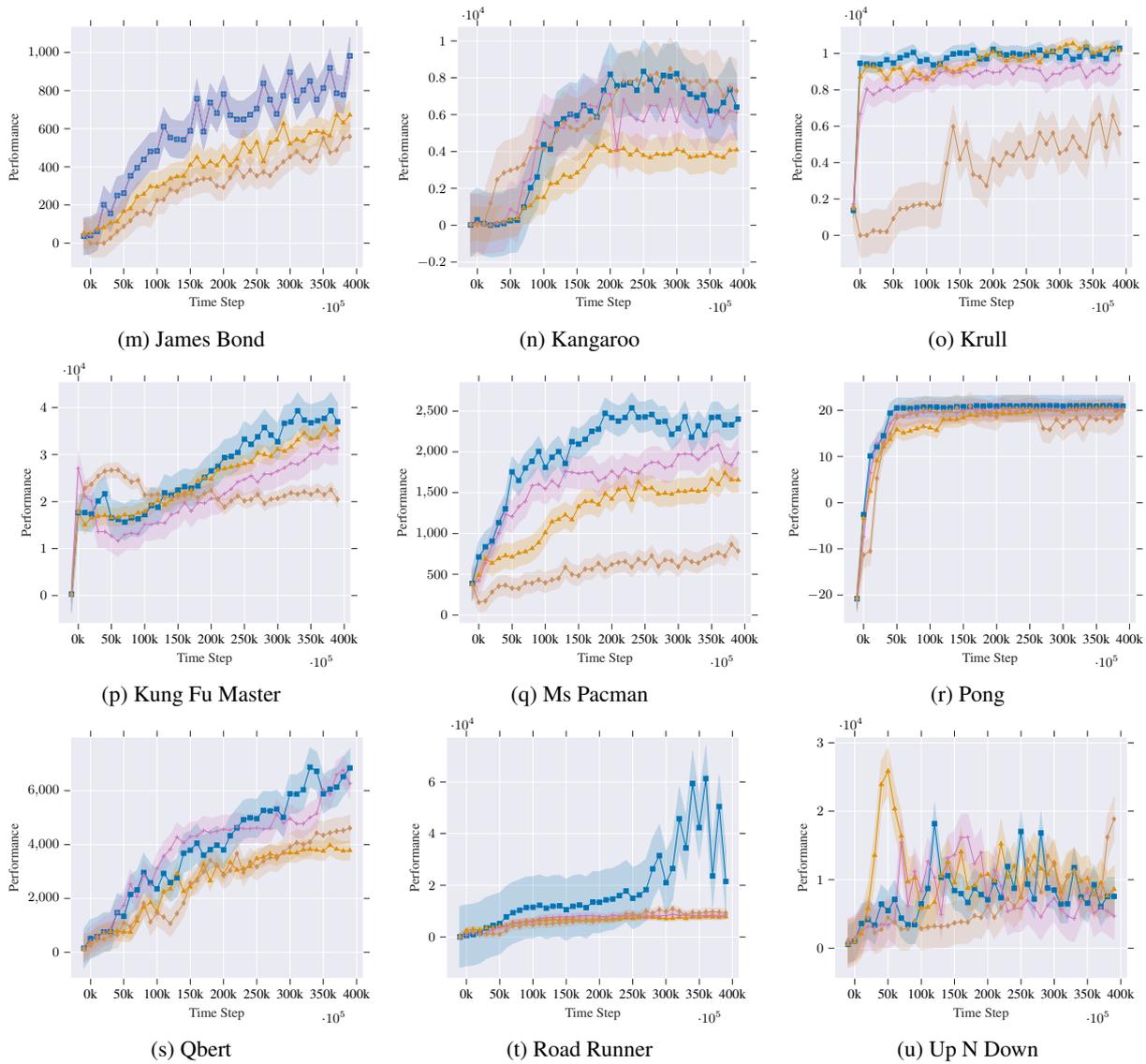

\begin{figure*}[!htbp]
\centering
    \includegraphics[scale=0.5]{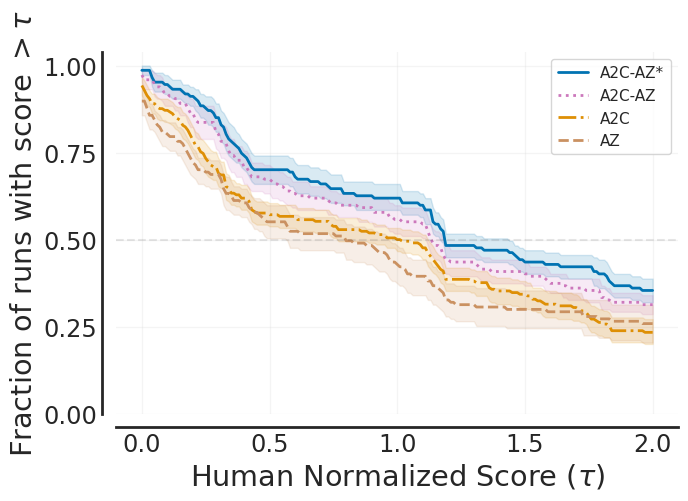}
    \bigskip
    \caption{\centering Human Normalised Score distribution on Atari100k Benchmarks. The shaded area shows $95\%$ stratified bootstrap confidence interval (CI) over $5$ seeds, following methodology. A higher score is better. } 
    \bigskip

~\label{fig:scoreDistribution}
\end{figure*}

\begin{figure*}
    \centering
        \subcaptionbox{X is A2C-AZ*, Y is A2C agent}[0.3\textwidth]{\includegraphics[width=\linewidth]{Image/RL_MCTS_Opti_IncreaseVSRL.png}}
        \hfill
        \subcaptionbox{X is A2C-AZ, Y is A2C agent}[0.3\textwidth]{\includegraphics[width=\linewidth]{Image/RL_MCTS_IncreaseVSRL.png}}
        \hfill
        \subcaptionbox{X is AlphaZero, Y is A2C agent}[0.3\textwidth]{\includegraphics[width=\linewidth]{Image/MCTS_IncreaseVSRL.png}}
        \bigskip

        \subcaptionbox{X is A2C-AZ*, Y is AlphaZero}[0.3\textwidth]{\includegraphics[width=\linewidth]{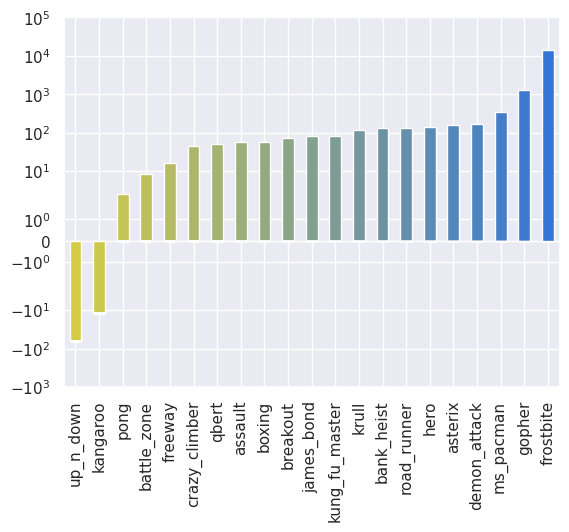}}
        \subcaptionbox{X is A2C-AZ, Y is AlphaZero}[0.3\textwidth]{\includegraphics[width=\linewidth]{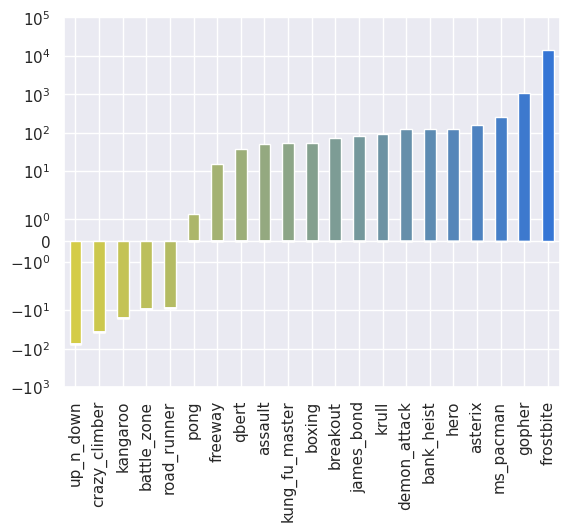}}
        \bigskip

        \caption{Percentage improvement of algorithm X compared to algorithm Y on Atari100k Benchmarks. Improvement is measured as a percentage of mean human-normalized return. }
        \bigskip

        ~\label{fig:increase_performance_mcts_annexe}
\end{figure*}

In Tables~\ref{tab:all_data_mb},~\ref{tab:all_data_mb2} and ~\ref{tab:all_data_mb3}, we observe the score obtained at the end of the training for the different algorithms tested on the Atari100k benchmarks. 
In Figure~\ref{fig:full_curves}, we observe all the learning curves of all the algorithms on the Atari100k benchmarks.
In Figure~\ref{fig:increase_performance_mcts_annexe}, we observe the percentage improvement of the different algorithms. 
In Figure~\ref{fig:scoreDistribution}, we observe the score distribution of the different algorithms, following methodology from~\cite{agarwal2021deep}.
 
\section{Questions and Answers}

A few questions were asked during the review process, which could also be of interest to the reader. 
\begin{itemize}
    \item \textbf{Q1:} Links with Offline RL?
    \item \textbf{A1:} There is no direct link with offline RL in our work, but it inspired our approach. Therefore, we discussed it in the Related Work section to highlight alternative methods of penalization for deviations from the expert. 
    \item \textbf{Q2:} Extending the Proposed Method to Value-Based Approaches like DQN?
    \item \textbf{A2:} For value-based approaches like DQN, which lack an actor loss function, two solutions are: (\romannumeral 1) remove the actor loss function entirely and retain only the critic loss, which still benefits from the expert's guidance, or (\romannumeral 2) retain the actor loss function so that it can still be used by AlphaZero, but only $\lossActorSub{\worldstate}$ of equation~\ref{equa:lossActorMCTS} would remain. 
    \item \textbf{Q3:} Useful/Useless Information from Expert:?
    \item \textbf{A3:} Utilizing a search algorithm as an expert generally avoids harmful guidance. However, as noted, lower performance was observed in one game, though this issue was resolved in A2C\_AZ*. In our case, the hyperparameters in Equations~\ref{equa:NewlossCritic} and \ref{equa:actor_final} were determined through experimentation with various values. Moving forward, it would be valuable to explore neural network-based weighting methods, which could dynamically adjust the weights based on both the expert's input and the current state. This approach could potentially minimize costs while maximizing the beneficial impact of expert guidance.
\end{itemize}

%% file: Image/assault.tex
\begin{tikzpicture}

\definecolor{darkcyan1115178}{RGB}{1,115,178}
\definecolor{darkorange2221435}{RGB}{222,143,5}
\definecolor{darkslategray38}{RGB}{38,38,38}
\definecolor{lavender234234242}{RGB}{234,234,242}
\definecolor{lightgray204}{RGB}{204,204,204}
\definecolor{orchid204120188}{RGB}{204,120,188}
\definecolor{peru20214597}{RGB}{202,145,97}

\begin{axis}[
axis background/.style={fill=lavender234234242},
axis line style={white},
legend cell align={left},
legend style={
  fill opacity=0.8,
  draw opacity=1,
  text opacity=1,
  at={(0.03,0.97)},
  anchor=north west,
  draw=lightgray204,
  fill=lavender234234242
},
tick align=outside,
x grid style={white},
xlabel=\textcolor{darkslategray38}{Time Step},
xmajorgrids,
xmajorticks=true,
xmin=-30000, xmax=410000,
xtick style={color=darkslategray38},
xtick={0,50000,100000,150000,200000,250000,300000,350000,400000},
xticklabels={0k,50k,100k,150k,200k,250k,300k,350k,400k},
y grid style={white},
ylabel=\textcolor{darkslategray38}{Performance},
ymajorgrids,
ymajorticks=true,
ymin=-261.542256132146, ymax=2678.68073477183,
ytick style={color=darkslategray38}
]
\path [draw=darkcyan1115178, fill=darkcyan1115178, opacity=0.2]
(axis cs:-10000,318.73423518529)
--(axis cs:-10000,-62.5342351852902)
--(axis cs:0,219.16576481471)
--(axis cs:10000,286.66576481471)
--(axis cs:20000,452.86576481471)
--(axis cs:30000,479.26576481471)
--(axis cs:40000,541.96576481471)
--(axis cs:50000,574.66576481471)
--(axis cs:60000,600.56576481471)
--(axis cs:70000,613.76576481471)
--(axis cs:80000,659.66576481471)
--(axis cs:90000,973.76576481471)
--(axis cs:100000,845.96576481471)
--(axis cs:110000,877.96576481471)
--(axis cs:120000,945.36576481471)
--(axis cs:130000,903.06576481471)
--(axis cs:140000,985.46576481471)
--(axis cs:150000,1077.86576481471)
--(axis cs:160000,1241.86576481471)
--(axis cs:170000,1135.36576481471)
--(axis cs:180000,1187.96576481471)
--(axis cs:190000,1414.06576481471)
--(axis cs:200000,1277.56576481471)
--(axis cs:210000,1280.26576481471)
--(axis cs:220000,1594.46576481471)
--(axis cs:230000,1398.66576481471)
--(axis cs:240000,1632.96576481471)
--(axis cs:250000,1378.76576481471)
--(axis cs:260000,1640.56576481471)
--(axis cs:270000,1625.06576481471)
--(axis cs:280000,1749.76576481471)
--(axis cs:290000,1647.86576481471)
--(axis cs:300000,1802.16576481471)
--(axis cs:310000,1890.26576481471)
--(axis cs:320000,1873.46576481471)
--(axis cs:330000,1759.46576481471)
--(axis cs:340000,1934.36576481471)
--(axis cs:350000,2076.86576481471)
--(axis cs:360000,2039.36576481471)
--(axis cs:370000,1952.56576481471)
--(axis cs:380000,2163.76576481471)
--(axis cs:390000,2126.06576481471)
--(axis cs:390000,2507.33423518529)
--(axis cs:390000,2507.33423518529)
--(axis cs:380000,2545.03423518529)
--(axis cs:370000,2333.83423518529)
--(axis cs:360000,2420.63423518529)
--(axis cs:350000,2458.13423518529)
--(axis cs:340000,2315.63423518529)
--(axis cs:330000,2140.73423518529)
--(axis cs:320000,2254.73423518529)
--(axis cs:310000,2271.53423518529)
--(axis cs:300000,2183.43423518529)
--(axis cs:290000,2029.13423518529)
--(axis cs:280000,2131.03423518529)
--(axis cs:270000,2006.33423518529)
--(axis cs:260000,2021.83423518529)
--(axis cs:250000,1760.03423518529)
--(axis cs:240000,2014.23423518529)
--(axis cs:230000,1779.93423518529)
--(axis cs:220000,1975.73423518529)
--(axis cs:210000,1661.53423518529)
--(axis cs:200000,1658.83423518529)
--(axis cs:190000,1795.33423518529)
--(axis cs:180000,1569.23423518529)
--(axis cs:170000,1516.63423518529)
--(axis cs:160000,1623.13423518529)
--(axis cs:150000,1459.13423518529)
--(axis cs:140000,1366.73423518529)
--(axis cs:130000,1284.33423518529)
--(axis cs:120000,1326.63423518529)
--(axis cs:110000,1259.23423518529)
--(axis cs:100000,1227.23423518529)
--(axis cs:90000,1355.03423518529)
--(axis cs:80000,1040.93423518529)
--(axis cs:70000,995.03423518529)
--(axis cs:60000,981.83423518529)
--(axis cs:50000,955.93423518529)
--(axis cs:40000,923.23423518529)
--(axis cs:30000,860.53423518529)
--(axis cs:20000,834.13423518529)
--(axis cs:10000,667.93423518529)
--(axis cs:0,600.43423518529)
--(axis cs:-10000,318.73423518529)
--cycle;

\path [draw=darkorange2221435, fill=darkorange2221435, opacity=0.2]
(axis cs:-10000,272.776488129916)
--(axis cs:-10000,152.223511870084)
--(axis cs:0,305.723511870084)
--(axis cs:10000,385.723511870084)
--(axis cs:20000,590.923511870084)
--(axis cs:30000,665.023511870084)
--(axis cs:40000,708.523511870084)
--(axis cs:50000,715.923511870084)
--(axis cs:60000,755.623511870084)
--(axis cs:70000,733.723511870084)
--(axis cs:80000,760.223511870084)
--(axis cs:90000,750.523511870084)
--(axis cs:100000,786.823511870084)
--(axis cs:110000,784.723511870084)
--(axis cs:120000,749.523511870084)
--(axis cs:130000,791.323511870084)
--(axis cs:140000,821.923511870084)
--(axis cs:150000,794.023511870084)
--(axis cs:160000,844.823511870084)
--(axis cs:170000,844.623511870084)
--(axis cs:180000,820.423511870084)
--(axis cs:190000,845.923511870084)
--(axis cs:200000,858.223511870084)
--(axis cs:210000,855.723511870084)
--(axis cs:220000,864.123511870084)
--(axis cs:230000,897.123511870084)
--(axis cs:240000,921.023511870084)
--(axis cs:250000,919.823511870084)
--(axis cs:260000,918.323511870084)
--(axis cs:270000,938.923511870084)
--(axis cs:280000,1064.42351187008)
--(axis cs:290000,1076.92351187008)
--(axis cs:300000,1009.42351187008)
--(axis cs:310000,1035.12351187008)
--(axis cs:320000,1119.42351187008)
--(axis cs:330000,1140.22351187008)
--(axis cs:340000,1098.72351187008)
--(axis cs:350000,1069.82351187008)
--(axis cs:360000,1020.22351187008)
--(axis cs:370000,1101.72351187008)
--(axis cs:380000,1037.32351187008)
--(axis cs:390000,1053.32351187008)
--(axis cs:390000,1173.87648812992)
--(axis cs:390000,1173.87648812992)
--(axis cs:380000,1157.87648812992)
--(axis cs:370000,1222.27648812992)
--(axis cs:360000,1140.77648812992)
--(axis cs:350000,1190.37648812992)
--(axis cs:340000,1219.27648812992)
--(axis cs:330000,1260.77648812992)
--(axis cs:320000,1239.97648812992)
--(axis cs:310000,1155.67648812992)
--(axis cs:300000,1129.97648812992)
--(axis cs:290000,1197.47648812992)
--(axis cs:280000,1184.97648812992)
--(axis cs:270000,1059.47648812992)
--(axis cs:260000,1038.87648812992)
--(axis cs:250000,1040.37648812992)
--(axis cs:240000,1041.57648812992)
--(axis cs:230000,1017.67648812992)
--(axis cs:220000,984.676488129916)
--(axis cs:210000,976.276488129916)
--(axis cs:200000,978.776488129916)
--(axis cs:190000,966.476488129916)
--(axis cs:180000,940.976488129916)
--(axis cs:170000,965.176488129916)
--(axis cs:160000,965.376488129916)
--(axis cs:150000,914.576488129916)
--(axis cs:140000,942.476488129916)
--(axis cs:130000,911.876488129916)
--(axis cs:120000,870.076488129916)
--(axis cs:110000,905.276488129916)
--(axis cs:100000,907.376488129916)
--(axis cs:90000,871.076488129916)
--(axis cs:80000,880.776488129916)
--(axis cs:70000,854.276488129916)
--(axis cs:60000,876.176488129916)
--(axis cs:50000,836.476488129916)
--(axis cs:40000,829.076488129916)
--(axis cs:30000,785.576488129916)
--(axis cs:20000,711.476488129916)
--(axis cs:10000,506.276488129916)
--(axis cs:0,426.276488129916)
--(axis cs:-10000,272.776488129916)
--cycle;

\path [draw=orchid204120188, fill=orchid204120188, opacity=0.2]
(axis cs:-10000,443.989181096645)
--(axis cs:-10000,57.6108189033548)
--(axis cs:0,184.510818903355)
--(axis cs:10000,185.410818903355)
--(axis cs:20000,261.310818903355)
--(axis cs:30000,452.410818903355)
--(axis cs:40000,495.010818903355)
--(axis cs:50000,528.910818903355)
--(axis cs:60000,571.610818903355)
--(axis cs:70000,598.810818903355)
--(axis cs:80000,635.510818903355)
--(axis cs:90000,727.610818903355)
--(axis cs:100000,781.310818903355)
--(axis cs:110000,827.110818903355)
--(axis cs:120000,936.010818903355)
--(axis cs:130000,1116.11081890335)
--(axis cs:140000,1181.21081890335)
--(axis cs:150000,1109.11081890335)
--(axis cs:160000,1207.81081890335)
--(axis cs:170000,1115.31081890335)
--(axis cs:180000,1327.01081890335)
--(axis cs:190000,1501.81081890335)
--(axis cs:200000,1456.31081890335)
--(axis cs:210000,1441.51081890335)
--(axis cs:220000,1430.31081890335)
--(axis cs:230000,1703.41081890335)
--(axis cs:240000,1886.81081890335)
--(axis cs:250000,1965.11081890336)
--(axis cs:260000,1998.81081890335)
--(axis cs:270000,2004.81081890335)
--(axis cs:280000,2081.51081890335)
--(axis cs:290000,1822.41081890335)
--(axis cs:300000,1974.81081890335)
--(axis cs:310000,1871.11081890336)
--(axis cs:320000,2064.61081890335)
--(axis cs:330000,1969.21081890335)
--(axis cs:340000,1948.41081890335)
--(axis cs:350000,1945.41081890335)
--(axis cs:360000,2135.61081890335)
--(axis cs:370000,1985.21081890335)
--(axis cs:380000,1941.11081890336)
--(axis cs:390000,2056.51081890335)
--(axis cs:390000,2442.88918109665)
--(axis cs:390000,2442.88918109665)
--(axis cs:380000,2327.48918109665)
--(axis cs:370000,2371.58918109665)
--(axis cs:360000,2521.98918109665)
--(axis cs:350000,2331.78918109665)
--(axis cs:340000,2334.78918109665)
--(axis cs:330000,2355.58918109665)
--(axis cs:320000,2450.98918109665)
--(axis cs:310000,2257.48918109665)
--(axis cs:300000,2361.18918109665)
--(axis cs:290000,2208.78918109665)
--(axis cs:280000,2467.88918109665)
--(axis cs:270000,2391.18918109665)
--(axis cs:260000,2385.18918109665)
--(axis cs:250000,2351.48918109665)
--(axis cs:240000,2273.18918109665)
--(axis cs:230000,2089.78918109665)
--(axis cs:220000,1816.68918109665)
--(axis cs:210000,1827.88918109665)
--(axis cs:200000,1842.68918109665)
--(axis cs:190000,1888.18918109665)
--(axis cs:180000,1713.38918109665)
--(axis cs:170000,1501.68918109665)
--(axis cs:160000,1594.18918109665)
--(axis cs:150000,1495.48918109665)
--(axis cs:140000,1567.58918109665)
--(axis cs:130000,1502.48918109665)
--(axis cs:120000,1322.38918109665)
--(axis cs:110000,1213.48918109665)
--(axis cs:100000,1167.68918109665)
--(axis cs:90000,1113.98918109665)
--(axis cs:80000,1021.88918109665)
--(axis cs:70000,985.189181096645)
--(axis cs:60000,957.989181096645)
--(axis cs:50000,915.289181096645)
--(axis cs:40000,881.389181096645)
--(axis cs:30000,838.789181096645)
--(axis cs:20000,647.689181096645)
--(axis cs:10000,571.789181096645)
--(axis cs:0,570.889181096645)
--(axis cs:-10000,443.989181096645)
--cycle;

\path [draw=peru20214597, fill=peru20214597, opacity=0.2]
(axis cs:-10000,372.395756545601)
--(axis cs:-10000,10.4042434543987)
--(axis cs:0,-127.895756545601)
--(axis cs:10000,311.604243454399)
--(axis cs:20000,350.904243454399)
--(axis cs:30000,380.304243454399)
--(axis cs:40000,407.304243454399)
--(axis cs:50000,428.004243454399)
--(axis cs:60000,438.204243454399)
--(axis cs:70000,475.704243454399)
--(axis cs:80000,531.904243454399)
--(axis cs:90000,602.504243454399)
--(axis cs:100000,618.004243454399)
--(axis cs:110000,695.004243454399)
--(axis cs:120000,797.704243454399)
--(axis cs:130000,828.904243454399)
--(axis cs:140000,849.004243454399)
--(axis cs:150000,846.804243454399)
--(axis cs:160000,901.304243454399)
--(axis cs:170000,971.804243454399)
--(axis cs:180000,1022.9042434544)
--(axis cs:190000,1145.7042434544)
--(axis cs:200000,1116.4042434544)
--(axis cs:210000,1164.5042434544)
--(axis cs:220000,1323.0042434544)
--(axis cs:230000,1534.8042434544)
--(axis cs:240000,1268.6042434544)
--(axis cs:250000,1351.1042434544)
--(axis cs:260000,1398.9042434544)
--(axis cs:270000,1350.4042434544)
--(axis cs:280000,1206.9042434544)
--(axis cs:290000,1453.5042434544)
--(axis cs:300000,1518.5042434544)
--(axis cs:310000,1544.5042434544)
--(axis cs:320000,1458.8042434544)
--(axis cs:330000,1377.0042434544)
--(axis cs:340000,1343.3042434544)
--(axis cs:350000,1477.2042434544)
--(axis cs:360000,1479.8042434544)
--(axis cs:370000,1468.8042434544)
--(axis cs:380000,1523.0042434544)
--(axis cs:390000,1378.7042434544)
--(axis cs:390000,1740.6957565456)
--(axis cs:390000,1740.6957565456)
--(axis cs:380000,1884.9957565456)
--(axis cs:370000,1830.7957565456)
--(axis cs:360000,1841.7957565456)
--(axis cs:350000,1839.1957565456)
--(axis cs:340000,1705.2957565456)
--(axis cs:330000,1738.9957565456)
--(axis cs:320000,1820.7957565456)
--(axis cs:310000,1906.4957565456)
--(axis cs:300000,1880.4957565456)
--(axis cs:290000,1815.4957565456)
--(axis cs:280000,1568.8957565456)
--(axis cs:270000,1712.3957565456)
--(axis cs:260000,1760.8957565456)
--(axis cs:250000,1713.0957565456)
--(axis cs:240000,1630.5957565456)
--(axis cs:230000,1896.7957565456)
--(axis cs:220000,1684.9957565456)
--(axis cs:210000,1526.4957565456)
--(axis cs:200000,1478.3957565456)
--(axis cs:190000,1507.6957565456)
--(axis cs:180000,1384.8957565456)
--(axis cs:170000,1333.7957565456)
--(axis cs:160000,1263.2957565456)
--(axis cs:150000,1208.7957565456)
--(axis cs:140000,1210.9957565456)
--(axis cs:130000,1190.8957565456)
--(axis cs:120000,1159.6957565456)
--(axis cs:110000,1056.9957565456)
--(axis cs:100000,979.995756545601)
--(axis cs:90000,964.495756545601)
--(axis cs:80000,893.895756545601)
--(axis cs:70000,837.695756545601)
--(axis cs:60000,800.195756545601)
--(axis cs:50000,789.995756545601)
--(axis cs:40000,769.295756545601)
--(axis cs:30000,742.295756545601)
--(axis cs:20000,712.895756545601)
--(axis cs:10000,673.595756545601)
--(axis cs:0,234.095756545601)
--(axis cs:-10000,372.395756545601)
--cycle;

\addplot [semithick, darkcyan1115178, mark=square*, mark size=1.5, mark options={solid}]
table {%
-10000 128.1
0 409.8
10000 477.3
20000 643.5
30000 669.9
40000 732.6
50000 765.3
60000 791.2
70000 804.4
80000 850.3
90000 1164.4
100000 1036.6
110000 1068.6
120000 1136
130000 1093.7
140000 1176.1
150000 1268.5
160000 1432.5
170000 1326
180000 1378.6
190000 1604.7
200000 1468.2
210000 1470.9
220000 1785.1
230000 1589.3
240000 1823.6
250000 1569.4
260000 1831.2
270000 1815.7
280000 1940.4
290000 1838.5
300000 1992.8
310000 2080.9
320000 2064.1
330000 1950.1
340000 2125
350000 2267.5
360000 2230
370000 2143.2
380000 2354.4
390000 2316.7
};
\addlegendentry{A2C-AZ*}
\addplot [semithick, darkorange2221435, mark=triangle*, mark size=1.5, mark options={solid}]
table {%
-10000 212.5
0 366
10000 446
20000 651.2
30000 725.3
40000 768.8
50000 776.2
60000 815.9
70000 794
80000 820.5
90000 810.8
100000 847.1
110000 845
120000 809.8
130000 851.6
140000 882.2
150000 854.3
160000 905.1
170000 904.9
180000 880.7
190000 906.2
200000 918.5
210000 916
220000 924.4
230000 957.4
240000 981.3
250000 980.1
260000 978.6
270000 999.2
280000 1124.7
290000 1137.2
300000 1069.7
310000 1095.4
320000 1179.7
330000 1200.5
340000 1159
350000 1130.1
360000 1080.5
370000 1162
380000 1097.6
390000 1113.6
};
\addlegendentry{A2C}
\addplot [semithick, orchid204120188, mark=+, mark size=1.5, mark options={solid}]
table {%
-10000 250.8
0 377.7
10000 378.6
20000 454.5
30000 645.6
40000 688.2
50000 722.1
60000 764.8
70000 792
80000 828.7
90000 920.8
100000 974.5
110000 1020.3
120000 1129.2
130000 1309.3
140000 1374.4
150000 1302.3
160000 1401
170000 1308.5
180000 1520.2
190000 1695
200000 1649.5
210000 1634.7
220000 1623.5
230000 1896.6
240000 2080
250000 2158.3
260000 2192
270000 2198
280000 2274.7
290000 2015.6
300000 2168
310000 2064.3
320000 2257.8
330000 2162.4
340000 2141.6
350000 2138.6
360000 2328.8
370000 2178.4
380000 2134.3
390000 2249.7
};
\addlegendentry{A2C-AZ}
\addplot [semithick, peru20214597, mark=diamond*, mark size=1.5, mark options={solid}]
table {%
-10000 191.4
0 53.1
10000 492.6
20000 531.9
30000 561.3
40000 588.3
50000 609
60000 619.2
70000 656.7
80000 712.9
90000 783.5
100000 799
110000 876
120000 978.7
130000 1009.9
140000 1030
150000 1027.8
160000 1082.3
170000 1152.8
180000 1203.9
190000 1326.7
200000 1297.4
210000 1345.5
220000 1504
230000 1715.8
240000 1449.6
250000 1532.1
260000 1579.9
270000 1531.4
280000 1387.9
290000 1634.5
300000 1699.5
310000 1725.5
320000 1639.8
330000 1558
340000 1524.3
350000 1658.2
360000 1660.8
370000 1649.8
380000 1704
390000 1559.7
};
\addlegendentry{AZ}
\end{axis}

\end{tikzpicture}

%% file: Image/asterix.tex
\begin{tikzpicture}

\definecolor{darkcyan1115178}{RGB}{1,115,178}
\definecolor{darkorange2221435}{RGB}{222,143,5}
\definecolor{darkslategray38}{RGB}{38,38,38}
\definecolor{lavender234234242}{RGB}{234,234,242}
\definecolor{lightgray204}{RGB}{204,204,204}
\definecolor{orchid204120188}{RGB}{204,120,188}
\definecolor{peru20214597}{RGB}{202,145,97}

\begin{axis}[
axis background/.style={fill=lavender234234242},
axis line style={white},
legend cell align={left},
legend style={
  fill opacity=0.8,
  draw opacity=1,
  text opacity=1,
  at={(0.03,0.97)},
  anchor=north west,
  draw=lightgray204,
  fill=lavender234234242
},
tick align=outside,
x grid style={white},
xlabel=\textcolor{darkslategray38}{Time Step},
xmajorgrids,
xmajorticks=true,
xmin=-30000, xmax=410000,
xtick style={color=darkslategray38},
xtick={0,50000,100000,150000,200000,250000,300000,350000,400000},
xticklabels={0k,50k,100k,150k,200k,250k,300k,350k,400k},
y grid style={white},
ylabel=\textcolor{darkslategray38}{Performance},
ymajorgrids,
ymajorticks=true,
ymin=-128.590670201245, ymax=3204.99067020124,
ytick style={color=darkslategray38}
]
\path [draw=darkcyan1115178, fill=darkcyan1115178, opacity=0.2]
(axis cs:-10000,408.464245637496)
--(axis cs:-10000,22.9357543625044)
--(axis cs:0,855.135754362504)
--(axis cs:10000,1108.6357543625)
--(axis cs:20000,1170.1357543625)
--(axis cs:30000,1064.3357543625)
--(axis cs:40000,999.335754362504)
--(axis cs:50000,1224.3357543625)
--(axis cs:60000,1494.3357543625)
--(axis cs:70000,1582.2357543625)
--(axis cs:80000,1682.2357543625)
--(axis cs:90000,1715.8357543625)
--(axis cs:100000,1775.8357543625)
--(axis cs:110000,1801.5357543625)
--(axis cs:120000,2016.5357543625)
--(axis cs:130000,2034.3357543625)
--(axis cs:140000,2127.9357543625)
--(axis cs:150000,1971.5357543625)
--(axis cs:160000,2015.8357543625)
--(axis cs:170000,2326.5357543625)
--(axis cs:180000,2106.5357543625)
--(axis cs:190000,2264.3357543625)
--(axis cs:200000,2322.9357543625)
--(axis cs:210000,2128.6357543625)
--(axis cs:220000,2198.6357543625)
--(axis cs:230000,2611.5357543625)
--(axis cs:240000,2515.1357543625)
--(axis cs:250000,2601.5357543625)
--(axis cs:260000,2506.5357543625)
--(axis cs:270000,2497.9357543625)
--(axis cs:280000,2667.9357543625)
--(axis cs:290000,2520.8357543625)
--(axis cs:300000,2451.5357543625)
--(axis cs:310000,2424.3357543625)
--(axis cs:320000,2554.3357543625)
--(axis cs:330000,2446.5357543625)
--(axis cs:340000,2525.8357543625)
--(axis cs:350000,2522.9357543625)
--(axis cs:360000,2470.1357543625)
--(axis cs:370000,2597.2357543625)
--(axis cs:380000,2582.9357543625)
--(axis cs:390000,2626.5357543625)
--(axis cs:390000,3012.0642456375)
--(axis cs:390000,3012.0642456375)
--(axis cs:380000,2968.4642456375)
--(axis cs:370000,2982.7642456375)
--(axis cs:360000,2855.6642456375)
--(axis cs:350000,2908.4642456375)
--(axis cs:340000,2911.3642456375)
--(axis cs:330000,2832.0642456375)
--(axis cs:320000,2939.8642456375)
--(axis cs:310000,2809.8642456375)
--(axis cs:300000,2837.0642456375)
--(axis cs:290000,2906.3642456375)
--(axis cs:280000,3053.4642456375)
--(axis cs:270000,2883.4642456375)
--(axis cs:260000,2892.0642456375)
--(axis cs:250000,2987.0642456375)
--(axis cs:240000,2900.6642456375)
--(axis cs:230000,2997.0642456375)
--(axis cs:220000,2584.1642456375)
--(axis cs:210000,2514.1642456375)
--(axis cs:200000,2708.4642456375)
--(axis cs:190000,2649.8642456375)
--(axis cs:180000,2492.0642456375)
--(axis cs:170000,2712.0642456375)
--(axis cs:160000,2401.3642456375)
--(axis cs:150000,2357.0642456375)
--(axis cs:140000,2513.4642456375)
--(axis cs:130000,2419.8642456375)
--(axis cs:120000,2402.0642456375)
--(axis cs:110000,2187.0642456375)
--(axis cs:100000,2161.3642456375)
--(axis cs:90000,2101.3642456375)
--(axis cs:80000,2067.7642456375)
--(axis cs:70000,1967.7642456375)
--(axis cs:60000,1879.8642456375)
--(axis cs:50000,1609.8642456375)
--(axis cs:40000,1384.8642456375)
--(axis cs:30000,1449.8642456375)
--(axis cs:20000,1555.6642456375)
--(axis cs:10000,1494.1642456375)
--(axis cs:0,1240.6642456375)
--(axis cs:-10000,408.464245637496)
--cycle;

\path [draw=darkorange2221435, fill=darkorange2221435, opacity=0.2]
(axis cs:-10000,299.41567007528)
--(axis cs:-10000,65.5843299247201)
--(axis cs:0,839.08432992472)
--(axis cs:10000,1007.08432992472)
--(axis cs:20000,1004.58432992472)
--(axis cs:30000,1180.58432992472)
--(axis cs:40000,1268.58432992472)
--(axis cs:50000,1260.58432992472)
--(axis cs:60000,1406.08432992472)
--(axis cs:70000,1509.08432992472)
--(axis cs:80000,1583.08432992472)
--(axis cs:90000,1658.08432992472)
--(axis cs:100000,1636.08432992472)
--(axis cs:110000,1760.08432992472)
--(axis cs:120000,1800.58432992472)
--(axis cs:130000,1769.58432992472)
--(axis cs:140000,1807.58432992472)
--(axis cs:150000,2009.58432992472)
--(axis cs:160000,1926.58432992472)
--(axis cs:170000,2003.08432992472)
--(axis cs:180000,1949.08432992472)
--(axis cs:190000,2021.58432992472)
--(axis cs:200000,2022.08432992472)
--(axis cs:210000,2056.08432992472)
--(axis cs:220000,1987.08432992472)
--(axis cs:230000,2017.08432992472)
--(axis cs:240000,1963.58432992472)
--(axis cs:250000,2029.58432992472)
--(axis cs:260000,2018.08432992472)
--(axis cs:270000,2110.58432992472)
--(axis cs:280000,2121.08432992472)
--(axis cs:290000,2148.58432992472)
--(axis cs:300000,2012.58432992472)
--(axis cs:310000,2108.58432992472)
--(axis cs:320000,2222.58432992472)
--(axis cs:330000,2198.58432992472)
--(axis cs:340000,2229.58432992472)
--(axis cs:350000,2228.08432992472)
--(axis cs:360000,2284.08432992472)
--(axis cs:370000,2275.58432992472)
--(axis cs:380000,2190.58432992472)
--(axis cs:390000,2177.58432992472)
--(axis cs:390000,2411.41567007528)
--(axis cs:390000,2411.41567007528)
--(axis cs:380000,2424.41567007528)
--(axis cs:370000,2509.41567007528)
--(axis cs:360000,2517.91567007528)
--(axis cs:350000,2461.91567007528)
--(axis cs:340000,2463.41567007528)
--(axis cs:330000,2432.41567007528)
--(axis cs:320000,2456.41567007528)
--(axis cs:310000,2342.41567007528)
--(axis cs:300000,2246.41567007528)
--(axis cs:290000,2382.41567007528)
--(axis cs:280000,2354.91567007528)
--(axis cs:270000,2344.41567007528)
--(axis cs:260000,2251.91567007528)
--(axis cs:250000,2263.41567007528)
--(axis cs:240000,2197.41567007528)
--(axis cs:230000,2250.91567007528)
--(axis cs:220000,2220.91567007528)
--(axis cs:210000,2289.91567007528)
--(axis cs:200000,2255.91567007528)
--(axis cs:190000,2255.41567007528)
--(axis cs:180000,2182.91567007528)
--(axis cs:170000,2236.91567007528)
--(axis cs:160000,2160.41567007528)
--(axis cs:150000,2243.41567007528)
--(axis cs:140000,2041.41567007528)
--(axis cs:130000,2003.41567007528)
--(axis cs:120000,2034.41567007528)
--(axis cs:110000,1993.91567007528)
--(axis cs:100000,1869.91567007528)
--(axis cs:90000,1891.91567007528)
--(axis cs:80000,1816.91567007528)
--(axis cs:70000,1742.91567007528)
--(axis cs:60000,1639.91567007528)
--(axis cs:50000,1494.41567007528)
--(axis cs:40000,1502.41567007528)
--(axis cs:30000,1414.41567007528)
--(axis cs:20000,1238.41567007528)
--(axis cs:10000,1240.91567007528)
--(axis cs:0,1072.91567007528)
--(axis cs:-10000,299.41567007528)
--cycle;

\path [draw=orchid204120188, fill=orchid204120188, opacity=0.2]
(axis cs:-10000,408.464245637496)
--(axis cs:-10000,22.9357543625044)
--(axis cs:0,855.135754362504)
--(axis cs:10000,1108.6357543625)
--(axis cs:20000,1170.1357543625)
--(axis cs:30000,1064.3357543625)
--(axis cs:40000,999.335754362504)
--(axis cs:50000,1224.3357543625)
--(axis cs:60000,1494.3357543625)
--(axis cs:70000,1582.2357543625)
--(axis cs:80000,1682.2357543625)
--(axis cs:90000,1715.8357543625)
--(axis cs:100000,1775.8357543625)
--(axis cs:110000,1801.5357543625)
--(axis cs:120000,2016.5357543625)
--(axis cs:130000,2034.3357543625)
--(axis cs:140000,2127.9357543625)
--(axis cs:150000,1971.5357543625)
--(axis cs:160000,2015.8357543625)
--(axis cs:170000,2326.5357543625)
--(axis cs:180000,2106.5357543625)
--(axis cs:190000,2264.3357543625)
--(axis cs:200000,2322.9357543625)
--(axis cs:210000,2128.6357543625)
--(axis cs:220000,2198.6357543625)
--(axis cs:230000,2611.5357543625)
--(axis cs:240000,2515.1357543625)
--(axis cs:250000,2601.5357543625)
--(axis cs:260000,2506.5357543625)
--(axis cs:270000,2497.9357543625)
--(axis cs:280000,2667.9357543625)
--(axis cs:290000,2520.8357543625)
--(axis cs:300000,2451.5357543625)
--(axis cs:310000,2424.3357543625)
--(axis cs:320000,2554.3357543625)
--(axis cs:330000,2446.5357543625)
--(axis cs:340000,2525.8357543625)
--(axis cs:350000,2522.9357543625)
--(axis cs:360000,2470.1357543625)
--(axis cs:370000,2597.2357543625)
--(axis cs:380000,2582.9357543625)
--(axis cs:390000,2626.5357543625)
--(axis cs:390000,3012.0642456375)
--(axis cs:390000,3012.0642456375)
--(axis cs:380000,2968.4642456375)
--(axis cs:370000,2982.7642456375)
--(axis cs:360000,2855.6642456375)
--(axis cs:350000,2908.4642456375)
--(axis cs:340000,2911.3642456375)
--(axis cs:330000,2832.0642456375)
--(axis cs:320000,2939.8642456375)
--(axis cs:310000,2809.8642456375)
--(axis cs:300000,2837.0642456375)
--(axis cs:290000,2906.3642456375)
--(axis cs:280000,3053.4642456375)
--(axis cs:270000,2883.4642456375)
--(axis cs:260000,2892.0642456375)
--(axis cs:250000,2987.0642456375)
--(axis cs:240000,2900.6642456375)
--(axis cs:230000,2997.0642456375)
--(axis cs:220000,2584.1642456375)
--(axis cs:210000,2514.1642456375)
--(axis cs:200000,2708.4642456375)
--(axis cs:190000,2649.8642456375)
--(axis cs:180000,2492.0642456375)
--(axis cs:170000,2712.0642456375)
--(axis cs:160000,2401.3642456375)
--(axis cs:150000,2357.0642456375)
--(axis cs:140000,2513.4642456375)
--(axis cs:130000,2419.8642456375)
--(axis cs:120000,2402.0642456375)
--(axis cs:110000,2187.0642456375)
--(axis cs:100000,2161.3642456375)
--(axis cs:90000,2101.3642456375)
--(axis cs:80000,2067.7642456375)
--(axis cs:70000,1967.7642456375)
--(axis cs:60000,1879.8642456375)
--(axis cs:50000,1609.8642456375)
--(axis cs:40000,1384.8642456375)
--(axis cs:30000,1449.8642456375)
--(axis cs:20000,1555.6642456375)
--(axis cs:10000,1494.1642456375)
--(axis cs:0,1240.6642456375)
--(axis cs:-10000,408.464245637496)
--cycle;

\path [draw=peru20214597, fill=peru20214597, opacity=0.2]
(axis cs:-10000,373.563177557596)
--(axis cs:-10000,95.0368224424043)
--(axis cs:0,195.036822442404)
--(axis cs:10000,498.636822442404)
--(axis cs:20000,557.136822442404)
--(axis cs:30000,566.436822442404)
--(axis cs:40000,686.436822442404)
--(axis cs:50000,738.636822442404)
--(axis cs:60000,777.136822442404)
--(axis cs:70000,780.036822442404)
--(axis cs:80000,753.636822442404)
--(axis cs:90000,725.736822442404)
--(axis cs:100000,879.336822442404)
--(axis cs:110000,791.436822442404)
--(axis cs:120000,787.136822442404)
--(axis cs:130000,897.136822442404)
--(axis cs:140000,838.636822442404)
--(axis cs:150000,912.136822442404)
--(axis cs:160000,895.036822442404)
--(axis cs:170000,882.136822442404)
--(axis cs:180000,811.436822442404)
--(axis cs:190000,903.636822442404)
--(axis cs:200000,933.636822442404)
--(axis cs:210000,849.336822442404)
--(axis cs:220000,853.636822442404)
--(axis cs:230000,919.336822442404)
--(axis cs:240000,922.836822442404)
--(axis cs:250000,899.336822442404)
--(axis cs:260000,998.636822442404)
--(axis cs:270000,801.436822442404)
--(axis cs:280000,875.036822442404)
--(axis cs:290000,817.836822442404)
--(axis cs:300000,948.636822442404)
--(axis cs:310000,1010.0368224424)
--(axis cs:320000,1045.7368224424)
--(axis cs:330000,1016.4368224424)
--(axis cs:340000,1091.4368224424)
--(axis cs:350000,990.036822442404)
--(axis cs:360000,1144.3368224424)
--(axis cs:370000,1131.4368224424)
--(axis cs:380000,1101.4368224424)
--(axis cs:390000,1083.6368224424)
--(axis cs:390000,1362.1631775576)
--(axis cs:390000,1362.1631775576)
--(axis cs:380000,1379.9631775576)
--(axis cs:370000,1409.9631775576)
--(axis cs:360000,1422.8631775576)
--(axis cs:350000,1268.5631775576)
--(axis cs:340000,1369.9631775576)
--(axis cs:330000,1294.9631775576)
--(axis cs:320000,1324.2631775576)
--(axis cs:310000,1288.5631775576)
--(axis cs:300000,1227.1631775576)
--(axis cs:290000,1096.3631775576)
--(axis cs:280000,1153.5631775576)
--(axis cs:270000,1079.9631775576)
--(axis cs:260000,1277.1631775576)
--(axis cs:250000,1177.8631775576)
--(axis cs:240000,1201.3631775576)
--(axis cs:230000,1197.8631775576)
--(axis cs:220000,1132.1631775576)
--(axis cs:210000,1127.8631775576)
--(axis cs:200000,1212.1631775576)
--(axis cs:190000,1182.1631775576)
--(axis cs:180000,1089.9631775576)
--(axis cs:170000,1160.6631775576)
--(axis cs:160000,1173.5631775576)
--(axis cs:150000,1190.6631775576)
--(axis cs:140000,1117.1631775576)
--(axis cs:130000,1175.6631775576)
--(axis cs:120000,1065.6631775576)
--(axis cs:110000,1069.9631775576)
--(axis cs:100000,1157.8631775576)
--(axis cs:90000,1004.2631775576)
--(axis cs:80000,1032.1631775576)
--(axis cs:70000,1058.5631775576)
--(axis cs:60000,1055.6631775576)
--(axis cs:50000,1017.1631775576)
--(axis cs:40000,964.963177557596)
--(axis cs:30000,844.963177557596)
--(axis cs:20000,835.663177557596)
--(axis cs:10000,777.163177557596)
--(axis cs:0,473.563177557596)
--(axis cs:-10000,373.563177557596)
--cycle;

\addplot [semithick, darkcyan1115178, mark=square*, mark size=1.5, mark options={solid}]
table {%
-10000 215.7
0 1047.9
10000 1301.4
20000 1362.9
30000 1257.1
40000 1192.1
50000 1417.1
60000 1687.1
70000 1775
80000 1875
90000 1908.6
100000 1968.6
110000 1994.3
120000 2209.3
130000 2227.1
140000 2320.7
150000 2164.3
160000 2208.6
170000 2519.3
180000 2299.3
190000 2457.1
200000 2515.7
210000 2321.4
220000 2391.4
230000 2804.3
240000 2707.9
250000 2794.3
260000 2699.3
270000 2690.7
280000 2860.7
290000 2713.6
300000 2644.3
310000 2617.1
320000 2747.1
330000 2639.3
340000 2718.6
350000 2715.7
360000 2662.9
370000 2790
380000 2775.7
390000 2819.3
};
\addplot [semithick, darkorange2221435, mark=triangle*, mark size=1.5, mark options={solid}]
table {%
-10000 182.5
0 956
10000 1124
20000 1121.5
30000 1297.5
40000 1385.5
50000 1377.5
60000 1523
70000 1626
80000 1700
90000 1775
100000 1753
110000 1877
120000 1917.5
130000 1886.5
140000 1924.5
150000 2126.5
160000 2043.5
170000 2120
180000 2066
190000 2138.5
200000 2139
210000 2173
220000 2104
230000 2134
240000 2080.5
250000 2146.5
260000 2135
270000 2227.5
280000 2238
290000 2265.5
300000 2129.5
310000 2225.5
320000 2339.5
330000 2315.5
340000 2346.5
350000 2345
360000 2401
370000 2392.5
380000 2307.5
390000 2294.5
};
\addplot [semithick, orchid204120188, mark=+, mark size=1.5, mark options={solid}]
table {%
-10000 215.7
0 1047.9
10000 1301.4
20000 1362.9
30000 1257.1
40000 1192.1
50000 1417.1
60000 1687.1
70000 1775
80000 1875
90000 1908.6
100000 1968.6
110000 1994.3
120000 2209.3
130000 2227.1
140000 2320.7
150000 2164.3
160000 2208.6
170000 2519.3
180000 2299.3
190000 2457.1
200000 2515.7
210000 2321.4
220000 2391.4
230000 2804.3
240000 2707.9
250000 2794.3
260000 2699.3
270000 2690.7
280000 2860.7
290000 2713.6
300000 2644.3
310000 2617.1
320000 2747.1
330000 2639.3
340000 2718.6
350000 2715.7
360000 2662.9
370000 2790
380000 2775.7
390000 2819.3
};
\addplot [semithick, peru20214597, mark=diamond*, mark size=1.5, mark options={solid}]
table {%
-10000 234.3
0 334.3
10000 637.9
20000 696.4
30000 705.7
40000 825.7
50000 877.9
60000 916.4
70000 919.3
80000 892.9
90000 865
100000 1018.6
110000 930.7
120000 926.4
130000 1036.4
140000 977.9
150000 1051.4
160000 1034.3
170000 1021.4
180000 950.7
190000 1042.9
200000 1072.9
210000 988.6
220000 992.9
230000 1058.6
240000 1062.1
250000 1038.6
260000 1137.9
270000 940.7
280000 1014.3
290000 957.1
300000 1087.9
310000 1149.3
320000 1185
330000 1155.7
340000 1230.7
350000 1129.3
360000 1283.6
370000 1270.7
380000 1240.7
390000 1222.9
};
\end{axis}

\end{tikzpicture}

%% file: Image/bank_heist.tex
\begin{tikzpicture}

\definecolor{darkcyan1115178}{RGB}{1,115,178}
\definecolor{darkorange2221435}{RGB}{222,143,5}
\definecolor{darkslategray38}{RGB}{38,38,38}
\definecolor{lavender234234242}{RGB}{234,234,242}
\definecolor{lightgray204}{RGB}{204,204,204}
\definecolor{orchid204120188}{RGB}{204,120,188}
\definecolor{peru20214597}{RGB}{202,145,97}

\begin{axis}[
axis background/.style={fill=lavender234234242},
axis line style={white},
legend cell align={left},
legend style={
  fill opacity=0.8,
  draw opacity=1,
  text opacity=1,
  at={(0.97,0.03)},
  anchor=south east,
  draw=lightgray204,
  fill=lavender234234242
},
tick align=outside,
x grid style={white},
xlabel=\textcolor{darkslategray38}{Time Step},
xmajorgrids,
xmajorticks=true,
xmin=-30000, xmax=410000,
xtick style={color=darkslategray38},
xtick={0,50000,100000,150000,200000,250000,300000,350000,400000},
xticklabels={0k,50k,100k,150k,200k,250k,300k,350k,400k},
y grid style={white},
ylabel=\textcolor{darkslategray38}{Performance},
ymajorgrids,
ymajorticks=true,
ymin=-173.631222775146, ymax=1519.93700179446,
ytick style={color=darkslategray38}
]
\path [draw=darkcyan1115178, fill=darkcyan1115178, opacity=0.2]
(axis cs:-10000,108.256627950387)
--(axis cs:-10000,-85.6566279503872)
--(axis cs:0,484.343372049613)
--(axis cs:10000,484.743372049613)
--(axis cs:20000,524.443372049613)
--(axis cs:30000,682.643372049613)
--(axis cs:40000,671.743372049613)
--(axis cs:50000,792.043372049613)
--(axis cs:60000,884.343372049613)
--(axis cs:70000,923.743372049613)
--(axis cs:80000,1075.74337204961)
--(axis cs:90000,1133.64337204961)
--(axis cs:100000,1151.04337204961)
--(axis cs:110000,1170.34337204961)
--(axis cs:120000,1213.74337204961)
--(axis cs:130000,1207.34337204961)
--(axis cs:140000,1201.44337204961)
--(axis cs:150000,1207.04337204961)
--(axis cs:160000,1200.64337204961)
--(axis cs:170000,1197.04337204961)
--(axis cs:180000,1219.44337204961)
--(axis cs:190000,1207.44337204961)
--(axis cs:200000,1221.14337204961)
--(axis cs:210000,1193.64337204961)
--(axis cs:220000,1217.44337204961)
--(axis cs:230000,1222.64337204961)
--(axis cs:240000,1234.64337204961)
--(axis cs:250000,1233.14337204961)
--(axis cs:260000,1230.14337204961)
--(axis cs:270000,1231.04337204961)
--(axis cs:280000,1232.04337204961)
--(axis cs:290000,1238.14337204961)
--(axis cs:300000,1238.44337204961)
--(axis cs:310000,1236.94337204961)
--(axis cs:320000,1237.04337204961)
--(axis cs:330000,1242.04337204961)
--(axis cs:340000,1243.04337204961)
--(axis cs:350000,1244.64337204961)
--(axis cs:360000,1240.04337204961)
--(axis cs:370000,1246.34337204961)
--(axis cs:380000,1232.14337204961)
--(axis cs:390000,1249.04337204961)
--(axis cs:390000,1442.95662795039)
--(axis cs:390000,1442.95662795039)
--(axis cs:380000,1426.05662795039)
--(axis cs:370000,1440.25662795039)
--(axis cs:360000,1433.95662795039)
--(axis cs:350000,1438.55662795039)
--(axis cs:340000,1436.95662795039)
--(axis cs:330000,1435.95662795039)
--(axis cs:320000,1430.95662795039)
--(axis cs:310000,1430.85662795039)
--(axis cs:300000,1432.35662795039)
--(axis cs:290000,1432.05662795039)
--(axis cs:280000,1425.95662795039)
--(axis cs:270000,1424.95662795039)
--(axis cs:260000,1424.05662795039)
--(axis cs:250000,1427.05662795039)
--(axis cs:240000,1428.55662795039)
--(axis cs:230000,1416.55662795039)
--(axis cs:220000,1411.35662795039)
--(axis cs:210000,1387.55662795039)
--(axis cs:200000,1415.05662795039)
--(axis cs:190000,1401.35662795039)
--(axis cs:180000,1413.35662795039)
--(axis cs:170000,1390.95662795039)
--(axis cs:160000,1394.55662795039)
--(axis cs:150000,1400.95662795039)
--(axis cs:140000,1395.35662795039)
--(axis cs:130000,1401.25662795039)
--(axis cs:120000,1407.65662795039)
--(axis cs:110000,1364.25662795039)
--(axis cs:100000,1344.95662795039)
--(axis cs:90000,1327.55662795039)
--(axis cs:80000,1269.65662795039)
--(axis cs:70000,1117.65662795039)
--(axis cs:60000,1078.25662795039)
--(axis cs:50000,985.956627950387)
--(axis cs:40000,865.656627950387)
--(axis cs:30000,876.556627950387)
--(axis cs:20000,718.356627950387)
--(axis cs:10000,678.656627950387)
--(axis cs:0,678.256627950387)
--(axis cs:-10000,108.256627950387)
--cycle;

\path [draw=darkorange2221435, fill=darkorange2221435, opacity=0.2]
(axis cs:-10000,71.7323142878752)
--(axis cs:-10000,-53.5323142878752)
--(axis cs:0,640.567685712125)
--(axis cs:10000,725.667685712125)
--(axis cs:20000,851.567685712125)
--(axis cs:30000,850.667685712125)
--(axis cs:40000,901.667685712125)
--(axis cs:50000,919.267685712125)
--(axis cs:60000,995.067685712125)
--(axis cs:70000,1041.66768571212)
--(axis cs:80000,1068.76768571212)
--(axis cs:90000,1095.46768571212)
--(axis cs:100000,1144.26768571212)
--(axis cs:110000,1155.66768571212)
--(axis cs:120000,1187.16768571212)
--(axis cs:130000,1179.36768571212)
--(axis cs:140000,1192.26768571212)
--(axis cs:150000,1209.36768571212)
--(axis cs:160000,1225.96768571212)
--(axis cs:170000,1239.66768571212)
--(axis cs:180000,1242.86768571212)
--(axis cs:190000,1247.86768571212)
--(axis cs:200000,1234.86768571212)
--(axis cs:210000,1247.36768571212)
--(axis cs:220000,1234.96768571212)
--(axis cs:230000,1244.36768571212)
--(axis cs:240000,1255.36768571212)
--(axis cs:250000,1252.46768571212)
--(axis cs:260000,1255.76768571212)
--(axis cs:270000,1238.86768571212)
--(axis cs:280000,1244.46768571212)
--(axis cs:290000,1258.16768571212)
--(axis cs:300000,1257.56768571212)
--(axis cs:310000,1255.06768571212)
--(axis cs:320000,1250.16768571212)
--(axis cs:330000,1259.16768571212)
--(axis cs:340000,1260.96768571212)
--(axis cs:350000,1263.36768571212)
--(axis cs:360000,1251.36768571212)
--(axis cs:370000,1257.36768571212)
--(axis cs:380000,1260.46768571212)
--(axis cs:390000,1265.36768571212)
--(axis cs:390000,1390.63231428788)
--(axis cs:390000,1390.63231428788)
--(axis cs:380000,1385.73231428788)
--(axis cs:370000,1382.63231428788)
--(axis cs:360000,1376.63231428788)
--(axis cs:350000,1388.63231428788)
--(axis cs:340000,1386.23231428788)
--(axis cs:330000,1384.43231428788)
--(axis cs:320000,1375.43231428788)
--(axis cs:310000,1380.33231428788)
--(axis cs:300000,1382.83231428788)
--(axis cs:290000,1383.43231428788)
--(axis cs:280000,1369.73231428788)
--(axis cs:270000,1364.13231428788)
--(axis cs:260000,1381.03231428788)
--(axis cs:250000,1377.73231428788)
--(axis cs:240000,1380.63231428788)
--(axis cs:230000,1369.63231428788)
--(axis cs:220000,1360.23231428788)
--(axis cs:210000,1372.63231428788)
--(axis cs:200000,1360.13231428788)
--(axis cs:190000,1373.13231428788)
--(axis cs:180000,1368.13231428788)
--(axis cs:170000,1364.93231428788)
--(axis cs:160000,1351.23231428788)
--(axis cs:150000,1334.63231428788)
--(axis cs:140000,1317.53231428788)
--(axis cs:130000,1304.63231428788)
--(axis cs:120000,1312.43231428788)
--(axis cs:110000,1280.93231428788)
--(axis cs:100000,1269.53231428788)
--(axis cs:90000,1220.73231428788)
--(axis cs:80000,1194.03231428788)
--(axis cs:70000,1166.93231428788)
--(axis cs:60000,1120.33231428788)
--(axis cs:50000,1044.53231428788)
--(axis cs:40000,1026.93231428788)
--(axis cs:30000,975.932314287875)
--(axis cs:20000,976.832314287875)
--(axis cs:10000,850.932314287875)
--(axis cs:0,765.832314287875)
--(axis cs:-10000,71.7323142878752)
--cycle;

\path [draw=orchid204120188, fill=orchid204120188, opacity=0.2]
(axis cs:-10000,121.450848931073)
--(axis cs:-10000,-96.6508489310732)
--(axis cs:0,170.249151068927)
--(axis cs:10000,233.649151068927)
--(axis cs:20000,369.049151068927)
--(axis cs:30000,384.649151068927)
--(axis cs:40000,421.349151068927)
--(axis cs:50000,623.549151068927)
--(axis cs:60000,668.049151068927)
--(axis cs:70000,799.049151068927)
--(axis cs:80000,881.349151068927)
--(axis cs:90000,1075.24915106893)
--(axis cs:100000,1157.84915106893)
--(axis cs:110000,1165.04915106893)
--(axis cs:120000,1190.54915106893)
--(axis cs:130000,1197.94915106893)
--(axis cs:140000,1203.94915106893)
--(axis cs:150000,1200.24915106893)
--(axis cs:160000,1208.34915106893)
--(axis cs:170000,1187.64915106893)
--(axis cs:180000,1203.94915106893)
--(axis cs:190000,1210.34915106893)
--(axis cs:200000,1205.94915106893)
--(axis cs:210000,1196.64915106893)
--(axis cs:220000,1186.04915106893)
--(axis cs:230000,1188.84915106893)
--(axis cs:240000,1197.54915106893)
--(axis cs:250000,1192.04915106893)
--(axis cs:260000,1193.64915106893)
--(axis cs:270000,1197.04915106893)
--(axis cs:280000,1185.54915106893)
--(axis cs:290000,1195.84915106893)
--(axis cs:300000,1195.84915106893)
--(axis cs:310000,1202.94915106893)
--(axis cs:320000,1206.04915106893)
--(axis cs:330000,1197.84915106893)
--(axis cs:340000,1203.64915106893)
--(axis cs:350000,1202.94915106893)
--(axis cs:360000,1206.24915106893)
--(axis cs:370000,1210.34915106893)
--(axis cs:380000,1216.04915106893)
--(axis cs:390000,1215.34915106893)
--(axis cs:390000,1433.45084893107)
--(axis cs:390000,1433.45084893107)
--(axis cs:380000,1434.15084893107)
--(axis cs:370000,1428.45084893107)
--(axis cs:360000,1424.35084893107)
--(axis cs:350000,1421.05084893107)
--(axis cs:340000,1421.75084893107)
--(axis cs:330000,1415.95084893107)
--(axis cs:320000,1424.15084893107)
--(axis cs:310000,1421.05084893107)
--(axis cs:300000,1413.95084893107)
--(axis cs:290000,1413.95084893107)
--(axis cs:280000,1403.65084893107)
--(axis cs:270000,1415.15084893107)
--(axis cs:260000,1411.75084893107)
--(axis cs:250000,1410.15084893107)
--(axis cs:240000,1415.65084893107)
--(axis cs:230000,1406.95084893107)
--(axis cs:220000,1404.15084893107)
--(axis cs:210000,1414.75084893107)
--(axis cs:200000,1424.05084893107)
--(axis cs:190000,1428.45084893107)
--(axis cs:180000,1422.05084893107)
--(axis cs:170000,1405.75084893107)
--(axis cs:160000,1426.45084893107)
--(axis cs:150000,1418.35084893107)
--(axis cs:140000,1422.05084893107)
--(axis cs:130000,1416.05084893107)
--(axis cs:120000,1408.65084893107)
--(axis cs:110000,1383.15084893107)
--(axis cs:100000,1375.95084893107)
--(axis cs:90000,1293.35084893107)
--(axis cs:80000,1099.45084893107)
--(axis cs:70000,1017.15084893107)
--(axis cs:60000,886.150848931073)
--(axis cs:50000,841.650848931073)
--(axis cs:40000,639.450848931073)
--(axis cs:30000,602.750848931073)
--(axis cs:20000,587.150848931073)
--(axis cs:10000,451.750848931073)
--(axis cs:0,388.350848931073)
--(axis cs:-10000,121.450848931073)
--cycle;

\path [draw=peru20214597, fill=peru20214597, opacity=0.2]
(axis cs:-10000,117.398615770727)
--(axis cs:-10000,-93.9986157707267)
--(axis cs:0,-2.79861577072666)
--(axis cs:10000,297.901384229273)
--(axis cs:20000,334.901384229273)
--(axis cs:30000,318.401384229273)
--(axis cs:40000,319.201384229273)
--(axis cs:50000,330.401384229273)
--(axis cs:60000,342.201384229273)
--(axis cs:70000,346.701384229273)
--(axis cs:80000,349.901384229273)
--(axis cs:90000,370.001384229273)
--(axis cs:100000,374.601384229273)
--(axis cs:110000,386.301384229273)
--(axis cs:120000,374.701384229273)
--(axis cs:130000,387.201384229273)
--(axis cs:140000,395.301384229273)
--(axis cs:150000,393.901384229273)
--(axis cs:160000,406.301384229273)
--(axis cs:170000,401.701384229273)
--(axis cs:180000,416.601384229273)
--(axis cs:190000,412.301384229273)
--(axis cs:200000,396.001384229273)
--(axis cs:210000,404.401384229273)
--(axis cs:220000,403.901384229273)
--(axis cs:230000,411.601384229273)
--(axis cs:240000,401.401384229273)
--(axis cs:250000,391.401384229273)
--(axis cs:260000,417.001384229273)
--(axis cs:270000,414.601384229273)
--(axis cs:280000,418.001384229273)
--(axis cs:290000,426.001384229273)
--(axis cs:300000,436.401384229273)
--(axis cs:310000,435.701384229273)
--(axis cs:320000,464.701384229273)
--(axis cs:330000,483.701384229273)
--(axis cs:340000,477.401384229273)
--(axis cs:350000,472.601384229273)
--(axis cs:360000,475.201384229273)
--(axis cs:370000,486.301384229273)
--(axis cs:380000,492.701384229273)
--(axis cs:390000,489.401384229273)
--(axis cs:390000,700.798615770727)
--(axis cs:390000,700.798615770727)
--(axis cs:380000,704.098615770727)
--(axis cs:370000,697.698615770727)
--(axis cs:360000,686.598615770727)
--(axis cs:350000,683.998615770727)
--(axis cs:340000,688.798615770727)
--(axis cs:330000,695.098615770727)
--(axis cs:320000,676.098615770727)
--(axis cs:310000,647.098615770727)
--(axis cs:300000,647.798615770727)
--(axis cs:290000,637.398615770727)
--(axis cs:280000,629.398615770727)
--(axis cs:270000,625.998615770727)
--(axis cs:260000,628.398615770727)
--(axis cs:250000,602.798615770727)
--(axis cs:240000,612.798615770727)
--(axis cs:230000,622.998615770727)
--(axis cs:220000,615.298615770727)
--(axis cs:210000,615.798615770727)
--(axis cs:200000,607.398615770727)
--(axis cs:190000,623.698615770727)
--(axis cs:180000,627.998615770727)
--(axis cs:170000,613.098615770727)
--(axis cs:160000,617.698615770727)
--(axis cs:150000,605.298615770727)
--(axis cs:140000,606.698615770727)
--(axis cs:130000,598.598615770727)
--(axis cs:120000,586.098615770727)
--(axis cs:110000,597.698615770727)
--(axis cs:100000,585.998615770727)
--(axis cs:90000,581.398615770727)
--(axis cs:80000,561.298615770727)
--(axis cs:70000,558.098615770727)
--(axis cs:60000,553.598615770727)
--(axis cs:50000,541.798615770727)
--(axis cs:40000,530.598615770727)
--(axis cs:30000,529.798615770727)
--(axis cs:20000,546.298615770727)
--(axis cs:10000,509.298615770727)
--(axis cs:0,208.598615770727)
--(axis cs:-10000,117.398615770727)
--cycle;

\addplot [semithick, darkcyan1115178, mark=square*, mark size=1.5, mark options={solid}]
table {%
-10000 11.3
0 581.3
10000 581.7
20000 621.4
30000 779.6
40000 768.7
50000 889
60000 981.3
70000 1020.7
80000 1172.7
90000 1230.6
100000 1248
110000 1267.3
120000 1310.7
130000 1304.3
140000 1298.4
150000 1304
160000 1297.6
170000 1294
180000 1316.4
190000 1304.4
200000 1318.1
210000 1290.6
220000 1314.4
230000 1319.6
240000 1331.6
250000 1330.1
260000 1327.1
270000 1328
280000 1329
290000 1335.1
300000 1335.4
310000 1333.9
320000 1334
330000 1339
340000 1340
350000 1341.6
360000 1337
370000 1343.3
380000 1329.1
390000 1346
};
\addplot [semithick, darkorange2221435, mark=triangle*, mark size=1.5, mark options={solid}]
table {%
-10000 9.1
0 703.2
10000 788.3
20000 914.2
30000 913.3
40000 964.3
50000 981.9
60000 1057.7
70000 1104.3
80000 1131.4
90000 1158.1
100000 1206.9
110000 1218.3
120000 1249.8
130000 1242
140000 1254.9
150000 1272
160000 1288.6
170000 1302.3
180000 1305.5
190000 1310.5
200000 1297.5
210000 1310
220000 1297.6
230000 1307
240000 1318
250000 1315.1
260000 1318.4
270000 1301.5
280000 1307.1
290000 1320.8
300000 1320.2
310000 1317.7
320000 1312.8
330000 1321.8
340000 1323.6
350000 1326
360000 1314
370000 1320
380000 1323.1
390000 1328
};
\addplot [semithick, orchid204120188, mark=+, mark size=1.5, mark options={solid}]
table {%
-10000 12.4
0 279.3
10000 342.7
20000 478.1
30000 493.7
40000 530.4
50000 732.6
60000 777.1
70000 908.1
80000 990.4
90000 1184.3
100000 1266.9
110000 1274.1
120000 1299.6
130000 1307
140000 1313
150000 1309.3
160000 1317.4
170000 1296.7
180000 1313
190000 1319.4
200000 1315
210000 1305.7
220000 1295.1
230000 1297.9
240000 1306.6
250000 1301.1
260000 1302.7
270000 1306.1
280000 1294.6
290000 1304.9
300000 1304.9
310000 1312
320000 1315.1
330000 1306.9
340000 1312.7
350000 1312
360000 1315.3
370000 1319.4
380000 1325.1
390000 1324.4
};
\addplot [semithick, peru20214597, mark=diamond*, mark size=1.5, mark options={solid}]
table {%
-10000 11.7
0 102.9
10000 403.6
20000 440.6
30000 424.1
40000 424.9
50000 436.1
60000 447.9
70000 452.4
80000 455.6
90000 475.7
100000 480.3
110000 492
120000 480.4
130000 492.9
140000 501
150000 499.6
160000 512
170000 507.4
180000 522.3
190000 518
200000 501.7
210000 510.1
220000 509.6
230000 517.3
240000 507.1
250000 497.1
260000 522.7
270000 520.3
280000 523.7
290000 531.7
300000 542.1
310000 541.4
320000 570.4
330000 589.4
340000 583.1
350000 578.3
360000 580.9
370000 592
380000 598.4
390000 595.1
};
\end{axis}

\end{tikzpicture}

%% file: Image/battle_zone.tex
\begin{tikzpicture}

\definecolor{darkcyan1115178}{RGB}{1,115,178}
\definecolor{darkorange2221435}{RGB}{222,143,5}
\definecolor{darkslategray38}{RGB}{38,38,38}
\definecolor{lavender234234242}{RGB}{234,234,242}
\definecolor{lightgray204}{RGB}{204,204,204}
\definecolor{orchid204120188}{RGB}{204,120,188}
\definecolor{peru20214597}{RGB}{202,145,97}

\begin{axis}[
axis background/.style={fill=lavender234234242},
axis line style={white},
legend cell align={left},
legend style={
  fill opacity=0.8,
  draw opacity=1,
  text opacity=1,
  at={(0.03,0.97)},
  anchor=north west,
  draw=lightgray204,
  fill=lavender234234242
},
tick align=outside,
x grid style={white},
xlabel=\textcolor{darkslategray38}{Time Step},
xmajorgrids,
xmajorticks=true,
xmin=-30000, xmax=410000,
xtick style={color=darkslategray38},
xtick={0,50000,100000,150000,200000,250000,300000,350000,400000},
xticklabels={0k,50k,100k,150k,200k,250k,300k,350k,400k},
y grid style={white},
ylabel=\textcolor{darkslategray38}{Performance},
ymajorgrids,
ymajorticks=true,
ymin=-3909.27036733611, ymax=22823.5703673361,
ytick style={color=darkslategray38}
]
\path [draw=darkcyan1115178, fill=darkcyan1115178, opacity=0.2]
(axis cs:-10000,6337.04124303283)
--(axis cs:-10000,-79.8412430328299)
--(axis cs:0,-2694.14124303283)
--(axis cs:10000,-1937.04124303283)
--(axis cs:20000,-265.54124303283)
--(axis cs:30000,-565.54124303283)
--(axis cs:40000,420.15875696717)
--(axis cs:50000,-351.34124303283)
--(axis cs:60000,1805.85875696717)
--(axis cs:70000,6162.95875696717)
--(axis cs:80000,6048.65875696717)
--(axis cs:90000,6162.95875696717)
--(axis cs:100000,5977.25875696717)
--(axis cs:110000,8577.25875696717)
--(axis cs:120000,9905.85875696717)
--(axis cs:130000,9434.45875696717)
--(axis cs:140000,12548.6587569672)
--(axis cs:150000,11962.9587569672)
--(axis cs:160000,12505.8587569672)
--(axis cs:170000,13605.8587569672)
--(axis cs:180000,14462.9587569672)
--(axis cs:190000,13905.8587569672)
--(axis cs:200000,13405.8587569672)
--(axis cs:210000,14905.8587569672)
--(axis cs:220000,13505.8587569672)
--(axis cs:230000,13277.2587569672)
--(axis cs:240000,14891.5587569672)
--(axis cs:250000,14662.9587569672)
--(axis cs:260000,15105.8587569672)
--(axis cs:270000,15191.5587569672)
--(axis cs:280000,13762.9587569672)
--(axis cs:290000,13777.2587569672)
--(axis cs:300000,13334.4587569672)
--(axis cs:310000,13262.9587569672)
--(axis cs:320000,13648.6587569672)
--(axis cs:330000,14162.9587569672)
--(axis cs:340000,13848.6587569672)
--(axis cs:350000,14934.4587569672)
--(axis cs:360000,13991.5587569672)
--(axis cs:370000,14605.8587569672)
--(axis cs:380000,14520.1587569672)
--(axis cs:390000,15005.8587569672)
--(axis cs:390000,21422.7412430328)
--(axis cs:390000,21422.7412430328)
--(axis cs:380000,20937.0412430328)
--(axis cs:370000,21022.7412430328)
--(axis cs:360000,20408.4412430328)
--(axis cs:350000,21351.3412430328)
--(axis cs:340000,20265.5412430328)
--(axis cs:330000,20579.8412430328)
--(axis cs:320000,20065.5412430328)
--(axis cs:310000,19679.8412430328)
--(axis cs:300000,19751.3412430328)
--(axis cs:290000,20194.1412430328)
--(axis cs:280000,20179.8412430328)
--(axis cs:270000,21608.4412430328)
--(axis cs:260000,21522.7412430328)
--(axis cs:250000,21079.8412430328)
--(axis cs:240000,21308.4412430328)
--(axis cs:230000,19694.1412430328)
--(axis cs:220000,19922.7412430328)
--(axis cs:210000,21322.7412430328)
--(axis cs:200000,19822.7412430328)
--(axis cs:190000,20322.7412430328)
--(axis cs:180000,20879.8412430328)
--(axis cs:170000,20022.7412430328)
--(axis cs:160000,18922.7412430328)
--(axis cs:150000,18379.8412430328)
--(axis cs:140000,18965.5412430328)
--(axis cs:130000,15851.3412430328)
--(axis cs:120000,16322.7412430328)
--(axis cs:110000,14994.1412430328)
--(axis cs:100000,12394.1412430328)
--(axis cs:90000,12579.8412430328)
--(axis cs:80000,12465.5412430328)
--(axis cs:70000,12579.8412430328)
--(axis cs:60000,8222.74124303283)
--(axis cs:50000,6065.54124303283)
--(axis cs:40000,6837.04124303283)
--(axis cs:30000,5851.34124303283)
--(axis cs:20000,6151.34124303283)
--(axis cs:10000,4479.84124303283)
--(axis cs:0,3722.74124303283)
--(axis cs:-10000,6337.04124303283)
--cycle;

\path [draw=darkorange2221435, fill=darkorange2221435, opacity=0.2]
(axis cs:-10000,4735.59565022582)
--(axis cs:-10000,404.404349774183)
--(axis cs:0,434.404349774183)
--(axis cs:10000,1444.40434977418)
--(axis cs:20000,884.404349774183)
--(axis cs:30000,1384.40434977418)
--(axis cs:40000,1274.40434977418)
--(axis cs:50000,1014.40434977418)
--(axis cs:60000,1154.40434977418)
--(axis cs:70000,1014.40434977418)
--(axis cs:80000,894.404349774183)
--(axis cs:90000,3184.40434977418)
--(axis cs:100000,3044.40434977418)
--(axis cs:110000,2994.40434977418)
--(axis cs:120000,3484.40434977418)
--(axis cs:130000,5524.40434977418)
--(axis cs:140000,5844.40434977418)
--(axis cs:150000,5354.40434977418)
--(axis cs:160000,6204.40434977418)
--(axis cs:170000,6164.40434977418)
--(axis cs:180000,5944.40434977418)
--(axis cs:190000,7234.40434977418)
--(axis cs:200000,7484.40434977418)
--(axis cs:210000,7654.40434977418)
--(axis cs:220000,6454.40434977418)
--(axis cs:230000,7294.40434977418)
--(axis cs:240000,7064.40434977418)
--(axis cs:250000,7244.40434977418)
--(axis cs:260000,7734.40434977418)
--(axis cs:270000,8794.40434977418)
--(axis cs:280000,10554.4043497742)
--(axis cs:290000,10544.4043497742)
--(axis cs:300000,10444.4043497742)
--(axis cs:310000,10574.4043497742)
--(axis cs:320000,10244.4043497742)
--(axis cs:330000,10254.4043497742)
--(axis cs:340000,9914.40434977418)
--(axis cs:350000,11624.4043497742)
--(axis cs:360000,12754.4043497742)
--(axis cs:370000,11834.4043497742)
--(axis cs:380000,11844.4043497742)
--(axis cs:390000,12134.4043497742)
--(axis cs:390000,16465.5956502258)
--(axis cs:390000,16465.5956502258)
--(axis cs:380000,16175.5956502258)
--(axis cs:370000,16165.5956502258)
--(axis cs:360000,17085.5956502258)
--(axis cs:350000,15955.5956502258)
--(axis cs:340000,14245.5956502258)
--(axis cs:330000,14585.5956502258)
--(axis cs:320000,14575.5956502258)
--(axis cs:310000,14905.5956502258)
--(axis cs:300000,14775.5956502258)
--(axis cs:290000,14875.5956502258)
--(axis cs:280000,14885.5956502258)
--(axis cs:270000,13125.5956502258)
--(axis cs:260000,12065.5956502258)
--(axis cs:250000,11575.5956502258)
--(axis cs:240000,11395.5956502258)
--(axis cs:230000,11625.5956502258)
--(axis cs:220000,10785.5956502258)
--(axis cs:210000,11985.5956502258)
--(axis cs:200000,11815.5956502258)
--(axis cs:190000,11565.5956502258)
--(axis cs:180000,10275.5956502258)
--(axis cs:170000,10495.5956502258)
--(axis cs:160000,10535.5956502258)
--(axis cs:150000,9685.59565022582)
--(axis cs:140000,10175.5956502258)
--(axis cs:130000,9855.59565022582)
--(axis cs:120000,7815.59565022582)
--(axis cs:110000,7325.59565022582)
--(axis cs:100000,7375.59565022582)
--(axis cs:90000,7515.59565022582)
--(axis cs:80000,5225.59565022582)
--(axis cs:70000,5345.59565022582)
--(axis cs:60000,5485.59565022582)
--(axis cs:50000,5345.59565022582)
--(axis cs:40000,5605.59565022582)
--(axis cs:30000,5715.59565022582)
--(axis cs:20000,5215.59565022582)
--(axis cs:10000,5775.59565022582)
--(axis cs:0,4765.59565022582)
--(axis cs:-10000,4735.59565022582)
--cycle;

\path [draw=orchid204120188, fill=orchid204120188, opacity=0.2]
(axis cs:-10000,4660.00862897913)
--(axis cs:-10000,-888.608628979133)
--(axis cs:0,-1702.90862897913)
--(axis cs:10000,-1188.60862897913)
--(axis cs:20000,-517.208628979133)
--(axis cs:30000,368.591371020867)
--(axis cs:40000,839.991371020868)
--(axis cs:50000,2268.59137102087)
--(axis cs:60000,2925.69137102087)
--(axis cs:70000,2739.99137102087)
--(axis cs:80000,2825.69137102087)
--(axis cs:90000,2939.99137102087)
--(axis cs:100000,3839.99137102087)
--(axis cs:110000,4011.39137102087)
--(axis cs:120000,3311.39137102087)
--(axis cs:130000,3639.99137102087)
--(axis cs:140000,3325.69137102087)
--(axis cs:150000,6182.79137102087)
--(axis cs:160000,6982.79137102087)
--(axis cs:170000,7225.69137102087)
--(axis cs:180000,6639.99137102087)
--(axis cs:190000,9939.99137102087)
--(axis cs:200000,9925.69137102087)
--(axis cs:210000,9739.99137102087)
--(axis cs:220000,10339.9913710209)
--(axis cs:230000,10182.7913710209)
--(axis cs:240000,9639.99137102087)
--(axis cs:250000,8225.69137102087)
--(axis cs:260000,8368.59137102087)
--(axis cs:270000,7625.69137102087)
--(axis cs:280000,8839.99137102087)
--(axis cs:290000,8682.79137102087)
--(axis cs:300000,8282.79137102087)
--(axis cs:310000,7639.99137102087)
--(axis cs:320000,8011.39137102087)
--(axis cs:330000,9411.39137102087)
--(axis cs:340000,11039.9913710209)
--(axis cs:350000,11982.7913710209)
--(axis cs:360000,12539.9913710209)
--(axis cs:370000,12325.6913710209)
--(axis cs:380000,12382.7913710209)
--(axis cs:390000,12868.5913710209)
--(axis cs:390000,18417.2086289791)
--(axis cs:390000,18417.2086289791)
--(axis cs:380000,17931.4086289791)
--(axis cs:370000,17874.3086289791)
--(axis cs:360000,18088.6086289791)
--(axis cs:350000,17531.4086289791)
--(axis cs:340000,16588.6086289791)
--(axis cs:330000,14960.0086289791)
--(axis cs:320000,13560.0086289791)
--(axis cs:310000,13188.6086289791)
--(axis cs:300000,13831.4086289791)
--(axis cs:290000,14231.4086289791)
--(axis cs:280000,14388.6086289791)
--(axis cs:270000,13174.3086289791)
--(axis cs:260000,13917.2086289791)
--(axis cs:250000,13774.3086289791)
--(axis cs:240000,15188.6086289791)
--(axis cs:230000,15731.4086289791)
--(axis cs:220000,15888.6086289791)
--(axis cs:210000,15288.6086289791)
--(axis cs:200000,15474.3086289791)
--(axis cs:190000,15488.6086289791)
--(axis cs:180000,12188.6086289791)
--(axis cs:170000,12774.3086289791)
--(axis cs:160000,12531.4086289791)
--(axis cs:150000,11731.4086289791)
--(axis cs:140000,8874.30862897913)
--(axis cs:130000,9188.60862897913)
--(axis cs:120000,8860.00862897913)
--(axis cs:110000,9560.00862897913)
--(axis cs:100000,9388.60862897913)
--(axis cs:90000,8488.60862897913)
--(axis cs:80000,8374.30862897913)
--(axis cs:70000,8288.60862897913)
--(axis cs:60000,8474.30862897913)
--(axis cs:50000,7817.20862897913)
--(axis cs:40000,6388.60862897913)
--(axis cs:30000,5917.20862897913)
--(axis cs:20000,5031.40862897913)
--(axis cs:10000,4360.00862897913)
--(axis cs:0,3845.70862897913)
--(axis cs:-10000,4660.00862897913)
--cycle;

\path [draw=peru20214597, fill=peru20214597, opacity=0.2]
(axis cs:-10000,4520.70331285505)
--(axis cs:-10000,165.096687144945)
--(axis cs:0,-1734.90331285505)
--(axis cs:10000,-1420.70331285505)
--(axis cs:20000,1422.19668714495)
--(axis cs:30000,2065.09668714495)
--(axis cs:40000,3407.89668714495)
--(axis cs:50000,3036.49668714495)
--(axis cs:60000,3079.29668714495)
--(axis cs:70000,5693.59668714495)
--(axis cs:80000,5893.59668714495)
--(axis cs:90000,7407.89668714495)
--(axis cs:100000,5879.29668714495)
--(axis cs:110000,5750.79668714495)
--(axis cs:120000,6722.19668714495)
--(axis cs:130000,8993.59668714495)
--(axis cs:140000,8965.09668714495)
--(axis cs:150000,9179.29668714495)
--(axis cs:160000,7922.19668714495)
--(axis cs:170000,8450.79668714495)
--(axis cs:180000,9065.09668714495)
--(axis cs:190000,9650.79668714495)
--(axis cs:200000,9407.89668714495)
--(axis cs:210000,12165.0966871449)
--(axis cs:220000,10450.7966871449)
--(axis cs:230000,10836.4966871449)
--(axis cs:240000,10950.7966871449)
--(axis cs:250000,11522.1966871449)
--(axis cs:260000,11793.5966871449)
--(axis cs:270000,13150.7966871449)
--(axis cs:280000,13707.8966871449)
--(axis cs:290000,11936.4966871449)
--(axis cs:300000,13950.7966871449)
--(axis cs:310000,13665.0966871449)
--(axis cs:320000,14650.7966871449)
--(axis cs:330000,12736.4966871449)
--(axis cs:340000,14707.8966871449)
--(axis cs:350000,14650.7966871449)
--(axis cs:360000,13550.7966871449)
--(axis cs:370000,15750.7966871449)
--(axis cs:380000,14565.0966871449)
--(axis cs:390000,14807.8966871449)
--(axis cs:390000,19163.5033128551)
--(axis cs:390000,19163.5033128551)
--(axis cs:380000,18920.7033128551)
--(axis cs:370000,20106.4033128551)
--(axis cs:360000,17906.4033128551)
--(axis cs:350000,19006.4033128551)
--(axis cs:340000,19063.5033128551)
--(axis cs:330000,17092.1033128551)
--(axis cs:320000,19006.4033128551)
--(axis cs:310000,18020.7033128551)
--(axis cs:300000,18306.4033128551)
--(axis cs:290000,16292.1033128551)
--(axis cs:280000,18063.5033128551)
--(axis cs:270000,17506.4033128551)
--(axis cs:260000,16149.2033128551)
--(axis cs:250000,15877.8033128551)
--(axis cs:240000,15306.4033128551)
--(axis cs:230000,15192.1033128551)
--(axis cs:220000,14806.4033128551)
--(axis cs:210000,16520.7033128551)
--(axis cs:200000,13763.5033128551)
--(axis cs:190000,14006.4033128551)
--(axis cs:180000,13420.7033128551)
--(axis cs:170000,12806.4033128551)
--(axis cs:160000,12277.8033128551)
--(axis cs:150000,13534.9033128551)
--(axis cs:140000,13320.7033128551)
--(axis cs:130000,13349.2033128551)
--(axis cs:120000,11077.8033128551)
--(axis cs:110000,10106.4033128551)
--(axis cs:100000,10234.9033128551)
--(axis cs:90000,11763.5033128551)
--(axis cs:80000,10249.2033128551)
--(axis cs:70000,10049.2033128551)
--(axis cs:60000,7434.90331285505)
--(axis cs:50000,7392.10331285506)
--(axis cs:40000,7763.50331285505)
--(axis cs:30000,6420.70331285505)
--(axis cs:20000,5777.80331285505)
--(axis cs:10000,2934.90331285505)
--(axis cs:0,2620.70331285505)
--(axis cs:-10000,4520.70331285505)
--cycle;

\addplot [semithick, darkcyan1115178, mark=square*, mark size=1.5, mark options={solid}]
table {%
-10000 3128.6
0 514.3
10000 1271.4
20000 2942.9
30000 2642.9
40000 3628.6
50000 2857.1
60000 5014.3
70000 9371.4
80000 9257.1
90000 9371.4
100000 9185.7
110000 11785.7
120000 13114.3
130000 12642.9
140000 15757.1
150000 15171.4
160000 15714.3
170000 16814.3
180000 17671.4
190000 17114.3
200000 16614.3
210000 18114.3
220000 16714.3
230000 16485.7
240000 18100
250000 17871.4
260000 18314.3
270000 18400
280000 16971.4
290000 16985.7
300000 16542.9
310000 16471.4
320000 16857.1
330000 17371.4
340000 17057.1
350000 18142.9
360000 17200
370000 17814.3
380000 17728.6
390000 18214.3
};
\addplot [semithick, darkorange2221435, mark=triangle*, mark size=1.5, mark options={solid}]
table {%
-10000 2570
0 2600
10000 3610
20000 3050
30000 3550
40000 3440
50000 3180
60000 3320
70000 3180
80000 3060
90000 5350
100000 5210
110000 5160
120000 5650
130000 7690
140000 8010
150000 7520
160000 8370
170000 8330
180000 8110
190000 9400
200000 9650
210000 9820
220000 8620
230000 9460
240000 9230
250000 9410
260000 9900
270000 10960
280000 12720
290000 12710
300000 12610
310000 12740
320000 12410
330000 12420
340000 12080
350000 13790
360000 14920
370000 14000
380000 14010
390000 14300
};
\addplot [semithick, orchid204120188, mark=+, mark size=1.5, mark options={solid}]
table {%
-10000 1885.7
0 1071.4
10000 1585.7
20000 2257.1
30000 3142.9
40000 3614.3
50000 5042.9
60000 5700
70000 5514.3
80000 5600
90000 5714.3
100000 6614.3
110000 6785.7
120000 6085.7
130000 6414.3
140000 6100
150000 8957.1
160000 9757.1
170000 10000
180000 9414.3
190000 12714.3
200000 12700
210000 12514.3
220000 13114.3
230000 12957.1
240000 12414.3
250000 11000
260000 11142.9
270000 10400
280000 11614.3
290000 11457.1
300000 11057.1
310000 10414.3
320000 10785.7
330000 12185.7
340000 13814.3
350000 14757.1
360000 15314.3
370000 15100
380000 15157.1
390000 15642.9
};
\addplot [semithick, peru20214597, mark=diamond*, mark size=1.5, mark options={solid}]
table {%
-10000 2342.9
0 442.9
10000 757.1
20000 3600
30000 4242.9
40000 5585.7
50000 5214.3
60000 5257.1
70000 7871.4
80000 8071.4
90000 9585.7
100000 8057.1
110000 7928.6
120000 8900
130000 11171.4
140000 11142.9
150000 11357.1
160000 10100
170000 10628.6
180000 11242.9
190000 11828.6
200000 11585.7
210000 14342.9
220000 12628.6
230000 13014.3
240000 13128.6
250000 13700
260000 13971.4
270000 15328.6
280000 15885.7
290000 14114.3
300000 16128.6
310000 15842.9
320000 16828.6
330000 14914.3
340000 16885.7
350000 16828.6
360000 15728.6
370000 17928.6
380000 16742.9
390000 16985.7
};
\end{axis}

\end{tikzpicture}

%% file: Image/boxing.tex
\begin{tikzpicture}

\definecolor{darkcyan1115178}{RGB}{1,115,178}
\definecolor{darkorange2221435}{RGB}{222,143,5}
\definecolor{darkslategray38}{RGB}{38,38,38}
\definecolor{lavender234234242}{RGB}{234,234,242}
\definecolor{lightgray204}{RGB}{204,204,204}
\definecolor{orchid204120188}{RGB}{204,120,188}
\definecolor{peru20214597}{RGB}{202,145,97}

\begin{axis}[
axis background/.style={fill=lavender234234242},
axis line style={white},
legend cell align={left},
legend style={
  fill opacity=0.8,
  draw opacity=1,
  text opacity=1,
  at={(0.97,0.03)},
  anchor=south east,
  draw=lightgray204,
  fill=lavender234234242
},
tick align=outside,
x grid style={white},
xlabel=\textcolor{darkslategray38}{Time Step},
xmajorgrids,
xmajorticks=true,
xmin=-30000, xmax=410000,
xtick style={color=darkslategray38},
xtick={0,50000,100000,150000,200000,250000,300000,350000,400000},
xticklabels={0k,50k,100k,150k,200k,250k,300k,350k,400k},
y grid style={white},
ylabel=\textcolor{darkslategray38}{Performance},
ymajorgrids,
ymajorticks=true,
ymin=-19.4177011089159, ymax=110.280038933334,
ytick style={color=darkslategray38}
]
\path [draw=darkcyan1115178, fill=darkcyan1115178, opacity=0.2]
(axis cs:-10000,2.18468711323187)
--(axis cs:-10000,-6.98468711323187)
--(axis cs:0,71.6153128867681)
--(axis cs:10000,91.8153128867681)
--(axis cs:20000,93.7153128867681)
--(axis cs:30000,94.5153128867681)
--(axis cs:40000,94.8153128867681)
--(axis cs:50000,94.9153128867681)
--(axis cs:60000,94.2153128867681)
--(axis cs:70000,94.3153128867681)
--(axis cs:80000,94.2153128867681)
--(axis cs:90000,94.4153128867681)
--(axis cs:100000,94.8153128867681)
--(axis cs:110000,94.8153128867681)
--(axis cs:120000,94.7153128867681)
--(axis cs:130000,94.7153128867681)
--(axis cs:140000,94.9153128867681)
--(axis cs:150000,94.7153128867681)
--(axis cs:160000,94.8153128867681)
--(axis cs:170000,94.8153128867681)
--(axis cs:180000,95.1153128867681)
--(axis cs:190000,95.1153128867681)
--(axis cs:200000,94.9153128867681)
--(axis cs:210000,95.2153128867681)
--(axis cs:220000,94.7153128867681)
--(axis cs:230000,94.9153128867681)
--(axis cs:240000,94.9153128867681)
--(axis cs:250000,94.7153128867681)
--(axis cs:260000,94.9153128867681)
--(axis cs:270000,94.4153128867681)
--(axis cs:280000,94.7153128867681)
--(axis cs:290000,94.9153128867681)
--(axis cs:300000,94.9153128867681)
--(axis cs:310000,94.9153128867681)
--(axis cs:320000,94.8153128867681)
--(axis cs:330000,95.0153128867681)
--(axis cs:340000,94.6153128867681)
--(axis cs:350000,95.0153128867681)
--(axis cs:360000,94.0153128867681)
--(axis cs:370000,95.0153128867681)
--(axis cs:380000,95.1153128867681)
--(axis cs:390000,94.8153128867681)
--(axis cs:390000,103.984687113232)
--(axis cs:390000,103.984687113232)
--(axis cs:380000,104.284687113232)
--(axis cs:370000,104.184687113232)
--(axis cs:360000,103.184687113232)
--(axis cs:350000,104.184687113232)
--(axis cs:340000,103.784687113232)
--(axis cs:330000,104.184687113232)
--(axis cs:320000,103.984687113232)
--(axis cs:310000,104.084687113232)
--(axis cs:300000,104.084687113232)
--(axis cs:290000,104.084687113232)
--(axis cs:280000,103.884687113232)
--(axis cs:270000,103.584687113232)
--(axis cs:260000,104.084687113232)
--(axis cs:250000,103.884687113232)
--(axis cs:240000,104.084687113232)
--(axis cs:230000,104.084687113232)
--(axis cs:220000,103.884687113232)
--(axis cs:210000,104.384687113232)
--(axis cs:200000,104.084687113232)
--(axis cs:190000,104.284687113232)
--(axis cs:180000,104.284687113232)
--(axis cs:170000,103.984687113232)
--(axis cs:160000,103.984687113232)
--(axis cs:150000,103.884687113232)
--(axis cs:140000,104.084687113232)
--(axis cs:130000,103.884687113232)
--(axis cs:120000,103.884687113232)
--(axis cs:110000,103.984687113232)
--(axis cs:100000,103.984687113232)
--(axis cs:90000,103.584687113232)
--(axis cs:80000,103.384687113232)
--(axis cs:70000,103.484687113232)
--(axis cs:60000,103.384687113232)
--(axis cs:50000,104.084687113232)
--(axis cs:40000,103.984687113232)
--(axis cs:30000,103.684687113232)
--(axis cs:20000,102.884687113232)
--(axis cs:10000,100.984687113232)
--(axis cs:0,80.7846871132319)
--(axis cs:-10000,2.18468711323187)
--cycle;

\path [draw=darkorange2221435, fill=darkorange2221435, opacity=0.2]
(axis cs:-10000,-1.34173904381332)
--(axis cs:-10000,-7.65826095618668)
--(axis cs:0,88.5417390438133)
--(axis cs:10000,95.0417390438133)
--(axis cs:20000,95.8417390438133)
--(axis cs:30000,96.0417390438133)
--(axis cs:40000,95.8417390438133)
--(axis cs:50000,96.3417390438133)
--(axis cs:60000,96.0417390438133)
--(axis cs:70000,96.5417390438133)
--(axis cs:80000,96.4417390438133)
--(axis cs:90000,96.4417390438133)
--(axis cs:100000,96.3417390438133)
--(axis cs:110000,96.2417390438133)
--(axis cs:120000,96.4417390438133)
--(axis cs:130000,96.6417390438133)
--(axis cs:140000,96.4417390438133)
--(axis cs:150000,96.5417390438133)
--(axis cs:160000,96.1417390438133)
--(axis cs:170000,96.6417390438133)
--(axis cs:180000,96.3417390438133)
--(axis cs:190000,96.3417390438133)
--(axis cs:200000,96.4417390438133)
--(axis cs:210000,96.5417390438133)
--(axis cs:220000,96.5417390438133)
--(axis cs:230000,96.5417390438133)
--(axis cs:240000,96.6417390438133)
--(axis cs:250000,96.5417390438133)
--(axis cs:260000,96.4417390438133)
--(axis cs:270000,96.5417390438133)
--(axis cs:280000,96.5417390438133)
--(axis cs:290000,96.5417390438133)
--(axis cs:300000,96.4417390438133)
--(axis cs:310000,96.6417390438133)
--(axis cs:320000,96.5417390438133)
--(axis cs:330000,96.7417390438133)
--(axis cs:340000,96.4417390438133)
--(axis cs:350000,96.4417390438133)
--(axis cs:360000,96.6417390438133)
--(axis cs:370000,96.3417390438133)
--(axis cs:380000,96.4417390438133)
--(axis cs:390000,96.4417390438133)
--(axis cs:390000,102.758260956187)
--(axis cs:390000,102.758260956187)
--(axis cs:380000,102.758260956187)
--(axis cs:370000,102.658260956187)
--(axis cs:360000,102.958260956187)
--(axis cs:350000,102.758260956187)
--(axis cs:340000,102.758260956187)
--(axis cs:330000,103.058260956187)
--(axis cs:320000,102.858260956187)
--(axis cs:310000,102.958260956187)
--(axis cs:300000,102.758260956187)
--(axis cs:290000,102.858260956187)
--(axis cs:280000,102.858260956187)
--(axis cs:270000,102.858260956187)
--(axis cs:260000,102.758260956187)
--(axis cs:250000,102.858260956187)
--(axis cs:240000,102.958260956187)
--(axis cs:230000,102.858260956187)
--(axis cs:220000,102.858260956187)
--(axis cs:210000,102.858260956187)
--(axis cs:200000,102.758260956187)
--(axis cs:190000,102.658260956187)
--(axis cs:180000,102.658260956187)
--(axis cs:170000,102.958260956187)
--(axis cs:160000,102.458260956187)
--(axis cs:150000,102.858260956187)
--(axis cs:140000,102.758260956187)
--(axis cs:130000,102.958260956187)
--(axis cs:120000,102.758260956187)
--(axis cs:110000,102.558260956187)
--(axis cs:100000,102.658260956187)
--(axis cs:90000,102.758260956187)
--(axis cs:80000,102.758260956187)
--(axis cs:70000,102.858260956187)
--(axis cs:60000,102.358260956187)
--(axis cs:50000,102.658260956187)
--(axis cs:40000,102.158260956187)
--(axis cs:30000,102.358260956187)
--(axis cs:20000,102.158260956187)
--(axis cs:10000,101.358260956187)
--(axis cs:0,94.8582609561867)
--(axis cs:-10000,-1.34173904381332)
--cycle;

\path [draw=orchid204120188, fill=orchid204120188, opacity=0.2]
(axis cs:-10000,1.75666935308717)
--(axis cs:-10000,-8.75666935308717)
--(axis cs:0,34.3433306469128)
--(axis cs:10000,73.9433306469128)
--(axis cs:20000,82.8433306469128)
--(axis cs:30000,85.1433306469128)
--(axis cs:40000,86.7433306469128)
--(axis cs:50000,89.7433306469128)
--(axis cs:60000,90.1433306469128)
--(axis cs:70000,90.8433306469128)
--(axis cs:80000,90.7433306469128)
--(axis cs:90000,91.4433306469128)
--(axis cs:100000,90.9433306469128)
--(axis cs:110000,92.2433306469128)
--(axis cs:120000,92.3433306469128)
--(axis cs:130000,92.7433306469128)
--(axis cs:140000,92.7433306469128)
--(axis cs:150000,92.7433306469128)
--(axis cs:160000,92.6433306469128)
--(axis cs:170000,92.6433306469128)
--(axis cs:180000,92.6433306469128)
--(axis cs:190000,92.4433306469128)
--(axis cs:200000,92.8433306469128)
--(axis cs:210000,93.5433306469128)
--(axis cs:220000,92.5433306469128)
--(axis cs:230000,93.2433306469128)
--(axis cs:240000,92.5433306469128)
--(axis cs:250000,92.6433306469128)
--(axis cs:260000,93.6433306469128)
--(axis cs:270000,92.0433306469128)
--(axis cs:280000,92.3433306469128)
--(axis cs:290000,92.6433306469128)
--(axis cs:300000,91.8433306469128)
--(axis cs:310000,92.9433306469128)
--(axis cs:320000,93.2433306469128)
--(axis cs:330000,93.7433306469128)
--(axis cs:340000,93.3433306469128)
--(axis cs:350000,92.2433306469128)
--(axis cs:360000,92.1433306469128)
--(axis cs:370000,91.0433306469128)
--(axis cs:380000,93.2433306469128)
--(axis cs:390000,93.3433306469128)
--(axis cs:390000,103.856669353087)
--(axis cs:390000,103.856669353087)
--(axis cs:380000,103.756669353087)
--(axis cs:370000,101.556669353087)
--(axis cs:360000,102.656669353087)
--(axis cs:350000,102.756669353087)
--(axis cs:340000,103.856669353087)
--(axis cs:330000,104.256669353087)
--(axis cs:320000,103.756669353087)
--(axis cs:310000,103.456669353087)
--(axis cs:300000,102.356669353087)
--(axis cs:290000,103.156669353087)
--(axis cs:280000,102.856669353087)
--(axis cs:270000,102.556669353087)
--(axis cs:260000,104.156669353087)
--(axis cs:250000,103.156669353087)
--(axis cs:240000,103.056669353087)
--(axis cs:230000,103.756669353087)
--(axis cs:220000,103.056669353087)
--(axis cs:210000,104.056669353087)
--(axis cs:200000,103.356669353087)
--(axis cs:190000,102.956669353087)
--(axis cs:180000,103.156669353087)
--(axis cs:170000,103.156669353087)
--(axis cs:160000,103.156669353087)
--(axis cs:150000,103.256669353087)
--(axis cs:140000,103.256669353087)
--(axis cs:130000,103.256669353087)
--(axis cs:120000,102.856669353087)
--(axis cs:110000,102.756669353087)
--(axis cs:100000,101.456669353087)
--(axis cs:90000,101.956669353087)
--(axis cs:80000,101.256669353087)
--(axis cs:70000,101.356669353087)
--(axis cs:60000,100.656669353087)
--(axis cs:50000,100.256669353087)
--(axis cs:40000,97.2566693530872)
--(axis cs:30000,95.6566693530872)
--(axis cs:20000,93.3566693530872)
--(axis cs:10000,84.4566693530872)
--(axis cs:0,44.8566693530872)
--(axis cs:-10000,1.75666935308717)
--cycle;

\path [draw=peru20214597, fill=peru20214597, opacity=0.2]
(axis cs:-10000,6.12234928881362)
--(axis cs:-10000,-13.5223492888136)
--(axis cs:0,-8.92234928881362)
--(axis cs:10000,15.6776507111864)
--(axis cs:20000,24.4776507111864)
--(axis cs:30000,23.9776507111864)
--(axis cs:40000,34.4776507111864)
--(axis cs:50000,38.2776507111864)
--(axis cs:60000,37.3776507111864)
--(axis cs:70000,36.2776507111864)
--(axis cs:80000,36.1776507111864)
--(axis cs:90000,36.1776507111864)
--(axis cs:100000,39.0776507111864)
--(axis cs:110000,38.4776507111864)
--(axis cs:120000,41.2776507111864)
--(axis cs:130000,45.1776507111864)
--(axis cs:140000,43.0776507111864)
--(axis cs:150000,40.8776507111864)
--(axis cs:160000,42.9776507111864)
--(axis cs:170000,43.9776507111864)
--(axis cs:180000,38.8776507111864)
--(axis cs:190000,38.5776507111864)
--(axis cs:200000,41.2776507111864)
--(axis cs:210000,42.9776507111864)
--(axis cs:220000,44.0776507111864)
--(axis cs:230000,47.8776507111864)
--(axis cs:240000,44.6776507111864)
--(axis cs:250000,45.9776507111864)
--(axis cs:260000,46.5776507111864)
--(axis cs:270000,47.4776507111864)
--(axis cs:280000,48.9776507111864)
--(axis cs:290000,46.7776507111864)
--(axis cs:300000,46.7776507111864)
--(axis cs:310000,45.9776507111864)
--(axis cs:320000,47.1776507111864)
--(axis cs:330000,47.6776507111864)
--(axis cs:340000,45.9776507111864)
--(axis cs:350000,48.0776507111864)
--(axis cs:360000,47.8776507111864)
--(axis cs:370000,53.2776507111864)
--(axis cs:380000,48.8776507111864)
--(axis cs:390000,53.5776507111864)
--(axis cs:390000,73.2223492888136)
--(axis cs:390000,73.2223492888136)
--(axis cs:380000,68.5223492888136)
--(axis cs:370000,72.9223492888136)
--(axis cs:360000,67.5223492888136)
--(axis cs:350000,67.7223492888136)
--(axis cs:340000,65.6223492888136)
--(axis cs:330000,67.3223492888136)
--(axis cs:320000,66.8223492888136)
--(axis cs:310000,65.6223492888136)
--(axis cs:300000,66.4223492888136)
--(axis cs:290000,66.4223492888136)
--(axis cs:280000,68.6223492888136)
--(axis cs:270000,67.1223492888136)
--(axis cs:260000,66.2223492888136)
--(axis cs:250000,65.6223492888136)
--(axis cs:240000,64.3223492888136)
--(axis cs:230000,67.5223492888136)
--(axis cs:220000,63.7223492888136)
--(axis cs:210000,62.6223492888136)
--(axis cs:200000,60.9223492888136)
--(axis cs:190000,58.2223492888136)
--(axis cs:180000,58.5223492888136)
--(axis cs:170000,63.6223492888136)
--(axis cs:160000,62.6223492888136)
--(axis cs:150000,60.5223492888136)
--(axis cs:140000,62.7223492888136)
--(axis cs:130000,64.8223492888136)
--(axis cs:120000,60.9223492888136)
--(axis cs:110000,58.1223492888136)
--(axis cs:100000,58.7223492888136)
--(axis cs:90000,55.8223492888136)
--(axis cs:80000,55.8223492888136)
--(axis cs:70000,55.9223492888136)
--(axis cs:60000,57.0223492888136)
--(axis cs:50000,57.9223492888136)
--(axis cs:40000,54.1223492888136)
--(axis cs:30000,43.6223492888136)
--(axis cs:20000,44.1223492888136)
--(axis cs:10000,35.3223492888136)
--(axis cs:0,10.7223492888136)
--(axis cs:-10000,6.12234928881362)
--cycle;

\addplot [semithick, darkcyan1115178, mark=square*, mark size=1.5, mark options={solid}]
table {%
-10000 -2.4
0 76.2
10000 96.4
20000 98.3
30000 99.1
40000 99.4
50000 99.5
60000 98.8
70000 98.9
80000 98.8
90000 99
100000 99.4
110000 99.4
120000 99.3
130000 99.3
140000 99.5
150000 99.3
160000 99.4
170000 99.4
180000 99.7
190000 99.7
200000 99.5
210000 99.8
220000 99.3
230000 99.5
240000 99.5
250000 99.3
260000 99.5
270000 99
280000 99.3
290000 99.5
300000 99.5
310000 99.5
320000 99.4
330000 99.6
340000 99.2
350000 99.6
360000 98.6
370000 99.6
380000 99.7
390000 99.4
};
\addplot [semithick, darkorange2221435, mark=triangle*, mark size=1.5, mark options={solid}]
table {%
-10000 -4.5
0 91.7
10000 98.2
20000 99
30000 99.2
40000 99
50000 99.5
60000 99.2
70000 99.7
80000 99.6
90000 99.6
100000 99.5
110000 99.4
120000 99.6
130000 99.8
140000 99.6
150000 99.7
160000 99.3
170000 99.8
180000 99.5
190000 99.5
200000 99.6
210000 99.7
220000 99.7
230000 99.7
240000 99.8
250000 99.7
260000 99.6
270000 99.7
280000 99.7
290000 99.7
300000 99.6
310000 99.8
320000 99.7
330000 99.9
340000 99.6
350000 99.6
360000 99.8
370000 99.5
380000 99.6
390000 99.6
};
\addplot [semithick, orchid204120188, mark=+, mark size=1.5, mark options={solid}]
table {%
-10000 -3.5
0 39.6
10000 79.2
20000 88.1
30000 90.4
40000 92
50000 95
60000 95.4
70000 96.1
80000 96
90000 96.7
100000 96.2
110000 97.5
120000 97.6
130000 98
140000 98
150000 98
160000 97.9
170000 97.9
180000 97.9
190000 97.7
200000 98.1
210000 98.8
220000 97.8
230000 98.5
240000 97.8
250000 97.9
260000 98.9
270000 97.3
280000 97.6
290000 97.9
300000 97.1
310000 98.2
320000 98.5
330000 99
340000 98.6
350000 97.5
360000 97.4
370000 96.3
380000 98.5
390000 98.6
};
\addplot [semithick, peru20214597, mark=diamond*, mark size=1.5, mark options={solid}]
table {%
-10000 -3.7
0 0.9
10000 25.5
20000 34.3
30000 33.8
40000 44.3
50000 48.1
60000 47.2
70000 46.1
80000 46
90000 46
100000 48.9
110000 48.3
120000 51.1
130000 55
140000 52.9
150000 50.7
160000 52.8
170000 53.8
180000 48.7
190000 48.4
200000 51.1
210000 52.8
220000 53.9
230000 57.7
240000 54.5
250000 55.8
260000 56.4
270000 57.3
280000 58.8
290000 56.6
300000 56.6
310000 55.8
320000 57
330000 57.5
340000 55.8
350000 57.9
360000 57.7
370000 63.1
380000 58.7
390000 63.4
};
\end{axis}

\end{tikzpicture}

%% file: Image/breakout.tex
\begin{tikzpicture}

\definecolor{darkcyan1115178}{RGB}{1,115,178}
\definecolor{darkorange2221435}{RGB}{222,143,5}
\definecolor{darkslategray38}{RGB}{38,38,38}
\definecolor{lavender234234242}{RGB}{234,234,242}
\definecolor{lightgray204}{RGB}{204,204,204}
\definecolor{orchid204120188}{RGB}{204,120,188}
\definecolor{peru20214597}{RGB}{202,145,97}

\begin{axis}[
axis background/.style={fill=lavender234234242},
axis line style={white},
legend cell align={left},
legend style={
  fill opacity=0.8,
  draw opacity=1,
  text opacity=1,
  at={(0.03,0.97)},
  anchor=north west,
  draw=lightgray204,
  fill=lavender234234242
},
tick align=outside,
x grid style={white},
xlabel=\textcolor{darkslategray38}{Time Step},
xmajorgrids,
xmajorticks=true,
xmin=-30000, xmax=410000,
xtick style={color=darkslategray38},
xtick={0,50000,100000,150000,200000,250000,300000,350000,400000},
xticklabels={0k,50k,100k,150k,200k,250k,300k,350k,400k},
y grid style={white},
ylabel=\textcolor{darkslategray38}{Performance},
ymajorgrids,
ymajorticks=true,
ymin=-29.6614228012333, ymax=204.861422801233,
ytick style={color=darkslategray38}
]
\path [draw=darkcyan1115178, fill=darkcyan1115178, opacity=0.2]
(axis cs:-10000,20.4012934556667)
--(axis cs:-10000,-19.0012934556667)
--(axis cs:0,-16.6012934556667)
--(axis cs:10000,-13.9012934556667)
--(axis cs:20000,-12.9012934556667)
--(axis cs:30000,-12.4012934556667)
--(axis cs:40000,-11.6012934556667)
--(axis cs:50000,-9.90129345566665)
--(axis cs:60000,-9.90129345566665)
--(axis cs:70000,-9.90129345566665)
--(axis cs:80000,-8.90129345566665)
--(axis cs:90000,-9.20129345566665)
--(axis cs:100000,1.89870654433335)
--(axis cs:110000,9.29870654433335)
--(axis cs:120000,6.09870654433335)
--(axis cs:130000,5.09870654433335)
--(axis cs:140000,36.9987065443333)
--(axis cs:150000,45.4987065443333)
--(axis cs:160000,34.4987065443333)
--(axis cs:170000,51.0987065443333)
--(axis cs:180000,57.1987065443333)
--(axis cs:190000,63.7987065443333)
--(axis cs:200000,75.9987065443333)
--(axis cs:210000,56.1987065443333)
--(axis cs:220000,62.7987065443333)
--(axis cs:230000,77.6987065443333)
--(axis cs:240000,85.9987065443333)
--(axis cs:250000,96.4987065443333)
--(axis cs:260000,79.8987065443333)
--(axis cs:270000,101.498706544333)
--(axis cs:280000,131.298706544333)
--(axis cs:290000,108.698706544333)
--(axis cs:300000,111.398706544333)
--(axis cs:310000,99.8987065443333)
--(axis cs:320000,121.398706544333)
--(axis cs:330000,91.6987065443333)
--(axis cs:340000,133.198706544333)
--(axis cs:350000,154.798706544333)
--(axis cs:360000,99.6987065443333)
--(axis cs:370000,106.998706544333)
--(axis cs:380000,79.6987065443333)
--(axis cs:390000,142.698706544333)
--(axis cs:390000,182.101293455667)
--(axis cs:390000,182.101293455667)
--(axis cs:380000,119.101293455667)
--(axis cs:370000,146.401293455667)
--(axis cs:360000,139.101293455667)
--(axis cs:350000,194.201293455667)
--(axis cs:340000,172.601293455667)
--(axis cs:330000,131.101293455667)
--(axis cs:320000,160.801293455667)
--(axis cs:310000,139.301293455667)
--(axis cs:300000,150.801293455667)
--(axis cs:290000,148.101293455667)
--(axis cs:280000,170.701293455667)
--(axis cs:270000,140.901293455667)
--(axis cs:260000,119.301293455667)
--(axis cs:250000,135.901293455667)
--(axis cs:240000,125.401293455667)
--(axis cs:230000,117.101293455667)
--(axis cs:220000,102.201293455667)
--(axis cs:210000,95.6012934556667)
--(axis cs:200000,115.401293455667)
--(axis cs:190000,103.201293455667)
--(axis cs:180000,96.6012934556667)
--(axis cs:170000,90.5012934556667)
--(axis cs:160000,73.9012934556667)
--(axis cs:150000,84.9012934556667)
--(axis cs:140000,76.4012934556667)
--(axis cs:130000,44.5012934556667)
--(axis cs:120000,45.5012934556667)
--(axis cs:110000,48.7012934556667)
--(axis cs:100000,41.3012934556667)
--(axis cs:90000,30.2012934556667)
--(axis cs:80000,30.5012934556667)
--(axis cs:70000,29.5012934556667)
--(axis cs:60000,29.5012934556667)
--(axis cs:50000,29.5012934556667)
--(axis cs:40000,27.8012934556667)
--(axis cs:30000,27.0012934556667)
--(axis cs:20000,26.5012934556667)
--(axis cs:10000,25.5012934556667)
--(axis cs:0,22.8012934556667)
--(axis cs:-10000,20.4012934556667)
--cycle;

\path [draw=darkorange2221435, fill=darkorange2221435, opacity=0.2]
(axis cs:-10000,1.64414643404615)
--(axis cs:-10000,0.755853565953854)
--(axis cs:0,1.95585356595385)
--(axis cs:10000,3.05585356595385)
--(axis cs:20000,4.45585356595385)
--(axis cs:30000,4.35585356595385)
--(axis cs:40000,4.15585356595385)
--(axis cs:50000,4.35585356595385)
--(axis cs:60000,4.35585356595385)
--(axis cs:70000,5.05585356595385)
--(axis cs:80000,5.15585356595385)
--(axis cs:90000,5.15585356595385)
--(axis cs:100000,5.95585356595385)
--(axis cs:110000,5.05585356595385)
--(axis cs:120000,5.15585356595385)
--(axis cs:130000,5.75585356595385)
--(axis cs:140000,5.35585356595385)
--(axis cs:150000,4.75585356595385)
--(axis cs:160000,5.15585356595385)
--(axis cs:170000,5.75585356595385)
--(axis cs:180000,5.75585356595385)
--(axis cs:190000,5.85585356595385)
--(axis cs:200000,6.35585356595385)
--(axis cs:210000,5.05585356595385)
--(axis cs:220000,5.35585356595385)
--(axis cs:230000,5.75585356595385)
--(axis cs:240000,5.55585356595385)
--(axis cs:250000,5.95585356595385)
--(axis cs:260000,5.75585356595385)
--(axis cs:270000,5.85585356595385)
--(axis cs:280000,5.75585356595385)
--(axis cs:290000,5.65585356595385)
--(axis cs:300000,6.25585356595385)
--(axis cs:310000,6.15585356595385)
--(axis cs:320000,5.15585356595385)
--(axis cs:330000,5.35585356595385)
--(axis cs:340000,5.55585356595385)
--(axis cs:350000,5.55585356595385)
--(axis cs:360000,5.25585356595385)
--(axis cs:370000,5.05585356595385)
--(axis cs:380000,4.55585356595385)
--(axis cs:390000,5.35585356595385)
--(axis cs:390000,6.24414643404615)
--(axis cs:390000,6.24414643404615)
--(axis cs:380000,5.44414643404615)
--(axis cs:370000,5.94414643404615)
--(axis cs:360000,6.14414643404615)
--(axis cs:350000,6.44414643404615)
--(axis cs:340000,6.44414643404615)
--(axis cs:330000,6.24414643404615)
--(axis cs:320000,6.04414643404615)
--(axis cs:310000,7.04414643404615)
--(axis cs:300000,7.14414643404615)
--(axis cs:290000,6.54414643404615)
--(axis cs:280000,6.64414643404615)
--(axis cs:270000,6.74414643404615)
--(axis cs:260000,6.64414643404615)
--(axis cs:250000,6.84414643404615)
--(axis cs:240000,6.44414643404615)
--(axis cs:230000,6.64414643404615)
--(axis cs:220000,6.24414643404615)
--(axis cs:210000,5.94414643404615)
--(axis cs:200000,7.24414643404615)
--(axis cs:190000,6.74414643404615)
--(axis cs:180000,6.64414643404615)
--(axis cs:170000,6.64414643404615)
--(axis cs:160000,6.04414643404615)
--(axis cs:150000,5.64414643404615)
--(axis cs:140000,6.24414643404615)
--(axis cs:130000,6.64414643404615)
--(axis cs:120000,6.04414643404615)
--(axis cs:110000,5.94414643404615)
--(axis cs:100000,6.84414643404615)
--(axis cs:90000,6.04414643404615)
--(axis cs:80000,6.04414643404615)
--(axis cs:70000,5.94414643404615)
--(axis cs:60000,5.24414643404615)
--(axis cs:50000,5.24414643404615)
--(axis cs:40000,5.04414643404615)
--(axis cs:30000,5.24414643404615)
--(axis cs:20000,5.34414643404615)
--(axis cs:10000,3.94414643404615)
--(axis cs:0,2.84414643404615)
--(axis cs:-10000,1.64414643404615)
--cycle;

\path [draw=orchid204120188, fill=orchid204120188, opacity=0.2]
(axis cs:-10000,20.4012934556667)
--(axis cs:-10000,-19.0012934556667)
--(axis cs:0,-16.6012934556667)
--(axis cs:10000,-13.9012934556667)
--(axis cs:20000,-12.9012934556667)
--(axis cs:30000,-12.4012934556667)
--(axis cs:40000,-11.6012934556667)
--(axis cs:50000,-9.90129345566665)
--(axis cs:60000,-9.90129345566665)
--(axis cs:70000,-9.90129345566665)
--(axis cs:80000,-8.90129345566665)
--(axis cs:90000,-9.20129345566665)
--(axis cs:100000,1.89870654433335)
--(axis cs:110000,9.29870654433335)
--(axis cs:120000,6.09870654433335)
--(axis cs:130000,5.09870654433335)
--(axis cs:140000,36.9987065443333)
--(axis cs:150000,45.4987065443333)
--(axis cs:160000,34.4987065443333)
--(axis cs:170000,51.0987065443333)
--(axis cs:180000,57.1987065443333)
--(axis cs:190000,63.7987065443333)
--(axis cs:200000,75.9987065443333)
--(axis cs:210000,56.1987065443333)
--(axis cs:220000,62.7987065443333)
--(axis cs:230000,77.6987065443333)
--(axis cs:240000,85.9987065443333)
--(axis cs:250000,96.4987065443333)
--(axis cs:260000,79.8987065443333)
--(axis cs:270000,101.498706544333)
--(axis cs:280000,131.298706544333)
--(axis cs:290000,108.698706544333)
--(axis cs:300000,111.398706544333)
--(axis cs:310000,99.8987065443333)
--(axis cs:320000,121.398706544333)
--(axis cs:330000,91.6987065443333)
--(axis cs:340000,133.198706544333)
--(axis cs:350000,154.798706544333)
--(axis cs:360000,99.6987065443333)
--(axis cs:370000,106.998706544333)
--(axis cs:380000,79.6987065443333)
--(axis cs:390000,142.698706544333)
--(axis cs:390000,182.101293455667)
--(axis cs:390000,182.101293455667)
--(axis cs:380000,119.101293455667)
--(axis cs:370000,146.401293455667)
--(axis cs:360000,139.101293455667)
--(axis cs:350000,194.201293455667)
--(axis cs:340000,172.601293455667)
--(axis cs:330000,131.101293455667)
--(axis cs:320000,160.801293455667)
--(axis cs:310000,139.301293455667)
--(axis cs:300000,150.801293455667)
--(axis cs:290000,148.101293455667)
--(axis cs:280000,170.701293455667)
--(axis cs:270000,140.901293455667)
--(axis cs:260000,119.301293455667)
--(axis cs:250000,135.901293455667)
--(axis cs:240000,125.401293455667)
--(axis cs:230000,117.101293455667)
--(axis cs:220000,102.201293455667)
--(axis cs:210000,95.6012934556667)
--(axis cs:200000,115.401293455667)
--(axis cs:190000,103.201293455667)
--(axis cs:180000,96.6012934556667)
--(axis cs:170000,90.5012934556667)
--(axis cs:160000,73.9012934556667)
--(axis cs:150000,84.9012934556667)
--(axis cs:140000,76.4012934556667)
--(axis cs:130000,44.5012934556667)
--(axis cs:120000,45.5012934556667)
--(axis cs:110000,48.7012934556667)
--(axis cs:100000,41.3012934556667)
--(axis cs:90000,30.2012934556667)
--(axis cs:80000,30.5012934556667)
--(axis cs:70000,29.5012934556667)
--(axis cs:60000,29.5012934556667)
--(axis cs:50000,29.5012934556667)
--(axis cs:40000,27.8012934556667)
--(axis cs:30000,27.0012934556667)
--(axis cs:20000,26.5012934556667)
--(axis cs:10000,25.5012934556667)
--(axis cs:0,22.8012934556667)
--(axis cs:-10000,20.4012934556667)
--cycle;

\path [draw=peru20214597, fill=peru20214597, opacity=0.2]
(axis cs:-10000,16.7507449810821)
--(axis cs:-10000,-15.1507449810821)
--(axis cs:0,-8.55074498108214)
--(axis cs:10000,-6.35074498108214)
--(axis cs:20000,-2.85074498108214)
--(axis cs:30000,-1.25074498108214)
--(axis cs:40000,13.9492550189179)
--(axis cs:50000,3.14925501891786)
--(axis cs:60000,5.84925501891786)
--(axis cs:70000,8.54925501891786)
--(axis cs:80000,18.7492550189179)
--(axis cs:90000,17.2492550189179)
--(axis cs:100000,17.6492550189179)
--(axis cs:110000,21.5492550189179)
--(axis cs:120000,30.1492550189179)
--(axis cs:130000,62.8492550189179)
--(axis cs:140000,50.3492550189179)
--(axis cs:150000,43.2492550189179)
--(axis cs:160000,33.9492550189179)
--(axis cs:170000,61.4492550189179)
--(axis cs:180000,51.0492550189179)
--(axis cs:190000,48.1492550189179)
--(axis cs:200000,47.0492550189179)
--(axis cs:210000,14.3492550189179)
--(axis cs:220000,26.1492550189179)
--(axis cs:230000,37.7492550189179)
--(axis cs:240000,59.1492550189179)
--(axis cs:250000,38.2492550189179)
--(axis cs:260000,85.5492550189179)
--(axis cs:270000,53.2492550189179)
--(axis cs:280000,63.8492550189179)
--(axis cs:290000,58.5492550189179)
--(axis cs:300000,90.2492550189179)
--(axis cs:310000,60.0492550189179)
--(axis cs:320000,75.2492550189179)
--(axis cs:330000,74.9492550189179)
--(axis cs:340000,45.8492550189179)
--(axis cs:350000,46.5492550189179)
--(axis cs:360000,115.949255018918)
--(axis cs:370000,89.2492550189179)
--(axis cs:380000,88.7492550189179)
--(axis cs:390000,78.4492550189179)
--(axis cs:390000,110.350744981082)
--(axis cs:390000,110.350744981082)
--(axis cs:380000,120.650744981082)
--(axis cs:370000,121.150744981082)
--(axis cs:360000,147.850744981082)
--(axis cs:350000,78.4507449810821)
--(axis cs:340000,77.7507449810821)
--(axis cs:330000,106.850744981082)
--(axis cs:320000,107.150744981082)
--(axis cs:310000,91.9507449810821)
--(axis cs:300000,122.150744981082)
--(axis cs:290000,90.4507449810821)
--(axis cs:280000,95.7507449810821)
--(axis cs:270000,85.1507449810821)
--(axis cs:260000,117.450744981082)
--(axis cs:250000,70.1507449810821)
--(axis cs:240000,91.0507449810821)
--(axis cs:230000,69.6507449810821)
--(axis cs:220000,58.0507449810821)
--(axis cs:210000,46.2507449810821)
--(axis cs:200000,78.9507449810821)
--(axis cs:190000,80.0507449810821)
--(axis cs:180000,82.9507449810821)
--(axis cs:170000,93.3507449810821)
--(axis cs:160000,65.8507449810821)
--(axis cs:150000,75.1507449810821)
--(axis cs:140000,82.2507449810821)
--(axis cs:130000,94.7507449810821)
--(axis cs:120000,62.0507449810821)
--(axis cs:110000,53.4507449810821)
--(axis cs:100000,49.5507449810821)
--(axis cs:90000,49.1507449810821)
--(axis cs:80000,50.6507449810821)
--(axis cs:70000,40.4507449810821)
--(axis cs:60000,37.7507449810821)
--(axis cs:50000,35.0507449810821)
--(axis cs:40000,45.8507449810821)
--(axis cs:30000,30.6507449810821)
--(axis cs:20000,29.0507449810821)
--(axis cs:10000,25.5507449810821)
--(axis cs:0,23.3507449810821)
--(axis cs:-10000,16.7507449810821)
--cycle;

\addplot [semithick, darkcyan1115178, mark=square*, mark size=1.5, mark options={solid}]
table {%
-10000 0.7
0 3.1
10000 5.8
20000 6.8
30000 7.3
40000 8.1
50000 9.8
60000 9.8
70000 9.8
80000 10.8
90000 10.5
100000 21.6
110000 29
120000 25.8
130000 24.8
140000 56.7
150000 65.2
160000 54.2
170000 70.8
180000 76.9
190000 83.5
200000 95.7
210000 75.9
220000 82.5
230000 97.4
240000 105.7
250000 116.2
260000 99.6
270000 121.2
280000 151
290000 128.4
300000 131.1
310000 119.6
320000 141.1
330000 111.4
340000 152.9
350000 174.5
360000 119.4
370000 126.7
380000 99.4
390000 162.4
};
\addplot [semithick, darkorange2221435, mark=triangle*, mark size=1.5, mark options={solid}]
table {%
-10000 1.2
0 2.4
10000 3.5
20000 4.9
30000 4.8
40000 4.6
50000 4.8
60000 4.8
70000 5.5
80000 5.6
90000 5.6
100000 6.4
110000 5.5
120000 5.6
130000 6.2
140000 5.8
150000 5.2
160000 5.6
170000 6.2
180000 6.2
190000 6.3
200000 6.8
210000 5.5
220000 5.8
230000 6.2
240000 6
250000 6.4
260000 6.2
270000 6.3
280000 6.2
290000 6.1
300000 6.7
310000 6.6
320000 5.6
330000 5.8
340000 6
350000 6
360000 5.7
370000 5.5
380000 5
390000 5.8
};
\addplot [semithick, orchid204120188, mark=+, mark size=1.5, mark options={solid}]
table {%
-10000 0.7
0 3.1
10000 5.8
20000 6.8
30000 7.3
40000 8.1
50000 9.8
60000 9.8
70000 9.8
80000 10.8
90000 10.5
100000 21.6
110000 29
120000 25.8
130000 24.8
140000 56.7
150000 65.2
160000 54.2
170000 70.8
180000 76.9
190000 83.5
200000 95.7
210000 75.9
220000 82.5
230000 97.4
240000 105.7
250000 116.2
260000 99.6
270000 121.2
280000 151
290000 128.4
300000 131.1
310000 119.6
320000 141.1
330000 111.4
340000 152.9
350000 174.5
360000 119.4
370000 126.7
380000 99.4
390000 162.4
};
\addplot [semithick, peru20214597, mark=diamond*, mark size=1.5, mark options={solid}]
table {%
-10000 0.8
0 7.4
10000 9.6
20000 13.1
30000 14.7
40000 29.9
50000 19.1
60000 21.8
70000 24.5
80000 34.7
90000 33.2
100000 33.6
110000 37.5
120000 46.1
130000 78.8
140000 66.3
150000 59.2
160000 49.9
170000 77.4
180000 67
190000 64.1
200000 63
210000 30.3
220000 42.1
230000 53.7
240000 75.1
250000 54.2
260000 101.5
270000 69.2
280000 79.8
290000 74.5
300000 106.2
310000 76
320000 91.2
330000 90.9
340000 61.8
350000 62.5
360000 131.9
370000 105.2
380000 104.7
390000 94.4
};
\end{axis}

\end{tikzpicture}

%% file: Image/crazy_climber.tex
\begin{tikzpicture}

\definecolor{darkcyan1115178}{RGB}{1,115,178}
\definecolor{darkorange2221435}{RGB}{222,143,5}
\definecolor{darkslategray38}{RGB}{38,38,38}
\definecolor{lavender234234242}{RGB}{234,234,242}
\definecolor{lightgray204}{RGB}{204,204,204}
\definecolor{orchid204120188}{RGB}{204,120,188}
\definecolor{peru20214597}{RGB}{202,145,97}

\begin{axis}[
axis background/.style={fill=lavender234234242},
axis line style={white},
legend cell align={left},
legend style={
  fill opacity=0.8,
  draw opacity=1,
  text opacity=1,
  at={(0.03,0.97)},
  anchor=north west,
  draw=lightgray204,
  fill=lavender234234242
},
tick align=outside,
x grid style={white},
xlabel=\textcolor{darkslategray38}{Time Step},
xmajorgrids,
xmajorticks=true,
xmin=-30000, xmax=410000,
xtick style={color=darkslategray38},
xtick={0,50000,100000,150000,200000,250000,300000,350000,400000},
xticklabels={0k,50k,100k,150k,200k,250k,300k,350k,400k},
y grid style={white},
ylabel=\textcolor{darkslategray38}{Performance},
ymajorgrids,
ymajorticks=true,
ymin=-12488.730874197, ymax=101070.130874197,
ytick style={color=darkslategray38}
]
\path [draw=darkcyan1115178, fill=darkcyan1115178, opacity=0.2]
(axis cs:-10000,14095.5644310882)
--(axis cs:-10000,-7326.96443108818)
--(axis cs:0,23113.0355689118)
--(axis cs:10000,30013.0355689118)
--(axis cs:20000,33968.7355689118)
--(axis cs:30000,32270.1355689118)
--(axis cs:40000,42184.4355689118)
--(axis cs:50000,46063.0355689118)
--(axis cs:60000,45347.3355689118)
--(axis cs:70000,50647.3355689118)
--(axis cs:80000,55661.6355689118)
--(axis cs:90000,56393.0355689118)
--(axis cs:100000,60408.7355689118)
--(axis cs:110000,57985.8355689118)
--(axis cs:120000,55215.8355689118)
--(axis cs:130000,58303.0355689118)
--(axis cs:140000,52440.1355689118)
--(axis cs:150000,59210.1355689118)
--(axis cs:160000,52904.4355689118)
--(axis cs:170000,55344.4355689118)
--(axis cs:180000,54038.7355689118)
--(axis cs:190000,55625.8355689118)
--(axis cs:200000,61433.0355689118)
--(axis cs:210000,61698.7355689118)
--(axis cs:220000,64580.1355689118)
--(axis cs:230000,55343.0355689118)
--(axis cs:240000,64485.8355689118)
--(axis cs:250000,53633.0355689118)
--(axis cs:260000,52410.1355689118)
--(axis cs:270000,55307.3355689118)
--(axis cs:280000,59798.7355689118)
--(axis cs:290000,58973.0355689118)
--(axis cs:300000,63803.0355689118)
--(axis cs:310000,60507.3355689118)
--(axis cs:320000,56803.0355689118)
--(axis cs:330000,66225.8355689118)
--(axis cs:340000,61125.8355689118)
--(axis cs:350000,52751.6355689118)
--(axis cs:360000,53620.1355689118)
--(axis cs:370000,61201.6355689118)
--(axis cs:380000,66453.0355689118)
--(axis cs:390000,74485.8355689118)
--(axis cs:390000,95908.3644310882)
--(axis cs:390000,95908.3644310882)
--(axis cs:380000,87875.5644310882)
--(axis cs:370000,82624.1644310882)
--(axis cs:360000,75042.6644310882)
--(axis cs:350000,74174.1644310882)
--(axis cs:340000,82548.3644310882)
--(axis cs:330000,87648.3644310882)
--(axis cs:320000,78225.5644310882)
--(axis cs:310000,81929.8644310882)
--(axis cs:300000,85225.5644310882)
--(axis cs:290000,80395.5644310882)
--(axis cs:280000,81221.2644310882)
--(axis cs:270000,76729.8644310882)
--(axis cs:260000,73832.6644310882)
--(axis cs:250000,75055.5644310882)
--(axis cs:240000,85908.3644310882)
--(axis cs:230000,76765.5644310882)
--(axis cs:220000,86002.6644310882)
--(axis cs:210000,83121.2644310882)
--(axis cs:200000,82855.5644310882)
--(axis cs:190000,77048.3644310882)
--(axis cs:180000,75461.2644310882)
--(axis cs:170000,76766.9644310882)
--(axis cs:160000,74326.9644310882)
--(axis cs:150000,80632.6644310882)
--(axis cs:140000,73862.6644310882)
--(axis cs:130000,79725.5644310882)
--(axis cs:120000,76638.3644310882)
--(axis cs:110000,79408.3644310882)
--(axis cs:100000,81831.2644310882)
--(axis cs:90000,77815.5644310882)
--(axis cs:80000,77084.1644310882)
--(axis cs:70000,72069.8644310882)
--(axis cs:60000,66769.8644310882)
--(axis cs:50000,67485.5644310882)
--(axis cs:40000,63606.9644310882)
--(axis cs:30000,53692.6644310882)
--(axis cs:20000,55391.2644310882)
--(axis cs:10000,51435.5644310882)
--(axis cs:0,44535.5644310882)
--(axis cs:-10000,14095.5644310882)
--cycle;

\path [draw=darkorange2221435, fill=darkorange2221435, opacity=0.2]
(axis cs:-10000,7470.16629747932)
--(axis cs:-10000,-2924.16629747932)
--(axis cs:0,7324.83370252068)
--(axis cs:10000,17130.8337025207)
--(axis cs:20000,16491.8337025207)
--(axis cs:30000,24598.8337025207)
--(axis cs:40000,28684.8337025207)
--(axis cs:50000,38626.8337025207)
--(axis cs:60000,42395.8337025207)
--(axis cs:70000,51756.8337025207)
--(axis cs:80000,38844.8337025207)
--(axis cs:90000,46901.8337025207)
--(axis cs:100000,38986.8337025207)
--(axis cs:110000,45815.8337025207)
--(axis cs:120000,49504.8337025207)
--(axis cs:130000,46731.8337025207)
--(axis cs:140000,49435.8337025207)
--(axis cs:150000,51440.8337025207)
--(axis cs:160000,43106.8337025207)
--(axis cs:170000,38892.8337025207)
--(axis cs:180000,45616.8337025207)
--(axis cs:190000,43433.8337025207)
--(axis cs:200000,51248.8337025207)
--(axis cs:210000,50801.8337025207)
--(axis cs:220000,53418.8337025207)
--(axis cs:230000,46960.8337025207)
--(axis cs:240000,49045.8337025207)
--(axis cs:250000,47975.8337025207)
--(axis cs:260000,47711.8337025207)
--(axis cs:270000,50240.8337025207)
--(axis cs:280000,47137.8337025207)
--(axis cs:290000,51245.8337025207)
--(axis cs:300000,50105.8337025207)
--(axis cs:310000,52530.8337025207)
--(axis cs:320000,45512.8337025207)
--(axis cs:330000,51096.8337025207)
--(axis cs:340000,49845.8337025207)
--(axis cs:350000,50056.8337025207)
--(axis cs:360000,48564.8337025207)
--(axis cs:370000,54206.8337025207)
--(axis cs:380000,52998.8337025207)
--(axis cs:390000,52070.8337025207)
--(axis cs:390000,62465.1662974793)
--(axis cs:390000,62465.1662974793)
--(axis cs:380000,63393.1662974793)
--(axis cs:370000,64601.1662974793)
--(axis cs:360000,58959.1662974793)
--(axis cs:350000,60451.1662974793)
--(axis cs:340000,60240.1662974793)
--(axis cs:330000,61491.1662974793)
--(axis cs:320000,55907.1662974793)
--(axis cs:310000,62925.1662974793)
--(axis cs:300000,60500.1662974793)
--(axis cs:290000,61640.1662974793)
--(axis cs:280000,57532.1662974793)
--(axis cs:270000,60635.1662974793)
--(axis cs:260000,58106.1662974793)
--(axis cs:250000,58370.1662974793)
--(axis cs:240000,59440.1662974793)
--(axis cs:230000,57355.1662974793)
--(axis cs:220000,63813.1662974793)
--(axis cs:210000,61196.1662974793)
--(axis cs:200000,61643.1662974793)
--(axis cs:190000,53828.1662974793)
--(axis cs:180000,56011.1662974793)
--(axis cs:170000,49287.1662974793)
--(axis cs:160000,53501.1662974793)
--(axis cs:150000,61835.1662974793)
--(axis cs:140000,59830.1662974793)
--(axis cs:130000,57126.1662974793)
--(axis cs:120000,59899.1662974793)
--(axis cs:110000,56210.1662974793)
--(axis cs:100000,49381.1662974793)
--(axis cs:90000,57296.1662974793)
--(axis cs:80000,49239.1662974793)
--(axis cs:70000,62151.1662974793)
--(axis cs:60000,52790.1662974793)
--(axis cs:50000,49021.1662974793)
--(axis cs:40000,39079.1662974793)
--(axis cs:30000,34993.1662974793)
--(axis cs:20000,26886.1662974793)
--(axis cs:10000,27525.1662974793)
--(axis cs:0,17719.1662974793)
--(axis cs:-10000,7470.16629747932)
--cycle;

\path [draw=orchid204120188, fill=orchid204120188, opacity=0.2]
(axis cs:-10000,9728.56048979324)
--(axis cs:-10000,-5211.36048979323)
--(axis cs:0,16598.6395102068)
--(axis cs:10000,14100.0395102068)
--(axis cs:20000,32992.9395102068)
--(axis cs:30000,38062.9395102068)
--(axis cs:40000,37075.7395102068)
--(axis cs:50000,31625.7395102068)
--(axis cs:60000,53050.0395102068)
--(axis cs:70000,38058.6395102068)
--(axis cs:80000,32401.4395102068)
--(axis cs:90000,28207.1395102068)
--(axis cs:100000,37700.0395102068)
--(axis cs:110000,34291.4395102068)
--(axis cs:120000,33627.1395102068)
--(axis cs:130000,37827.1395102068)
--(axis cs:140000,41067.1395102068)
--(axis cs:150000,38261.4395102068)
--(axis cs:160000,39154.3395102068)
--(axis cs:170000,34680.0395102068)
--(axis cs:180000,34285.7395102068)
--(axis cs:190000,38811.4395102068)
--(axis cs:200000,37575.7395102068)
--(axis cs:210000,38844.3395102068)
--(axis cs:220000,34208.6395102068)
--(axis cs:230000,41142.9395102068)
--(axis cs:240000,38197.1395102068)
--(axis cs:250000,34638.6395102068)
--(axis cs:260000,34250.0395102068)
--(axis cs:270000,37447.1395102068)
--(axis cs:280000,43065.7395102068)
--(axis cs:290000,35681.4395102068)
--(axis cs:300000,39211.4395102068)
--(axis cs:310000,33627.1395102068)
--(axis cs:320000,31351.4395102068)
--(axis cs:330000,33621.4395102068)
--(axis cs:340000,34345.7395102068)
--(axis cs:350000,34230.0395102068)
--(axis cs:360000,33307.1395102068)
--(axis cs:370000,36288.6395102068)
--(axis cs:380000,40855.7395102068)
--(axis cs:390000,35687.1395102068)
--(axis cs:390000,50627.0604897932)
--(axis cs:390000,50627.0604897932)
--(axis cs:380000,55795.6604897932)
--(axis cs:370000,51228.5604897932)
--(axis cs:360000,48247.0604897932)
--(axis cs:350000,49169.9604897932)
--(axis cs:340000,49285.6604897932)
--(axis cs:330000,48561.3604897932)
--(axis cs:320000,46291.3604897932)
--(axis cs:310000,48567.0604897932)
--(axis cs:300000,54151.3604897932)
--(axis cs:290000,50621.3604897932)
--(axis cs:280000,58005.6604897932)
--(axis cs:270000,52387.0604897932)
--(axis cs:260000,49189.9604897932)
--(axis cs:250000,49578.5604897932)
--(axis cs:240000,53137.0604897932)
--(axis cs:230000,56082.8604897932)
--(axis cs:220000,49148.5604897932)
--(axis cs:210000,53784.2604897932)
--(axis cs:200000,52515.6604897932)
--(axis cs:190000,53751.3604897932)
--(axis cs:180000,49225.6604897932)
--(axis cs:170000,49619.9604897932)
--(axis cs:160000,54094.2604897932)
--(axis cs:150000,53201.3604897932)
--(axis cs:140000,56007.0604897932)
--(axis cs:130000,52767.0604897932)
--(axis cs:120000,48567.0604897932)
--(axis cs:110000,49231.3604897932)
--(axis cs:100000,52639.9604897932)
--(axis cs:90000,43147.0604897932)
--(axis cs:80000,47341.3604897932)
--(axis cs:70000,52998.5604897932)
--(axis cs:60000,67989.9604897932)
--(axis cs:50000,46565.6604897932)
--(axis cs:40000,52015.6604897932)
--(axis cs:30000,53002.8604897932)
--(axis cs:20000,47932.8604897932)
--(axis cs:10000,29039.9604897932)
--(axis cs:0,31538.5604897932)
--(axis cs:-10000,9728.56048979324)
--cycle;

\path [draw=peru20214597, fill=peru20214597, opacity=0.2]
(axis cs:-10000,10633.4683500496)
--(axis cs:-10000,-4967.66835004959)
--(axis cs:0,3215.13164995041)
--(axis cs:10000,12618.0316499504)
--(axis cs:20000,12089.4316499504)
--(axis cs:30000,16172.3316499504)
--(axis cs:40000,13858.0316499504)
--(axis cs:50000,11735.1316499504)
--(axis cs:60000,21459.4316499504)
--(axis cs:70000,22856.5316499504)
--(axis cs:80000,27580.8316499504)
--(axis cs:90000,30628.0316499504)
--(axis cs:100000,27670.8316499504)
--(axis cs:110000,30192.3316499504)
--(axis cs:120000,31433.7316499504)
--(axis cs:130000,32052.3316499504)
--(axis cs:140000,30465.1316499504)
--(axis cs:150000,31305.1316499504)
--(axis cs:160000,41513.7316499504)
--(axis cs:170000,44989.4316499504)
--(axis cs:180000,40989.4316499504)
--(axis cs:190000,43772.3316499504)
--(axis cs:200000,38629.4316499504)
--(axis cs:210000,41912.3316499504)
--(axis cs:220000,46738.0316499504)
--(axis cs:230000,41256.5316499504)
--(axis cs:240000,40855.1316499504)
--(axis cs:250000,43629.4316499504)
--(axis cs:260000,43795.1316499504)
--(axis cs:270000,46823.7316499504)
--(axis cs:280000,43143.7316499504)
--(axis cs:290000,46025.1316499504)
--(axis cs:300000,49060.8316499504)
--(axis cs:310000,51310.8316499504)
--(axis cs:320000,48650.8316499504)
--(axis cs:330000,52435.1316499504)
--(axis cs:340000,55339.4316499504)
--(axis cs:350000,58532.3316499504)
--(axis cs:360000,55455.1316499504)
--(axis cs:370000,51610.8316499504)
--(axis cs:380000,52542.3316499504)
--(axis cs:390000,54592.3316499504)
--(axis cs:390000,70193.4683500496)
--(axis cs:390000,70193.4683500496)
--(axis cs:380000,68143.4683500496)
--(axis cs:370000,67211.9683500496)
--(axis cs:360000,71056.2683500496)
--(axis cs:350000,74133.4683500496)
--(axis cs:340000,70940.5683500496)
--(axis cs:330000,68036.2683500496)
--(axis cs:320000,64251.9683500496)
--(axis cs:310000,66911.9683500496)
--(axis cs:300000,64661.9683500496)
--(axis cs:290000,61626.2683500496)
--(axis cs:280000,58744.8683500496)
--(axis cs:270000,62424.8683500496)
--(axis cs:260000,59396.2683500496)
--(axis cs:250000,59230.5683500496)
--(axis cs:240000,56456.2683500496)
--(axis cs:230000,56857.6683500496)
--(axis cs:220000,62339.1683500496)
--(axis cs:210000,57513.4683500496)
--(axis cs:200000,54230.5683500496)
--(axis cs:190000,59373.4683500496)
--(axis cs:180000,56590.5683500496)
--(axis cs:170000,60590.5683500496)
--(axis cs:160000,57114.8683500496)
--(axis cs:150000,46906.2683500496)
--(axis cs:140000,46066.2683500496)
--(axis cs:130000,47653.4683500496)
--(axis cs:120000,47034.8683500496)
--(axis cs:110000,45793.4683500496)
--(axis cs:100000,43271.9683500496)
--(axis cs:90000,46229.1683500496)
--(axis cs:80000,43181.9683500496)
--(axis cs:70000,38457.6683500496)
--(axis cs:60000,37060.5683500496)
--(axis cs:50000,27336.2683500496)
--(axis cs:40000,29459.1683500496)
--(axis cs:30000,31773.4683500496)
--(axis cs:20000,27690.5683500496)
--(axis cs:10000,28219.1683500496)
--(axis cs:0,18816.2683500496)
--(axis cs:-10000,10633.4683500496)
--cycle;

\addplot [semithick, darkcyan1115178, mark=square*, mark size=1.5, mark options={solid}]
table {%
-10000 3384.3
0 33824.3
10000 40724.3
20000 44680
30000 42981.4
40000 52895.7
50000 56774.3
60000 56058.6
70000 61358.6
80000 66372.9
90000 67104.3
100000 71120
110000 68697.1
120000 65927.1
130000 69014.3
140000 63151.4
150000 69921.4
160000 63615.7
170000 66055.7
180000 64750
190000 66337.1
200000 72144.3
210000 72410
220000 75291.4
230000 66054.3
240000 75197.1
250000 64344.3
260000 63121.4
270000 66018.6
280000 70510
290000 69684.3
300000 74514.3
310000 71218.6
320000 67514.3
330000 76937.1
340000 71837.1
350000 63462.9
360000 64331.4
370000 71912.9
380000 77164.3
390000 85197.1
};
\addplot [semithick, darkorange2221435, mark=triangle*, mark size=1.5, mark options={solid}]
table {%
-10000 2273
0 12522
10000 22328
20000 21689
30000 29796
40000 33882
50000 43824
60000 47593
70000 56954
80000 44042
90000 52099
100000 44184
110000 51013
120000 54702
130000 51929
140000 54633
150000 56638
160000 48304
170000 44090
180000 50814
190000 48631
200000 56446
210000 55999
220000 58616
230000 52158
240000 54243
250000 53173
260000 52909
270000 55438
280000 52335
290000 56443
300000 55303
310000 57728
320000 50710
330000 56294
340000 55043
350000 55254
360000 53762
370000 59404
380000 58196
390000 57268
};
\addplot [semithick, orchid204120188, mark=+, mark size=1.5, mark options={solid}]
table {%
-10000 2258.6
0 24068.6
10000 21570
20000 40462.9
30000 45532.9
40000 44545.7
50000 39095.7
60000 60520
70000 45528.6
80000 39871.4
90000 35677.1
100000 45170
110000 41761.4
120000 41097.1
130000 45297.1
140000 48537.1
150000 45731.4
160000 46624.3
170000 42150
180000 41755.7
190000 46281.4
200000 45045.7
210000 46314.3
220000 41678.6
230000 48612.9
240000 45667.1
250000 42108.6
260000 41720
270000 44917.1
280000 50535.7
290000 43151.4
300000 46681.4
310000 41097.1
320000 38821.4
330000 41091.4
340000 41815.7
350000 41700
360000 40777.1
370000 43758.6
380000 48325.7
390000 43157.1
};
\addplot [semithick, peru20214597, mark=diamond*, mark size=1.5, mark options={solid}]
table {%
-10000 2832.9
0 11015.7
10000 20418.6
20000 19890
30000 23972.9
40000 21658.6
50000 19535.7
60000 29260
70000 30657.1
80000 35381.4
90000 38428.6
100000 35471.4
110000 37992.9
120000 39234.3
130000 39852.9
140000 38265.7
150000 39105.7
160000 49314.3
170000 52790
180000 48790
190000 51572.9
200000 46430
210000 49712.9
220000 54538.6
230000 49057.1
240000 48655.7
250000 51430
260000 51595.7
270000 54624.3
280000 50944.3
290000 53825.7
300000 56861.4
310000 59111.4
320000 56451.4
330000 60235.7
340000 63140
350000 66332.9
360000 63255.7
370000 59411.4
380000 60342.9
390000 62392.9
};
\end{axis}

\end{tikzpicture}

%% file: Image/demon_attack.tex
\begin{tikzpicture}

\definecolor{darkcyan1115178}{RGB}{1,115,178}
\definecolor{darkorange2221435}{RGB}{222,143,5}
\definecolor{darkslategray38}{RGB}{38,38,38}
\definecolor{lavender234234242}{RGB}{234,234,242}
\definecolor{lightgray204}{RGB}{204,204,204}
\definecolor{orchid204120188}{RGB}{204,120,188}
\definecolor{peru20214597}{RGB}{202,145,97}

\begin{axis}[
axis background/.style={fill=lavender234234242},
axis line style={white},
legend cell align={left},
legend style={
  fill opacity=0.8,
  draw opacity=1,
  text opacity=1,
  at={(0.03,0.97)},
  anchor=north west,
  draw=lightgray204,
  fill=lavender234234242
},
tick align=outside,
x grid style={white},
xlabel=\textcolor{darkslategray38}{Time Step},
xmajorgrids,
xmajorticks=true,
xmin=-30000, xmax=410000,
xtick style={color=darkslategray38},
xtick={0,50000,100000,150000,200000,250000,300000,350000,400000},
xticklabels={0k,50k,100k,150k,200k,250k,300k,350k,400k},
y grid style={white},
ylabel=\textcolor{darkslategray38}{Performance},
ymajorgrids,
ymajorticks=true,
ymin=-1113.52381722838, ymax=9616.92381722837,
ytick style={color=darkslategray38}
]
\path [draw=darkcyan1115178, fill=darkcyan1115178, opacity=0.2]
(axis cs:-10000,892.976197480341)
--(axis cs:-10000,-625.776197480341)
--(axis cs:0,-558.776197480341)
--(axis cs:10000,-433.776197480341)
--(axis cs:20000,-310.776197480341)
--(axis cs:30000,-82.9761974803414)
--(axis cs:40000,98.4238025196586)
--(axis cs:50000,539.023802519659)
--(axis cs:60000,419.323802519659)
--(axis cs:70000,436.223802519659)
--(axis cs:80000,746.023802519659)
--(axis cs:90000,1188.52380251966)
--(axis cs:100000,1465.22380251966)
--(axis cs:110000,1565.02380251966)
--(axis cs:120000,1446.82380251966)
--(axis cs:130000,1668.52380251966)
--(axis cs:140000,1966.02380251966)
--(axis cs:150000,2217.62380251966)
--(axis cs:160000,2326.12380251966)
--(axis cs:170000,2098.62380251966)
--(axis cs:180000,2870.02380251966)
--(axis cs:190000,3128.02380251966)
--(axis cs:200000,3847.32380251966)
--(axis cs:210000,3909.82380251966)
--(axis cs:220000,3928.02380251966)
--(axis cs:230000,4538.12380251966)
--(axis cs:240000,4787.42380251966)
--(axis cs:250000,4193.32380251966)
--(axis cs:260000,5140.72380251966)
--(axis cs:270000,5428.52380251966)
--(axis cs:280000,6010.22380251966)
--(axis cs:290000,5012.82380251966)
--(axis cs:300000,6058.92380251966)
--(axis cs:310000,5117.72380251966)
--(axis cs:320000,5314.82380251966)
--(axis cs:330000,6391.02380251966)
--(axis cs:340000,6049.02380251966)
--(axis cs:350000,6441.52380251966)
--(axis cs:360000,6601.62380251966)
--(axis cs:370000,6333.72380251966)
--(axis cs:380000,7374.72380251966)
--(axis cs:390000,7610.42380251966)
--(axis cs:390000,9129.17619748034)
--(axis cs:390000,9129.17619748034)
--(axis cs:380000,8893.47619748034)
--(axis cs:370000,7852.47619748034)
--(axis cs:360000,8120.37619748034)
--(axis cs:350000,7960.27619748034)
--(axis cs:340000,7567.77619748034)
--(axis cs:330000,7909.77619748034)
--(axis cs:320000,6833.57619748034)
--(axis cs:310000,6636.47619748034)
--(axis cs:300000,7577.67619748034)
--(axis cs:290000,6531.57619748034)
--(axis cs:280000,7528.97619748034)
--(axis cs:270000,6947.27619748034)
--(axis cs:260000,6659.47619748034)
--(axis cs:250000,5712.07619748034)
--(axis cs:240000,6306.17619748034)
--(axis cs:230000,6056.87619748034)
--(axis cs:220000,5446.77619748034)
--(axis cs:210000,5428.57619748034)
--(axis cs:200000,5366.07619748034)
--(axis cs:190000,4646.77619748034)
--(axis cs:180000,4388.77619748034)
--(axis cs:170000,3617.37619748034)
--(axis cs:160000,3844.87619748034)
--(axis cs:150000,3736.37619748034)
--(axis cs:140000,3484.77619748034)
--(axis cs:130000,3187.27619748034)
--(axis cs:120000,2965.57619748034)
--(axis cs:110000,3083.77619748034)
--(axis cs:100000,2983.97619748034)
--(axis cs:90000,2707.27619748034)
--(axis cs:80000,2264.77619748034)
--(axis cs:70000,1954.97619748034)
--(axis cs:60000,1938.07619748034)
--(axis cs:50000,2057.77619748034)
--(axis cs:40000,1617.17619748034)
--(axis cs:30000,1435.77619748034)
--(axis cs:20000,1207.97619748034)
--(axis cs:10000,1084.97619748034)
--(axis cs:0,959.976197480341)
--(axis cs:-10000,892.976197480341)
--cycle;

\path [draw=darkorange2221435, fill=darkorange2221435, opacity=0.2]
(axis cs:-10000,449.608768376134)
--(axis cs:-10000,-204.408768376134)
--(axis cs:0,-155.308768376134)
--(axis cs:10000,41.9912316238657)
--(axis cs:20000,108.191231623866)
--(axis cs:30000,267.291231623866)
--(axis cs:40000,443.791231623866)
--(axis cs:50000,580.591231623866)
--(axis cs:60000,482.391231623866)
--(axis cs:70000,815.391231623866)
--(axis cs:80000,880.491231623866)
--(axis cs:90000,978.891231623866)
--(axis cs:100000,1155.59123162387)
--(axis cs:110000,1284.19123162387)
--(axis cs:120000,1025.39123162387)
--(axis cs:130000,1161.09123162387)
--(axis cs:140000,1274.19123162387)
--(axis cs:150000,1342.69123162387)
--(axis cs:160000,1523.29123162387)
--(axis cs:170000,1657.49123162387)
--(axis cs:180000,1490.59123162387)
--(axis cs:190000,1478.99123162387)
--(axis cs:200000,2219.99123162387)
--(axis cs:210000,2491.99123162387)
--(axis cs:220000,2190.59123162387)
--(axis cs:230000,1838.19123162387)
--(axis cs:240000,2265.19123162387)
--(axis cs:250000,2683.89123162387)
--(axis cs:260000,2350.39123162387)
--(axis cs:270000,2838.29123162387)
--(axis cs:280000,3262.19123162387)
--(axis cs:290000,2374.99123162387)
--(axis cs:300000,3270.99123162387)
--(axis cs:310000,3448.59123162387)
--(axis cs:320000,3675.99123162387)
--(axis cs:330000,4070.39123162387)
--(axis cs:340000,4014.99123162387)
--(axis cs:350000,4138.99123162387)
--(axis cs:360000,4765.99123162387)
--(axis cs:370000,4406.79123162387)
--(axis cs:380000,4673.79123162387)
--(axis cs:390000,4841.39123162387)
--(axis cs:390000,5495.40876837613)
--(axis cs:390000,5495.40876837613)
--(axis cs:380000,5327.80876837613)
--(axis cs:370000,5060.80876837613)
--(axis cs:360000,5420.00876837613)
--(axis cs:350000,4793.00876837613)
--(axis cs:340000,4669.00876837613)
--(axis cs:330000,4724.40876837613)
--(axis cs:320000,4330.00876837613)
--(axis cs:310000,4102.60876837613)
--(axis cs:300000,3925.00876837613)
--(axis cs:290000,3029.00876837613)
--(axis cs:280000,3916.20876837613)
--(axis cs:270000,3492.30876837613)
--(axis cs:260000,3004.40876837613)
--(axis cs:250000,3337.90876837613)
--(axis cs:240000,2919.20876837613)
--(axis cs:230000,2492.20876837613)
--(axis cs:220000,2844.60876837613)
--(axis cs:210000,3146.00876837613)
--(axis cs:200000,2874.00876837613)
--(axis cs:190000,2133.00876837613)
--(axis cs:180000,2144.60876837613)
--(axis cs:170000,2311.50876837613)
--(axis cs:160000,2177.30876837613)
--(axis cs:150000,1996.70876837613)
--(axis cs:140000,1928.20876837613)
--(axis cs:130000,1815.10876837613)
--(axis cs:120000,1679.40876837613)
--(axis cs:110000,1938.20876837613)
--(axis cs:100000,1809.60876837613)
--(axis cs:90000,1632.90876837613)
--(axis cs:80000,1534.50876837613)
--(axis cs:70000,1469.40876837613)
--(axis cs:60000,1136.40876837613)
--(axis cs:50000,1234.60876837613)
--(axis cs:40000,1097.80876837613)
--(axis cs:30000,921.308768376134)
--(axis cs:20000,762.208768376134)
--(axis cs:10000,696.008768376134)
--(axis cs:0,498.708768376134)
--(axis cs:-10000,449.608768376134)
--cycle;

\path [draw=orchid204120188, fill=orchid204120188, opacity=0.2]
(axis cs:-10000,779.43996275062)
--(axis cs:-10000,-538.039962750619)
--(axis cs:0,-381.639962750619)
--(axis cs:10000,-344.139962750619)
--(axis cs:20000,-183.739962750619)
--(axis cs:30000,-94.9399627506195)
--(axis cs:40000,58.6600372493805)
--(axis cs:50000,308.960037249381)
--(axis cs:60000,526.160037249381)
--(axis cs:70000,632.760037249381)
--(axis cs:80000,1041.46003724938)
--(axis cs:90000,1548.66003724938)
--(axis cs:100000,1526.86003724938)
--(axis cs:110000,1956.76003724938)
--(axis cs:120000,2136.76003724938)
--(axis cs:130000,2982.66003724938)
--(axis cs:140000,2366.46003724938)
--(axis cs:150000,3587.76003724938)
--(axis cs:160000,3722.86003724938)
--(axis cs:170000,3737.56003724938)
--(axis cs:180000,4472.66003724938)
--(axis cs:190000,4656.66003724938)
--(axis cs:200000,2820.76003724938)
--(axis cs:210000,3270.86003724938)
--(axis cs:220000,3809.36003724938)
--(axis cs:230000,4112.86003724938)
--(axis cs:240000,3464.26003724938)
--(axis cs:250000,3799.36003724938)
--(axis cs:260000,4235.06003724938)
--(axis cs:270000,4895.86003724938)
--(axis cs:280000,4834.36003724938)
--(axis cs:290000,5334.46003724938)
--(axis cs:300000,5280.06003724938)
--(axis cs:310000,5814.16003724938)
--(axis cs:320000,4912.16003724938)
--(axis cs:330000,5771.56003724938)
--(axis cs:340000,6014.16003724938)
--(axis cs:350000,6421.56003724938)
--(axis cs:360000,4937.86003724938)
--(axis cs:370000,5146.16003724938)
--(axis cs:380000,6381.96003724938)
--(axis cs:390000,6308.56003724938)
--(axis cs:390000,7626.03996275062)
--(axis cs:390000,7626.03996275062)
--(axis cs:380000,7699.43996275062)
--(axis cs:370000,6463.63996275062)
--(axis cs:360000,6255.33996275062)
--(axis cs:350000,7739.03996275062)
--(axis cs:340000,7331.63996275062)
--(axis cs:330000,7089.03996275062)
--(axis cs:320000,6229.63996275062)
--(axis cs:310000,7131.63996275062)
--(axis cs:300000,6597.53996275062)
--(axis cs:290000,6651.93996275062)
--(axis cs:280000,6151.83996275062)
--(axis cs:270000,6213.33996275062)
--(axis cs:260000,5552.53996275062)
--(axis cs:250000,5116.83996275062)
--(axis cs:240000,4781.73996275062)
--(axis cs:230000,5430.33996275062)
--(axis cs:220000,5126.83996275062)
--(axis cs:210000,4588.33996275062)
--(axis cs:200000,4138.23996275062)
--(axis cs:190000,5974.13996275062)
--(axis cs:180000,5790.13996275062)
--(axis cs:170000,5055.03996275062)
--(axis cs:160000,5040.33996275062)
--(axis cs:150000,4905.23996275062)
--(axis cs:140000,3683.93996275062)
--(axis cs:130000,4300.13996275062)
--(axis cs:120000,3454.23996275062)
--(axis cs:110000,3274.23996275062)
--(axis cs:100000,2844.33996275062)
--(axis cs:90000,2866.13996275062)
--(axis cs:80000,2358.93996275062)
--(axis cs:70000,1950.23996275062)
--(axis cs:60000,1843.63996275062)
--(axis cs:50000,1626.43996275062)
--(axis cs:40000,1376.13996275062)
--(axis cs:30000,1222.53996275062)
--(axis cs:20000,1133.73996275062)
--(axis cs:10000,973.33996275062)
--(axis cs:0,935.83996275062)
--(axis cs:-10000,779.43996275062)
--cycle;

\path [draw=peru20214597, fill=peru20214597, opacity=0.2]
(axis cs:-10000,503.438977540963)
--(axis cs:-10000,-253.838977540963)
--(axis cs:0,-363.738977540963)
--(axis cs:10000,-318.238977540963)
--(axis cs:20000,-279.738977540963)
--(axis cs:30000,-204.438977540963)
--(axis cs:40000,-139.938977540963)
--(axis cs:50000,-41.038977540963)
--(axis cs:60000,40.761022459037)
--(axis cs:70000,65.961022459037)
--(axis cs:80000,204.761022459037)
--(axis cs:90000,370.261022459037)
--(axis cs:100000,306.461022459037)
--(axis cs:110000,444.461022459037)
--(axis cs:120000,696.861022459037)
--(axis cs:130000,672.761022459037)
--(axis cs:140000,743.461022459037)
--(axis cs:150000,947.761022459037)
--(axis cs:160000,1076.76102245904)
--(axis cs:170000,1078.46102245904)
--(axis cs:180000,1156.06102245904)
--(axis cs:190000,1201.26102245904)
--(axis cs:200000,564.461022459037)
--(axis cs:210000,627.961022459037)
--(axis cs:220000,980.261022459037)
--(axis cs:230000,834.961022459037)
--(axis cs:240000,1478.36102245904)
--(axis cs:250000,1183.06102245904)
--(axis cs:260000,1065.26102245904)
--(axis cs:270000,1296.76102245904)
--(axis cs:280000,1386.26102245904)
--(axis cs:290000,1357.36102245904)
--(axis cs:300000,1896.26102245904)
--(axis cs:310000,2153.46102245904)
--(axis cs:320000,1935.66102245904)
--(axis cs:330000,2187.36102245904)
--(axis cs:340000,2116.46102245904)
--(axis cs:350000,2111.26102245904)
--(axis cs:360000,2201.26102245904)
--(axis cs:370000,2661.46102245904)
--(axis cs:380000,2794.26102245904)
--(axis cs:390000,2819.06102245904)
--(axis cs:390000,3576.33897754096)
--(axis cs:390000,3576.33897754096)
--(axis cs:380000,3551.53897754096)
--(axis cs:370000,3418.73897754096)
--(axis cs:360000,2958.53897754096)
--(axis cs:350000,2868.53897754096)
--(axis cs:340000,2873.73897754096)
--(axis cs:330000,2944.63897754096)
--(axis cs:320000,2692.93897754096)
--(axis cs:310000,2910.73897754096)
--(axis cs:300000,2653.53897754096)
--(axis cs:290000,2114.63897754096)
--(axis cs:280000,2143.53897754096)
--(axis cs:270000,2054.03897754096)
--(axis cs:260000,1822.53897754096)
--(axis cs:250000,1940.33897754096)
--(axis cs:240000,2235.63897754096)
--(axis cs:230000,1592.23897754096)
--(axis cs:220000,1737.53897754096)
--(axis cs:210000,1385.23897754096)
--(axis cs:200000,1321.73897754096)
--(axis cs:190000,1958.53897754096)
--(axis cs:180000,1913.33897754096)
--(axis cs:170000,1835.73897754096)
--(axis cs:160000,1834.03897754096)
--(axis cs:150000,1705.03897754096)
--(axis cs:140000,1500.73897754096)
--(axis cs:130000,1430.03897754096)
--(axis cs:120000,1454.13897754096)
--(axis cs:110000,1201.73897754096)
--(axis cs:100000,1063.73897754096)
--(axis cs:90000,1127.53897754096)
--(axis cs:80000,962.038977540963)
--(axis cs:70000,823.238977540963)
--(axis cs:60000,798.038977540963)
--(axis cs:50000,716.238977540963)
--(axis cs:40000,617.338977540963)
--(axis cs:30000,552.838977540963)
--(axis cs:20000,477.538977540963)
--(axis cs:10000,439.038977540963)
--(axis cs:0,393.538977540963)
--(axis cs:-10000,503.438977540963)
--cycle;

\addplot [semithick, darkcyan1115178, mark=square*, mark size=1.5, mark options={solid}]
table {%
-10000 133.6
0 200.6
10000 325.6
20000 448.6
30000 676.4
40000 857.8
50000 1298.4
60000 1178.7
70000 1195.6
80000 1505.4
90000 1947.9
100000 2224.6
110000 2324.4
120000 2206.2
130000 2427.9
140000 2725.4
150000 2977
160000 3085.5
170000 2858
180000 3629.4
190000 3887.4
200000 4606.7
210000 4669.2
220000 4687.4
230000 5297.5
240000 5546.8
250000 4952.7
260000 5900.1
270000 6187.9
280000 6769.6
290000 5772.2
300000 6818.3
310000 5877.1
320000 6074.2
330000 7150.4
340000 6808.4
350000 7200.9
360000 7361
370000 7093.1
380000 8134.1
390000 8369.8
};
\addplot [semithick, darkorange2221435, mark=triangle*, mark size=1.5, mark options={solid}]
table {%
-10000 122.6
0 171.7
10000 369
20000 435.2
30000 594.3
40000 770.8
50000 907.6
60000 809.4
70000 1142.4
80000 1207.5
90000 1305.9
100000 1482.6
110000 1611.2
120000 1352.4
130000 1488.1
140000 1601.2
150000 1669.7
160000 1850.3
170000 1984.5
180000 1817.6
190000 1806
200000 2547
210000 2819
220000 2517.6
230000 2165.2
240000 2592.2
250000 3010.9
260000 2677.4
270000 3165.3
280000 3589.2
290000 2702
300000 3598
310000 3775.6
320000 4003
330000 4397.4
340000 4342
350000 4466
360000 5093
370000 4733.8
380000 5000.8
390000 5168.4
};
\addplot [semithick, orchid204120188, mark=+, mark size=1.5, mark options={solid}]
table {%
-10000 120.7
0 277.1
10000 314.6
20000 475
30000 563.8
40000 717.4
50000 967.7
60000 1184.9
70000 1291.5
80000 1700.2
90000 2207.4
100000 2185.6
110000 2615.5
120000 2795.5
130000 3641.4
140000 3025.2
150000 4246.5
160000 4381.6
170000 4396.3
180000 5131.4
190000 5315.4
200000 3479.5
210000 3929.6
220000 4468.1
230000 4771.6
240000 4123
250000 4458.1
260000 4893.8
270000 5554.6
280000 5493.1
290000 5993.2
300000 5938.8
310000 6472.9
320000 5570.9
330000 6430.3
340000 6672.9
350000 7080.3
360000 5596.6
370000 5804.9
380000 7040.7
390000 6967.3
};
\addplot [semithick, peru20214597, mark=diamond*, mark size=1.5, mark options={solid}]
table {%
-10000 124.8
0 14.9
10000 60.4
20000 98.9
30000 174.2
40000 238.7
50000 337.6
60000 419.4
70000 444.6
80000 583.4
90000 748.9
100000 685.1
110000 823.1
120000 1075.5
130000 1051.4
140000 1122.1
150000 1326.4
160000 1455.4
170000 1457.1
180000 1534.7
190000 1579.9
200000 943.1
210000 1006.6
220000 1358.9
230000 1213.6
240000 1857
250000 1561.7
260000 1443.9
270000 1675.4
280000 1764.9
290000 1736
300000 2274.9
310000 2532.1
320000 2314.3
330000 2566
340000 2495.1
350000 2489.9
360000 2579.9
370000 3040.1
380000 3172.9
390000 3197.7
};
\end{axis}

\end{tikzpicture}

%% file: Image/freeway.tex
\begin{tikzpicture}

\definecolor{darkcyan1115178}{RGB}{1,115,178}
\definecolor{darkorange2221435}{RGB}{222,143,5}
\definecolor{darkslategray38}{RGB}{38,38,38}
\definecolor{lavender234234242}{RGB}{234,234,242}
\definecolor{lightgray204}{RGB}{204,204,204}
\definecolor{orchid204120188}{RGB}{204,120,188}
\definecolor{peru20214597}{RGB}{202,145,97}

\begin{axis}[
axis background/.style={fill=lavender234234242},
axis line style={white},
legend cell align={left},
legend style={
  fill opacity=0.8,
  draw opacity=1,
  text opacity=1,
  at={(0.97,0.03)},
  anchor=south east,
  draw=lightgray204,
  fill=lavender234234242
},
tick align=outside,
x grid style={white},
xlabel=\textcolor{darkslategray38}{Time Step},
xmajorgrids,
xmajorticks=true,
xmin=-30000, xmax=410000,
xtick style={color=darkslategray38},
xtick={0,50000,100000,150000,200000,250000,300000,350000,400000},
xticklabels={0k,50k,100k,150k,200k,250k,300k,350k,400k},
y grid style={white},
ylabel=\textcolor{darkslategray38}{Performance},
ymajorgrids,
ymajorticks=true,
ymin=-3.96966977379445, ymax=36.8536760020196,
ytick style={color=darkslategray38}
]
\path [draw=darkcyan1115178, fill=darkcyan1115178, opacity=0.2]
(axis cs:-10000,2.59806937584623)
--(axis cs:-10000,-0.398069375846227)
--(axis cs:0,20.0019306241538)
--(axis cs:10000,24.5019306241538)
--(axis cs:20000,26.8019306241538)
--(axis cs:30000,27.4019306241538)
--(axis cs:40000,28.4019306241538)
--(axis cs:50000,29.3019306241538)
--(axis cs:60000,28.7019306241538)
--(axis cs:70000,29.2019306241538)
--(axis cs:80000,29.4019306241538)
--(axis cs:90000,29.8019306241538)
--(axis cs:100000,29.6019306241538)
--(axis cs:110000,30.2019306241538)
--(axis cs:120000,30.3019306241538)
--(axis cs:130000,30.4019306241538)
--(axis cs:140000,30.3019306241538)
--(axis cs:150000,30.7019306241538)
--(axis cs:160000,31.0019306241538)
--(axis cs:170000,31.0019306241538)
--(axis cs:180000,31.3019306241538)
--(axis cs:190000,31.1019306241538)
--(axis cs:200000,31.5019306241538)
--(axis cs:210000,31.4019306241538)
--(axis cs:220000,31.2019306241538)
--(axis cs:230000,31.2019306241538)
--(axis cs:240000,31.8019306241538)
--(axis cs:250000,31.8019306241538)
--(axis cs:260000,31.7019306241538)
--(axis cs:270000,31.9019306241538)
--(axis cs:280000,31.8019306241538)
--(axis cs:290000,32.0019306241538)
--(axis cs:300000,31.7019306241538)
--(axis cs:310000,32.0019306241538)
--(axis cs:320000,31.9019306241538)
--(axis cs:330000,31.7019306241538)
--(axis cs:340000,32.0019306241538)
--(axis cs:350000,32.0019306241538)
--(axis cs:360000,31.8019306241538)
--(axis cs:370000,32.0019306241538)
--(axis cs:380000,31.9019306241538)
--(axis cs:390000,32.0019306241538)
--(axis cs:390000,34.9980693758462)
--(axis cs:390000,34.9980693758462)
--(axis cs:380000,34.8980693758462)
--(axis cs:370000,34.9980693758462)
--(axis cs:360000,34.7980693758462)
--(axis cs:350000,34.9980693758462)
--(axis cs:340000,34.9980693758462)
--(axis cs:330000,34.6980693758462)
--(axis cs:320000,34.8980693758462)
--(axis cs:310000,34.9980693758462)
--(axis cs:300000,34.6980693758462)
--(axis cs:290000,34.9980693758462)
--(axis cs:280000,34.7980693758462)
--(axis cs:270000,34.8980693758462)
--(axis cs:260000,34.6980693758462)
--(axis cs:250000,34.7980693758462)
--(axis cs:240000,34.7980693758462)
--(axis cs:230000,34.1980693758462)
--(axis cs:220000,34.1980693758462)
--(axis cs:210000,34.3980693758462)
--(axis cs:200000,34.4980693758462)
--(axis cs:190000,34.0980693758462)
--(axis cs:180000,34.2980693758462)
--(axis cs:170000,33.9980693758462)
--(axis cs:160000,33.9980693758462)
--(axis cs:150000,33.6980693758462)
--(axis cs:140000,33.2980693758462)
--(axis cs:130000,33.3980693758462)
--(axis cs:120000,33.2980693758462)
--(axis cs:110000,33.1980693758462)
--(axis cs:100000,32.5980693758462)
--(axis cs:90000,32.7980693758462)
--(axis cs:80000,32.3980693758462)
--(axis cs:70000,32.1980693758462)
--(axis cs:60000,31.6980693758462)
--(axis cs:50000,32.2980693758462)
--(axis cs:40000,31.3980693758462)
--(axis cs:30000,30.3980693758462)
--(axis cs:20000,29.7980693758462)
--(axis cs:10000,27.4980693758462)
--(axis cs:0,22.9980693758462)
--(axis cs:-10000,2.59806937584623)
--cycle;

\path [draw=darkorange2221435, fill=darkorange2221435, opacity=0.2]
(axis cs:-10000,1.81141000450743)
--(axis cs:-10000,-0.411410004507434)
--(axis cs:0,20.2885899954926)
--(axis cs:10000,20.6885899954926)
--(axis cs:20000,22.6885899954926)
--(axis cs:30000,25.5885899954926)
--(axis cs:40000,27.0885899954926)
--(axis cs:50000,29.1885899954926)
--(axis cs:60000,29.4885899954926)
--(axis cs:70000,29.4885899954926)
--(axis cs:80000,29.7885899954926)
--(axis cs:90000,30.1885899954926)
--(axis cs:100000,30.1885899954926)
--(axis cs:110000,30.3885899954926)
--(axis cs:120000,30.8885899954926)
--(axis cs:130000,30.7885899954926)
--(axis cs:140000,30.4885899954926)
--(axis cs:150000,30.7885899954926)
--(axis cs:160000,31.0885899954926)
--(axis cs:170000,31.0885899954926)
--(axis cs:180000,31.1885899954926)
--(axis cs:190000,31.5885899954926)
--(axis cs:200000,31.3885899954926)
--(axis cs:210000,31.7885899954926)
--(axis cs:220000,31.5885899954926)
--(axis cs:230000,31.5885899954926)
--(axis cs:240000,31.7885899954926)
--(axis cs:250000,31.7885899954926)
--(axis cs:260000,31.8885899954926)
--(axis cs:270000,31.9885899954926)
--(axis cs:280000,31.9885899954926)
--(axis cs:290000,31.8885899954926)
--(axis cs:300000,31.9885899954926)
--(axis cs:310000,32.0885899954926)
--(axis cs:320000,32.0885899954926)
--(axis cs:330000,32.0885899954926)
--(axis cs:340000,31.9885899954926)
--(axis cs:350000,32.0885899954926)
--(axis cs:360000,31.9885899954926)
--(axis cs:370000,32.0885899954926)
--(axis cs:380000,32.0885899954926)
--(axis cs:390000,32.1885899954926)
--(axis cs:390000,34.4114100045074)
--(axis cs:390000,34.4114100045074)
--(axis cs:380000,34.3114100045074)
--(axis cs:370000,34.3114100045074)
--(axis cs:360000,34.2114100045074)
--(axis cs:350000,34.3114100045074)
--(axis cs:340000,34.2114100045074)
--(axis cs:330000,34.3114100045074)
--(axis cs:320000,34.3114100045074)
--(axis cs:310000,34.3114100045074)
--(axis cs:300000,34.2114100045074)
--(axis cs:290000,34.1114100045074)
--(axis cs:280000,34.2114100045074)
--(axis cs:270000,34.2114100045074)
--(axis cs:260000,34.1114100045074)
--(axis cs:250000,34.0114100045074)
--(axis cs:240000,34.0114100045074)
--(axis cs:230000,33.8114100045074)
--(axis cs:220000,33.8114100045074)
--(axis cs:210000,34.0114100045074)
--(axis cs:200000,33.6114100045074)
--(axis cs:190000,33.8114100045074)
--(axis cs:180000,33.4114100045074)
--(axis cs:170000,33.3114100045074)
--(axis cs:160000,33.3114100045074)
--(axis cs:150000,33.0114100045074)
--(axis cs:140000,32.7114100045074)
--(axis cs:130000,33.0114100045074)
--(axis cs:120000,33.1114100045074)
--(axis cs:110000,32.6114100045074)
--(axis cs:100000,32.4114100045074)
--(axis cs:90000,32.4114100045074)
--(axis cs:80000,32.0114100045074)
--(axis cs:70000,31.7114100045074)
--(axis cs:60000,31.7114100045074)
--(axis cs:50000,31.4114100045074)
--(axis cs:40000,29.3114100045074)
--(axis cs:30000,27.8114100045074)
--(axis cs:20000,24.9114100045074)
--(axis cs:10000,22.9114100045074)
--(axis cs:0,22.5114100045074)
--(axis cs:-10000,1.81141000450743)
--cycle;

\path [draw=orchid204120188, fill=orchid204120188, opacity=0.2]
(axis cs:-10000,2.04740401392838)
--(axis cs:-10000,-0.847404013928381)
--(axis cs:0,21.5525959860716)
--(axis cs:10000,26.0525959860716)
--(axis cs:20000,27.3525959860716)
--(axis cs:30000,29.0525959860716)
--(axis cs:40000,29.2525959860716)
--(axis cs:50000,30.0525959860716)
--(axis cs:60000,29.8525959860716)
--(axis cs:70000,30.1525959860716)
--(axis cs:80000,30.0525959860716)
--(axis cs:90000,30.2525959860716)
--(axis cs:100000,30.2525959860716)
--(axis cs:110000,30.3525959860716)
--(axis cs:120000,30.5525959860716)
--(axis cs:130000,30.6525959860716)
--(axis cs:140000,30.7525959860716)
--(axis cs:150000,30.7525959860716)
--(axis cs:160000,30.6525959860716)
--(axis cs:170000,30.6525959860716)
--(axis cs:180000,30.6525959860716)
--(axis cs:190000,30.7525959860716)
--(axis cs:200000,30.6525959860716)
--(axis cs:210000,30.7525959860716)
--(axis cs:220000,30.8525959860716)
--(axis cs:230000,31.0525959860716)
--(axis cs:240000,31.1525959860716)
--(axis cs:250000,31.1525959860716)
--(axis cs:260000,31.2525959860716)
--(axis cs:270000,31.2525959860716)
--(axis cs:280000,31.3525959860716)
--(axis cs:290000,31.5525959860716)
--(axis cs:300000,31.5525959860716)
--(axis cs:310000,31.5525959860716)
--(axis cs:320000,31.6525959860716)
--(axis cs:330000,31.8525959860716)
--(axis cs:340000,31.7525959860716)
--(axis cs:350000,31.6525959860716)
--(axis cs:360000,31.6525959860716)
--(axis cs:370000,31.7525959860716)
--(axis cs:380000,31.7525959860716)
--(axis cs:390000,31.6525959860716)
--(axis cs:390000,34.5474040139284)
--(axis cs:390000,34.5474040139284)
--(axis cs:380000,34.6474040139284)
--(axis cs:370000,34.6474040139284)
--(axis cs:360000,34.5474040139284)
--(axis cs:350000,34.5474040139284)
--(axis cs:340000,34.6474040139284)
--(axis cs:330000,34.7474040139284)
--(axis cs:320000,34.5474040139284)
--(axis cs:310000,34.4474040139284)
--(axis cs:300000,34.4474040139284)
--(axis cs:290000,34.4474040139284)
--(axis cs:280000,34.2474040139284)
--(axis cs:270000,34.1474040139284)
--(axis cs:260000,34.1474040139284)
--(axis cs:250000,34.0474040139284)
--(axis cs:240000,34.0474040139284)
--(axis cs:230000,33.9474040139284)
--(axis cs:220000,33.7474040139284)
--(axis cs:210000,33.6474040139284)
--(axis cs:200000,33.5474040139284)
--(axis cs:190000,33.6474040139284)
--(axis cs:180000,33.5474040139284)
--(axis cs:170000,33.5474040139284)
--(axis cs:160000,33.5474040139284)
--(axis cs:150000,33.6474040139284)
--(axis cs:140000,33.6474040139284)
--(axis cs:130000,33.5474040139284)
--(axis cs:120000,33.4474040139284)
--(axis cs:110000,33.2474040139284)
--(axis cs:100000,33.1474040139284)
--(axis cs:90000,33.1474040139284)
--(axis cs:80000,32.9474040139284)
--(axis cs:70000,33.0474040139284)
--(axis cs:60000,32.7474040139284)
--(axis cs:50000,32.9474040139284)
--(axis cs:40000,32.1474040139284)
--(axis cs:30000,31.9474040139284)
--(axis cs:20000,30.2474040139284)
--(axis cs:10000,28.9474040139284)
--(axis cs:0,24.4474040139284)
--(axis cs:-10000,2.04740401392838)
--cycle;

\path [draw=peru20214597, fill=peru20214597, opacity=0.2]
(axis cs:-10000,2.91406314762108)
--(axis cs:-10000,-1.51406314762108)
--(axis cs:0,-2.11406314762108)
--(axis cs:10000,0.185936852378917)
--(axis cs:20000,4.98593685237892)
--(axis cs:30000,14.8859368523789)
--(axis cs:40000,21.7859368523789)
--(axis cs:50000,21.0859368523789)
--(axis cs:60000,22.8859368523789)
--(axis cs:70000,24.5859368523789)
--(axis cs:80000,24.5859368523789)
--(axis cs:90000,25.4859368523789)
--(axis cs:100000,25.7859368523789)
--(axis cs:110000,25.8859368523789)
--(axis cs:120000,26.3859368523789)
--(axis cs:130000,27.1859368523789)
--(axis cs:140000,26.4859368523789)
--(axis cs:150000,26.7859368523789)
--(axis cs:160000,26.5859368523789)
--(axis cs:170000,26.7859368523789)
--(axis cs:180000,26.6859368523789)
--(axis cs:190000,26.7859368523789)
--(axis cs:200000,23.4859368523789)
--(axis cs:210000,23.3859368523789)
--(axis cs:220000,23.5859368523789)
--(axis cs:230000,23.4859368523789)
--(axis cs:240000,23.7859368523789)
--(axis cs:250000,23.6859368523789)
--(axis cs:260000,23.9859368523789)
--(axis cs:270000,24.1859368523789)
--(axis cs:280000,24.1859368523789)
--(axis cs:290000,23.8859368523789)
--(axis cs:300000,24.2859368523789)
--(axis cs:310000,24.3859368523789)
--(axis cs:320000,24.2859368523789)
--(axis cs:330000,24.4859368523789)
--(axis cs:340000,25.0859368523789)
--(axis cs:350000,25.6859368523789)
--(axis cs:360000,25.3859368523789)
--(axis cs:370000,25.8859368523789)
--(axis cs:380000,26.0859368523789)
--(axis cs:390000,26.5859368523789)
--(axis cs:390000,31.0140631476211)
--(axis cs:390000,31.0140631476211)
--(axis cs:380000,30.5140631476211)
--(axis cs:370000,30.3140631476211)
--(axis cs:360000,29.8140631476211)
--(axis cs:350000,30.1140631476211)
--(axis cs:340000,29.5140631476211)
--(axis cs:330000,28.9140631476211)
--(axis cs:320000,28.7140631476211)
--(axis cs:310000,28.8140631476211)
--(axis cs:300000,28.7140631476211)
--(axis cs:290000,28.3140631476211)
--(axis cs:280000,28.6140631476211)
--(axis cs:270000,28.6140631476211)
--(axis cs:260000,28.4140631476211)
--(axis cs:250000,28.1140631476211)
--(axis cs:240000,28.2140631476211)
--(axis cs:230000,27.9140631476211)
--(axis cs:220000,28.0140631476211)
--(axis cs:210000,27.8140631476211)
--(axis cs:200000,27.9140631476211)
--(axis cs:190000,31.2140631476211)
--(axis cs:180000,31.1140631476211)
--(axis cs:170000,31.2140631476211)
--(axis cs:160000,31.0140631476211)
--(axis cs:150000,31.2140631476211)
--(axis cs:140000,30.9140631476211)
--(axis cs:130000,31.6140631476211)
--(axis cs:120000,30.8140631476211)
--(axis cs:110000,30.3140631476211)
--(axis cs:100000,30.2140631476211)
--(axis cs:90000,29.9140631476211)
--(axis cs:80000,29.0140631476211)
--(axis cs:70000,29.0140631476211)
--(axis cs:60000,27.3140631476211)
--(axis cs:50000,25.5140631476211)
--(axis cs:40000,26.2140631476211)
--(axis cs:30000,19.3140631476211)
--(axis cs:20000,9.41406314762108)
--(axis cs:10000,4.61406314762108)
--(axis cs:0,2.31406314762108)
--(axis cs:-10000,2.91406314762108)
--cycle;

\addplot [semithick, darkcyan1115178, mark=square*, mark size=1.5, mark options={solid}]
table {%
-10000 1.1
0 21.5
10000 26
20000 28.3
30000 28.9
40000 29.9
50000 30.8
60000 30.2
70000 30.7
80000 30.9
90000 31.3
100000 31.1
110000 31.7
120000 31.8
130000 31.9
140000 31.8
150000 32.2
160000 32.5
170000 32.5
180000 32.8
190000 32.6
200000 33
210000 32.9
220000 32.7
230000 32.7
240000 33.3
250000 33.3
260000 33.2
270000 33.4
280000 33.3
290000 33.5
300000 33.2
310000 33.5
320000 33.4
330000 33.2
340000 33.5
350000 33.5
360000 33.3
370000 33.5
380000 33.4
390000 33.5
};
\addplot [semithick, darkorange2221435, mark=triangle*, mark size=1.5, mark options={solid}]
table {%
-10000 0.7
0 21.4
10000 21.8
20000 23.8
30000 26.7
40000 28.2
50000 30.3
60000 30.6
70000 30.6
80000 30.9
90000 31.3
100000 31.3
110000 31.5
120000 32
130000 31.9
140000 31.6
150000 31.9
160000 32.2
170000 32.2
180000 32.3
190000 32.7
200000 32.5
210000 32.9
220000 32.7
230000 32.7
240000 32.9
250000 32.9
260000 33
270000 33.1
280000 33.1
290000 33
300000 33.1
310000 33.2
320000 33.2
330000 33.2
340000 33.1
350000 33.2
360000 33.1
370000 33.2
380000 33.2
390000 33.3
};
\addplot [semithick, orchid204120188, mark=+, mark size=1.5, mark options={solid}]
table {%
-10000 0.6
0 23
10000 27.5
20000 28.8
30000 30.5
40000 30.7
50000 31.5
60000 31.3
70000 31.6
80000 31.5
90000 31.7
100000 31.7
110000 31.8
120000 32
130000 32.1
140000 32.2
150000 32.2
160000 32.1
170000 32.1
180000 32.1
190000 32.2
200000 32.1
210000 32.2
220000 32.3
230000 32.5
240000 32.6
250000 32.6
260000 32.7
270000 32.7
280000 32.8
290000 33
300000 33
310000 33
320000 33.1
330000 33.3
340000 33.2
350000 33.1
360000 33.1
370000 33.2
380000 33.2
390000 33.1
};
\addplot [semithick, peru20214597, mark=diamond*, mark size=1.5, mark options={solid}]
table {%
-10000 0.7
0 0.1
10000 2.4
20000 7.2
30000 17.1
40000 24
50000 23.3
60000 25.1
70000 26.8
80000 26.8
90000 27.7
100000 28
110000 28.1
120000 28.6
130000 29.4
140000 28.7
150000 29
160000 28.8
170000 29
180000 28.9
190000 29
200000 25.7
210000 25.6
220000 25.8
230000 25.7
240000 26
250000 25.9
260000 26.2
270000 26.4
280000 26.4
290000 26.1
300000 26.5
310000 26.6
320000 26.5
330000 26.7
340000 27.3
350000 27.9
360000 27.6
370000 28.1
380000 28.3
390000 28.8
};
\end{axis}

\end{tikzpicture}

%% file: Image/frostbite.tex
\begin{tikzpicture}

\definecolor{darkcyan1115178}{RGB}{1,115,178}
\definecolor{darkorange2221435}{RGB}{222,143,5}
\definecolor{darkslategray38}{RGB}{38,38,38}
\definecolor{lavender234234242}{RGB}{234,234,242}
\definecolor{lightgray204}{RGB}{204,204,204}
\definecolor{orchid204120188}{RGB}{204,120,188}
\definecolor{peru20214597}{RGB}{202,145,97}

\begin{axis}[
axis background/.style={fill=lavender234234242},
axis line style={white},
legend cell align={left},
legend style={
  fill opacity=0.8,
  draw opacity=1,
  text opacity=1,
  at={(0.03,0.97)},
  anchor=north west,
  draw=lightgray204,
  fill=lavender234234242
},
tick align=outside,
x grid style={white},
xlabel=\textcolor{darkslategray38}{Time Step},
xmajorgrids,
xmajorticks=true,
xmin=-30000, xmax=410000,
xtick style={color=darkslategray38},
xtick={0,50000,100000,150000,200000,250000,300000,350000,400000},
xticklabels={0k,50k,100k,150k,200k,250k,300k,350k,400k},
y grid style={white},
ylabel=\textcolor{darkslategray38}{Performance},
ymajorgrids,
ymajorticks=true,
ymin=-535.64153328406, ymax=4205.44153328406,
ytick style={color=darkslategray38}
]
\path [draw=darkcyan1115178, fill=darkcyan1115178, opacity=0.2]
(axis cs:-10000,483.537757530964)
--(axis cs:-10000,-320.137757530964)
--(axis cs:0,-200.837757530964)
--(axis cs:10000,-10.7377575309635)
--(axis cs:20000,657.062242469037)
--(axis cs:30000,788.862242469037)
--(axis cs:40000,543.562242469036)
--(axis cs:50000,379.262242469036)
--(axis cs:60000,461.762242469036)
--(axis cs:70000,543.262242469036)
--(axis cs:80000,741.762242469036)
--(axis cs:90000,1086.26224246904)
--(axis cs:100000,1188.86224246904)
--(axis cs:110000,1220.56224246904)
--(axis cs:120000,1270.46224246904)
--(axis cs:130000,1251.46224246904)
--(axis cs:140000,1268.16224246904)
--(axis cs:150000,1357.26224246904)
--(axis cs:160000,1520.26224246904)
--(axis cs:170000,1697.76224246904)
--(axis cs:180000,1745.56224246904)
--(axis cs:190000,1667.06224246904)
--(axis cs:200000,2037.06224246904)
--(axis cs:210000,2067.26224246904)
--(axis cs:220000,1819.06224246904)
--(axis cs:230000,1864.06224246904)
--(axis cs:240000,1777.46224246904)
--(axis cs:250000,1879.86224246904)
--(axis cs:260000,2216.56224246904)
--(axis cs:270000,2306.26224246904)
--(axis cs:280000,2037.16224246904)
--(axis cs:290000,2262.76224246904)
--(axis cs:300000,2370.76224246904)
--(axis cs:310000,2548.76224246904)
--(axis cs:320000,2495.86224246904)
--(axis cs:330000,2429.26224246904)
--(axis cs:340000,2461.46224246904)
--(axis cs:350000,2413.86224246904)
--(axis cs:360000,2853.76224246904)
--(axis cs:370000,2602.16224246904)
--(axis cs:380000,2576.16224246904)
--(axis cs:390000,3186.26224246904)
--(axis cs:390000,3989.93775753096)
--(axis cs:390000,3989.93775753096)
--(axis cs:380000,3379.83775753096)
--(axis cs:370000,3405.83775753096)
--(axis cs:360000,3657.43775753096)
--(axis cs:350000,3217.53775753096)
--(axis cs:340000,3265.13775753096)
--(axis cs:330000,3232.93775753096)
--(axis cs:320000,3299.53775753096)
--(axis cs:310000,3352.43775753096)
--(axis cs:300000,3174.43775753096)
--(axis cs:290000,3066.43775753096)
--(axis cs:280000,2840.83775753096)
--(axis cs:270000,3109.93775753096)
--(axis cs:260000,3020.23775753096)
--(axis cs:250000,2683.53775753096)
--(axis cs:240000,2581.13775753096)
--(axis cs:230000,2667.73775753096)
--(axis cs:220000,2622.73775753096)
--(axis cs:210000,2870.93775753096)
--(axis cs:200000,2840.73775753096)
--(axis cs:190000,2470.73775753096)
--(axis cs:180000,2549.23775753096)
--(axis cs:170000,2501.43775753096)
--(axis cs:160000,2323.93775753096)
--(axis cs:150000,2160.93775753096)
--(axis cs:140000,2071.83775753096)
--(axis cs:130000,2055.13775753096)
--(axis cs:120000,2074.13775753096)
--(axis cs:110000,2024.23775753096)
--(axis cs:100000,1992.53775753096)
--(axis cs:90000,1889.93775753096)
--(axis cs:80000,1545.43775753096)
--(axis cs:70000,1346.93775753096)
--(axis cs:60000,1265.43775753096)
--(axis cs:50000,1182.93775753096)
--(axis cs:40000,1347.23775753096)
--(axis cs:30000,1592.53775753096)
--(axis cs:20000,1460.73775753096)
--(axis cs:10000,792.937757530964)
--(axis cs:0,602.837757530963)
--(axis cs:-10000,483.537757530964)
--cycle;

\path [draw=darkorange2221435, fill=darkorange2221435, opacity=0.2]
(axis cs:-10000,365.922127481209)
--(axis cs:-10000,-268.922127481209)
--(axis cs:0,-80.9221274812091)
--(axis cs:10000,-72.5221274812091)
--(axis cs:20000,405.277872518791)
--(axis cs:30000,971.877872518791)
--(axis cs:40000,1059.17787251879)
--(axis cs:50000,1165.87787251879)
--(axis cs:60000,1660.47787251879)
--(axis cs:70000,1602.67787251879)
--(axis cs:80000,1296.87787251879)
--(axis cs:90000,1397.07787251879)
--(axis cs:100000,1292.17787251879)
--(axis cs:110000,1370.57787251879)
--(axis cs:120000,1450.97787251879)
--(axis cs:130000,1484.67787251879)
--(axis cs:140000,1550.37787251879)
--(axis cs:150000,1555.27787251879)
--(axis cs:160000,1577.57787251879)
--(axis cs:170000,1775.77787251879)
--(axis cs:180000,1752.67787251879)
--(axis cs:190000,1707.57787251879)
--(axis cs:200000,1561.07787251879)
--(axis cs:210000,1486.97787251879)
--(axis cs:220000,1833.87787251879)
--(axis cs:230000,1745.67787251879)
--(axis cs:240000,1513.17787251879)
--(axis cs:250000,1655.37787251879)
--(axis cs:260000,1721.57787251879)
--(axis cs:270000,1750.47787251879)
--(axis cs:280000,1753.27787251879)
--(axis cs:290000,1738.57787251879)
--(axis cs:300000,1846.07787251879)
--(axis cs:310000,1678.57787251879)
--(axis cs:320000,1602.47787251879)
--(axis cs:330000,1756.97787251879)
--(axis cs:340000,1881.97787251879)
--(axis cs:350000,1871.67787251879)
--(axis cs:360000,1728.97787251879)
--(axis cs:370000,1718.67787251879)
--(axis cs:380000,2002.27787251879)
--(axis cs:390000,2103.27787251879)
--(axis cs:390000,2738.12212748121)
--(axis cs:390000,2738.12212748121)
--(axis cs:380000,2637.12212748121)
--(axis cs:370000,2353.52212748121)
--(axis cs:360000,2363.82212748121)
--(axis cs:350000,2506.52212748121)
--(axis cs:340000,2516.82212748121)
--(axis cs:330000,2391.82212748121)
--(axis cs:320000,2237.32212748121)
--(axis cs:310000,2313.42212748121)
--(axis cs:300000,2480.92212748121)
--(axis cs:290000,2373.42212748121)
--(axis cs:280000,2388.12212748121)
--(axis cs:270000,2385.32212748121)
--(axis cs:260000,2356.42212748121)
--(axis cs:250000,2290.22212748121)
--(axis cs:240000,2148.02212748121)
--(axis cs:230000,2380.52212748121)
--(axis cs:220000,2468.72212748121)
--(axis cs:210000,2121.82212748121)
--(axis cs:200000,2195.92212748121)
--(axis cs:190000,2342.42212748121)
--(axis cs:180000,2387.52212748121)
--(axis cs:170000,2410.62212748121)
--(axis cs:160000,2212.42212748121)
--(axis cs:150000,2190.12212748121)
--(axis cs:140000,2185.22212748121)
--(axis cs:130000,2119.52212748121)
--(axis cs:120000,2085.82212748121)
--(axis cs:110000,2005.42212748121)
--(axis cs:100000,1927.02212748121)
--(axis cs:90000,2031.92212748121)
--(axis cs:80000,1931.72212748121)
--(axis cs:70000,2237.52212748121)
--(axis cs:60000,2295.32212748121)
--(axis cs:50000,1800.72212748121)
--(axis cs:40000,1694.02212748121)
--(axis cs:30000,1606.72212748121)
--(axis cs:20000,1040.12212748121)
--(axis cs:10000,562.322127481209)
--(axis cs:0,553.922127481209)
--(axis cs:-10000,365.922127481209)
--cycle;

\path [draw=orchid204120188, fill=orchid204120188, opacity=0.2]
(axis cs:-10000,483.537757530964)
--(axis cs:-10000,-320.137757530964)
--(axis cs:0,-200.837757530964)
--(axis cs:10000,-10.7377575309635)
--(axis cs:20000,657.062242469037)
--(axis cs:30000,788.862242469037)
--(axis cs:40000,543.562242469036)
--(axis cs:50000,379.262242469036)
--(axis cs:60000,461.762242469036)
--(axis cs:70000,543.262242469036)
--(axis cs:80000,741.762242469036)
--(axis cs:90000,1086.26224246904)
--(axis cs:100000,1188.86224246904)
--(axis cs:110000,1220.56224246904)
--(axis cs:120000,1270.46224246904)
--(axis cs:130000,1251.46224246904)
--(axis cs:140000,1268.16224246904)
--(axis cs:150000,1357.26224246904)
--(axis cs:160000,1520.26224246904)
--(axis cs:170000,1697.76224246904)
--(axis cs:180000,1745.56224246904)
--(axis cs:190000,1667.06224246904)
--(axis cs:200000,2037.06224246904)
--(axis cs:210000,2067.26224246904)
--(axis cs:220000,1819.06224246904)
--(axis cs:230000,1864.06224246904)
--(axis cs:240000,1777.46224246904)
--(axis cs:250000,1879.86224246904)
--(axis cs:260000,2216.56224246904)
--(axis cs:270000,2306.26224246904)
--(axis cs:280000,2037.16224246904)
--(axis cs:290000,2262.76224246904)
--(axis cs:300000,2370.76224246904)
--(axis cs:310000,2548.76224246904)
--(axis cs:320000,2495.86224246904)
--(axis cs:330000,2429.26224246904)
--(axis cs:340000,2461.46224246904)
--(axis cs:350000,2413.86224246904)
--(axis cs:360000,2853.76224246904)
--(axis cs:370000,2602.16224246904)
--(axis cs:380000,2576.16224246904)
--(axis cs:390000,3186.26224246904)
--(axis cs:390000,3989.93775753096)
--(axis cs:390000,3989.93775753096)
--(axis cs:380000,3379.83775753096)
--(axis cs:370000,3405.83775753096)
--(axis cs:360000,3657.43775753096)
--(axis cs:350000,3217.53775753096)
--(axis cs:340000,3265.13775753096)
--(axis cs:330000,3232.93775753096)
--(axis cs:320000,3299.53775753096)
--(axis cs:310000,3352.43775753096)
--(axis cs:300000,3174.43775753096)
--(axis cs:290000,3066.43775753096)
--(axis cs:280000,2840.83775753096)
--(axis cs:270000,3109.93775753096)
--(axis cs:260000,3020.23775753096)
--(axis cs:250000,2683.53775753096)
--(axis cs:240000,2581.13775753096)
--(axis cs:230000,2667.73775753096)
--(axis cs:220000,2622.73775753096)
--(axis cs:210000,2870.93775753096)
--(axis cs:200000,2840.73775753096)
--(axis cs:190000,2470.73775753096)
--(axis cs:180000,2549.23775753096)
--(axis cs:170000,2501.43775753096)
--(axis cs:160000,2323.93775753096)
--(axis cs:150000,2160.93775753096)
--(axis cs:140000,2071.83775753096)
--(axis cs:130000,2055.13775753096)
--(axis cs:120000,2074.13775753096)
--(axis cs:110000,2024.23775753096)
--(axis cs:100000,1992.53775753096)
--(axis cs:90000,1889.93775753096)
--(axis cs:80000,1545.43775753096)
--(axis cs:70000,1346.93775753096)
--(axis cs:60000,1265.43775753096)
--(axis cs:50000,1182.93775753096)
--(axis cs:40000,1347.23775753096)
--(axis cs:30000,1592.53775753096)
--(axis cs:20000,1460.73775753096)
--(axis cs:10000,792.937757530964)
--(axis cs:0,602.837757530963)
--(axis cs:-10000,483.537757530964)
--cycle;

\path [draw=peru20214597, fill=peru20214597, opacity=0.2]
(axis cs:-10000,72.1419794529538)
--(axis cs:-10000,32.0580205470462)
--(axis cs:0,2.95802054704617)
--(axis cs:10000,29.3580205470462)
--(axis cs:20000,44.9580205470462)
--(axis cs:30000,56.3580205470462)
--(axis cs:40000,66.8580205470462)
--(axis cs:50000,71.3580205470462)
--(axis cs:60000,79.3580205470462)
--(axis cs:70000,81.8580205470462)
--(axis cs:80000,81.5580205470462)
--(axis cs:90000,82.0580205470462)
--(axis cs:100000,82.6580205470462)
--(axis cs:110000,80.9580205470462)
--(axis cs:120000,83.8580205470462)
--(axis cs:130000,79.3580205470462)
--(axis cs:140000,73.9580205470462)
--(axis cs:150000,76.0580205470462)
--(axis cs:160000,78.2580205470462)
--(axis cs:170000,80.6580205470462)
--(axis cs:180000,76.8580205470462)
--(axis cs:190000,72.9580205470462)
--(axis cs:200000,77.8580205470462)
--(axis cs:210000,73.3580205470462)
--(axis cs:220000,73.9580205470462)
--(axis cs:230000,80.2580205470462)
--(axis cs:240000,82.5580205470462)
--(axis cs:250000,79.8580205470462)
--(axis cs:260000,76.6580205470462)
--(axis cs:270000,75.0580205470462)
--(axis cs:280000,75.9580205470462)
--(axis cs:290000,73.8580205470462)
--(axis cs:300000,71.6580205470462)
--(axis cs:310000,83.3580205470462)
--(axis cs:320000,77.0580205470462)
--(axis cs:330000,71.9580205470462)
--(axis cs:340000,72.9580205470462)
--(axis cs:350000,72.9580205470462)
--(axis cs:360000,70.0580205470462)
--(axis cs:370000,74.0580205470462)
--(axis cs:380000,76.2580205470462)
--(axis cs:390000,69.3580205470462)
--(axis cs:390000,109.441979452954)
--(axis cs:390000,109.441979452954)
--(axis cs:380000,116.341979452954)
--(axis cs:370000,114.141979452954)
--(axis cs:360000,110.141979452954)
--(axis cs:350000,113.041979452954)
--(axis cs:340000,113.041979452954)
--(axis cs:330000,112.041979452954)
--(axis cs:320000,117.141979452954)
--(axis cs:310000,123.441979452954)
--(axis cs:300000,111.741979452954)
--(axis cs:290000,113.941979452954)
--(axis cs:280000,116.041979452954)
--(axis cs:270000,115.141979452954)
--(axis cs:260000,116.741979452954)
--(axis cs:250000,119.941979452954)
--(axis cs:240000,122.641979452954)
--(axis cs:230000,120.341979452954)
--(axis cs:220000,114.041979452954)
--(axis cs:210000,113.441979452954)
--(axis cs:200000,117.941979452954)
--(axis cs:190000,113.041979452954)
--(axis cs:180000,116.941979452954)
--(axis cs:170000,120.741979452954)
--(axis cs:160000,118.341979452954)
--(axis cs:150000,116.141979452954)
--(axis cs:140000,114.041979452954)
--(axis cs:130000,119.441979452954)
--(axis cs:120000,123.941979452954)
--(axis cs:110000,121.041979452954)
--(axis cs:100000,122.741979452954)
--(axis cs:90000,122.141979452954)
--(axis cs:80000,121.641979452954)
--(axis cs:70000,121.941979452954)
--(axis cs:60000,119.441979452954)
--(axis cs:50000,111.441979452954)
--(axis cs:40000,106.941979452954)
--(axis cs:30000,96.4419794529538)
--(axis cs:20000,85.0419794529538)
--(axis cs:10000,69.4419794529538)
--(axis cs:0,43.0419794529538)
--(axis cs:-10000,72.1419794529538)
--cycle;

\addplot [semithick, darkcyan1115178, mark=square*, mark size=1.5, mark options={solid}]
table {%
-10000 81.7
0 201
10000 391.1
20000 1058.9
30000 1190.7
40000 945.4
50000 781.1
60000 863.6
70000 945.1
80000 1143.6
90000 1488.1
100000 1590.7
110000 1622.4
120000 1672.3
130000 1653.3
140000 1670
150000 1759.1
160000 1922.1
170000 2099.6
180000 2147.4
190000 2068.9
200000 2438.9
210000 2469.1
220000 2220.9
230000 2265.9
240000 2179.3
250000 2281.7
260000 2618.4
270000 2708.1
280000 2439
290000 2664.6
300000 2772.6
310000 2950.6
320000 2897.7
330000 2831.1
340000 2863.3
350000 2815.7
360000 3255.6
370000 3004
380000 2978
390000 3588.1
};
\addplot [semithick, darkorange2221435, mark=triangle*, mark size=1.5, mark options={solid}]
table {%
-10000 48.5
0 236.5
10000 244.9
20000 722.7
30000 1289.3
40000 1376.6
50000 1483.3
60000 1977.9
70000 1920.1
80000 1614.3
90000 1714.5
100000 1609.6
110000 1688
120000 1768.4
130000 1802.1
140000 1867.8
150000 1872.7
160000 1895
170000 2093.2
180000 2070.1
190000 2025
200000 1878.5
210000 1804.4
220000 2151.3
230000 2063.1
240000 1830.6
250000 1972.8
260000 2039
270000 2067.9
280000 2070.7
290000 2056
300000 2163.5
310000 1996
320000 1919.9
330000 2074.4
340000 2199.4
350000 2189.1
360000 2046.4
370000 2036.1
380000 2319.7
390000 2420.7
};
\addplot [semithick, orchid204120188, mark=+, mark size=1.5, mark options={solid}]
table {%
-10000 81.7
0 201
10000 391.1
20000 1058.9
30000 1190.7
40000 945.4
50000 781.1
60000 863.6
70000 945.1
80000 1143.6
90000 1488.1
100000 1590.7
110000 1622.4
120000 1672.3
130000 1653.3
140000 1670
150000 1759.1
160000 1922.1
170000 2099.6
180000 2147.4
190000 2068.9
200000 2438.9
210000 2469.1
220000 2220.9
230000 2265.9
240000 2179.3
250000 2281.7
260000 2618.4
270000 2708.1
280000 2439
290000 2664.6
300000 2772.6
310000 2950.6
320000 2897.7
330000 2831.1
340000 2863.3
350000 2815.7
360000 3255.6
370000 3004
380000 2978
390000 3588.1
};
\addplot [semithick, peru20214597, mark=diamond*, mark size=1.5, mark options={solid}]
table {%
-10000 52.1
0 23
10000 49.4
20000 65
30000 76.4
40000 86.9
50000 91.4
60000 99.4
70000 101.9
80000 101.6
90000 102.1
100000 102.7
110000 101
120000 103.9
130000 99.4
140000 94
150000 96.1
160000 98.3
170000 100.7
180000 96.9
190000 93
200000 97.9
210000 93.4
220000 94
230000 100.3
240000 102.6
250000 99.9
260000 96.7
270000 95.1
280000 96
290000 93.9
300000 91.7
310000 103.4
320000 97.1
330000 92
340000 93
350000 93
360000 90.1
370000 94.1
380000 96.3
390000 89.4
};
\end{axis}

\end{tikzpicture}

%% file: Image/gopher.tex
\begin{tikzpicture}

\definecolor{darkcyan1115178}{RGB}{1,115,178}
\definecolor{darkorange2221435}{RGB}{222,143,5}
\definecolor{darkslategray38}{RGB}{38,38,38}
\definecolor{lavender234234242}{RGB}{234,234,242}
\definecolor{lightgray204}{RGB}{204,204,204}
\definecolor{orchid204120188}{RGB}{204,120,188}
\definecolor{peru20214597}{RGB}{202,145,97}

\begin{axis}[
axis background/.style={fill=lavender234234242},
axis line style={white},
legend cell align={left},
legend style={
  fill opacity=0.8,
  draw opacity=1,
  text opacity=1,
  at={(0.03,0.97)},
  anchor=north west,
  draw=lightgray204,
  fill=lavender234234242
},
tick align=outside,
x grid style={white},
xlabel=\textcolor{darkslategray38}{Time Step},
xmajorgrids,
xmajorticks=true,
xmin=-30000, xmax=410000,
xtick style={color=darkslategray38},
xtick={0,50000,100000,150000,200000,250000,300000,350000,400000},
xticklabels={0k,50k,100k,150k,200k,250k,300k,350k,400k},
y grid style={white},
ylabel=\textcolor{darkslategray38}{Performance},
ymajorgrids,
ymajorticks=true,
ymin=82.0191592863259, ymax=530.969382418883,
ytick style={color=darkslategray38}
]
\path [draw=darkcyan1115178, fill=darkcyan1115178, opacity=0.2]
(axis cs:-10000,309.662554094676)
--(axis cs:-10000,248.537445905324)
--(axis cs:0,328.837445905324)
--(axis cs:10000,329.437445905324)
--(axis cs:20000,218.037445905324)
--(axis cs:30000,167.737445905324)
--(axis cs:40000,172.337445905324)
--(axis cs:50000,182.537445905324)
--(axis cs:60000,195.437445905324)
--(axis cs:70000,234.837445905324)
--(axis cs:80000,247.437445905324)
--(axis cs:90000,266.037445905324)
--(axis cs:100000,292.037445905324)
--(axis cs:110000,276.837445905324)
--(axis cs:120000,270.337445905324)
--(axis cs:130000,270.837445905324)
--(axis cs:140000,306.337445905324)
--(axis cs:150000,286.537445905324)
--(axis cs:160000,335.437445905324)
--(axis cs:170000,354.537445905324)
--(axis cs:180000,297.437445905324)
--(axis cs:190000,335.137445905324)
--(axis cs:200000,327.437445905324)
--(axis cs:210000,371.737445905324)
--(axis cs:220000,319.437445905324)
--(axis cs:230000,349.737445905324)
--(axis cs:240000,355.737445905324)
--(axis cs:250000,374.537445905324)
--(axis cs:260000,328.337445905324)
--(axis cs:270000,356.837445905324)
--(axis cs:280000,410.537445905324)
--(axis cs:290000,330.337445905324)
--(axis cs:300000,370.037445905324)
--(axis cs:310000,436.837445905324)
--(axis cs:320000,449.437445905324)
--(axis cs:330000,414.837445905324)
--(axis cs:340000,351.137445905324)
--(axis cs:350000,370.537445905324)
--(axis cs:360000,336.337445905324)
--(axis cs:370000,299.437445905324)
--(axis cs:380000,328.837445905324)
--(axis cs:390000,404.037445905324)
--(axis cs:390000,465.162554094676)
--(axis cs:390000,465.162554094676)
--(axis cs:380000,389.962554094676)
--(axis cs:370000,360.562554094676)
--(axis cs:360000,397.462554094676)
--(axis cs:350000,431.662554094676)
--(axis cs:340000,412.262554094676)
--(axis cs:330000,475.962554094676)
--(axis cs:320000,510.562554094676)
--(axis cs:310000,497.962554094676)
--(axis cs:300000,431.162554094676)
--(axis cs:290000,391.462554094676)
--(axis cs:280000,471.662554094676)
--(axis cs:270000,417.962554094676)
--(axis cs:260000,389.462554094676)
--(axis cs:250000,435.662554094676)
--(axis cs:240000,416.862554094676)
--(axis cs:230000,410.862554094676)
--(axis cs:220000,380.562554094676)
--(axis cs:210000,432.862554094676)
--(axis cs:200000,388.562554094676)
--(axis cs:190000,396.262554094676)
--(axis cs:180000,358.562554094676)
--(axis cs:170000,415.662554094676)
--(axis cs:160000,396.562554094676)
--(axis cs:150000,347.662554094676)
--(axis cs:140000,367.462554094676)
--(axis cs:130000,331.962554094676)
--(axis cs:120000,331.462554094676)
--(axis cs:110000,337.962554094676)
--(axis cs:100000,353.162554094676)
--(axis cs:90000,327.162554094676)
--(axis cs:80000,308.562554094676)
--(axis cs:70000,295.962554094676)
--(axis cs:60000,256.562554094676)
--(axis cs:50000,243.662554094676)
--(axis cs:40000,233.462554094676)
--(axis cs:30000,228.862554094676)
--(axis cs:20000,279.162554094676)
--(axis cs:10000,390.562554094676)
--(axis cs:0,389.962554094676)
--(axis cs:-10000,309.662554094676)
--cycle;

\path [draw=darkorange2221435, fill=darkorange2221435, opacity=0.2]
(axis cs:-10000,329.504923952268)
--(axis cs:-10000,296.895076047732)
--(axis cs:0,376.695076047732)
--(axis cs:10000,352.095076047732)
--(axis cs:20000,313.495076047732)
--(axis cs:30000,274.895076047732)
--(axis cs:40000,296.095076047732)
--(axis cs:50000,286.095076047732)
--(axis cs:60000,275.295076047732)
--(axis cs:70000,272.495076047732)
--(axis cs:80000,278.695076047732)
--(axis cs:90000,314.095076047732)
--(axis cs:100000,307.695076047732)
--(axis cs:110000,276.895076047732)
--(axis cs:120000,286.295076047732)
--(axis cs:130000,272.495076047732)
--(axis cs:140000,316.695076047732)
--(axis cs:150000,327.095076047732)
--(axis cs:160000,330.095076047732)
--(axis cs:170000,325.295076047732)
--(axis cs:180000,300.095076047732)
--(axis cs:190000,294.895076047732)
--(axis cs:200000,266.695076047732)
--(axis cs:210000,302.495076047732)
--(axis cs:220000,302.895076047732)
--(axis cs:230000,308.095076047732)
--(axis cs:240000,284.895076047732)
--(axis cs:250000,311.495076047732)
--(axis cs:260000,306.895076047732)
--(axis cs:270000,297.895076047732)
--(axis cs:280000,300.895076047732)
--(axis cs:290000,288.295076047732)
--(axis cs:300000,303.495076047732)
--(axis cs:310000,310.895076047732)
--(axis cs:320000,315.695076047732)
--(axis cs:330000,274.095076047732)
--(axis cs:340000,312.495076047732)
--(axis cs:350000,328.295076047732)
--(axis cs:360000,294.895076047732)
--(axis cs:370000,317.095076047732)
--(axis cs:380000,321.895076047732)
--(axis cs:390000,273.295076047732)
--(axis cs:390000,305.904923952268)
--(axis cs:390000,305.904923952268)
--(axis cs:380000,354.504923952268)
--(axis cs:370000,349.704923952268)
--(axis cs:360000,327.504923952268)
--(axis cs:350000,360.904923952268)
--(axis cs:340000,345.104923952268)
--(axis cs:330000,306.704923952268)
--(axis cs:320000,348.304923952268)
--(axis cs:310000,343.504923952268)
--(axis cs:300000,336.104923952268)
--(axis cs:290000,320.904923952268)
--(axis cs:280000,333.504923952268)
--(axis cs:270000,330.504923952268)
--(axis cs:260000,339.504923952268)
--(axis cs:250000,344.104923952268)
--(axis cs:240000,317.504923952268)
--(axis cs:230000,340.704923952268)
--(axis cs:220000,335.504923952268)
--(axis cs:210000,335.104923952268)
--(axis cs:200000,299.304923952268)
--(axis cs:190000,327.504923952268)
--(axis cs:180000,332.704923952268)
--(axis cs:170000,357.904923952268)
--(axis cs:160000,362.704923952268)
--(axis cs:150000,359.704923952268)
--(axis cs:140000,349.304923952268)
--(axis cs:130000,305.104923952268)
--(axis cs:120000,318.904923952268)
--(axis cs:110000,309.504923952268)
--(axis cs:100000,340.304923952268)
--(axis cs:90000,346.704923952268)
--(axis cs:80000,311.304923952268)
--(axis cs:70000,305.104923952268)
--(axis cs:60000,307.904923952268)
--(axis cs:50000,318.704923952268)
--(axis cs:40000,328.704923952268)
--(axis cs:30000,307.504923952268)
--(axis cs:20000,346.104923952268)
--(axis cs:10000,384.704923952268)
--(axis cs:0,409.304923952268)
--(axis cs:-10000,329.504923952268)
--cycle;

\path [draw=orchid204120188, fill=orchid204120188, opacity=0.2]
(axis cs:-10000,336.178696948688)
--(axis cs:-10000,273.621303051312)
--(axis cs:0,340.121303051312)
--(axis cs:10000,340.421303051312)
--(axis cs:20000,201.621303051312)
--(axis cs:30000,181.621303051312)
--(axis cs:40000,150.721303051312)
--(axis cs:50000,154.121303051312)
--(axis cs:60000,226.121303051312)
--(axis cs:70000,178.421303051312)
--(axis cs:80000,189.821303051312)
--(axis cs:90000,207.821303051312)
--(axis cs:100000,243.621303051312)
--(axis cs:110000,257.621303051312)
--(axis cs:120000,245.821303051312)
--(axis cs:130000,244.421303051312)
--(axis cs:140000,265.321303051312)
--(axis cs:150000,212.121303051312)
--(axis cs:160000,247.321303051312)
--(axis cs:170000,279.021303051312)
--(axis cs:180000,251.321303051312)
--(axis cs:190000,247.021303051312)
--(axis cs:200000,251.321303051312)
--(axis cs:210000,326.721303051312)
--(axis cs:220000,247.321303051312)
--(axis cs:230000,421.821303051312)
--(axis cs:240000,370.721303051312)
--(axis cs:250000,316.721303051312)
--(axis cs:260000,325.021303051312)
--(axis cs:270000,383.321303051312)
--(axis cs:280000,377.321303051312)
--(axis cs:290000,355.621303051312)
--(axis cs:300000,319.621303051312)
--(axis cs:310000,295.621303051312)
--(axis cs:320000,307.621303051312)
--(axis cs:330000,349.621303051312)
--(axis cs:340000,359.021303051312)
--(axis cs:350000,396.421303051312)
--(axis cs:360000,319.021303051312)
--(axis cs:370000,349.021303051312)
--(axis cs:380000,366.121303051312)
--(axis cs:390000,367.321303051312)
--(axis cs:390000,429.878696948688)
--(axis cs:390000,429.878696948688)
--(axis cs:380000,428.678696948688)
--(axis cs:370000,411.578696948688)
--(axis cs:360000,381.578696948688)
--(axis cs:350000,458.978696948688)
--(axis cs:340000,421.578696948688)
--(axis cs:330000,412.178696948688)
--(axis cs:320000,370.178696948688)
--(axis cs:310000,358.178696948688)
--(axis cs:300000,382.178696948688)
--(axis cs:290000,418.178696948688)
--(axis cs:280000,439.878696948688)
--(axis cs:270000,445.878696948688)
--(axis cs:260000,387.578696948688)
--(axis cs:250000,379.278696948688)
--(axis cs:240000,433.278696948688)
--(axis cs:230000,484.378696948688)
--(axis cs:220000,309.878696948688)
--(axis cs:210000,389.278696948688)
--(axis cs:200000,313.878696948688)
--(axis cs:190000,309.578696948688)
--(axis cs:180000,313.878696948688)
--(axis cs:170000,341.578696948688)
--(axis cs:160000,309.878696948688)
--(axis cs:150000,274.678696948688)
--(axis cs:140000,327.878696948688)
--(axis cs:130000,306.978696948688)
--(axis cs:120000,308.378696948688)
--(axis cs:110000,320.178696948688)
--(axis cs:100000,306.178696948688)
--(axis cs:90000,270.378696948688)
--(axis cs:80000,252.378696948688)
--(axis cs:70000,240.978696948688)
--(axis cs:60000,288.678696948688)
--(axis cs:50000,216.678696948688)
--(axis cs:40000,213.278696948688)
--(axis cs:30000,244.178696948688)
--(axis cs:20000,264.178696948688)
--(axis cs:10000,402.978696948688)
--(axis cs:0,402.678696948688)
--(axis cs:-10000,336.178696948688)
--cycle;

\path [draw=peru20214597, fill=peru20214597, opacity=0.2]
(axis cs:-10000,304.174012389467)
--(axis cs:-10000,253.025987610533)
--(axis cs:0,251.325987610533)
--(axis cs:10000,245.025987610533)
--(axis cs:20000,176.125987610533)
--(axis cs:30000,149.825987610533)
--(axis cs:40000,142.425987610533)
--(axis cs:50000,124.725987610533)
--(axis cs:60000,102.425987610533)
--(axis cs:70000,152.425987610533)
--(axis cs:80000,181.525987610533)
--(axis cs:90000,145.325987610533)
--(axis cs:100000,134.125987610533)
--(axis cs:110000,187.525987610533)
--(axis cs:120000,156.725987610533)
--(axis cs:130000,158.725987610533)
--(axis cs:140000,130.125987610533)
--(axis cs:150000,148.125987610533)
--(axis cs:160000,171.825987610533)
--(axis cs:170000,202.125987610533)
--(axis cs:180000,198.125987610533)
--(axis cs:190000,175.825987610533)
--(axis cs:200000,208.725987610533)
--(axis cs:210000,233.825987610533)
--(axis cs:220000,234.125987610533)
--(axis cs:230000,208.125987610533)
--(axis cs:240000,195.825987610533)
--(axis cs:250000,192.425987610533)
--(axis cs:260000,204.125987610533)
--(axis cs:270000,127.025987610533)
--(axis cs:280000,207.025987610533)
--(axis cs:290000,199.525987610533)
--(axis cs:300000,187.325987610533)
--(axis cs:310000,179.025987610533)
--(axis cs:320000,175.325987610533)
--(axis cs:330000,254.125987610533)
--(axis cs:340000,194.125987610533)
--(axis cs:350000,228.425987610533)
--(axis cs:360000,225.525987610533)
--(axis cs:370000,239.825987610533)
--(axis cs:380000,204.125987610533)
--(axis cs:390000,217.325987610533)
--(axis cs:390000,268.474012389467)
--(axis cs:390000,268.474012389467)
--(axis cs:380000,255.274012389467)
--(axis cs:370000,290.974012389467)
--(axis cs:360000,276.674012389467)
--(axis cs:350000,279.574012389467)
--(axis cs:340000,245.274012389467)
--(axis cs:330000,305.274012389467)
--(axis cs:320000,226.474012389467)
--(axis cs:310000,230.174012389467)
--(axis cs:300000,238.474012389467)
--(axis cs:290000,250.674012389467)
--(axis cs:280000,258.174012389467)
--(axis cs:270000,178.174012389467)
--(axis cs:260000,255.274012389467)
--(axis cs:250000,243.574012389467)
--(axis cs:240000,246.974012389467)
--(axis cs:230000,259.274012389467)
--(axis cs:220000,285.274012389467)
--(axis cs:210000,284.974012389467)
--(axis cs:200000,259.874012389467)
--(axis cs:190000,226.974012389467)
--(axis cs:180000,249.274012389467)
--(axis cs:170000,253.274012389467)
--(axis cs:160000,222.974012389467)
--(axis cs:150000,199.274012389467)
--(axis cs:140000,181.274012389467)
--(axis cs:130000,209.874012389467)
--(axis cs:120000,207.874012389467)
--(axis cs:110000,238.674012389467)
--(axis cs:100000,185.274012389467)
--(axis cs:90000,196.474012389467)
--(axis cs:80000,232.674012389467)
--(axis cs:70000,203.574012389467)
--(axis cs:60000,153.574012389467)
--(axis cs:50000,175.874012389467)
--(axis cs:40000,193.574012389467)
--(axis cs:30000,200.974012389467)
--(axis cs:20000,227.274012389467)
--(axis cs:10000,296.174012389467)
--(axis cs:0,302.474012389467)
--(axis cs:-10000,304.174012389467)
--cycle;

\addplot [semithick, darkcyan1115178, mark=square*, mark size=1.5, mark options={solid}]
table {%
-10000 279.1
0 359.4
10000 360
20000 248.6
30000 198.3
40000 202.9
50000 213.1
60000 226
70000 265.4
80000 278
90000 296.6
100000 322.6
110000 307.4
120000 300.9
130000 301.4
140000 336.9
150000 317.1
160000 366
170000 385.1
180000 328
190000 365.7
200000 358
210000 402.3
220000 350
230000 380.3
240000 386.3
250000 405.1
260000 358.9
270000 387.4
280000 441.1
290000 360.9
300000 400.6
310000 467.4
320000 480
330000 445.4
340000 381.7
350000 401.1
360000 366.9
370000 330
380000 359.4
390000 434.6
};
\addplot [semithick, darkorange2221435, mark=triangle*, mark size=1.5, mark options={solid}]
table {%
-10000 313.2
0 393
10000 368.4
20000 329.8
30000 291.2
40000 312.4
50000 302.4
60000 291.6
70000 288.8
80000 295
90000 330.4
100000 324
110000 293.2
120000 302.6
130000 288.8
140000 333
150000 343.4
160000 346.4
170000 341.6
180000 316.4
190000 311.2
200000 283
210000 318.8
220000 319.2
230000 324.4
240000 301.2
250000 327.8
260000 323.2
270000 314.2
280000 317.2
290000 304.6
300000 319.8
310000 327.2
320000 332
330000 290.4
340000 328.8
350000 344.6
360000 311.2
370000 333.4
380000 338.2
390000 289.6
};
\addplot [semithick, orchid204120188, mark=+, mark size=1.5, mark options={solid}]
table {%
-10000 304.9
0 371.4
10000 371.7
20000 232.9
30000 212.9
40000 182
50000 185.4
60000 257.4
70000 209.7
80000 221.1
90000 239.1
100000 274.9
110000 288.9
120000 277.1
130000 275.7
140000 296.6
150000 243.4
160000 278.6
170000 310.3
180000 282.6
190000 278.3
200000 282.6
210000 358
220000 278.6
230000 453.1
240000 402
250000 348
260000 356.3
270000 414.6
280000 408.6
290000 386.9
300000 350.9
310000 326.9
320000 338.9
330000 380.9
340000 390.3
350000 427.7
360000 350.3
370000 380.3
380000 397.4
390000 398.6
};
\addplot [semithick, peru20214597, mark=diamond*, mark size=1.5, mark options={solid}]
table {%
-10000 278.6
0 276.9
10000 270.6
20000 201.7
30000 175.4
40000 168
50000 150.3
60000 128
70000 178
80000 207.1
90000 170.9
100000 159.7
110000 213.1
120000 182.3
130000 184.3
140000 155.7
150000 173.7
160000 197.4
170000 227.7
180000 223.7
190000 201.4
200000 234.3
210000 259.4
220000 259.7
230000 233.7
240000 221.4
250000 218
260000 229.7
270000 152.6
280000 232.6
290000 225.1
300000 212.9
310000 204.6
320000 200.9
330000 279.7
340000 219.7
350000 254
360000 251.1
370000 265.4
380000 229.7
390000 242.9
};
\end{axis}

\end{tikzpicture}

%% file: Image/hero.tex
\begin{tikzpicture}

\definecolor{darkcyan1115178}{RGB}{1,115,178}
\definecolor{darkorange2221435}{RGB}{222,143,5}
\definecolor{darkslategray38}{RGB}{38,38,38}
\definecolor{lavender234234242}{RGB}{234,234,242}
\definecolor{lightgray204}{RGB}{204,204,204}
\definecolor{orchid204120188}{RGB}{204,120,188}
\definecolor{peru20214597}{RGB}{202,145,97}

\begin{axis}[
axis background/.style={fill=lavender234234242},
axis line style={white},
legend cell align={left},
legend style={
  fill opacity=0.8,
  draw opacity=1,
  text opacity=1,
  at={(0.97,0.03)},
  anchor=south east,
  draw=lightgray204,
  fill=lavender234234242
},
tick align=outside,
x grid style={white},
xlabel=\textcolor{darkslategray38}{Time Step},
xmajorgrids,
xmajorticks=true,
xmin=-30000, xmax=410000,
xtick style={color=darkslategray38},
xtick={0,50000,100000,150000,200000,250000,300000,350000,400000},
xticklabels={0k,50k,100k,150k,200k,250k,300k,350k,400k},
y grid style={white},
ylabel=\textcolor{darkslategray38}{Performance},
ymajorgrids,
ymajorticks=true,
ymin=-1712.89024212877, ymax=14183.1902421288,
ytick style={color=darkslategray38}
]
\path [draw=darkcyan1115178, fill=darkcyan1115178, opacity=0.2]
(axis cs:-10000,1243.34112920797)
--(axis cs:-10000,-990.341129207973)
--(axis cs:0,-228.341129207973)
--(axis cs:10000,2072.75887079203)
--(axis cs:20000,3535.15887079203)
--(axis cs:30000,4605.35887079203)
--(axis cs:40000,5523.45887079203)
--(axis cs:50000,4649.05887079203)
--(axis cs:60000,7422.45887079203)
--(axis cs:70000,8079.35887079203)
--(axis cs:80000,8040.05887079203)
--(axis cs:90000,8165.75887079203)
--(axis cs:100000,8894.05887079203)
--(axis cs:110000,9414.05887079203)
--(axis cs:120000,10148.558870792)
--(axis cs:130000,9318.75887079203)
--(axis cs:140000,9958.95887079203)
--(axis cs:150000,10184.058870792)
--(axis cs:160000,9600.35887079203)
--(axis cs:170000,9223.75887079203)
--(axis cs:180000,9776.25887079203)
--(axis cs:190000,10165.058870792)
--(axis cs:200000,9280.05887079203)
--(axis cs:210000,10340.258870792)
--(axis cs:220000,10519.958870792)
--(axis cs:230000,8439.15887079203)
--(axis cs:240000,9624.75887079203)
--(axis cs:250000,10655.058870792)
--(axis cs:260000,9660.75887079203)
--(axis cs:270000,10501.258870792)
--(axis cs:280000,10832.758870792)
--(axis cs:290000,10634.758870792)
--(axis cs:300000,10727.558870792)
--(axis cs:310000,10741.058870792)
--(axis cs:320000,10872.958870792)
--(axis cs:330000,10901.558870792)
--(axis cs:340000,11120.558870792)
--(axis cs:350000,10572.458870792)
--(axis cs:360000,10967.558870792)
--(axis cs:370000,10837.358870792)
--(axis cs:380000,10881.258870792)
--(axis cs:390000,11226.958870792)
--(axis cs:390000,13460.641129208)
--(axis cs:390000,13460.641129208)
--(axis cs:380000,13114.941129208)
--(axis cs:370000,13071.041129208)
--(axis cs:360000,13201.241129208)
--(axis cs:350000,12806.141129208)
--(axis cs:340000,13354.241129208)
--(axis cs:330000,13135.241129208)
--(axis cs:320000,13106.641129208)
--(axis cs:310000,12974.741129208)
--(axis cs:300000,12961.241129208)
--(axis cs:290000,12868.441129208)
--(axis cs:280000,13066.441129208)
--(axis cs:270000,12734.941129208)
--(axis cs:260000,11894.441129208)
--(axis cs:250000,12888.741129208)
--(axis cs:240000,11858.441129208)
--(axis cs:230000,10672.841129208)
--(axis cs:220000,12753.641129208)
--(axis cs:210000,12573.941129208)
--(axis cs:200000,11513.741129208)
--(axis cs:190000,12398.741129208)
--(axis cs:180000,12009.941129208)
--(axis cs:170000,11457.441129208)
--(axis cs:160000,11834.041129208)
--(axis cs:150000,12417.741129208)
--(axis cs:140000,12192.641129208)
--(axis cs:130000,11552.441129208)
--(axis cs:120000,12382.241129208)
--(axis cs:110000,11647.741129208)
--(axis cs:100000,11127.741129208)
--(axis cs:90000,10399.441129208)
--(axis cs:80000,10273.741129208)
--(axis cs:70000,10313.041129208)
--(axis cs:60000,9656.14112920797)
--(axis cs:50000,6882.74112920797)
--(axis cs:40000,7757.14112920797)
--(axis cs:30000,6839.04112920797)
--(axis cs:20000,5768.84112920797)
--(axis cs:10000,4306.44112920797)
--(axis cs:0,2005.34112920797)
--(axis cs:-10000,1243.34112920797)
--cycle;

\path [draw=darkorange2221435, fill=darkorange2221435, opacity=0.2]
(axis cs:-10000,842.863899239431)
--(axis cs:-10000,-732.863899239431)
--(axis cs:0,2328.53610076057)
--(axis cs:10000,7254.73610076057)
--(axis cs:20000,9368.53610076057)
--(axis cs:30000,9329.23610076057)
--(axis cs:40000,10118.5361007606)
--(axis cs:50000,9482.13610076057)
--(axis cs:60000,9475.33610076057)
--(axis cs:70000,10148.3361007606)
--(axis cs:80000,10680.8361007606)
--(axis cs:90000,10470.7361007606)
--(axis cs:100000,10594.4361007606)
--(axis cs:110000,10833.4361007606)
--(axis cs:120000,10976.5361007606)
--(axis cs:130000,11548.7361007606)
--(axis cs:140000,11326.9361007606)
--(axis cs:150000,11199.5361007606)
--(axis cs:160000,11536.9361007606)
--(axis cs:170000,11009.9361007606)
--(axis cs:180000,11097.8361007606)
--(axis cs:190000,11121.8361007606)
--(axis cs:200000,11317.1361007606)
--(axis cs:210000,11186.6361007606)
--(axis cs:220000,10827.9361007606)
--(axis cs:230000,10792.9361007606)
--(axis cs:240000,10776.0361007606)
--(axis cs:250000,10480.9361007606)
--(axis cs:260000,10808.1361007606)
--(axis cs:270000,10773.1361007606)
--(axis cs:280000,10568.3361007606)
--(axis cs:290000,11111.1361007606)
--(axis cs:300000,11743.7361007606)
--(axis cs:310000,11758.5361007606)
--(axis cs:320000,11752.7361007606)
--(axis cs:330000,11768.9361007606)
--(axis cs:340000,11804.6361007606)
--(axis cs:350000,11800.9361007606)
--(axis cs:360000,11807.7361007606)
--(axis cs:370000,11816.2361007606)
--(axis cs:380000,11864.5361007606)
--(axis cs:390000,11810.5361007606)
--(axis cs:390000,13386.2638992394)
--(axis cs:390000,13386.2638992394)
--(axis cs:380000,13440.2638992394)
--(axis cs:370000,13391.9638992394)
--(axis cs:360000,13383.4638992394)
--(axis cs:350000,13376.6638992394)
--(axis cs:340000,13380.3638992394)
--(axis cs:330000,13344.6638992394)
--(axis cs:320000,13328.4638992394)
--(axis cs:310000,13334.2638992394)
--(axis cs:300000,13319.4638992394)
--(axis cs:290000,12686.8638992394)
--(axis cs:280000,12144.0638992394)
--(axis cs:270000,12348.8638992394)
--(axis cs:260000,12383.8638992394)
--(axis cs:250000,12056.6638992394)
--(axis cs:240000,12351.7638992394)
--(axis cs:230000,12368.6638992394)
--(axis cs:220000,12403.6638992394)
--(axis cs:210000,12762.3638992394)
--(axis cs:200000,12892.8638992394)
--(axis cs:190000,12697.5638992394)
--(axis cs:180000,12673.5638992394)
--(axis cs:170000,12585.6638992394)
--(axis cs:160000,13112.6638992394)
--(axis cs:150000,12775.2638992394)
--(axis cs:140000,12902.6638992394)
--(axis cs:130000,13124.4638992394)
--(axis cs:120000,12552.2638992394)
--(axis cs:110000,12409.1638992394)
--(axis cs:100000,12170.1638992394)
--(axis cs:90000,12046.4638992394)
--(axis cs:80000,12256.5638992394)
--(axis cs:70000,11724.0638992394)
--(axis cs:60000,11051.0638992394)
--(axis cs:50000,11057.8638992394)
--(axis cs:40000,11694.2638992394)
--(axis cs:30000,10904.9638992394)
--(axis cs:20000,10944.2638992394)
--(axis cs:10000,8830.46389923943)
--(axis cs:0,3904.26389923943)
--(axis cs:-10000,842.863899239431)
--cycle;

\path [draw=orchid204120188, fill=orchid204120188, opacity=0.2]
(axis cs:-10000,2071.55139460755)
--(axis cs:-10000,-311.551394607549)
--(axis cs:0,-714.751394607549)
--(axis cs:10000,2276.84860539245)
--(axis cs:20000,2376.64860539245)
--(axis cs:30000,5124.54860539245)
--(axis cs:40000,4050.14860539245)
--(axis cs:50000,4083.54860539245)
--(axis cs:60000,5088.04860539245)
--(axis cs:70000,6577.54860539245)
--(axis cs:80000,8391.94860539245)
--(axis cs:90000,8314.14860539245)
--(axis cs:100000,7838.74860539245)
--(axis cs:110000,8258.34860539245)
--(axis cs:120000,9512.54860539245)
--(axis cs:130000,9624.34860539245)
--(axis cs:140000,9565.74860539245)
--(axis cs:150000,9631.94860539245)
--(axis cs:160000,9723.34860539245)
--(axis cs:170000,9526.14860539245)
--(axis cs:180000,9212.54860539245)
--(axis cs:190000,8668.54860539245)
--(axis cs:200000,9308.44860539245)
--(axis cs:210000,9459.14860539245)
--(axis cs:220000,10216.4486053925)
--(axis cs:230000,10086.8486053924)
--(axis cs:240000,10062.0486053925)
--(axis cs:250000,10179.1486053925)
--(axis cs:260000,10232.7486053924)
--(axis cs:270000,10325.6486053925)
--(axis cs:280000,10437.0486053925)
--(axis cs:290000,10301.6486053925)
--(axis cs:300000,10241.2486053924)
--(axis cs:310000,10339.3486053924)
--(axis cs:320000,10330.0486053925)
--(axis cs:330000,10365.8486053924)
--(axis cs:340000,10326.0486053925)
--(axis cs:350000,10456.5486053925)
--(axis cs:360000,10045.2486053924)
--(axis cs:370000,10279.9486053925)
--(axis cs:380000,10210.8486053924)
--(axis cs:390000,10339.8486053924)
--(axis cs:390000,12722.9513946075)
--(axis cs:390000,12722.9513946075)
--(axis cs:380000,12593.9513946075)
--(axis cs:370000,12663.0513946075)
--(axis cs:360000,12428.3513946075)
--(axis cs:350000,12839.6513946076)
--(axis cs:340000,12709.1513946076)
--(axis cs:330000,12748.9513946075)
--(axis cs:320000,12713.1513946076)
--(axis cs:310000,12722.4513946075)
--(axis cs:300000,12624.3513946075)
--(axis cs:290000,12684.7513946076)
--(axis cs:280000,12820.1513946076)
--(axis cs:270000,12708.7513946076)
--(axis cs:260000,12615.8513946075)
--(axis cs:250000,12562.2513946076)
--(axis cs:240000,12445.1513946076)
--(axis cs:230000,12469.9513946075)
--(axis cs:220000,12599.5513946075)
--(axis cs:210000,11842.2513946076)
--(axis cs:200000,11691.5513946075)
--(axis cs:190000,11051.6513946076)
--(axis cs:180000,11595.6513946076)
--(axis cs:170000,11909.2513946076)
--(axis cs:160000,12106.4513946075)
--(axis cs:150000,12015.0513946075)
--(axis cs:140000,11948.8513946075)
--(axis cs:130000,12007.4513946075)
--(axis cs:120000,11895.6513946076)
--(axis cs:110000,10641.4513946075)
--(axis cs:100000,10221.8513946075)
--(axis cs:90000,10697.2513946076)
--(axis cs:80000,10775.0513946075)
--(axis cs:70000,8960.65139460755)
--(axis cs:60000,7471.15139460755)
--(axis cs:50000,6466.65139460755)
--(axis cs:40000,6433.25139460755)
--(axis cs:30000,7507.65139460755)
--(axis cs:20000,4759.75139460755)
--(axis cs:10000,4659.95139460755)
--(axis cs:0,1668.35139460755)
--(axis cs:-10000,2071.55139460755)
--cycle;

\path [draw=peru20214597, fill=peru20214597, opacity=0.2]
(axis cs:-10000,1107.61855702085)
--(axis cs:-10000,-954.41855702085)
--(axis cs:0,-141.71855702085)
--(axis cs:10000,699.88144297915)
--(axis cs:20000,2515.18144297915)
--(axis cs:30000,2944.38144297915)
--(axis cs:40000,3521.68144297915)
--(axis cs:50000,4032.58144297915)
--(axis cs:60000,4133.68144297915)
--(axis cs:70000,4322.08144297915)
--(axis cs:80000,4123.18144297915)
--(axis cs:90000,4104.08144297915)
--(axis cs:100000,4181.68144297915)
--(axis cs:110000,4043.48144297915)
--(axis cs:120000,4199.48144297915)
--(axis cs:130000,4192.18144297915)
--(axis cs:140000,4042.08144297915)
--(axis cs:150000,4176.68144297915)
--(axis cs:160000,4132.18144297915)
--(axis cs:170000,4102.18144297915)
--(axis cs:180000,4138.08144297915)
--(axis cs:190000,4253.68144297915)
--(axis cs:200000,4303.78144297915)
--(axis cs:210000,4212.28144297915)
--(axis cs:220000,4173.88144297915)
--(axis cs:230000,4244.08144297915)
--(axis cs:240000,4218.78144297915)
--(axis cs:250000,4239.68144297915)
--(axis cs:260000,4222.08144297915)
--(axis cs:270000,4304.88144297915)
--(axis cs:280000,4837.18144297915)
--(axis cs:290000,5033.18144297915)
--(axis cs:300000,4856.28144297915)
--(axis cs:310000,4869.98144297915)
--(axis cs:320000,4880.08144297915)
--(axis cs:330000,4696.08144297915)
--(axis cs:340000,4834.98144297915)
--(axis cs:350000,4631.88144297915)
--(axis cs:360000,4804.58144297915)
--(axis cs:370000,4293.18144297915)
--(axis cs:380000,4757.58144297915)
--(axis cs:390000,4638.08144297915)
--(axis cs:390000,6700.11855702085)
--(axis cs:390000,6700.11855702085)
--(axis cs:380000,6819.61855702085)
--(axis cs:370000,6355.21855702085)
--(axis cs:360000,6866.61855702085)
--(axis cs:350000,6693.91855702085)
--(axis cs:340000,6897.01855702085)
--(axis cs:330000,6758.11855702085)
--(axis cs:320000,6942.11855702085)
--(axis cs:310000,6932.01855702085)
--(axis cs:300000,6918.31855702085)
--(axis cs:290000,7095.21855702085)
--(axis cs:280000,6899.21855702085)
--(axis cs:270000,6366.91855702085)
--(axis cs:260000,6284.11855702085)
--(axis cs:250000,6301.71855702085)
--(axis cs:240000,6280.81855702085)
--(axis cs:230000,6306.11855702085)
--(axis cs:220000,6235.91855702085)
--(axis cs:210000,6274.31855702085)
--(axis cs:200000,6365.81855702085)
--(axis cs:190000,6315.71855702085)
--(axis cs:180000,6200.11855702085)
--(axis cs:170000,6164.21855702085)
--(axis cs:160000,6194.21855702085)
--(axis cs:150000,6238.71855702085)
--(axis cs:140000,6104.11855702085)
--(axis cs:130000,6254.21855702085)
--(axis cs:120000,6261.51855702085)
--(axis cs:110000,6105.51855702085)
--(axis cs:100000,6243.71855702085)
--(axis cs:90000,6166.11855702085)
--(axis cs:80000,6185.21855702085)
--(axis cs:70000,6384.11855702085)
--(axis cs:60000,6195.71855702085)
--(axis cs:50000,6094.61855702085)
--(axis cs:40000,5583.71855702085)
--(axis cs:30000,5006.41855702085)
--(axis cs:20000,4577.21855702085)
--(axis cs:10000,2761.91855702085)
--(axis cs:0,1920.31855702085)
--(axis cs:-10000,1107.61855702085)
--cycle;

\addplot [semithick, darkcyan1115178, mark=square*, mark size=1.5, mark options={solid}]
table {%
-10000 126.5
0 888.5
10000 3189.6
20000 4652
30000 5722.2
40000 6640.3
50000 5765.9
60000 8539.3
70000 9196.2
80000 9156.9
90000 9282.6
100000 10010.9
110000 10530.9
120000 11265.4
130000 10435.6
140000 11075.8
150000 11300.9
160000 10717.2
170000 10340.6
180000 10893.1
190000 11281.9
200000 10396.9
210000 11457.1
220000 11636.8
230000 9556
240000 10741.6
250000 11771.9
260000 10777.6
270000 11618.1
280000 11949.6
290000 11751.6
300000 11844.4
310000 11857.9
320000 11989.8
330000 12018.4
340000 12237.4
350000 11689.3
360000 12084.4
370000 11954.2
380000 11998.1
390000 12343.8
};
\addplot [semithick, darkorange2221435, mark=triangle*, mark size=1.5, mark options={solid}]
table {%
-10000 55
0 3116.4
10000 8042.6
20000 10156.4
30000 10117.1
40000 10906.4
50000 10270
60000 10263.2
70000 10936.2
80000 11468.7
90000 11258.6
100000 11382.3
110000 11621.3
120000 11764.4
130000 12336.6
140000 12114.8
150000 11987.4
160000 12324.8
170000 11797.8
180000 11885.7
190000 11909.7
200000 12105
210000 11974.5
220000 11615.8
230000 11580.8
240000 11563.9
250000 11268.8
260000 11596
270000 11561
280000 11356.2
290000 11899
300000 12531.6
310000 12546.4
320000 12540.6
330000 12556.8
340000 12592.5
350000 12588.8
360000 12595.6
370000 12604.1
380000 12652.4
390000 12598.4
};
\addplot [semithick, orchid204120188, mark=+, mark size=1.5, mark options={solid}]
table {%
-10000 880
0 476.8
10000 3468.4
20000 3568.2
30000 6316.1
40000 5241.7
50000 5275.1
60000 6279.6
70000 7769.1
80000 9583.5
90000 9505.7
100000 9030.3
110000 9449.9
120000 10704.1
130000 10815.9
140000 10757.3
150000 10823.5
160000 10914.9
170000 10717.7
180000 10404.1
190000 9860.1
200000 10500
210000 10650.7
220000 11408
230000 11278.4
240000 11253.6
250000 11370.7
260000 11424.3
270000 11517.2
280000 11628.6
290000 11493.2
300000 11432.8
310000 11530.9
320000 11521.6
330000 11557.4
340000 11517.6
350000 11648.1
360000 11236.8
370000 11471.5
380000 11402.4
390000 11531.4
};
\addplot [semithick, peru20214597, mark=diamond*, mark size=1.5, mark options={solid}]
table {%
-10000 76.6
0 889.3
10000 1730.9
20000 3546.2
30000 3975.4
40000 4552.7
50000 5063.6
60000 5164.7
70000 5353.1
80000 5154.2
90000 5135.1
100000 5212.7
110000 5074.5
120000 5230.5
130000 5223.2
140000 5073.1
150000 5207.7
160000 5163.2
170000 5133.2
180000 5169.1
190000 5284.7
200000 5334.8
210000 5243.3
220000 5204.9
230000 5275.1
240000 5249.8
250000 5270.7
260000 5253.1
270000 5335.9
280000 5868.2
290000 6064.2
300000 5887.3
310000 5901
320000 5911.1
330000 5727.1
340000 5866
350000 5662.9
360000 5835.6
370000 5324.2
380000 5788.6
390000 5669.1
};
\end{axis}

\end{tikzpicture}

%% file: Image/james_bond.tex
\begin{tikzpicture}

\definecolor{darkcyan1115178}{RGB}{1,115,178}
\definecolor{darkorange2221435}{RGB}{222,143,5}
\definecolor{darkslategray38}{RGB}{38,38,38}
\definecolor{lavender234234242}{RGB}{234,234,242}
\definecolor{lightgray204}{RGB}{204,204,204}
\definecolor{orchid204120188}{RGB}{204,120,188}
\definecolor{peru20214597}{RGB}{202,145,97}

\begin{axis}[
axis background/.style={fill=lavender234234242},
axis line style={white},
legend cell align={left},
legend style={
  fill opacity=0.8,
  draw opacity=1,
  text opacity=1,
  at={(0.03,0.97)},
  anchor=north west,
  draw=lightgray204,
  fill=lavender234234242
},
tick align=outside,
x grid style={white},
xlabel=\textcolor{darkslategray38}{Time Step},
xmajorgrids,
xmajorticks=true,
xmin=-30000, xmax=410000,
xtick style={color=darkslategray38},
xtick={0,50000,100000,150000,200000,250000,300000,350000,400000},
xticklabels={0k,50k,100k,150k,200k,250k,300k,350k,400k},
y grid style={white},
ylabel=\textcolor{darkslategray38}{Performance},
ymajorgrids,
ymajorticks=true,
ymin=-129.165420745834, ymax=1135.48177232525,
ytick style={color=darkslategray38}
]
\path [draw=darkcyan1115178, fill=darkcyan1115178, opacity=0.2]
(axis cs:-10000,132.997809003842)
--(axis cs:-10000,-60.1978090038416)
--(axis cs:0,-55.8978090038416)
--(axis cs:10000,-34.4978090038416)
--(axis cs:20000,104.802190996158)
--(axis cs:30000,59.8021909961584)
--(axis cs:40000,153.402190996158)
--(axis cs:50000,165.502190996158)
--(axis cs:60000,256.302190996158)
--(axis cs:70000,299.102190996158)
--(axis cs:80000,341.302190996158)
--(axis cs:90000,384.102190996158)
--(axis cs:100000,387.002190996158)
--(axis cs:110000,514.802190996158)
--(axis cs:120000,457.002190996158)
--(axis cs:130000,447.702190996158)
--(axis cs:140000,445.502190996158)
--(axis cs:150000,492.002190996158)
--(axis cs:160000,661.302190996158)
--(axis cs:170000,487.702190996158)
--(axis cs:180000,640.502190996158)
--(axis cs:190000,584.802190996158)
--(axis cs:200000,685.502190996158)
--(axis cs:210000,574.102190996158)
--(axis cs:220000,552.002190996158)
--(axis cs:230000,552.002190996158)
--(axis cs:240000,576.302190996158)
--(axis cs:250000,608.402190996158)
--(axis cs:260000,741.302190996158)
--(axis cs:270000,655.502190996158)
--(axis cs:280000,580.502190996158)
--(axis cs:290000,676.302190996158)
--(axis cs:300000,799.802190996158)
--(axis cs:310000,649.802190996158)
--(axis cs:320000,705.502190996158)
--(axis cs:330000,753.402190996158)
--(axis cs:340000,656.302190996158)
--(axis cs:350000,716.302190996158)
--(axis cs:360000,822.002190996158)
--(axis cs:370000,689.102190996158)
--(axis cs:380000,681.302190996158)
--(axis cs:390000,884.802190996158)
--(axis cs:390000,1077.99780900384)
--(axis cs:390000,1077.99780900384)
--(axis cs:380000,874.497809003842)
--(axis cs:370000,882.297809003842)
--(axis cs:360000,1015.19780900384)
--(axis cs:350000,909.497809003842)
--(axis cs:340000,849.497809003842)
--(axis cs:330000,946.597809003842)
--(axis cs:320000,898.697809003842)
--(axis cs:310000,842.997809003842)
--(axis cs:300000,992.997809003842)
--(axis cs:290000,869.497809003842)
--(axis cs:280000,773.697809003842)
--(axis cs:270000,848.697809003842)
--(axis cs:260000,934.497809003842)
--(axis cs:250000,801.597809003842)
--(axis cs:240000,769.497809003842)
--(axis cs:230000,745.197809003842)
--(axis cs:220000,745.197809003842)
--(axis cs:210000,767.297809003842)
--(axis cs:200000,878.697809003842)
--(axis cs:190000,777.997809003842)
--(axis cs:180000,833.697809003842)
--(axis cs:170000,680.897809003842)
--(axis cs:160000,854.497809003842)
--(axis cs:150000,685.197809003842)
--(axis cs:140000,638.697809003842)
--(axis cs:130000,640.897809003842)
--(axis cs:120000,650.197809003842)
--(axis cs:110000,707.997809003842)
--(axis cs:100000,580.197809003842)
--(axis cs:90000,577.297809003842)
--(axis cs:80000,534.497809003842)
--(axis cs:70000,492.297809003842)
--(axis cs:60000,449.497809003842)
--(axis cs:50000,358.697809003842)
--(axis cs:40000,346.597809003842)
--(axis cs:30000,252.997809003842)
--(axis cs:20000,297.997809003842)
--(axis cs:10000,158.697809003842)
--(axis cs:0,137.297809003842)
--(axis cs:-10000,132.997809003842)
--cycle;

\path [draw=darkorange2221435, fill=darkorange2221435, opacity=0.2]
(axis cs:-10000,127.601962016428)
--(axis cs:-10000,-20.6019620164278)
--(axis cs:0,-26.1019620164278)
--(axis cs:10000,0.398037983572152)
--(axis cs:20000,7.89803798357215)
--(axis cs:30000,32.8980379835722)
--(axis cs:40000,38.3980379835722)
--(axis cs:50000,92.3980379835722)
--(axis cs:60000,106.898037983572)
--(axis cs:70000,169.398037983572)
--(axis cs:80000,183.398037983572)
--(axis cs:90000,222.898037983572)
--(axis cs:100000,220.898037983572)
--(axis cs:110000,235.398037983572)
--(axis cs:120000,263.898037983572)
--(axis cs:130000,274.898037983572)
--(axis cs:140000,276.898037983572)
--(axis cs:150000,335.898037983572)
--(axis cs:160000,376.398037983572)
--(axis cs:170000,325.398037983572)
--(axis cs:180000,353.398037983572)
--(axis cs:190000,332.398037983572)
--(axis cs:200000,379.898037983572)
--(axis cs:210000,332.898037983572)
--(axis cs:220000,367.398037983572)
--(axis cs:230000,450.898037983572)
--(axis cs:240000,403.898037983572)
--(axis cs:250000,454.398037983572)
--(axis cs:260000,354.398037983572)
--(axis cs:270000,450.898037983572)
--(axis cs:280000,467.398037983572)
--(axis cs:290000,552.398037983572)
--(axis cs:300000,445.898037983572)
--(axis cs:310000,469.398037983572)
--(axis cs:320000,460.398037983572)
--(axis cs:330000,503.398037983572)
--(axis cs:340000,512.398037983572)
--(axis cs:350000,504.398037983572)
--(axis cs:360000,488.398037983572)
--(axis cs:370000,597.398037983572)
--(axis cs:380000,556.898037983572)
--(axis cs:390000,596.898037983572)
--(axis cs:390000,745.101962016428)
--(axis cs:390000,745.101962016428)
--(axis cs:380000,705.101962016428)
--(axis cs:370000,745.601962016428)
--(axis cs:360000,636.601962016428)
--(axis cs:350000,652.601962016428)
--(axis cs:340000,660.601962016428)
--(axis cs:330000,651.601962016428)
--(axis cs:320000,608.601962016428)
--(axis cs:310000,617.601962016428)
--(axis cs:300000,594.101962016428)
--(axis cs:290000,700.601962016428)
--(axis cs:280000,615.601962016428)
--(axis cs:270000,599.101962016428)
--(axis cs:260000,502.601962016428)
--(axis cs:250000,602.601962016428)
--(axis cs:240000,552.101962016428)
--(axis cs:230000,599.101962016428)
--(axis cs:220000,515.601962016428)
--(axis cs:210000,481.101962016428)
--(axis cs:200000,528.101962016428)
--(axis cs:190000,480.601962016428)
--(axis cs:180000,501.601962016428)
--(axis cs:170000,473.601962016428)
--(axis cs:160000,524.601962016428)
--(axis cs:150000,484.101962016428)
--(axis cs:140000,425.101962016428)
--(axis cs:130000,423.101962016428)
--(axis cs:120000,412.101962016428)
--(axis cs:110000,383.601962016428)
--(axis cs:100000,369.101962016428)
--(axis cs:90000,371.101962016428)
--(axis cs:80000,331.601962016428)
--(axis cs:70000,317.601962016428)
--(axis cs:60000,255.101962016428)
--(axis cs:50000,240.601962016428)
--(axis cs:40000,186.601962016428)
--(axis cs:30000,181.101962016428)
--(axis cs:20000,156.101962016428)
--(axis cs:10000,148.601962016428)
--(axis cs:0,122.101962016428)
--(axis cs:-10000,127.601962016428)
--cycle;

\path [draw=orchid204120188, fill=orchid204120188, opacity=0.2]
(axis cs:-10000,132.997809003842)
--(axis cs:-10000,-60.1978090038416)
--(axis cs:0,-55.8978090038416)
--(axis cs:10000,-34.4978090038416)
--(axis cs:20000,104.802190996158)
--(axis cs:30000,59.8021909961584)
--(axis cs:40000,153.402190996158)
--(axis cs:50000,165.502190996158)
--(axis cs:60000,256.302190996158)
--(axis cs:70000,299.102190996158)
--(axis cs:80000,341.302190996158)
--(axis cs:90000,384.102190996158)
--(axis cs:100000,387.002190996158)
--(axis cs:110000,514.802190996158)
--(axis cs:120000,457.002190996158)
--(axis cs:130000,447.702190996158)
--(axis cs:140000,445.502190996158)
--(axis cs:150000,492.002190996158)
--(axis cs:160000,661.302190996158)
--(axis cs:170000,487.702190996158)
--(axis cs:180000,640.502190996158)
--(axis cs:190000,584.802190996158)
--(axis cs:200000,685.502190996158)
--(axis cs:210000,574.102190996158)
--(axis cs:220000,552.002190996158)
--(axis cs:230000,552.002190996158)
--(axis cs:240000,576.302190996158)
--(axis cs:250000,608.402190996158)
--(axis cs:260000,741.302190996158)
--(axis cs:270000,655.502190996158)
--(axis cs:280000,580.502190996158)
--(axis cs:290000,676.302190996158)
--(axis cs:300000,799.802190996158)
--(axis cs:310000,649.802190996158)
--(axis cs:320000,705.502190996158)
--(axis cs:330000,753.402190996158)
--(axis cs:340000,656.302190996158)
--(axis cs:350000,716.302190996158)
--(axis cs:360000,822.002190996158)
--(axis cs:370000,689.102190996158)
--(axis cs:380000,681.302190996158)
--(axis cs:390000,884.802190996158)
--(axis cs:390000,1077.99780900384)
--(axis cs:390000,1077.99780900384)
--(axis cs:380000,874.497809003842)
--(axis cs:370000,882.297809003842)
--(axis cs:360000,1015.19780900384)
--(axis cs:350000,909.497809003842)
--(axis cs:340000,849.497809003842)
--(axis cs:330000,946.597809003842)
--(axis cs:320000,898.697809003842)
--(axis cs:310000,842.997809003842)
--(axis cs:300000,992.997809003842)
--(axis cs:290000,869.497809003842)
--(axis cs:280000,773.697809003842)
--(axis cs:270000,848.697809003842)
--(axis cs:260000,934.497809003842)
--(axis cs:250000,801.597809003842)
--(axis cs:240000,769.497809003842)
--(axis cs:230000,745.197809003842)
--(axis cs:220000,745.197809003842)
--(axis cs:210000,767.297809003842)
--(axis cs:200000,878.697809003842)
--(axis cs:190000,777.997809003842)
--(axis cs:180000,833.697809003842)
--(axis cs:170000,680.897809003842)
--(axis cs:160000,854.497809003842)
--(axis cs:150000,685.197809003842)
--(axis cs:140000,638.697809003842)
--(axis cs:130000,640.897809003842)
--(axis cs:120000,650.197809003842)
--(axis cs:110000,707.997809003842)
--(axis cs:100000,580.197809003842)
--(axis cs:90000,577.297809003842)
--(axis cs:80000,534.497809003842)
--(axis cs:70000,492.297809003842)
--(axis cs:60000,449.497809003842)
--(axis cs:50000,358.697809003842)
--(axis cs:40000,346.597809003842)
--(axis cs:30000,252.997809003842)
--(axis cs:20000,297.997809003842)
--(axis cs:10000,158.697809003842)
--(axis cs:0,137.297809003842)
--(axis cs:-10000,132.997809003842)
--cycle;

\path [draw=peru20214597, fill=peru20214597, opacity=0.2]
(axis cs:-10000,114.58145742442)
--(axis cs:-10000,-28.7814574244204)
--(axis cs:0,-71.6814574244204)
--(axis cs:10000,-71.6814574244204)
--(axis cs:20000,-71.6814574244204)
--(axis cs:30000,-44.5814574244204)
--(axis cs:40000,-10.2814574244204)
--(axis cs:50000,16.2185425755796)
--(axis cs:60000,46.9185425755796)
--(axis cs:70000,84.7185425755796)
--(axis cs:80000,96.9185425755796)
--(axis cs:90000,81.9185425755796)
--(axis cs:100000,152.61854257558)
--(axis cs:110000,156.21854257558)
--(axis cs:120000,209.71854257558)
--(axis cs:130000,199.71854257558)
--(axis cs:140000,238.31854257558)
--(axis cs:150000,239.71854257558)
--(axis cs:160000,259.01854257558)
--(axis cs:170000,266.21854257558)
--(axis cs:180000,265.41854257558)
--(axis cs:190000,228.31854257558)
--(axis cs:200000,221.91854257558)
--(axis cs:210000,271.91854257558)
--(axis cs:220000,327.61854257558)
--(axis cs:230000,275.41854257558)
--(axis cs:240000,311.21854257558)
--(axis cs:250000,281.21854257558)
--(axis cs:260000,301.91854257558)
--(axis cs:270000,283.31854257558)
--(axis cs:280000,321.21854257558)
--(axis cs:290000,354.71854257558)
--(axis cs:300000,381.21854257558)
--(axis cs:310000,409.01854257558)
--(axis cs:320000,375.41854257558)
--(axis cs:330000,358.31854257558)
--(axis cs:340000,385.41854257558)
--(axis cs:350000,477.61854257558)
--(axis cs:360000,404.01854257558)
--(axis cs:370000,416.21854257558)
--(axis cs:380000,478.31854257558)
--(axis cs:390000,486.21854257558)
--(axis cs:390000,629.58145742442)
--(axis cs:390000,629.58145742442)
--(axis cs:380000,621.68145742442)
--(axis cs:370000,559.58145742442)
--(axis cs:360000,547.38145742442)
--(axis cs:350000,620.98145742442)
--(axis cs:340000,528.78145742442)
--(axis cs:330000,501.68145742442)
--(axis cs:320000,518.78145742442)
--(axis cs:310000,552.38145742442)
--(axis cs:300000,524.58145742442)
--(axis cs:290000,498.08145742442)
--(axis cs:280000,464.58145742442)
--(axis cs:270000,426.68145742442)
--(axis cs:260000,445.28145742442)
--(axis cs:250000,424.58145742442)
--(axis cs:240000,454.58145742442)
--(axis cs:230000,418.78145742442)
--(axis cs:220000,470.98145742442)
--(axis cs:210000,415.28145742442)
--(axis cs:200000,365.28145742442)
--(axis cs:190000,371.68145742442)
--(axis cs:180000,408.78145742442)
--(axis cs:170000,409.58145742442)
--(axis cs:160000,402.38145742442)
--(axis cs:150000,383.08145742442)
--(axis cs:140000,381.68145742442)
--(axis cs:130000,343.08145742442)
--(axis cs:120000,353.08145742442)
--(axis cs:110000,299.58145742442)
--(axis cs:100000,295.98145742442)
--(axis cs:90000,225.28145742442)
--(axis cs:80000,240.28145742442)
--(axis cs:70000,228.08145742442)
--(axis cs:60000,190.28145742442)
--(axis cs:50000,159.58145742442)
--(axis cs:40000,133.08145742442)
--(axis cs:30000,98.7814574244204)
--(axis cs:20000,71.6814574244204)
--(axis cs:10000,71.6814574244204)
--(axis cs:0,71.6814574244204)
--(axis cs:-10000,114.58145742442)
--cycle;

\addplot [semithick, darkcyan1115178, mark=square*, mark size=1.5, mark options={solid}]
table {%
-10000 36.4
0 40.7
10000 62.1
20000 201.4
30000 156.4
40000 250
50000 262.1
60000 352.9
70000 395.7
80000 437.9
90000 480.7
100000 483.6
110000 611.4
120000 553.6
130000 544.3
140000 542.1
150000 588.6
160000 757.9
170000 584.3
180000 737.1
190000 681.4
200000 782.1
210000 670.7
220000 648.6
230000 648.6
240000 672.9
250000 705
260000 837.9
270000 752.1
280000 677.1
290000 772.9
300000 896.4
310000 746.4
320000 802.1
330000 850
340000 752.9
350000 812.9
360000 918.6
370000 785.7
380000 777.9
390000 981.4
};
\addplot [semithick, darkorange2221435, mark=triangle*, mark size=1.5, mark options={solid}]
table {%
-10000 53.5
0 48
10000 74.5
20000 82
30000 107
40000 112.5
50000 166.5
60000 181
70000 243.5
80000 257.5
90000 297
100000 295
110000 309.5
120000 338
130000 349
140000 351
150000 410
160000 450.5
170000 399.5
180000 427.5
190000 406.5
200000 454
210000 407
220000 441.5
230000 525
240000 478
250000 528.5
260000 428.5
270000 525
280000 541.5
290000 626.5
300000 520
310000 543.5
320000 534.5
330000 577.5
340000 586.5
350000 578.5
360000 562.5
370000 671.5
380000 631
390000 671
};
\addplot [semithick, orchid204120188, mark=+, mark size=1.5, mark options={solid}]
table {%
-10000 36.4
0 40.7
10000 62.1
20000 201.4
30000 156.4
40000 250
50000 262.1
60000 352.9
70000 395.7
80000 437.9
90000 480.7
100000 483.6
110000 611.4
120000 553.6
130000 544.3
140000 542.1
150000 588.6
160000 757.9
170000 584.3
180000 737.1
190000 681.4
200000 782.1
210000 670.7
220000 648.6
230000 648.6
240000 672.9
250000 705
260000 837.9
270000 752.1
280000 677.1
290000 772.9
300000 896.4
310000 746.4
320000 802.1
330000 850
340000 752.9
350000 812.9
360000 918.6
370000 785.7
380000 777.9
390000 981.4
};
\addplot [semithick, peru20214597, mark=diamond*, mark size=1.5, mark options={solid}]
table {%
-10000 42.9
0 0
10000 0
20000 0
30000 27.1
40000 61.4
50000 87.9
60000 118.6
70000 156.4
80000 168.6
90000 153.6
100000 224.3
110000 227.9
120000 281.4
130000 271.4
140000 310
150000 311.4
160000 330.7
170000 337.9
180000 337.1
190000 300
200000 293.6
210000 343.6
220000 399.3
230000 347.1
240000 382.9
250000 352.9
260000 373.6
270000 355
280000 392.9
290000 426.4
300000 452.9
310000 480.7
320000 447.1
330000 430
340000 457.1
350000 549.3
360000 475.7
370000 487.9
380000 550
390000 557.9
};
\end{axis}

\end{tikzpicture}

%% file: Image/kangaroo.tex
\begin{tikzpicture}

\definecolor{darkcyan1115178}{RGB}{1,115,178}
\definecolor{darkorange2221435}{RGB}{222,143,5}
\definecolor{darkslategray38}{RGB}{38,38,38}
\definecolor{lavender234234242}{RGB}{234,234,242}
\definecolor{lightgray204}{RGB}{204,204,204}
\definecolor{orchid204120188}{RGB}{204,120,188}
\definecolor{peru20214597}{RGB}{202,145,97}

\begin{axis}[
axis background/.style={fill=lavender234234242},
axis line style={white},
legend cell align={left},
legend style={
  fill opacity=0.8,
  draw opacity=1,
  text opacity=1,
  at={(0.03,0.97)},
  anchor=north west,
  draw=lightgray204,
  fill=lavender234234242
},
tick align=outside,
x grid style={white},
xlabel=\textcolor{darkslategray38}{Time Step},
xmajorgrids,
xmajorticks=true,
xmin=-30000, xmax=410000,
xtick style={color=darkslategray38},
xtick={0,50000,100000,150000,200000,250000,300000,350000,400000},
xticklabels={0k,50k,100k,150k,200k,250k,300k,350k,400k},
y grid style={white},
ylabel=\textcolor{darkslategray38}{Performance},
ymajorgrids,
ymajorticks=true,
ymin=-2314.3969370488, ymax=10782.0831841706,
ytick style={color=darkslategray38}
]
\path [draw=darkcyan1115178, fill=darkcyan1115178, opacity=0.2]
(axis cs:-10000,1733.40238608428)
--(axis cs:-10000,-1710.60238608428)
--(axis cs:0,-1430.60238608428)
--(axis cs:10000,-1636.30238608428)
--(axis cs:20000,-1719.10238608428)
--(axis cs:30000,-1687.70238608428)
--(axis cs:40000,-1633.40238608428)
--(axis cs:50000,-1479.10238608428)
--(axis cs:60000,-1444.90238608428)
--(axis cs:70000,-733.40238608428)
--(axis cs:80000,309.39761391572)
--(axis cs:90000,893.69761391572)
--(axis cs:100000,2643.69761391572)
--(axis cs:110000,2402.29761391572)
--(axis cs:120000,3775.09761391572)
--(axis cs:130000,4050.89761391572)
--(axis cs:140000,4296.59761391572)
--(axis cs:150000,4225.09761391572)
--(axis cs:160000,4780.89761391572)
--(axis cs:170000,4472.29761391572)
--(axis cs:180000,4152.29761391572)
--(axis cs:190000,5592.29761391572)
--(axis cs:200000,6472.29761391572)
--(axis cs:210000,5842.29761391572)
--(axis cs:220000,5900.89761391572)
--(axis cs:230000,5993.69761391572)
--(axis cs:240000,5602.29761391572)
--(axis cs:250000,6626.59761391572)
--(axis cs:260000,6193.69761391572)
--(axis cs:270000,5605.09761391572)
--(axis cs:280000,6409.39761391572)
--(axis cs:290000,6357.99761391572)
--(axis cs:300000,6497.99761391572)
--(axis cs:310000,5763.69761391572)
--(axis cs:320000,5387.99761391572)
--(axis cs:330000,5212.29761391572)
--(axis cs:340000,5355.09761391572)
--(axis cs:350000,4492.29761391572)
--(axis cs:360000,4446.59761391572)
--(axis cs:370000,4930.89761391572)
--(axis cs:380000,5639.39761391572)
--(axis cs:390000,4689.39761391572)
--(axis cs:390000,8133.40238608428)
--(axis cs:390000,8133.40238608428)
--(axis cs:380000,9083.40238608428)
--(axis cs:370000,8374.90238608428)
--(axis cs:360000,7890.60238608428)
--(axis cs:350000,7936.30238608428)
--(axis cs:340000,8799.10238608428)
--(axis cs:330000,8656.30238608428)
--(axis cs:320000,8832.00238608428)
--(axis cs:310000,9207.70238608428)
--(axis cs:300000,9942.00238608428)
--(axis cs:290000,9802.00238608428)
--(axis cs:280000,9853.40238608428)
--(axis cs:270000,9049.10238608428)
--(axis cs:260000,9637.70238608428)
--(axis cs:250000,10070.6023860843)
--(axis cs:240000,9046.30238608428)
--(axis cs:230000,9437.70238608428)
--(axis cs:220000,9344.90238608428)
--(axis cs:210000,9286.30238608428)
--(axis cs:200000,9916.30238608428)
--(axis cs:190000,9036.30238608428)
--(axis cs:180000,7596.30238608428)
--(axis cs:170000,7916.30238608428)
--(axis cs:160000,8224.90238608428)
--(axis cs:150000,7669.10238608428)
--(axis cs:140000,7740.60238608428)
--(axis cs:130000,7494.90238608428)
--(axis cs:120000,7219.10238608428)
--(axis cs:110000,5846.30238608428)
--(axis cs:100000,6087.70238608428)
--(axis cs:90000,4337.70238608428)
--(axis cs:80000,3753.40238608428)
--(axis cs:70000,2710.60238608428)
--(axis cs:60000,1999.10238608428)
--(axis cs:50000,1964.90238608428)
--(axis cs:40000,1810.60238608428)
--(axis cs:30000,1756.30238608428)
--(axis cs:20000,1724.90238608428)
--(axis cs:10000,1807.70238608428)
--(axis cs:0,2013.40238608428)
--(axis cs:-10000,1733.40238608428)
--cycle;

\path [draw=darkorange2221435, fill=darkorange2221435, opacity=0.2]
(axis cs:-10000,963.700035411161)
--(axis cs:-10000,-871.700035411161)
--(axis cs:0,-793.700035411161)
--(axis cs:10000,-879.700035411161)
--(axis cs:20000,-901.700035411161)
--(axis cs:30000,-791.700035411161)
--(axis cs:40000,-725.700035411161)
--(axis cs:50000,-625.700035411161)
--(axis cs:60000,-551.700035411161)
--(axis cs:70000,26.2999645888387)
--(axis cs:80000,138.299964588839)
--(axis cs:90000,574.299964588839)
--(axis cs:100000,587.299964588839)
--(axis cs:110000,1332.29996458884)
--(axis cs:120000,1372.29996458884)
--(axis cs:130000,1799.29996458884)
--(axis cs:140000,1690.29996458884)
--(axis cs:150000,1929.29996458884)
--(axis cs:160000,2256.29996458884)
--(axis cs:170000,2874.29996458884)
--(axis cs:180000,3299.29996458884)
--(axis cs:190000,3389.29996458884)
--(axis cs:200000,3101.29996458884)
--(axis cs:210000,3147.29996458884)
--(axis cs:220000,3231.29996458884)
--(axis cs:230000,2898.29996458884)
--(axis cs:240000,3139.29996458884)
--(axis cs:250000,2870.29996458884)
--(axis cs:260000,2753.29996458884)
--(axis cs:270000,2938.29996458884)
--(axis cs:280000,2909.29996458884)
--(axis cs:290000,2947.29996458884)
--(axis cs:300000,3182.29996458884)
--(axis cs:310000,3093.29996458884)
--(axis cs:320000,2765.29996458884)
--(axis cs:330000,2816.29996458884)
--(axis cs:340000,2843.29996458884)
--(axis cs:350000,2984.29996458884)
--(axis cs:360000,2823.29996458884)
--(axis cs:370000,2760.29996458884)
--(axis cs:380000,3146.29996458884)
--(axis cs:390000,3166.29996458884)
--(axis cs:390000,5001.70003541116)
--(axis cs:390000,5001.70003541116)
--(axis cs:380000,4981.70003541116)
--(axis cs:370000,4595.70003541116)
--(axis cs:360000,4658.70003541116)
--(axis cs:350000,4819.70003541116)
--(axis cs:340000,4678.70003541116)
--(axis cs:330000,4651.70003541116)
--(axis cs:320000,4600.70003541116)
--(axis cs:310000,4928.70003541116)
--(axis cs:300000,5017.70003541116)
--(axis cs:290000,4782.70003541116)
--(axis cs:280000,4744.70003541116)
--(axis cs:270000,4773.70003541116)
--(axis cs:260000,4588.70003541116)
--(axis cs:250000,4705.70003541116)
--(axis cs:240000,4974.70003541116)
--(axis cs:230000,4733.70003541116)
--(axis cs:220000,5066.70003541116)
--(axis cs:210000,4982.70003541116)
--(axis cs:200000,4936.70003541116)
--(axis cs:190000,5224.70003541116)
--(axis cs:180000,5134.70003541116)
--(axis cs:170000,4709.70003541116)
--(axis cs:160000,4091.70003541116)
--(axis cs:150000,3764.70003541116)
--(axis cs:140000,3525.70003541116)
--(axis cs:130000,3634.70003541116)
--(axis cs:120000,3207.70003541116)
--(axis cs:110000,3167.70003541116)
--(axis cs:100000,2422.70003541116)
--(axis cs:90000,2409.70003541116)
--(axis cs:80000,1973.70003541116)
--(axis cs:70000,1861.70003541116)
--(axis cs:60000,1283.70003541116)
--(axis cs:50000,1209.70003541116)
--(axis cs:40000,1109.70003541116)
--(axis cs:30000,1043.70003541116)
--(axis cs:20000,933.700035411161)
--(axis cs:10000,955.700035411161)
--(axis cs:0,1041.70003541116)
--(axis cs:-10000,963.700035411161)
--cycle;

\path [draw=orchid204120188, fill=orchid204120188, opacity=0.2]
(axis cs:-10000,1626.31213036882)
--(axis cs:-10000,-1449.11213036882)
--(axis cs:0,-1377.71213036882)
--(axis cs:10000,-1529.11213036882)
--(axis cs:20000,-1520.61213036882)
--(axis cs:30000,-1457.71213036882)
--(axis cs:40000,-1406.31213036882)
--(axis cs:50000,-677.712130368816)
--(axis cs:60000,-897.712130368816)
--(axis cs:70000,783.687869631184)
--(axis cs:80000,953.687869631184)
--(axis cs:90000,3027.98786963118)
--(axis cs:100000,4065.18786963118)
--(axis cs:110000,3770.88786963118)
--(axis cs:120000,3733.68786963118)
--(axis cs:130000,3737.98786963118)
--(axis cs:140000,4503.68786963118)
--(axis cs:150000,4107.98786963118)
--(axis cs:160000,4930.88786963118)
--(axis cs:170000,4972.28786963118)
--(axis cs:180000,4877.98786963118)
--(axis cs:190000,5637.98786963118)
--(axis cs:200000,5573.68786963118)
--(axis cs:210000,2586.58786963118)
--(axis cs:220000,5277.98786963118)
--(axis cs:230000,4480.88786963118)
--(axis cs:240000,4210.88786963118)
--(axis cs:250000,4097.98786963118)
--(axis cs:260000,4957.98786963118)
--(axis cs:270000,4946.58786963118)
--(axis cs:280000,4162.28786963118)
--(axis cs:290000,5305.18786963118)
--(axis cs:300000,4079.38786963118)
--(axis cs:310000,5357.98786963118)
--(axis cs:320000,4460.88786963118)
--(axis cs:330000,4065.18786963118)
--(axis cs:340000,5230.88786963118)
--(axis cs:350000,3789.38786963118)
--(axis cs:360000,4525.18786963118)
--(axis cs:370000,4256.58786963118)
--(axis cs:380000,4659.38786963118)
--(axis cs:390000,4579.38786963118)
--(axis cs:390000,7654.81213036882)
--(axis cs:390000,7654.81213036882)
--(axis cs:380000,7734.81213036882)
--(axis cs:370000,7332.01213036882)
--(axis cs:360000,7600.61213036882)
--(axis cs:350000,6864.81213036882)
--(axis cs:340000,8306.31213036882)
--(axis cs:330000,7140.61213036882)
--(axis cs:320000,7536.31213036882)
--(axis cs:310000,8433.41213036882)
--(axis cs:300000,7154.81213036882)
--(axis cs:290000,8380.61213036882)
--(axis cs:280000,7237.71213036882)
--(axis cs:270000,8022.01213036882)
--(axis cs:260000,8033.41213036882)
--(axis cs:250000,7173.41213036882)
--(axis cs:240000,7286.31213036882)
--(axis cs:230000,7556.31213036882)
--(axis cs:220000,8353.41213036882)
--(axis cs:210000,5662.01213036882)
--(axis cs:200000,8649.11213036882)
--(axis cs:190000,8713.41213036882)
--(axis cs:180000,7953.41213036882)
--(axis cs:170000,8047.71213036882)
--(axis cs:160000,8006.31213036882)
--(axis cs:150000,7183.41213036882)
--(axis cs:140000,7579.11213036882)
--(axis cs:130000,6813.41213036882)
--(axis cs:120000,6809.11213036882)
--(axis cs:110000,6846.31213036882)
--(axis cs:100000,7140.61213036882)
--(axis cs:90000,6103.41213036882)
--(axis cs:80000,4029.11213036882)
--(axis cs:70000,3859.11213036882)
--(axis cs:60000,2177.71213036882)
--(axis cs:50000,2397.71213036882)
--(axis cs:40000,1669.11213036882)
--(axis cs:30000,1617.71213036882)
--(axis cs:20000,1554.81213036882)
--(axis cs:10000,1546.31213036882)
--(axis cs:0,1697.71213036882)
--(axis cs:-10000,1626.31213036882)
--cycle;

\path [draw=peru20214597, fill=peru20214597, opacity=0.2]
(axis cs:-10000,1712.48863320612)
--(axis cs:-10000,-1649.68863320612)
--(axis cs:0,-1681.08863320612)
--(axis cs:10000,-1569.68863320612)
--(axis cs:20000,-492.488633206121)
--(axis cs:30000,793.211366793879)
--(axis cs:40000,1158.91136679388)
--(axis cs:50000,1296.01136679388)
--(axis cs:60000,1418.91136679388)
--(axis cs:70000,1616.01136679388)
--(axis cs:80000,2493.21136679388)
--(axis cs:90000,2524.61136679388)
--(axis cs:100000,2438.91136679388)
--(axis cs:110000,2704.61136679388)
--(axis cs:120000,3458.91136679388)
--(axis cs:130000,3910.31136679388)
--(axis cs:140000,3616.01136679388)
--(axis cs:150000,3496.01136679388)
--(axis cs:160000,3707.51136679388)
--(axis cs:170000,4070.31136679388)
--(axis cs:180000,4170.31136679388)
--(axis cs:190000,4676.01136679388)
--(axis cs:200000,4864.61136679388)
--(axis cs:210000,5687.51136679388)
--(axis cs:220000,6121.81136679388)
--(axis cs:230000,6090.31136679388)
--(axis cs:240000,6208.91136679388)
--(axis cs:250000,5961.81136679388)
--(axis cs:260000,6418.91136679388)
--(axis cs:270000,6573.21136679388)
--(axis cs:280000,6041.81136679388)
--(axis cs:290000,6824.61136679388)
--(axis cs:300000,6181.81136679388)
--(axis cs:310000,6240.31136679388)
--(axis cs:320000,6078.91136679388)
--(axis cs:330000,6147.51136679388)
--(axis cs:340000,5770.31136679388)
--(axis cs:350000,6153.21136679388)
--(axis cs:360000,6094.61136679388)
--(axis cs:370000,5338.91136679388)
--(axis cs:380000,5791.81136679388)
--(axis cs:390000,5607.51136679388)
--(axis cs:390000,8969.68863320612)
--(axis cs:390000,8969.68863320612)
--(axis cs:380000,9153.98863320612)
--(axis cs:370000,8701.08863320612)
--(axis cs:360000,9456.78863320612)
--(axis cs:350000,9515.38863320612)
--(axis cs:340000,9132.48863320612)
--(axis cs:330000,9509.68863320612)
--(axis cs:320000,9441.08863320612)
--(axis cs:310000,9602.48863320612)
--(axis cs:300000,9543.98863320612)
--(axis cs:290000,10186.7886332061)
--(axis cs:280000,9403.98863320612)
--(axis cs:270000,9935.38863320612)
--(axis cs:260000,9781.08863320612)
--(axis cs:250000,9323.98863320612)
--(axis cs:240000,9571.08863320612)
--(axis cs:230000,9452.48863320612)
--(axis cs:220000,9483.98863320612)
--(axis cs:210000,9049.68863320612)
--(axis cs:200000,8226.78863320612)
--(axis cs:190000,8038.18863320612)
--(axis cs:180000,7532.48863320612)
--(axis cs:170000,7432.48863320612)
--(axis cs:160000,7069.68863320612)
--(axis cs:150000,6858.18863320612)
--(axis cs:140000,6978.18863320612)
--(axis cs:130000,7272.48863320612)
--(axis cs:120000,6821.08863320612)
--(axis cs:110000,6066.78863320612)
--(axis cs:100000,5801.08863320612)
--(axis cs:90000,5886.78863320612)
--(axis cs:80000,5855.38863320612)
--(axis cs:70000,4978.18863320612)
--(axis cs:60000,4781.08863320612)
--(axis cs:50000,4658.18863320612)
--(axis cs:40000,4521.08863320612)
--(axis cs:30000,4155.38863320612)
--(axis cs:20000,2869.68863320612)
--(axis cs:10000,1792.48863320612)
--(axis cs:0,1681.08863320612)
--(axis cs:-10000,1712.48863320612)
--cycle;

\addplot [semithick, darkcyan1115178, mark=square*, mark size=1.5, mark options={solid}]
table {%
-10000 11.4
0 291.4
10000 85.7
20000 2.9
30000 34.3
40000 88.6
50000 242.9
60000 277.1
70000 988.6
80000 2031.4
90000 2615.7
100000 4365.7
110000 4124.3
120000 5497.1
130000 5772.9
140000 6018.6
150000 5947.1
160000 6502.9
170000 6194.3
180000 5874.3
190000 7314.3
200000 8194.3
210000 7564.3
220000 7622.9
230000 7715.7
240000 7324.3
250000 8348.6
260000 7915.7
270000 7327.1
280000 8131.4
290000 8080
300000 8220
310000 7485.7
320000 7110
330000 6934.3
340000 7077.1
350000 6214.3
360000 6168.6
370000 6652.9
380000 7361.4
390000 6411.4
};
\addplot [semithick, darkorange2221435, mark=triangle*, mark size=1.5, mark options={solid}]
table {%
-10000 46
0 124
10000 38
20000 16
30000 126
40000 192
50000 292
60000 366
70000 944
80000 1056
90000 1492
100000 1505
110000 2250
120000 2290
130000 2717
140000 2608
150000 2847
160000 3174
170000 3792
180000 4217
190000 4307
200000 4019
210000 4065
220000 4149
230000 3816
240000 4057
250000 3788
260000 3671
270000 3856
280000 3827
290000 3865
300000 4100
310000 4011
320000 3683
330000 3734
340000 3761
350000 3902
360000 3741
370000 3678
380000 4064
390000 4084
};
\addplot [semithick, orchid204120188, mark=+, mark size=1.5, mark options={solid}]
table {%
-10000 88.6
0 160
10000 8.6
20000 17.1
30000 80
40000 131.4
50000 860
60000 640
70000 2321.4
80000 2491.4
90000 4565.7
100000 5602.9
110000 5308.6
120000 5271.4
130000 5275.7
140000 6041.4
150000 5645.7
160000 6468.6
170000 6510
180000 6415.7
190000 7175.7
200000 7111.4
210000 4124.3
220000 6815.7
230000 6018.6
240000 5748.6
250000 5635.7
260000 6495.7
270000 6484.3
280000 5700
290000 6842.9
300000 5617.1
310000 6895.7
320000 5998.6
330000 5602.9
340000 6768.6
350000 5327.1
360000 6062.9
370000 5794.3
380000 6197.1
390000 6117.1
};
\addplot [semithick, peru20214597, mark=diamond*, mark size=1.5, mark options={solid}]
table {%
-10000 31.4
0 0
10000 111.4
20000 1188.6
30000 2474.3
40000 2840
50000 2977.1
60000 3100
70000 3297.1
80000 4174.3
90000 4205.7
100000 4120
110000 4385.7
120000 5140
130000 5591.4
140000 5297.1
150000 5177.1
160000 5388.6
170000 5751.4
180000 5851.4
190000 6357.1
200000 6545.7
210000 7368.6
220000 7802.9
230000 7771.4
240000 7890
250000 7642.9
260000 8100
270000 8254.3
280000 7722.9
290000 8505.7
300000 7862.9
310000 7921.4
320000 7760
330000 7828.6
340000 7451.4
350000 7834.3
360000 7775.7
370000 7020
380000 7472.9
390000 7288.6
};
\end{axis}

\end{tikzpicture}

%% file: Image/krull.tex
\begin{tikzpicture}

\definecolor{darkcyan1115178}{RGB}{1,115,178}
\definecolor{darkorange2221435}{RGB}{222,143,5}
\definecolor{darkslategray38}{RGB}{38,38,38}
\definecolor{lavender234234242}{RGB}{234,234,242}
\definecolor{lightgray204}{RGB}{204,204,204}
\definecolor{orchid204120188}{RGB}{204,120,188}
\definecolor{peru20214597}{RGB}{202,145,97}

\begin{axis}[
axis background/.style={fill=lavender234234242},
axis line style={white},
legend cell align={left},
legend style={
  fill opacity=0.8,
  draw opacity=1,
  text opacity=1,
  at={(0.97,0.03)},
  anchor=south east,
  draw=lightgray204,
  fill=lavender234234242
},
tick align=outside,
x grid style={white},
xlabel=\textcolor{darkslategray38}{Time Step},
xmajorgrids,
xmajorticks=true,
xmin=-30000, xmax=410000,
xtick style={color=darkslategray38},
xtick={0,50000,100000,150000,200000,250000,300000,350000,400000},
xticklabels={0k,50k,100k,150k,200k,250k,300k,350k,400k},
y grid style={white},
ylabel=\textcolor{darkslategray38}{Performance},
ymajorgrids,
ymajorticks=true,
ymin=-1787.04254642809, ymax=11474.9048133935,
ytick style={color=darkslategray38}
]
\path [draw=darkcyan1115178, fill=darkcyan1115178, opacity=0.2]
(axis cs:-10000,1802.29760679666)
--(axis cs:-10000,929.90239320334)
--(axis cs:0,9020.80239320334)
--(axis cs:10000,8978.50239320334)
--(axis cs:20000,8935.40239320334)
--(axis cs:30000,8961.80239320334)
--(axis cs:40000,9207.90239320334)
--(axis cs:50000,9035.20239320334)
--(axis cs:60000,9316.90239320334)
--(axis cs:70000,9460.40239320334)
--(axis cs:80000,9615.40239320334)
--(axis cs:90000,9126.90239320334)
--(axis cs:100000,9209.90239320334)
--(axis cs:110000,8931.20239320334)
--(axis cs:120000,8986.10239320334)
--(axis cs:130000,9307.80239320334)
--(axis cs:140000,9528.90239320334)
--(axis cs:150000,9580.70239320334)
--(axis cs:160000,9567.90239320334)
--(axis cs:170000,9728.10239320334)
--(axis cs:180000,9260.10239320334)
--(axis cs:190000,9269.70239320334)
--(axis cs:200000,9780.40239320334)
--(axis cs:210000,9544.70239320334)
--(axis cs:220000,9480.10239320334)
--(axis cs:230000,9563.90239320334)
--(axis cs:240000,9514.40239320334)
--(axis cs:250000,9463.10239320334)
--(axis cs:260000,9593.90239320334)
--(axis cs:270000,9738.40239320334)
--(axis cs:280000,9450.10239320334)
--(axis cs:290000,9665.70239320334)
--(axis cs:300000,9338.40239320334)
--(axis cs:310000,9696.10239320334)
--(axis cs:320000,9241.50239320334)
--(axis cs:330000,9380.10239320334)
--(axis cs:340000,9881.20239320334)
--(axis cs:350000,9414.90239320334)
--(axis cs:360000,9510.50239320334)
--(axis cs:370000,9279.40239320334)
--(axis cs:380000,9796.50239320334)
--(axis cs:390000,9847.90239320334)
--(axis cs:390000,10720.2976067967)
--(axis cs:390000,10720.2976067967)
--(axis cs:380000,10668.8976067967)
--(axis cs:370000,10151.7976067967)
--(axis cs:360000,10382.8976067967)
--(axis cs:350000,10287.2976067967)
--(axis cs:340000,10753.5976067967)
--(axis cs:330000,10252.4976067967)
--(axis cs:320000,10113.8976067967)
--(axis cs:310000,10568.4976067967)
--(axis cs:300000,10210.7976067967)
--(axis cs:290000,10538.0976067967)
--(axis cs:280000,10322.4976067967)
--(axis cs:270000,10610.7976067967)
--(axis cs:260000,10466.2976067967)
--(axis cs:250000,10335.4976067967)
--(axis cs:240000,10386.7976067967)
--(axis cs:230000,10436.2976067967)
--(axis cs:220000,10352.4976067967)
--(axis cs:210000,10417.0976067967)
--(axis cs:200000,10652.7976067967)
--(axis cs:190000,10142.0976067967)
--(axis cs:180000,10132.4976067967)
--(axis cs:170000,10600.4976067967)
--(axis cs:160000,10440.2976067967)
--(axis cs:150000,10453.0976067967)
--(axis cs:140000,10401.2976067967)
--(axis cs:130000,10180.1976067967)
--(axis cs:120000,9858.49760679666)
--(axis cs:110000,9803.59760679666)
--(axis cs:100000,10082.2976067967)
--(axis cs:90000,9999.29760679666)
--(axis cs:80000,10487.7976067967)
--(axis cs:70000,10332.7976067967)
--(axis cs:60000,10189.2976067967)
--(axis cs:50000,9907.59760679666)
--(axis cs:40000,10080.2976067967)
--(axis cs:30000,9834.19760679666)
--(axis cs:20000,9807.79760679666)
--(axis cs:10000,9850.89760679666)
--(axis cs:0,9893.19760679666)
--(axis cs:-10000,1802.29760679666)
--cycle;

\path [draw=darkorange2221435, fill=darkorange2221435, opacity=0.2]
(axis cs:-10000,1797.88902431068)
--(axis cs:-10000,1127.91097568932)
--(axis cs:0,8383.61097568932)
--(axis cs:10000,8946.21097568932)
--(axis cs:20000,8927.61097568932)
--(axis cs:30000,8756.31097568932)
--(axis cs:40000,8247.71097568932)
--(axis cs:50000,8792.11097568932)
--(axis cs:60000,8856.61097568932)
--(axis cs:70000,8196.91097568932)
--(axis cs:80000,8709.91097568932)
--(axis cs:90000,8502.61097568932)
--(axis cs:100000,8245.41097568932)
--(axis cs:110000,8584.71097568932)
--(axis cs:120000,9060.41097568932)
--(axis cs:130000,8830.41097568932)
--(axis cs:140000,8651.81097568932)
--(axis cs:150000,8775.01097568932)
--(axis cs:160000,9013.41097568932)
--(axis cs:170000,9068.71097568932)
--(axis cs:180000,9436.51097568932)
--(axis cs:190000,9556.61097568932)
--(axis cs:200000,9745.21097568932)
--(axis cs:210000,9608.81097568932)
--(axis cs:220000,9260.11097568932)
--(axis cs:230000,9467.61097568932)
--(axis cs:240000,9280.01097568932)
--(axis cs:250000,9494.51097568932)
--(axis cs:260000,9545.41097568932)
--(axis cs:270000,9167.41097568932)
--(axis cs:280000,9747.21097568932)
--(axis cs:290000,9777.91097568932)
--(axis cs:300000,9953.41097568932)
--(axis cs:310000,10139.7109756893)
--(axis cs:320000,10202.1109756893)
--(axis cs:330000,9925.21097568932)
--(axis cs:340000,10068.8109756893)
--(axis cs:350000,9686.81097568932)
--(axis cs:360000,9796.71097568932)
--(axis cs:370000,9986.21097568932)
--(axis cs:380000,9976.51097568932)
--(axis cs:390000,9815.21097568932)
--(axis cs:390000,10485.1890243107)
--(axis cs:390000,10485.1890243107)
--(axis cs:380000,10646.4890243107)
--(axis cs:370000,10656.1890243107)
--(axis cs:360000,10466.6890243107)
--(axis cs:350000,10356.7890243107)
--(axis cs:340000,10738.7890243107)
--(axis cs:330000,10595.1890243107)
--(axis cs:320000,10872.0890243107)
--(axis cs:310000,10809.6890243107)
--(axis cs:300000,10623.3890243107)
--(axis cs:290000,10447.8890243107)
--(axis cs:280000,10417.1890243107)
--(axis cs:270000,9837.38902431068)
--(axis cs:260000,10215.3890243107)
--(axis cs:250000,10164.4890243107)
--(axis cs:240000,9949.98902431068)
--(axis cs:230000,10137.5890243107)
--(axis cs:220000,9930.08902431068)
--(axis cs:210000,10278.7890243107)
--(axis cs:200000,10415.1890243107)
--(axis cs:190000,10226.5890243107)
--(axis cs:180000,10106.4890243107)
--(axis cs:170000,9738.68902431068)
--(axis cs:160000,9683.38902431068)
--(axis cs:150000,9444.98902431068)
--(axis cs:140000,9321.78902431068)
--(axis cs:130000,9500.38902431068)
--(axis cs:120000,9730.38902431068)
--(axis cs:110000,9254.68902431068)
--(axis cs:100000,8915.38902431068)
--(axis cs:90000,9172.58902431068)
--(axis cs:80000,9379.88902431068)
--(axis cs:70000,8866.88902431068)
--(axis cs:60000,9526.58902431068)
--(axis cs:50000,9462.08902431068)
--(axis cs:40000,8917.68902431068)
--(axis cs:30000,9426.28902431068)
--(axis cs:20000,9597.58902431068)
--(axis cs:10000,9616.18902431068)
--(axis cs:0,9053.58902431068)
--(axis cs:-10000,1797.88902431068)
--cycle;

\path [draw=orchid204120188, fill=orchid204120188, opacity=0.2]
(axis cs:-10000,2408.25328579709)
--(axis cs:-10000,1012.54671420291)
--(axis cs:0,5975.14671420291)
--(axis cs:10000,7337.74671420291)
--(axis cs:20000,7027.04671420291)
--(axis cs:30000,7278.54671420291)
--(axis cs:40000,7474.84671420291)
--(axis cs:50000,7256.84671420291)
--(axis cs:60000,7441.24671420291)
--(axis cs:70000,7664.14671420291)
--(axis cs:80000,7923.44671420291)
--(axis cs:90000,7875.44671420291)
--(axis cs:100000,8064.24671420291)
--(axis cs:110000,7821.04671420291)
--(axis cs:120000,8240.04671420291)
--(axis cs:130000,8336.24671420291)
--(axis cs:140000,8195.54671420291)
--(axis cs:150000,8252.14671420291)
--(axis cs:160000,8322.24671420291)
--(axis cs:170000,8018.84671420291)
--(axis cs:180000,8203.14671420291)
--(axis cs:190000,8364.54671420291)
--(axis cs:200000,8243.24671420291)
--(axis cs:210000,8279.04671420291)
--(axis cs:220000,8611.84671420291)
--(axis cs:230000,8168.24671420291)
--(axis cs:240000,8607.74671420291)
--(axis cs:250000,8486.44671420291)
--(axis cs:260000,8459.74671420291)
--(axis cs:270000,8365.24671420291)
--(axis cs:280000,7807.44671420291)
--(axis cs:290000,8172.24671420291)
--(axis cs:300000,8173.14671420291)
--(axis cs:310000,8484.54671420291)
--(axis cs:320000,8517.44671420291)
--(axis cs:330000,8665.24671420291)
--(axis cs:340000,7977.04671420291)
--(axis cs:350000,8340.14671420291)
--(axis cs:360000,8510.44671420291)
--(axis cs:370000,8189.14671420291)
--(axis cs:380000,8224.14671420291)
--(axis cs:390000,8669.04671420291)
--(axis cs:390000,10064.7532857971)
--(axis cs:390000,10064.7532857971)
--(axis cs:380000,9619.85328579709)
--(axis cs:370000,9584.85328579709)
--(axis cs:360000,9906.15328579709)
--(axis cs:350000,9735.85328579709)
--(axis cs:340000,9372.75328579709)
--(axis cs:330000,10060.9532857971)
--(axis cs:320000,9913.15328579709)
--(axis cs:310000,9880.25328579709)
--(axis cs:300000,9568.85328579709)
--(axis cs:290000,9567.95328579709)
--(axis cs:280000,9203.15328579709)
--(axis cs:270000,9760.95328579709)
--(axis cs:260000,9855.45328579709)
--(axis cs:250000,9882.15328579709)
--(axis cs:240000,10003.4532857971)
--(axis cs:230000,9563.95328579709)
--(axis cs:220000,10007.5532857971)
--(axis cs:210000,9674.75328579709)
--(axis cs:200000,9638.95328579709)
--(axis cs:190000,9760.25328579709)
--(axis cs:180000,9598.85328579709)
--(axis cs:170000,9414.55328579709)
--(axis cs:160000,9717.95328579709)
--(axis cs:150000,9647.85328579709)
--(axis cs:140000,9591.25328579709)
--(axis cs:130000,9731.95328579709)
--(axis cs:120000,9635.75328579709)
--(axis cs:110000,9216.75328579709)
--(axis cs:100000,9459.95328579709)
--(axis cs:90000,9271.15328579709)
--(axis cs:80000,9319.15328579709)
--(axis cs:70000,9059.85328579709)
--(axis cs:60000,8836.95328579709)
--(axis cs:50000,8652.55328579709)
--(axis cs:40000,8870.55328579709)
--(axis cs:30000,8674.25328579709)
--(axis cs:20000,8422.75328579709)
--(axis cs:10000,8733.45328579709)
--(axis cs:0,7370.85328579709)
--(axis cs:-10000,2408.25328579709)
--cycle;

\path [draw=peru20214597, fill=peru20214597, opacity=0.2]
(axis cs:-10000,2744.52675734529)
--(axis cs:-10000,372.673242654713)
--(axis cs:0,-1183.92675734529)
--(axis cs:10000,-1184.22675734529)
--(axis cs:20000,-926.026757345287)
--(axis cs:30000,-976.826757345287)
--(axis cs:40000,-978.926757345287)
--(axis cs:50000,-271.226757345287)
--(axis cs:60000,266.773242654713)
--(axis cs:70000,297.773242654713)
--(axis cs:80000,449.673242654713)
--(axis cs:90000,517.673242654713)
--(axis cs:100000,527.473242654713)
--(axis cs:110000,346.073242654713)
--(axis cs:120000,512.073242654713)
--(axis cs:130000,2760.47324265471)
--(axis cs:140000,4777.37324265471)
--(axis cs:150000,2998.37324265471)
--(axis cs:160000,3943.37324265471)
--(axis cs:170000,2159.17324265471)
--(axis cs:180000,2021.37324265471)
--(axis cs:190000,1536.17324265471)
--(axis cs:200000,2998.37324265471)
--(axis cs:210000,2652.47324265471)
--(axis cs:220000,3341.37324265471)
--(axis cs:230000,3191.77324265471)
--(axis cs:240000,3882.67324265471)
--(axis cs:250000,3065.07324265471)
--(axis cs:260000,3189.67324265471)
--(axis cs:270000,4440.67324265471)
--(axis cs:280000,3840.37324265471)
--(axis cs:290000,4427.67324265471)
--(axis cs:300000,3334.37324265471)
--(axis cs:310000,3527.97324265471)
--(axis cs:320000,4242.17324265471)
--(axis cs:330000,3618.77324265471)
--(axis cs:340000,3148.17324265471)
--(axis cs:350000,4937.17324265471)
--(axis cs:360000,5415.67324265471)
--(axis cs:370000,3675.37324265471)
--(axis cs:380000,5397.47324265471)
--(axis cs:390000,4407.67324265471)
--(axis cs:390000,6779.52675734529)
--(axis cs:390000,6779.52675734529)
--(axis cs:380000,7769.32675734529)
--(axis cs:370000,6047.22675734529)
--(axis cs:360000,7787.52675734529)
--(axis cs:350000,7309.02675734529)
--(axis cs:340000,5520.02675734529)
--(axis cs:330000,5990.62675734529)
--(axis cs:320000,6614.02675734529)
--(axis cs:310000,5899.82675734529)
--(axis cs:300000,5706.22675734529)
--(axis cs:290000,6799.52675734529)
--(axis cs:280000,6212.22675734529)
--(axis cs:270000,6812.52675734529)
--(axis cs:260000,5561.52675734529)
--(axis cs:250000,5436.92675734529)
--(axis cs:240000,6254.52675734529)
--(axis cs:230000,5563.62675734529)
--(axis cs:220000,5713.22675734529)
--(axis cs:210000,5024.32675734529)
--(axis cs:200000,5370.22675734529)
--(axis cs:190000,3908.02675734529)
--(axis cs:180000,4393.22675734529)
--(axis cs:170000,4531.02675734529)
--(axis cs:160000,6315.22675734529)
--(axis cs:150000,5370.22675734529)
--(axis cs:140000,7149.22675734529)
--(axis cs:130000,5132.32675734529)
--(axis cs:120000,2883.92675734529)
--(axis cs:110000,2717.92675734529)
--(axis cs:100000,2899.32675734529)
--(axis cs:90000,2889.52675734529)
--(axis cs:80000,2821.52675734529)
--(axis cs:70000,2669.62675734529)
--(axis cs:60000,2638.62675734529)
--(axis cs:50000,2100.62675734529)
--(axis cs:40000,1392.92675734529)
--(axis cs:30000,1395.02675734529)
--(axis cs:20000,1445.82675734529)
--(axis cs:10000,1187.62675734529)
--(axis cs:0,1187.92675734529)
--(axis cs:-10000,2744.52675734529)
--cycle;

\addplot [semithick, darkcyan1115178, mark=square*, mark size=1.5, mark options={solid}]
table {%
-10000 1366.1
0 9457
10000 9414.7
20000 9371.6
30000 9398
40000 9644.1
50000 9471.4
60000 9753.1
70000 9896.6
80000 10051.6
90000 9563.1
100000 9646.1
110000 9367.4
120000 9422.3
130000 9744
140000 9965.1
150000 10016.9
160000 10004.1
170000 10164.3
180000 9696.3
190000 9705.9
200000 10216.6
210000 9980.9
220000 9916.3
230000 10000.1
240000 9950.6
250000 9899.3
260000 10030.1
270000 10174.6
280000 9886.3
290000 10101.9
300000 9774.6
310000 10132.3
320000 9677.7
330000 9816.3
340000 10317.4
350000 9851.1
360000 9946.7
370000 9715.6
380000 10232.7
390000 10284.1
};
\addplot [semithick, darkorange2221435, mark=triangle*, mark size=1.5, mark options={solid}]
table {%
-10000 1462.9
0 8718.6
10000 9281.2
20000 9262.6
30000 9091.3
40000 8582.7
50000 9127.1
60000 9191.6
70000 8531.9
80000 9044.9
90000 8837.6
100000 8580.4
110000 8919.7
120000 9395.4
130000 9165.4
140000 8986.8
150000 9110
160000 9348.4
170000 9403.7
180000 9771.5
190000 9891.6
200000 10080.2
210000 9943.8
220000 9595.1
230000 9802.6
240000 9615
250000 9829.5
260000 9880.4
270000 9502.4
280000 10082.2
290000 10112.9
300000 10288.4
310000 10474.7
320000 10537.1
330000 10260.2
340000 10403.8
350000 10021.8
360000 10131.7
370000 10321.2
380000 10311.5
390000 10150.2
};
\addplot [semithick, orchid204120188, mark=+, mark size=1.5, mark options={solid}]
table {%
-10000 1710.4
0 6673
10000 8035.6
20000 7724.9
30000 7976.4
40000 8172.7
50000 7954.7
60000 8139.1
70000 8362
80000 8621.3
90000 8573.3
100000 8762.1
110000 8518.9
120000 8937.9
130000 9034.1
140000 8893.4
150000 8950
160000 9020.1
170000 8716.7
180000 8901
190000 9062.4
200000 8941.1
210000 8976.9
220000 9309.7
230000 8866.1
240000 9305.6
250000 9184.3
260000 9157.6
270000 9063.1
280000 8505.3
290000 8870.1
300000 8871
310000 9182.4
320000 9215.3
330000 9363.1
340000 8674.9
350000 9038
360000 9208.3
370000 8887
380000 8922
390000 9366.9
};
\addplot [semithick, peru20214597, mark=diamond*, mark size=1.5, mark options={solid}]
table {%
-10000 1558.6
0 2
10000 1.7
20000 259.9
30000 209.1
40000 207
50000 914.7
60000 1452.7
70000 1483.7
80000 1635.6
90000 1703.6
100000 1713.4
110000 1532
120000 1698
130000 3946.4
140000 5963.3
150000 4184.3
160000 5129.3
170000 3345.1
180000 3207.3
190000 2722.1
200000 4184.3
210000 3838.4
220000 4527.3
230000 4377.7
240000 5068.6
250000 4251
260000 4375.6
270000 5626.6
280000 5026.3
290000 5613.6
300000 4520.3
310000 4713.9
320000 5428.1
330000 4804.7
340000 4334.1
350000 6123.1
360000 6601.6
370000 4861.3
380000 6583.4
390000 5593.6
};
\end{axis}

\end{tikzpicture}

%% file: Image/kung_fu_master.tex
\begin{tikzpicture}

\definecolor{darkcyan1115178}{RGB}{1,115,178}
\definecolor{darkorange2221435}{RGB}{222,143,5}
\definecolor{darkslategray38}{RGB}{38,38,38}
\definecolor{lavender234234242}{RGB}{234,234,242}
\definecolor{lightgray204}{RGB}{204,204,204}
\definecolor{orchid204120188}{RGB}{204,120,188}
\definecolor{peru20214597}{RGB}{202,145,97}

\begin{axis}[
axis background/.style={fill=lavender234234242},
axis line style={white},
legend cell align={left},
legend style={
  fill opacity=0.8,
  draw opacity=1,
  text opacity=1,
  at={(0.03,0.97)},
  anchor=north west,
  draw=lightgray204,
  fill=lavender234234242
},
tick align=outside,
x grid style={white},
xlabel=\textcolor{darkslategray38}{Time Step},
xmajorgrids,
xmajorticks=true,
xmin=-30000, xmax=410000,
xtick style={color=darkslategray38},
xtick={0,50000,100000,150000,200000,250000,300000,350000,400000},
xticklabels={0k,50k,100k,150k,200k,250k,300k,350k,400k},
y grid style={white},
ylabel=\textcolor{darkslategray38}{Performance},
ymajorgrids,
ymajorticks=true,
ymin=-5862.23670955173, ymax=45492.2367095517,
ytick style={color=darkslategray38}
]
\path [draw=darkcyan1115178, fill=darkcyan1115178, opacity=0.2]
(axis cs:-10000,4116.54246322884)
--(axis cs:-10000,-3527.94246322884)
--(axis cs:0,13826.3575367712)
--(axis cs:10000,13843.4575367712)
--(axis cs:20000,13506.3575367712)
--(axis cs:30000,16297.7575367712)
--(axis cs:40000,17820.6575367712)
--(axis cs:50000,12714.8575367712)
--(axis cs:60000,12360.6575367712)
--(axis cs:70000,11792.0575367712)
--(axis cs:80000,12723.4575367712)
--(axis cs:90000,12509.1575367712)
--(axis cs:100000,13403.4575367712)
--(axis cs:110000,15326.3575367712)
--(axis cs:120000,14964.8575367712)
--(axis cs:130000,18047.7575367712)
--(axis cs:140000,17524.8575367712)
--(axis cs:150000,18610.6575367712)
--(axis cs:160000,19342.0575367712)
--(axis cs:170000,19043.4575367712)
--(axis cs:180000,19497.7575367712)
--(axis cs:190000,21329.1575367712)
--(axis cs:200000,22706.3575367712)
--(axis cs:210000,23666.3575367712)
--(axis cs:220000,25552.0575367712)
--(axis cs:230000,25803.4575367712)
--(axis cs:240000,26634.8575367712)
--(axis cs:250000,29454.8575367712)
--(axis cs:260000,28387.7575367712)
--(axis cs:270000,29979.1575367712)
--(axis cs:280000,31946.3575367712)
--(axis cs:290000,30344.8575367712)
--(axis cs:300000,28914.8575367712)
--(axis cs:310000,32836.3575367712)
--(axis cs:320000,33144.8575367712)
--(axis cs:330000,35484.8575367712)
--(axis cs:340000,33523.4575367712)
--(axis cs:350000,32956.3575367712)
--(axis cs:360000,33394.8575367712)
--(axis cs:370000,33867.7575367712)
--(axis cs:380000,35513.4575367712)
--(axis cs:390000,33180.6575367712)
--(axis cs:390000,40825.1424632288)
--(axis cs:390000,40825.1424632288)
--(axis cs:380000,43157.9424632288)
--(axis cs:370000,41512.2424632288)
--(axis cs:360000,41039.3424632288)
--(axis cs:350000,40600.8424632288)
--(axis cs:340000,41167.9424632288)
--(axis cs:330000,43129.3424632288)
--(axis cs:320000,40789.3424632288)
--(axis cs:310000,40480.8424632288)
--(axis cs:300000,36559.3424632288)
--(axis cs:290000,37989.3424632288)
--(axis cs:280000,39590.8424632288)
--(axis cs:270000,37623.6424632288)
--(axis cs:260000,36032.2424632288)
--(axis cs:250000,37099.3424632288)
--(axis cs:240000,34279.3424632288)
--(axis cs:230000,33447.9424632288)
--(axis cs:220000,33196.5424632288)
--(axis cs:210000,31310.8424632288)
--(axis cs:200000,30350.8424632288)
--(axis cs:190000,28973.6424632288)
--(axis cs:180000,27142.2424632288)
--(axis cs:170000,26687.9424632288)
--(axis cs:160000,26986.5424632288)
--(axis cs:150000,26255.1424632288)
--(axis cs:140000,25169.3424632288)
--(axis cs:130000,25692.2424632288)
--(axis cs:120000,22609.3424632288)
--(axis cs:110000,22970.8424632288)
--(axis cs:100000,21047.9424632288)
--(axis cs:90000,20153.6424632288)
--(axis cs:80000,20367.9424632288)
--(axis cs:70000,19436.5424632288)
--(axis cs:60000,20005.1424632288)
--(axis cs:50000,20359.3424632288)
--(axis cs:40000,25465.1424632288)
--(axis cs:30000,23942.2424632288)
--(axis cs:20000,21150.8424632288)
--(axis cs:10000,21487.9424632288)
--(axis cs:0,21470.8424632288)
--(axis cs:-10000,4116.54246322884)
--cycle;

\path [draw=darkorange2221435, fill=darkorange2221435, opacity=0.2]
(axis cs:-10000,2233.63017297826)
--(axis cs:-10000,-1605.63017297826)
--(axis cs:0,15850.3698270217)
--(axis cs:10000,13158.3698270217)
--(axis cs:20000,14622.3698270217)
--(axis cs:30000,14965.3698270217)
--(axis cs:40000,15150.3698270217)
--(axis cs:50000,14720.3698270217)
--(axis cs:60000,14755.3698270217)
--(axis cs:70000,15243.3698270217)
--(axis cs:80000,14919.3698270217)
--(axis cs:90000,15708.3698270217)
--(axis cs:100000,15953.3698270217)
--(axis cs:110000,17451.3698270217)
--(axis cs:120000,18266.3698270217)
--(axis cs:130000,17377.3698270217)
--(axis cs:140000,18447.3698270217)
--(axis cs:150000,19576.3698270217)
--(axis cs:160000,19888.3698270217)
--(axis cs:170000,20478.3698270217)
--(axis cs:180000,22524.3698270217)
--(axis cs:190000,23157.3698270217)
--(axis cs:200000,22872.3698270217)
--(axis cs:210000,24823.3698270217)
--(axis cs:220000,25071.3698270217)
--(axis cs:230000,25427.3698270217)
--(axis cs:240000,25725.3698270217)
--(axis cs:250000,26129.3698270217)
--(axis cs:260000,26470.3698270217)
--(axis cs:270000,28362.3698270217)
--(axis cs:280000,28065.3698270217)
--(axis cs:290000,27705.3698270217)
--(axis cs:300000,29199.3698270217)
--(axis cs:310000,28808.3698270217)
--(axis cs:320000,29728.3698270217)
--(axis cs:330000,31201.3698270217)
--(axis cs:340000,32611.3698270217)
--(axis cs:350000,31418.3698270217)
--(axis cs:360000,31662.3698270217)
--(axis cs:370000,33822.3698270217)
--(axis cs:380000,32314.3698270217)
--(axis cs:390000,33287.3698270217)
--(axis cs:390000,37126.6301729783)
--(axis cs:390000,37126.6301729783)
--(axis cs:380000,36153.6301729783)
--(axis cs:370000,37661.6301729783)
--(axis cs:360000,35501.6301729783)
--(axis cs:350000,35257.6301729783)
--(axis cs:340000,36450.6301729783)
--(axis cs:330000,35040.6301729783)
--(axis cs:320000,33567.6301729783)
--(axis cs:310000,32647.6301729783)
--(axis cs:300000,33038.6301729783)
--(axis cs:290000,31544.6301729783)
--(axis cs:280000,31904.6301729783)
--(axis cs:270000,32201.6301729783)
--(axis cs:260000,30309.6301729783)
--(axis cs:250000,29968.6301729783)
--(axis cs:240000,29564.6301729783)
--(axis cs:230000,29266.6301729783)
--(axis cs:220000,28910.6301729783)
--(axis cs:210000,28662.6301729783)
--(axis cs:200000,26711.6301729783)
--(axis cs:190000,26996.6301729783)
--(axis cs:180000,26363.6301729783)
--(axis cs:170000,24317.6301729783)
--(axis cs:160000,23727.6301729783)
--(axis cs:150000,23415.6301729783)
--(axis cs:140000,22286.6301729783)
--(axis cs:130000,21216.6301729783)
--(axis cs:120000,22105.6301729783)
--(axis cs:110000,21290.6301729783)
--(axis cs:100000,19792.6301729783)
--(axis cs:90000,19547.6301729783)
--(axis cs:80000,18758.6301729783)
--(axis cs:70000,19082.6301729783)
--(axis cs:60000,18594.6301729783)
--(axis cs:50000,18559.6301729783)
--(axis cs:40000,18989.6301729783)
--(axis cs:30000,18804.6301729783)
--(axis cs:20000,18461.6301729783)
--(axis cs:10000,16997.6301729783)
--(axis cs:0,19689.6301729783)
--(axis cs:-10000,2233.63017297826)
--cycle;

\path [draw=orchid204120188, fill=orchid204120188, opacity=0.2]
(axis cs:-10000,3684.75467088873)
--(axis cs:-10000,-3064.75467088873)
--(axis cs:0,23662.3453291113)
--(axis cs:10000,17725.2453291113)
--(axis cs:20000,16699.5453291113)
--(axis cs:30000,10225.2453291113)
--(axis cs:40000,10215.2453291113)
--(axis cs:50000,9298.14532911127)
--(axis cs:60000,8275.24532911127)
--(axis cs:70000,9612.34532911127)
--(axis cs:80000,9948.14532911127)
--(axis cs:90000,9740.94532911127)
--(axis cs:100000,11800.9453291113)
--(axis cs:110000,11756.6453291113)
--(axis cs:120000,12165.2453291113)
--(axis cs:130000,12050.9453291113)
--(axis cs:140000,13855.2453291113)
--(axis cs:150000,14332.3453291113)
--(axis cs:160000,16425.2453291113)
--(axis cs:170000,14590.9453291113)
--(axis cs:180000,16456.6453291113)
--(axis cs:190000,16292.3453291113)
--(axis cs:200000,17263.8453291113)
--(axis cs:210000,17310.9453291113)
--(axis cs:220000,18293.8453291113)
--(axis cs:230000,19288.1453291113)
--(axis cs:240000,20819.5453291113)
--(axis cs:250000,21340.9453291113)
--(axis cs:260000,22125.2453291113)
--(axis cs:270000,20743.8453291113)
--(axis cs:280000,22499.5453291113)
--(axis cs:290000,22482.3453291113)
--(axis cs:300000,23136.6453291113)
--(axis cs:310000,23796.6453291113)
--(axis cs:320000,24778.1453291113)
--(axis cs:330000,24540.9453291113)
--(axis cs:340000,25426.6453291113)
--(axis cs:350000,26789.5453291113)
--(axis cs:360000,26828.1453291113)
--(axis cs:370000,28390.9453291113)
--(axis cs:380000,27712.3453291113)
--(axis cs:390000,28055.2453291113)
--(axis cs:390000,34804.7546708887)
--(axis cs:390000,34804.7546708887)
--(axis cs:380000,34461.8546708887)
--(axis cs:370000,35140.4546708887)
--(axis cs:360000,33577.6546708887)
--(axis cs:350000,33539.0546708887)
--(axis cs:340000,32176.1546708887)
--(axis cs:330000,31290.4546708887)
--(axis cs:320000,31527.6546708887)
--(axis cs:310000,30546.1546708887)
--(axis cs:300000,29886.1546708887)
--(axis cs:290000,29231.8546708887)
--(axis cs:280000,29249.0546708887)
--(axis cs:270000,27493.3546708887)
--(axis cs:260000,28874.7546708887)
--(axis cs:250000,28090.4546708887)
--(axis cs:240000,27569.0546708887)
--(axis cs:230000,26037.6546708887)
--(axis cs:220000,25043.3546708887)
--(axis cs:210000,24060.4546708887)
--(axis cs:200000,24013.3546708887)
--(axis cs:190000,23041.8546708887)
--(axis cs:180000,23206.1546708887)
--(axis cs:170000,21340.4546708887)
--(axis cs:160000,23174.7546708887)
--(axis cs:150000,21081.8546708887)
--(axis cs:140000,20604.7546708887)
--(axis cs:130000,18800.4546708887)
--(axis cs:120000,18914.7546708887)
--(axis cs:110000,18506.1546708887)
--(axis cs:100000,18550.4546708887)
--(axis cs:90000,16490.4546708887)
--(axis cs:80000,16697.6546708887)
--(axis cs:70000,16361.8546708887)
--(axis cs:60000,15024.7546708887)
--(axis cs:50000,16047.6546708887)
--(axis cs:40000,16964.7546708887)
--(axis cs:30000,16974.7546708887)
--(axis cs:20000,23449.0546708887)
--(axis cs:10000,24474.7546708887)
--(axis cs:0,30411.8546708887)
--(axis cs:-10000,3684.75467088873)
--cycle;

\path [draw=peru20214597, fill=peru20214597, opacity=0.2]
(axis cs:-10000,1752.14668965666)
--(axis cs:-10000,-1340.74668965666)
--(axis cs:0,16356.4533103433)
--(axis cs:10000,21304.9533103433)
--(axis cs:20000,22133.5533103433)
--(axis cs:30000,23884.9533103433)
--(axis cs:40000,24910.6533103433)
--(axis cs:50000,25119.2533103433)
--(axis cs:60000,25150.6533103433)
--(axis cs:70000,23790.6533103433)
--(axis cs:80000,22653.5533103433)
--(axis cs:90000,22967.8533103433)
--(axis cs:100000,19890.6533103433)
--(axis cs:110000,19827.8533103433)
--(axis cs:120000,20019.2533103433)
--(axis cs:130000,19293.5533103433)
--(axis cs:140000,19324.9533103433)
--(axis cs:150000,19473.5533103433)
--(axis cs:160000,19739.2533103433)
--(axis cs:170000,20087.8533103433)
--(axis cs:180000,19279.2533103433)
--(axis cs:190000,20202.1533103433)
--(axis cs:200000,20776.4533103433)
--(axis cs:210000,18913.5533103433)
--(axis cs:220000,17302.1533103433)
--(axis cs:230000,18604.9533103433)
--(axis cs:240000,19653.5533103433)
--(axis cs:250000,18607.8533103433)
--(axis cs:260000,18870.6533103433)
--(axis cs:270000,19982.1533103433)
--(axis cs:280000,18064.9533103433)
--(axis cs:290000,18953.5533103433)
--(axis cs:300000,19142.1533103433)
--(axis cs:310000,19659.2533103433)
--(axis cs:320000,20313.5533103433)
--(axis cs:330000,20019.2533103433)
--(axis cs:340000,20547.8533103433)
--(axis cs:350000,19984.9533103433)
--(axis cs:360000,20764.9533103433)
--(axis cs:370000,19924.9533103433)
--(axis cs:380000,20967.8533103433)
--(axis cs:390000,18907.8533103433)
--(axis cs:390000,22000.7466896567)
--(axis cs:390000,22000.7466896567)
--(axis cs:380000,24060.7466896567)
--(axis cs:370000,23017.8466896567)
--(axis cs:360000,23857.8466896567)
--(axis cs:350000,23077.8466896567)
--(axis cs:340000,23640.7466896567)
--(axis cs:330000,23112.1466896567)
--(axis cs:320000,23406.4466896567)
--(axis cs:310000,22752.1466896567)
--(axis cs:300000,22235.0466896567)
--(axis cs:290000,22046.4466896567)
--(axis cs:280000,21157.8466896567)
--(axis cs:270000,23075.0466896567)
--(axis cs:260000,21963.5466896567)
--(axis cs:250000,21700.7466896567)
--(axis cs:240000,22746.4466896567)
--(axis cs:230000,21697.8466896567)
--(axis cs:220000,20395.0466896567)
--(axis cs:210000,22006.4466896567)
--(axis cs:200000,23869.3466896567)
--(axis cs:190000,23295.0466896567)
--(axis cs:180000,22372.1466896567)
--(axis cs:170000,23180.7466896567)
--(axis cs:160000,22832.1466896567)
--(axis cs:150000,22566.4466896567)
--(axis cs:140000,22417.8466896567)
--(axis cs:130000,22386.4466896567)
--(axis cs:120000,23112.1466896567)
--(axis cs:110000,22920.7466896567)
--(axis cs:100000,22983.5466896567)
--(axis cs:90000,26060.7466896567)
--(axis cs:80000,25746.4466896567)
--(axis cs:70000,26883.5466896567)
--(axis cs:60000,28243.5466896567)
--(axis cs:50000,28212.1466896567)
--(axis cs:40000,28003.5466896567)
--(axis cs:30000,26977.8466896567)
--(axis cs:20000,25226.4466896567)
--(axis cs:10000,24397.8466896567)
--(axis cs:0,19449.3466896567)
--(axis cs:-10000,1752.14668965666)
--cycle;

\addplot [semithick, darkcyan1115178, mark=square*, mark size=1.5, mark options={solid}]
table {%
-10000 294.3
0 17648.6
10000 17665.7
20000 17328.6
30000 20120
40000 21642.9
50000 16537.1
60000 16182.9
70000 15614.3
80000 16545.7
90000 16331.4
100000 17225.7
110000 19148.6
120000 18787.1
130000 21870
140000 21347.1
150000 22432.9
160000 23164.3
170000 22865.7
180000 23320
190000 25151.4
200000 26528.6
210000 27488.6
220000 29374.3
230000 29625.7
240000 30457.1
250000 33277.1
260000 32210
270000 33801.4
280000 35768.6
290000 34167.1
300000 32737.1
310000 36658.6
320000 36967.1
330000 39307.1
340000 37345.7
350000 36778.6
360000 37217.1
370000 37690
380000 39335.7
390000 37002.9
};
\addplot [semithick, darkorange2221435, mark=triangle*, mark size=1.5, mark options={solid}]
table {%
-10000 314
0 17770
10000 15078
20000 16542
30000 16885
40000 17070
50000 16640
60000 16675
70000 17163
80000 16839
90000 17628
100000 17873
110000 19371
120000 20186
130000 19297
140000 20367
150000 21496
160000 21808
170000 22398
180000 24444
190000 25077
200000 24792
210000 26743
220000 26991
230000 27347
240000 27645
250000 28049
260000 28390
270000 30282
280000 29985
290000 29625
300000 31119
310000 30728
320000 31648
330000 33121
340000 34531
350000 33338
360000 33582
370000 35742
380000 34234
390000 35207
};
\addplot [semithick, orchid204120188, mark=+, mark size=1.5, mark options={solid}]
table {%
-10000 310
0 27037.1
10000 21100
20000 20074.3
30000 13600
40000 13590
50000 12672.9
60000 11650
70000 12987.1
80000 13322.9
90000 13115.7
100000 15175.7
110000 15131.4
120000 15540
130000 15425.7
140000 17230
150000 17707.1
160000 19800
170000 17965.7
180000 19831.4
190000 19667.1
200000 20638.6
210000 20685.7
220000 21668.6
230000 22662.9
240000 24194.3
250000 24715.7
260000 25500
270000 24118.6
280000 25874.3
290000 25857.1
300000 26511.4
310000 27171.4
320000 28152.9
330000 27915.7
340000 28801.4
350000 30164.3
360000 30202.9
370000 31765.7
380000 31087.1
390000 31430
};
\addplot [semithick, peru20214597, mark=diamond*, mark size=1.5, mark options={solid}]
table {%
-10000 205.7
0 17902.9
10000 22851.4
20000 23680
30000 25431.4
40000 26457.1
50000 26665.7
60000 26697.1
70000 25337.1
80000 24200
90000 24514.3
100000 21437.1
110000 21374.3
120000 21565.7
130000 20840
140000 20871.4
150000 21020
160000 21285.7
170000 21634.3
180000 20825.7
190000 21748.6
200000 22322.9
210000 20460
220000 18848.6
230000 20151.4
240000 21200
250000 20154.3
260000 20417.1
270000 21528.6
280000 19611.4
290000 20500
300000 20688.6
310000 21205.7
320000 21860
330000 21565.7
340000 22094.3
350000 21531.4
360000 22311.4
370000 21471.4
380000 22514.3
390000 20454.3
};
\end{axis}

\end{tikzpicture}

%% file: Image/ms_pacman.tex
\begin{tikzpicture}

\definecolor{darkcyan1115178}{RGB}{1,115,178}
\definecolor{darkorange2221435}{RGB}{222,143,5}
\definecolor{darkslategray38}{RGB}{38,38,38}
\definecolor{lavender234234242}{RGB}{234,234,242}
\definecolor{lightgray204}{RGB}{204,204,204}
\definecolor{orchid204120188}{RGB}{204,120,188}
\definecolor{peru20214597}{RGB}{202,145,97}

\begin{axis}[
axis background/.style={fill=lavender234234242},
axis line style={white},
legend cell align={left},
legend style={
  fill opacity=0.8,
  draw opacity=1,
  text opacity=1,
  at={(0.03,0.97)},
  anchor=north west,
  draw=lightgray204,
  fill=lavender234234242
},
tick align=outside,
x grid style={white},
xlabel=\textcolor{darkslategray38}{Time Step},
xmajorgrids,
xmajorticks=true,
xmin=-30000, xmax=410000,
xtick style={color=darkslategray38},
xtick={0,50000,100000,150000,200000,250000,300000,350000,400000},
xticklabels={0k,50k,100k,150k,200k,250k,300k,350k,400k},
y grid style={white},
ylabel=\textcolor{darkslategray38}{Performance},
ymajorgrids,
ymajorticks=true,
ymin=-98.4970214359015, ymax=2859.14649429919,
ytick style={color=darkslategray38}
]
\path [draw=darkcyan1115178, fill=darkcyan1115178, opacity=0.2]
(axis cs:-10000,571.308152674868)
--(axis cs:-10000,202.091847325132)
--(axis cs:0,527.091847325132)
--(axis cs:10000,652.491847325132)
--(axis cs:20000,720.991847325132)
--(axis cs:30000,948.791847325132)
--(axis cs:40000,1116.99184732513)
--(axis cs:50000,1569.79184732513)
--(axis cs:60000,1465.39184732513)
--(axis cs:70000,1616.99184732513)
--(axis cs:80000,1699.39184732513)
--(axis cs:90000,1820.99184732513)
--(axis cs:100000,1625.49184732513)
--(axis cs:110000,1749.79184732513)
--(axis cs:120000,1819.49184732513)
--(axis cs:130000,1673.49184732513)
--(axis cs:140000,1939.69184732513)
--(axis cs:150000,1911.49184732513)
--(axis cs:160000,1969.39184732513)
--(axis cs:170000,2067.39184732513)
--(axis cs:180000,2093.79184732513)
--(axis cs:190000,2289.09184732513)
--(axis cs:200000,2234.69184732513)
--(axis cs:210000,2192.29184732513)
--(axis cs:220000,2236.29184732513)
--(axis cs:230000,2355.49184732513)
--(axis cs:240000,2239.49184732513)
--(axis cs:250000,2243.99184732513)
--(axis cs:260000,2274.29184732513)
--(axis cs:270000,2182.99184732513)
--(axis cs:280000,2190.29184732513)
--(axis cs:290000,2031.09184732513)
--(axis cs:300000,2100.49184732513)
--(axis cs:310000,2246.39184732513)
--(axis cs:320000,1995.39184732513)
--(axis cs:330000,2132.69184732513)
--(axis cs:340000,2024.49184732513)
--(axis cs:350000,2236.49184732513)
--(axis cs:360000,2243.49184732513)
--(axis cs:370000,2145.99184732513)
--(axis cs:380000,2147.69184732513)
--(axis cs:390000,2217.99184732513)
--(axis cs:390000,2587.20815267487)
--(axis cs:390000,2587.20815267487)
--(axis cs:380000,2516.90815267487)
--(axis cs:370000,2515.20815267487)
--(axis cs:360000,2612.70815267487)
--(axis cs:350000,2605.70815267487)
--(axis cs:340000,2393.70815267487)
--(axis cs:330000,2501.90815267487)
--(axis cs:320000,2364.60815267487)
--(axis cs:310000,2615.60815267487)
--(axis cs:300000,2469.70815267487)
--(axis cs:290000,2400.30815267487)
--(axis cs:280000,2559.50815267487)
--(axis cs:270000,2552.20815267487)
--(axis cs:260000,2643.50815267487)
--(axis cs:250000,2613.20815267487)
--(axis cs:240000,2608.70815267487)
--(axis cs:230000,2724.70815267487)
--(axis cs:220000,2605.50815267487)
--(axis cs:210000,2561.50815267487)
--(axis cs:200000,2603.90815267487)
--(axis cs:190000,2658.30815267487)
--(axis cs:180000,2463.00815267487)
--(axis cs:170000,2436.60815267487)
--(axis cs:160000,2338.60815267487)
--(axis cs:150000,2280.70815267487)
--(axis cs:140000,2308.90815267487)
--(axis cs:130000,2042.70815267487)
--(axis cs:120000,2188.70815267487)
--(axis cs:110000,2119.00815267487)
--(axis cs:100000,1994.70815267487)
--(axis cs:90000,2190.20815267487)
--(axis cs:80000,2068.60815267487)
--(axis cs:70000,1986.20815267487)
--(axis cs:60000,1834.60815267487)
--(axis cs:50000,1939.00815267487)
--(axis cs:40000,1486.20815267487)
--(axis cs:30000,1318.00815267487)
--(axis cs:20000,1090.20815267487)
--(axis cs:10000,1021.70815267487)
--(axis cs:0,896.308152674868)
--(axis cs:-10000,571.308152674868)
--cycle;

\path [draw=darkorange2221435, fill=darkorange2221435, opacity=0.2]
(axis cs:-10000,527.473796637699)
--(axis cs:-10000,218.926203362301)
--(axis cs:0,334.226203362301)
--(axis cs:10000,533.726203362301)
--(axis cs:20000,483.026203362301)
--(axis cs:30000,538.426203362302)
--(axis cs:40000,575.326203362301)
--(axis cs:50000,557.826203362301)
--(axis cs:60000,605.426203362302)
--(axis cs:70000,623.026203362301)
--(axis cs:80000,661.526203362301)
--(axis cs:90000,732.326203362301)
--(axis cs:100000,856.326203362301)
--(axis cs:110000,988.326203362301)
--(axis cs:120000,1018.3262033623)
--(axis cs:130000,1072.4262033623)
--(axis cs:140000,1013.9262033623)
--(axis cs:150000,1175.8262033623)
--(axis cs:160000,1235.1262033623)
--(axis cs:170000,1240.4262033623)
--(axis cs:180000,1198.1262033623)
--(axis cs:190000,1332.1262033623)
--(axis cs:200000,1266.6262033623)
--(axis cs:210000,1387.2262033623)
--(axis cs:220000,1407.9262033623)
--(axis cs:230000,1253.7262033623)
--(axis cs:240000,1477.9262033623)
--(axis cs:250000,1391.3262033623)
--(axis cs:260000,1402.8262033623)
--(axis cs:270000,1329.3262033623)
--(axis cs:280000,1339.5262033623)
--(axis cs:290000,1328.4262033623)
--(axis cs:300000,1363.9262033623)
--(axis cs:310000,1359.8262033623)
--(axis cs:320000,1375.9262033623)
--(axis cs:330000,1362.6262033623)
--(axis cs:340000,1378.4262033623)
--(axis cs:350000,1510.6262033623)
--(axis cs:360000,1447.7262033623)
--(axis cs:370000,1585.9262033623)
--(axis cs:380000,1503.6262033623)
--(axis cs:390000,1502.2262033623)
--(axis cs:390000,1810.7737966377)
--(axis cs:390000,1810.7737966377)
--(axis cs:380000,1812.1737966377)
--(axis cs:370000,1894.4737966377)
--(axis cs:360000,1756.2737966377)
--(axis cs:350000,1819.1737966377)
--(axis cs:340000,1686.9737966377)
--(axis cs:330000,1671.1737966377)
--(axis cs:320000,1684.4737966377)
--(axis cs:310000,1668.3737966377)
--(axis cs:300000,1672.4737966377)
--(axis cs:290000,1636.9737966377)
--(axis cs:280000,1648.0737966377)
--(axis cs:270000,1637.8737966377)
--(axis cs:260000,1711.3737966377)
--(axis cs:250000,1699.8737966377)
--(axis cs:240000,1786.4737966377)
--(axis cs:230000,1562.2737966377)
--(axis cs:220000,1716.4737966377)
--(axis cs:210000,1695.7737966377)
--(axis cs:200000,1575.1737966377)
--(axis cs:190000,1640.6737966377)
--(axis cs:180000,1506.6737966377)
--(axis cs:170000,1548.9737966377)
--(axis cs:160000,1543.6737966377)
--(axis cs:150000,1484.3737966377)
--(axis cs:140000,1322.4737966377)
--(axis cs:130000,1380.9737966377)
--(axis cs:120000,1326.8737966377)
--(axis cs:110000,1296.8737966377)
--(axis cs:100000,1164.8737966377)
--(axis cs:90000,1040.8737966377)
--(axis cs:80000,970.073796637698)
--(axis cs:70000,931.573796637698)
--(axis cs:60000,913.973796637699)
--(axis cs:50000,866.373796637699)
--(axis cs:40000,883.873796637699)
--(axis cs:30000,846.973796637699)
--(axis cs:20000,791.573796637698)
--(axis cs:10000,842.273796637699)
--(axis cs:0,642.773796637699)
--(axis cs:-10000,527.473796637699)
--cycle;

\path [draw=orchid204120188, fill=orchid204120188, opacity=0.2]
(axis cs:-10000,570.870335137167)
--(axis cs:-10000,184.929664862833)
--(axis cs:0,228.429664862833)
--(axis cs:10000,447.329664862833)
--(axis cs:20000,665.129664862833)
--(axis cs:30000,813.029664862833)
--(axis cs:40000,1042.42966486283)
--(axis cs:50000,1012.72966486283)
--(axis cs:60000,1136.12966486283)
--(axis cs:70000,1201.32966486283)
--(axis cs:80000,1391.42966486283)
--(axis cs:90000,1412.02966486283)
--(axis cs:100000,1354.62966486283)
--(axis cs:110000,1452.92966486283)
--(axis cs:120000,1384.62966486283)
--(axis cs:130000,1568.42966486283)
--(axis cs:140000,1558.02966486283)
--(axis cs:150000,1541.32966486283)
--(axis cs:160000,1551.32966486283)
--(axis cs:170000,1559.72966486283)
--(axis cs:180000,1451.72966486283)
--(axis cs:190000,1490.12966486283)
--(axis cs:200000,1572.02966486283)
--(axis cs:210000,1499.02966486283)
--(axis cs:220000,1600.72966486283)
--(axis cs:230000,1557.92966486283)
--(axis cs:240000,1543.92966486283)
--(axis cs:250000,1637.32966486283)
--(axis cs:260000,1678.92966486283)
--(axis cs:270000,1678.12966486283)
--(axis cs:280000,1614.02966486283)
--(axis cs:290000,1634.62966486283)
--(axis cs:300000,1772.72966486283)
--(axis cs:310000,1782.02966486283)
--(axis cs:320000,1776.62966486283)
--(axis cs:330000,1711.92966486283)
--(axis cs:340000,1784.02966486283)
--(axis cs:350000,1849.72966486283)
--(axis cs:360000,1892.42966486283)
--(axis cs:370000,1679.92966486283)
--(axis cs:380000,1644.02966486283)
--(axis cs:390000,1792.92966486283)
--(axis cs:390000,2178.87033513717)
--(axis cs:390000,2178.87033513717)
--(axis cs:380000,2029.97033513717)
--(axis cs:370000,2065.87033513717)
--(axis cs:360000,2278.37033513717)
--(axis cs:350000,2235.67033513717)
--(axis cs:340000,2169.97033513717)
--(axis cs:330000,2097.87033513717)
--(axis cs:320000,2162.57033513717)
--(axis cs:310000,2167.97033513717)
--(axis cs:300000,2158.67033513717)
--(axis cs:290000,2020.57033513717)
--(axis cs:280000,1999.97033513717)
--(axis cs:270000,2064.07033513717)
--(axis cs:260000,2064.87033513717)
--(axis cs:250000,2023.27033513717)
--(axis cs:240000,1929.87033513717)
--(axis cs:230000,1943.87033513717)
--(axis cs:220000,1986.67033513717)
--(axis cs:210000,1884.97033513717)
--(axis cs:200000,1957.97033513717)
--(axis cs:190000,1876.07033513717)
--(axis cs:180000,1837.67033513717)
--(axis cs:170000,1945.67033513717)
--(axis cs:160000,1937.27033513717)
--(axis cs:150000,1927.27033513717)
--(axis cs:140000,1943.97033513717)
--(axis cs:130000,1954.37033513717)
--(axis cs:120000,1770.57033513717)
--(axis cs:110000,1838.87033513717)
--(axis cs:100000,1740.57033513717)
--(axis cs:90000,1797.97033513717)
--(axis cs:80000,1777.37033513717)
--(axis cs:70000,1587.27033513717)
--(axis cs:60000,1522.07033513717)
--(axis cs:50000,1398.67033513717)
--(axis cs:40000,1428.37033513717)
--(axis cs:30000,1198.97033513717)
--(axis cs:20000,1051.07033513717)
--(axis cs:10000,833.270335137167)
--(axis cs:0,614.370335137167)
--(axis cs:-10000,570.870335137167)
--cycle;

\path [draw=peru20214597, fill=peru20214597, opacity=0.2]
(axis cs:-10000,499.858679811579)
--(axis cs:-10000,256.341320188421)
--(axis cs:0,35.9413201884208)
--(axis cs:10000,51.6413201884208)
--(axis cs:20000,160.541320188421)
--(axis cs:30000,232.141320188421)
--(axis cs:40000,243.841320188421)
--(axis cs:50000,207.341320188421)
--(axis cs:60000,203.341320188421)
--(axis cs:70000,273.241320188421)
--(axis cs:80000,269.541320188421)
--(axis cs:90000,310.141320188421)
--(axis cs:100000,273.941320188421)
--(axis cs:110000,309.941320188421)
--(axis cs:120000,329.141320188421)
--(axis cs:130000,465.141320188421)
--(axis cs:140000,377.241320188421)
--(axis cs:150000,360.541320188421)
--(axis cs:160000,443.141320188421)
--(axis cs:170000,439.841320188421)
--(axis cs:180000,511.241320188421)
--(axis cs:190000,439.341320188421)
--(axis cs:200000,499.241320188421)
--(axis cs:210000,523.141320188421)
--(axis cs:220000,535.141320188421)
--(axis cs:230000,468.541320188421)
--(axis cs:240000,562.941320188421)
--(axis cs:250000,513.241320188421)
--(axis cs:260000,604.341320188421)
--(axis cs:270000,477.341320188421)
--(axis cs:280000,555.241320188421)
--(axis cs:290000,544.641320188421)
--(axis cs:300000,471.341320188421)
--(axis cs:310000,573.141320188421)
--(axis cs:320000,531.241320188421)
--(axis cs:330000,517.141320188421)
--(axis cs:340000,572.941320188421)
--(axis cs:350000,608.341320188421)
--(axis cs:360000,639.641320188421)
--(axis cs:370000,602.141320188421)
--(axis cs:380000,744.841320188421)
--(axis cs:390000,662.641320188421)
--(axis cs:390000,906.158679811579)
--(axis cs:390000,906.158679811579)
--(axis cs:380000,988.358679811579)
--(axis cs:370000,845.658679811579)
--(axis cs:360000,883.158679811579)
--(axis cs:350000,851.858679811579)
--(axis cs:340000,816.458679811579)
--(axis cs:330000,760.658679811579)
--(axis cs:320000,774.758679811579)
--(axis cs:310000,816.658679811579)
--(axis cs:300000,714.858679811579)
--(axis cs:290000,788.158679811579)
--(axis cs:280000,798.758679811579)
--(axis cs:270000,720.858679811579)
--(axis cs:260000,847.858679811579)
--(axis cs:250000,756.758679811579)
--(axis cs:240000,806.458679811579)
--(axis cs:230000,712.058679811579)
--(axis cs:220000,778.658679811579)
--(axis cs:210000,766.658679811579)
--(axis cs:200000,742.758679811579)
--(axis cs:190000,682.858679811579)
--(axis cs:180000,754.758679811579)
--(axis cs:170000,683.358679811579)
--(axis cs:160000,686.658679811579)
--(axis cs:150000,604.058679811579)
--(axis cs:140000,620.758679811579)
--(axis cs:130000,708.658679811579)
--(axis cs:120000,572.658679811579)
--(axis cs:110000,553.458679811579)
--(axis cs:100000,517.458679811579)
--(axis cs:90000,553.658679811579)
--(axis cs:80000,513.058679811579)
--(axis cs:70000,516.758679811579)
--(axis cs:60000,446.858679811579)
--(axis cs:50000,450.858679811579)
--(axis cs:40000,487.358679811579)
--(axis cs:30000,475.658679811579)
--(axis cs:20000,404.058679811579)
--(axis cs:10000,295.158679811579)
--(axis cs:0,279.458679811579)
--(axis cs:-10000,499.858679811579)
--cycle;

\addplot [semithick, darkcyan1115178, mark=square*, mark size=1.5, mark options={solid}]
table {%
-10000 386.7
0 711.7
10000 837.1
20000 905.6
30000 1133.4
40000 1301.6
50000 1754.4
60000 1650
70000 1801.6
80000 1884
90000 2005.6
100000 1810.1
110000 1934.4
120000 2004.1
130000 1858.1
140000 2124.3
150000 2096.1
160000 2154
170000 2252
180000 2278.4
190000 2473.7
200000 2419.3
210000 2376.9
220000 2420.9
230000 2540.1
240000 2424.1
250000 2428.6
260000 2458.9
270000 2367.6
280000 2374.9
290000 2215.7
300000 2285.1
310000 2431
320000 2180
330000 2317.3
340000 2209.1
350000 2421.1
360000 2428.1
370000 2330.6
380000 2332.3
390000 2402.6
};
\addplot [semithick, darkorange2221435, mark=triangle*, mark size=1.5, mark options={solid}]
table {%
-10000 373.2
0 488.5
10000 688
20000 637.3
30000 692.7
40000 729.6
50000 712.1
60000 759.7
70000 777.3
80000 815.8
90000 886.6
100000 1010.6
110000 1142.6
120000 1172.6
130000 1226.7
140000 1168.2
150000 1330.1
160000 1389.4
170000 1394.7
180000 1352.4
190000 1486.4
200000 1420.9
210000 1541.5
220000 1562.2
230000 1408
240000 1632.2
250000 1545.6
260000 1557.1
270000 1483.6
280000 1493.8
290000 1482.7
300000 1518.2
310000 1514.1
320000 1530.2
330000 1516.9
340000 1532.7
350000 1664.9
360000 1602
370000 1740.2
380000 1657.9
390000 1656.5
};
\addplot [semithick, orchid204120188, mark=+, mark size=1.5, mark options={solid}]
table {%
-10000 377.9
0 421.4
10000 640.3
20000 858.1
30000 1006
40000 1235.4
50000 1205.7
60000 1329.1
70000 1394.3
80000 1584.4
90000 1605
100000 1547.6
110000 1645.9
120000 1577.6
130000 1761.4
140000 1751
150000 1734.3
160000 1744.3
170000 1752.7
180000 1644.7
190000 1683.1
200000 1765
210000 1692
220000 1793.7
230000 1750.9
240000 1736.9
250000 1830.3
260000 1871.9
270000 1871.1
280000 1807
290000 1827.6
300000 1965.7
310000 1975
320000 1969.6
330000 1904.9
340000 1977
350000 2042.7
360000 2085.4
370000 1872.9
380000 1837
390000 1985.9
};
\addplot [semithick, peru20214597, mark=diamond*, mark size=1.5, mark options={solid}]
table {%
-10000 378.1
0 157.7
10000 173.4
20000 282.3
30000 353.9
40000 365.6
50000 329.1
60000 325.1
70000 395
80000 391.3
90000 431.9
100000 395.7
110000 431.7
120000 450.9
130000 586.9
140000 499
150000 482.3
160000 564.9
170000 561.6
180000 633
190000 561.1
200000 621
210000 644.9
220000 656.9
230000 590.3
240000 684.7
250000 635
260000 726.1
270000 599.1
280000 677
290000 666.4
300000 593.1
310000 694.9
320000 653
330000 638.9
340000 694.7
350000 730.1
360000 761.4
370000 723.9
380000 866.6
390000 784.4
};
\end{axis}

\end{tikzpicture}

%% file: Image/pong.tex
\begin{tikzpicture}

\definecolor{darkcyan1115178}{RGB}{1,115,178}
\definecolor{darkorange2221435}{RGB}{222,143,5}
\definecolor{darkslategray38}{RGB}{38,38,38}
\definecolor{lavender234234242}{RGB}{234,234,242}
\definecolor{lightgray204}{RGB}{204,204,204}
\definecolor{orchid204120188}{RGB}{204,120,188}
\definecolor{peru20214597}{RGB}{202,145,97}

\begin{axis}[
axis background/.style={fill=lavender234234242},
axis line style={white},
legend cell align={left},
legend style={
  fill opacity=0.8,
  draw opacity=1,
  text opacity=1,
  at={(0.97,0.03)},
  anchor=south east,
  draw=lightgray204,
  fill=lavender234234242
},
tick align=outside,
x grid style={white},
xlabel=\textcolor{darkslategray38}{Time Step},
xmajorgrids,
xmajorticks=true,
xmin=-30000, xmax=410000,
xtick style={color=darkslategray38},
xtick={0,50000,100000,150000,200000,250000,300000,350000,400000},
xticklabels={0k,50k,100k,150k,200k,250k,300k,350k,400k},
y grid style={white},
ylabel=\textcolor{darkslategray38}{Performance},
ymajorgrids,
ymajorticks=true,
ymin=-26.0768495541978, ymax=26.1768495541978,
ytick style={color=darkslategray38}
]
\path [draw=darkcyan1115178, fill=darkcyan1115178, opacity=0.2]
(axis cs:-10000,-18.6006183033157)
--(axis cs:-10000,-22.9993816966843)
--(axis cs:0,-4.79938169668425)
--(axis cs:10000,7.90061830331575)
--(axis cs:20000,9.90061830331575)
--(axis cs:30000,12.3006183033157)
--(axis cs:40000,17.2006183033157)
--(axis cs:50000,18.3006183033157)
--(axis cs:60000,18.3006183033157)
--(axis cs:70000,18.2006183033157)
--(axis cs:80000,18.4006183033157)
--(axis cs:90000,18.5006183033157)
--(axis cs:100000,18.5006183033157)
--(axis cs:110000,18.5006183033157)
--(axis cs:120000,18.4006183033157)
--(axis cs:130000,18.4006183033157)
--(axis cs:140000,18.6006183033157)
--(axis cs:150000,18.6006183033157)
--(axis cs:160000,18.6006183033157)
--(axis cs:170000,18.8006183033157)
--(axis cs:180000,18.7006183033157)
--(axis cs:190000,18.8006183033157)
--(axis cs:200000,18.7006183033157)
--(axis cs:210000,18.8006183033157)
--(axis cs:220000,18.8006183033157)
--(axis cs:230000,18.7006183033157)
--(axis cs:240000,18.8006183033157)
--(axis cs:250000,18.8006183033157)
--(axis cs:260000,18.7006183033157)
--(axis cs:270000,18.8006183033157)
--(axis cs:280000,18.8006183033157)
--(axis cs:290000,18.8006183033157)
--(axis cs:300000,18.7006183033157)
--(axis cs:310000,18.8006183033157)
--(axis cs:320000,18.8006183033157)
--(axis cs:330000,18.8006183033157)
--(axis cs:340000,18.8006183033157)
--(axis cs:350000,18.7006183033157)
--(axis cs:360000,18.8006183033157)
--(axis cs:370000,18.8006183033157)
--(axis cs:380000,18.8006183033157)
--(axis cs:390000,18.7006183033157)
--(axis cs:390000,23.0993816966843)
--(axis cs:390000,23.0993816966843)
--(axis cs:380000,23.1993816966843)
--(axis cs:370000,23.1993816966843)
--(axis cs:360000,23.1993816966843)
--(axis cs:350000,23.0993816966843)
--(axis cs:340000,23.1993816966843)
--(axis cs:330000,23.1993816966843)
--(axis cs:320000,23.1993816966843)
--(axis cs:310000,23.1993816966843)
--(axis cs:300000,23.0993816966843)
--(axis cs:290000,23.1993816966843)
--(axis cs:280000,23.1993816966843)
--(axis cs:270000,23.1993816966843)
--(axis cs:260000,23.0993816966843)
--(axis cs:250000,23.1993816966843)
--(axis cs:240000,23.1993816966843)
--(axis cs:230000,23.0993816966843)
--(axis cs:220000,23.1993816966843)
--(axis cs:210000,23.1993816966843)
--(axis cs:200000,23.0993816966843)
--(axis cs:190000,23.1993816966843)
--(axis cs:180000,23.0993816966843)
--(axis cs:170000,23.1993816966843)
--(axis cs:160000,22.9993816966843)
--(axis cs:150000,22.9993816966843)
--(axis cs:140000,22.9993816966843)
--(axis cs:130000,22.7993816966843)
--(axis cs:120000,22.7993816966843)
--(axis cs:110000,22.8993816966843)
--(axis cs:100000,22.8993816966843)
--(axis cs:90000,22.8993816966843)
--(axis cs:80000,22.7993816966843)
--(axis cs:70000,22.5993816966843)
--(axis cs:60000,22.6993816966843)
--(axis cs:50000,22.6993816966843)
--(axis cs:40000,21.5993816966843)
--(axis cs:30000,16.6993816966843)
--(axis cs:20000,14.2993816966843)
--(axis cs:10000,12.2993816966843)
--(axis cs:0,-0.400618303315749)
--(axis cs:-10000,-18.6006183033157)
--cycle;

\path [draw=darkorange2221435, fill=darkorange2221435, opacity=0.2]
(axis cs:-10000,-18.8901305822908)
--(axis cs:-10000,-22.3098694177092)
--(axis cs:0,-5.10986941770918)
--(axis cs:10000,0.79013058229082)
--(axis cs:20000,7.39013058229082)
--(axis cs:30000,10.3901305822908)
--(axis cs:40000,11.9901305822908)
--(axis cs:50000,14.0901305822908)
--(axis cs:60000,13.3901305822908)
--(axis cs:70000,13.7901305822908)
--(axis cs:80000,14.2901305822908)
--(axis cs:90000,14.7901305822908)
--(axis cs:100000,14.4901305822908)
--(axis cs:110000,13.9901305822908)
--(axis cs:120000,16.2901305822908)
--(axis cs:130000,16.2901305822908)
--(axis cs:140000,16.2901305822908)
--(axis cs:150000,16.6901305822908)
--(axis cs:160000,17.2901305822908)
--(axis cs:170000,17.0901305822908)
--(axis cs:180000,17.9901305822908)
--(axis cs:190000,17.6901305822908)
--(axis cs:200000,17.7901305822908)
--(axis cs:210000,17.5901305822908)
--(axis cs:220000,17.5901305822908)
--(axis cs:230000,17.6901305822908)
--(axis cs:240000,17.7901305822908)
--(axis cs:250000,17.9901305822908)
--(axis cs:260000,18.4901305822908)
--(axis cs:270000,18.2901305822908)
--(axis cs:280000,18.3901305822908)
--(axis cs:290000,18.2901305822908)
--(axis cs:300000,17.8901305822908)
--(axis cs:310000,18.5901305822908)
--(axis cs:320000,18.5901305822908)
--(axis cs:330000,18.6901305822908)
--(axis cs:340000,18.6901305822908)
--(axis cs:350000,18.7901305822908)
--(axis cs:360000,18.7901305822908)
--(axis cs:370000,18.7901305822908)
--(axis cs:380000,18.1901305822908)
--(axis cs:390000,18.3901305822908)
--(axis cs:390000,21.8098694177092)
--(axis cs:390000,21.8098694177092)
--(axis cs:380000,21.6098694177092)
--(axis cs:370000,22.2098694177092)
--(axis cs:360000,22.2098694177092)
--(axis cs:350000,22.2098694177092)
--(axis cs:340000,22.1098694177092)
--(axis cs:330000,22.1098694177092)
--(axis cs:320000,22.0098694177092)
--(axis cs:310000,22.0098694177092)
--(axis cs:300000,21.3098694177092)
--(axis cs:290000,21.7098694177092)
--(axis cs:280000,21.8098694177092)
--(axis cs:270000,21.7098694177092)
--(axis cs:260000,21.9098694177092)
--(axis cs:250000,21.4098694177092)
--(axis cs:240000,21.2098694177092)
--(axis cs:230000,21.1098694177092)
--(axis cs:220000,21.0098694177092)
--(axis cs:210000,21.0098694177092)
--(axis cs:200000,21.2098694177092)
--(axis cs:190000,21.1098694177092)
--(axis cs:180000,21.4098694177092)
--(axis cs:170000,20.5098694177092)
--(axis cs:160000,20.7098694177092)
--(axis cs:150000,20.1098694177092)
--(axis cs:140000,19.7098694177092)
--(axis cs:130000,19.7098694177092)
--(axis cs:120000,19.7098694177092)
--(axis cs:110000,17.4098694177092)
--(axis cs:100000,17.9098694177092)
--(axis cs:90000,18.2098694177092)
--(axis cs:80000,17.7098694177092)
--(axis cs:70000,17.2098694177092)
--(axis cs:60000,16.8098694177092)
--(axis cs:50000,17.5098694177092)
--(axis cs:40000,15.4098694177092)
--(axis cs:30000,13.8098694177092)
--(axis cs:20000,10.8098694177092)
--(axis cs:10000,4.20986941770918)
--(axis cs:0,-1.69013058229082)
--(axis cs:-10000,-18.8901305822908)
--cycle;

\path [draw=orchid204120188, fill=orchid204120188, opacity=0.2]
(axis cs:-10000,-18.6102596402798)
--(axis cs:-10000,-23.1897403597202)
--(axis cs:0,-9.6897403597202)
--(axis cs:10000,4.2102596402798)
--(axis cs:20000,10.6102596402798)
--(axis cs:30000,10.9102596402798)
--(axis cs:40000,14.8102596402798)
--(axis cs:50000,16.7102596402798)
--(axis cs:60000,16.6102596402798)
--(axis cs:70000,16.8102596402798)
--(axis cs:80000,17.5102596402798)
--(axis cs:90000,17.2102596402798)
--(axis cs:100000,17.4102596402798)
--(axis cs:110000,17.2102596402798)
--(axis cs:120000,17.3102596402798)
--(axis cs:130000,17.5102596402798)
--(axis cs:140000,17.1102596402798)
--(axis cs:150000,17.7102596402798)
--(axis cs:160000,17.9102596402798)
--(axis cs:170000,17.8102596402798)
--(axis cs:180000,17.8102596402798)
--(axis cs:190000,17.4102596402798)
--(axis cs:200000,18.0102596402798)
--(axis cs:210000,18.2102596402798)
--(axis cs:220000,17.5102596402798)
--(axis cs:230000,17.5102596402798)
--(axis cs:240000,17.8102596402798)
--(axis cs:250000,18.1102596402798)
--(axis cs:260000,17.8102596402798)
--(axis cs:270000,18.0102596402798)
--(axis cs:280000,17.9102596402798)
--(axis cs:290000,17.7102596402798)
--(axis cs:300000,17.8102596402798)
--(axis cs:310000,17.9102596402798)
--(axis cs:320000,17.6102596402798)
--(axis cs:330000,17.9102596402798)
--(axis cs:340000,17.9102596402798)
--(axis cs:350000,17.8102596402798)
--(axis cs:360000,17.9102596402798)
--(axis cs:370000,17.9102596402798)
--(axis cs:380000,17.8102596402798)
--(axis cs:390000,18.1102596402798)
--(axis cs:390000,22.6897403597202)
--(axis cs:390000,22.6897403597202)
--(axis cs:380000,22.3897403597202)
--(axis cs:370000,22.4897403597202)
--(axis cs:360000,22.4897403597202)
--(axis cs:350000,22.3897403597202)
--(axis cs:340000,22.4897403597202)
--(axis cs:330000,22.4897403597202)
--(axis cs:320000,22.1897403597202)
--(axis cs:310000,22.4897403597202)
--(axis cs:300000,22.3897403597202)
--(axis cs:290000,22.2897403597202)
--(axis cs:280000,22.4897403597202)
--(axis cs:270000,22.5897403597202)
--(axis cs:260000,22.3897403597202)
--(axis cs:250000,22.6897403597202)
--(axis cs:240000,22.3897403597202)
--(axis cs:230000,22.0897403597202)
--(axis cs:220000,22.0897403597202)
--(axis cs:210000,22.7897403597202)
--(axis cs:200000,22.5897403597202)
--(axis cs:190000,21.9897403597202)
--(axis cs:180000,22.3897403597202)
--(axis cs:170000,22.3897403597202)
--(axis cs:160000,22.4897403597202)
--(axis cs:150000,22.2897403597202)
--(axis cs:140000,21.6897403597202)
--(axis cs:130000,22.0897403597202)
--(axis cs:120000,21.8897403597202)
--(axis cs:110000,21.7897403597202)
--(axis cs:100000,21.9897403597202)
--(axis cs:90000,21.7897403597202)
--(axis cs:80000,22.0897403597202)
--(axis cs:70000,21.3897403597202)
--(axis cs:60000,21.1897403597202)
--(axis cs:50000,21.2897403597202)
--(axis cs:40000,19.3897403597202)
--(axis cs:30000,15.4897403597202)
--(axis cs:20000,15.1897403597202)
--(axis cs:10000,8.7897403597202)
--(axis cs:0,-5.1102596402798)
--(axis cs:-10000,-18.6102596402798)
--cycle;

\path [draw=peru20214597, fill=peru20214597, opacity=0.2]
(axis cs:-10000,-17.8983185870929)
--(axis cs:-10000,-23.7016814129071)
--(axis cs:0,-14.2016814129071)
--(axis cs:10000,-13.4016814129071)
--(axis cs:20000,2.39831858709287)
--(axis cs:30000,9.89831858709287)
--(axis cs:40000,11.9983185870929)
--(axis cs:50000,15.4983185870929)
--(axis cs:60000,15.7983185870929)
--(axis cs:70000,16.7983185870929)
--(axis cs:80000,16.1983185870929)
--(axis cs:90000,17.1983185870929)
--(axis cs:100000,17.0983185870929)
--(axis cs:110000,16.5983185870929)
--(axis cs:120000,16.7983185870929)
--(axis cs:130000,16.9983185870929)
--(axis cs:140000,17.5983185870929)
--(axis cs:150000,16.6983185870929)
--(axis cs:160000,17.9983185870929)
--(axis cs:170000,16.5983185870929)
--(axis cs:180000,16.2983185870929)
--(axis cs:190000,16.0983185870929)
--(axis cs:200000,15.6983185870929)
--(axis cs:210000,17.3983185870929)
--(axis cs:220000,17.3983185870929)
--(axis cs:230000,17.4983185870929)
--(axis cs:240000,17.3983185870929)
--(axis cs:250000,17.3983185870929)
--(axis cs:260000,17.2983185870929)
--(axis cs:270000,13.1983185870929)
--(axis cs:280000,12.8983185870929)
--(axis cs:290000,15.0983185870929)
--(axis cs:300000,13.4983185870929)
--(axis cs:310000,14.7983185870929)
--(axis cs:320000,15.4983185870929)
--(axis cs:330000,13.0983185870929)
--(axis cs:340000,15.6983185870929)
--(axis cs:350000,15.0983185870929)
--(axis cs:360000,15.3983185870929)
--(axis cs:370000,14.6983185870929)
--(axis cs:380000,15.3983185870929)
--(axis cs:390000,16.9983185870929)
--(axis cs:390000,22.8016814129071)
--(axis cs:390000,22.8016814129071)
--(axis cs:380000,21.2016814129071)
--(axis cs:370000,20.5016814129071)
--(axis cs:360000,21.2016814129071)
--(axis cs:350000,20.9016814129071)
--(axis cs:340000,21.5016814129071)
--(axis cs:330000,18.9016814129071)
--(axis cs:320000,21.3016814129071)
--(axis cs:310000,20.6016814129071)
--(axis cs:300000,19.3016814129071)
--(axis cs:290000,20.9016814129071)
--(axis cs:280000,18.7016814129071)
--(axis cs:270000,19.0016814129071)
--(axis cs:260000,23.1016814129071)
--(axis cs:250000,23.2016814129071)
--(axis cs:240000,23.2016814129071)
--(axis cs:230000,23.3016814129071)
--(axis cs:220000,23.2016814129071)
--(axis cs:210000,23.2016814129071)
--(axis cs:200000,21.5016814129071)
--(axis cs:190000,21.9016814129071)
--(axis cs:180000,22.1016814129071)
--(axis cs:170000,22.4016814129071)
--(axis cs:160000,23.8016814129071)
--(axis cs:150000,22.5016814129071)
--(axis cs:140000,23.4016814129071)
--(axis cs:130000,22.8016814129071)
--(axis cs:120000,22.6016814129071)
--(axis cs:110000,22.4016814129071)
--(axis cs:100000,22.9016814129071)
--(axis cs:90000,23.0016814129071)
--(axis cs:80000,22.0016814129071)
--(axis cs:70000,22.6016814129071)
--(axis cs:60000,21.6016814129071)
--(axis cs:50000,21.3016814129071)
--(axis cs:40000,17.8016814129071)
--(axis cs:30000,15.7016814129071)
--(axis cs:20000,8.20168141290713)
--(axis cs:10000,-7.59831858709287)
--(axis cs:0,-8.39831858709287)
--(axis cs:-10000,-17.8983185870929)
--cycle;

\addplot [semithick, darkcyan1115178, mark=square*, mark size=1.5, mark options={solid}]
table {%
-10000 -20.8
0 -2.6
10000 10.1
20000 12.1
30000 14.5
40000 19.4
50000 20.5
60000 20.5
70000 20.4
80000 20.6
90000 20.7
100000 20.7
110000 20.7
120000 20.6
130000 20.6
140000 20.8
150000 20.8
160000 20.8
170000 21
180000 20.9
190000 21
200000 20.9
210000 21
220000 21
230000 20.9
240000 21
250000 21
260000 20.9
270000 21
280000 21
290000 21
300000 20.9
310000 21
320000 21
330000 21
340000 21
350000 20.9
360000 21
370000 21
380000 21
390000 20.9
};
\addplot [semithick, darkorange2221435, mark=triangle*, mark size=1.5, mark options={solid}]
table {%
-10000 -20.6
0 -3.4
10000 2.5
20000 9.1
30000 12.1
40000 13.7
50000 15.8
60000 15.1
70000 15.5
80000 16
90000 16.5
100000 16.2
110000 15.7
120000 18
130000 18
140000 18
150000 18.4
160000 19
170000 18.8
180000 19.7
190000 19.4
200000 19.5
210000 19.3
220000 19.3
230000 19.4
240000 19.5
250000 19.7
260000 20.2
270000 20
280000 20.1
290000 20
300000 19.6
310000 20.3
320000 20.3
330000 20.4
340000 20.4
350000 20.5
360000 20.5
370000 20.5
380000 19.9
390000 20.1
};
\addplot [semithick, orchid204120188, mark=+, mark size=1.5, mark options={solid}]
table {%
-10000 -20.9
0 -7.4
10000 6.5
20000 12.9
30000 13.2
40000 17.1
50000 19
60000 18.9
70000 19.1
80000 19.8
90000 19.5
100000 19.7
110000 19.5
120000 19.6
130000 19.8
140000 19.4
150000 20
160000 20.2
170000 20.1
180000 20.1
190000 19.7
200000 20.3
210000 20.5
220000 19.8
230000 19.8
240000 20.1
250000 20.4
260000 20.1
270000 20.3
280000 20.2
290000 20
300000 20.1
310000 20.2
320000 19.9
330000 20.2
340000 20.2
350000 20.1
360000 20.2
370000 20.2
380000 20.1
390000 20.4
};
\addplot [semithick, peru20214597, mark=diamond*, mark size=1.5, mark options={solid}]
table {%
-10000 -20.8
0 -11.3
10000 -10.5
20000 5.3
30000 12.8
40000 14.9
50000 18.4
60000 18.7
70000 19.7
80000 19.1
90000 20.1
100000 20
110000 19.5
120000 19.7
130000 19.9
140000 20.5
150000 19.6
160000 20.9
170000 19.5
180000 19.2
190000 19
200000 18.6
210000 20.3
220000 20.3
230000 20.4
240000 20.3
250000 20.3
260000 20.2
270000 16.1
280000 15.8
290000 18
300000 16.4
310000 17.7
320000 18.4
330000 16
340000 18.6
350000 18
360000 18.3
370000 17.6
380000 18.3
390000 19.9
};
\end{axis}

\end{tikzpicture}

%% file: Image/qbert.tex
\begin{tikzpicture}

\definecolor{darkcyan1115178}{RGB}{1,115,178}
\definecolor{darkorange2221435}{RGB}{222,143,5}
\definecolor{darkslategray38}{RGB}{38,38,38}
\definecolor{lavender234234242}{RGB}{234,234,242}
\definecolor{lightgray204}{RGB}{204,204,204}
\definecolor{orchid204120188}{RGB}{204,120,188}
\definecolor{peru20214597}{RGB}{202,145,97}

\begin{axis}[
axis background/.style={fill=lavender234234242},
axis line style={white},
legend cell align={left},
legend style={
  fill opacity=0.8,
  draw opacity=1,
  text opacity=1,
  at={(0.03,0.97)},
  anchor=north west,
  draw=lightgray204,
  fill=lavender234234242
},
tick align=outside,
x grid style={white},
xlabel=\textcolor{darkslategray38}{Time Step},
xmajorgrids,
xmajorticks=true,
xmin=-30000, xmax=410000,
xtick style={color=darkslategray38},
xtick={0,50000,100000,150000,200000,250000,300000,350000,400000},
xticklabels={0k,50k,100k,150k,200k,250k,300k,350k,400k},
y grid style={white},
ylabel=\textcolor{darkslategray38}{Performance},
ymajorgrids,
ymajorticks=true,
ymin=-968.948593879945, ymax=7993.94859387994,
ytick style={color=darkslategray38}
]
\path [draw=darkcyan1115178, fill=darkcyan1115178, opacity=0.2]
(axis cs:-10000,860.144176254495)
--(axis cs:-10000,-561.544176254495)
--(axis cs:0,-202.944176254495)
--(axis cs:10000,-125.444176254495)
--(axis cs:20000,45.2558237455049)
--(axis cs:30000,56.2558237455049)
--(axis cs:40000,760.555823745505)
--(axis cs:50000,624.855823745505)
--(axis cs:60000,1438.45582374551)
--(axis cs:70000,1607.7558237455)
--(axis cs:80000,2250.25582374551)
--(axis cs:90000,1869.1558237455)
--(axis cs:100000,1646.6558237455)
--(axis cs:110000,2215.9558237455)
--(axis cs:120000,1878.45582374551)
--(axis cs:130000,2059.8558237455)
--(axis cs:140000,2969.8558237455)
--(axis cs:150000,3079.1558237455)
--(axis cs:160000,3344.8558237455)
--(axis cs:170000,2898.75582374551)
--(axis cs:180000,3103.05582374551)
--(axis cs:190000,3267.3558237455)
--(axis cs:200000,3090.9558237455)
--(axis cs:210000,3620.9558237455)
--(axis cs:220000,3909.5558237455)
--(axis cs:230000,4215.5558237455)
--(axis cs:240000,4288.0558237455)
--(axis cs:250000,4254.1558237455)
--(axis cs:260000,4559.5558237455)
--(axis cs:270000,4538.75582374551)
--(axis cs:280000,4607.3558237455)
--(axis cs:290000,4300.5558237455)
--(axis cs:300000,5174.1558237455)
--(axis cs:310000,5164.8558237455)
--(axis cs:320000,5327.75582374551)
--(axis cs:330000,6164.8558237455)
--(axis cs:340000,6014.8558237455)
--(axis cs:350000,5173.75582374551)
--(axis cs:360000,5343.4558237455)
--(axis cs:370000,5425.25582374551)
--(axis cs:380000,5814.8558237455)
--(axis cs:390000,6133.75582374551)
--(axis cs:390000,7555.4441762545)
--(axis cs:390000,7555.4441762545)
--(axis cs:380000,7236.5441762545)
--(axis cs:370000,6846.9441762545)
--(axis cs:360000,6765.1441762545)
--(axis cs:350000,6595.4441762545)
--(axis cs:340000,7436.5441762545)
--(axis cs:330000,7586.5441762545)
--(axis cs:320000,6749.4441762545)
--(axis cs:310000,6586.5441762545)
--(axis cs:300000,6595.8441762545)
--(axis cs:290000,5722.24417625449)
--(axis cs:280000,6029.0441762545)
--(axis cs:270000,5960.4441762545)
--(axis cs:260000,5981.24417625449)
--(axis cs:250000,5675.8441762545)
--(axis cs:240000,5709.74417625449)
--(axis cs:230000,5637.24417625449)
--(axis cs:220000,5331.24417625449)
--(axis cs:210000,5042.6441762545)
--(axis cs:200000,4512.6441762545)
--(axis cs:190000,4689.0441762545)
--(axis cs:180000,4524.74417625449)
--(axis cs:170000,4320.44417625449)
--(axis cs:160000,4766.5441762545)
--(axis cs:150000,4500.8441762545)
--(axis cs:140000,4391.5441762545)
--(axis cs:130000,3481.5441762545)
--(axis cs:120000,3300.1441762545)
--(axis cs:110000,3637.6441762545)
--(axis cs:100000,3068.3441762545)
--(axis cs:90000,3290.8441762545)
--(axis cs:80000,3671.94417625449)
--(axis cs:70000,3029.44417625449)
--(axis cs:60000,2860.1441762545)
--(axis cs:50000,2046.5441762545)
--(axis cs:40000,2182.24417625449)
--(axis cs:30000,1477.9441762545)
--(axis cs:20000,1466.9441762545)
--(axis cs:10000,1296.24417625449)
--(axis cs:0,1218.74417625449)
--(axis cs:-10000,860.144176254495)
--cycle;

\path [draw=darkorange2221435, fill=darkorange2221435, opacity=0.2]
(axis cs:-10000,459.225106388108)
--(axis cs:-10000,-238.825106388108)
--(axis cs:0,7.77489361189186)
--(axis cs:10000,240.474893611892)
--(axis cs:20000,355.174893611892)
--(axis cs:30000,433.974893611892)
--(axis cs:40000,435.974893611892)
--(axis cs:50000,405.174893611892)
--(axis cs:60000,400.474893611892)
--(axis cs:70000,809.774893611892)
--(axis cs:80000,1322.17489361189)
--(axis cs:90000,1497.47489361189)
--(axis cs:100000,1172.17489361189)
--(axis cs:110000,1895.77489361189)
--(axis cs:120000,2006.97489361189)
--(axis cs:130000,2558.77489361189)
--(axis cs:140000,1914.97489361189)
--(axis cs:150000,2081.47489361189)
--(axis cs:160000,2357.47489361189)
--(axis cs:170000,2915.97489361189)
--(axis cs:180000,2309.77489361189)
--(axis cs:190000,2855.97489361189)
--(axis cs:200000,2556.17489361189)
--(axis cs:210000,2992.17489361189)
--(axis cs:220000,2704.77489361189)
--(axis cs:230000,2753.77489361189)
--(axis cs:240000,2836.77489361189)
--(axis cs:250000,3131.77489361189)
--(axis cs:260000,3188.77489361189)
--(axis cs:270000,3308.97489361189)
--(axis cs:280000,3220.47489361189)
--(axis cs:290000,3336.17489361189)
--(axis cs:300000,3343.47489361189)
--(axis cs:310000,3345.97489361189)
--(axis cs:320000,3460.17489361189)
--(axis cs:330000,3477.47489361189)
--(axis cs:340000,3444.47489361189)
--(axis cs:350000,3407.97489361189)
--(axis cs:360000,3623.47489361189)
--(axis cs:370000,3481.77489361189)
--(axis cs:380000,3413.17489361189)
--(axis cs:390000,3431.17489361189)
--(axis cs:390000,4129.22510638811)
--(axis cs:390000,4129.22510638811)
--(axis cs:380000,4111.22510638811)
--(axis cs:370000,4179.82510638811)
--(axis cs:360000,4321.52510638811)
--(axis cs:350000,4106.02510638811)
--(axis cs:340000,4142.52510638811)
--(axis cs:330000,4175.52510638811)
--(axis cs:320000,4158.22510638811)
--(axis cs:310000,4044.02510638811)
--(axis cs:300000,4041.52510638811)
--(axis cs:290000,4034.22510638811)
--(axis cs:280000,3918.52510638811)
--(axis cs:270000,4007.02510638811)
--(axis cs:260000,3886.82510638811)
--(axis cs:250000,3829.82510638811)
--(axis cs:240000,3534.82510638811)
--(axis cs:230000,3451.82510638811)
--(axis cs:220000,3402.82510638811)
--(axis cs:210000,3690.22510638811)
--(axis cs:200000,3254.22510638811)
--(axis cs:190000,3554.02510638811)
--(axis cs:180000,3007.82510638811)
--(axis cs:170000,3614.02510638811)
--(axis cs:160000,3055.52510638811)
--(axis cs:150000,2779.52510638811)
--(axis cs:140000,2613.02510638811)
--(axis cs:130000,3256.82510638811)
--(axis cs:120000,2705.02510638811)
--(axis cs:110000,2593.82510638811)
--(axis cs:100000,1870.22510638811)
--(axis cs:90000,2195.52510638811)
--(axis cs:80000,2020.22510638811)
--(axis cs:70000,1507.82510638811)
--(axis cs:60000,1098.52510638811)
--(axis cs:50000,1103.22510638811)
--(axis cs:40000,1134.02510638811)
--(axis cs:30000,1132.02510638811)
--(axis cs:20000,1053.22510638811)
--(axis cs:10000,938.525106388108)
--(axis cs:0,705.825106388108)
--(axis cs:-10000,459.225106388108)
--cycle;

\path [draw=orchid204120188, fill=orchid204120188, opacity=0.2]
(axis cs:-10000,682.355934288066)
--(axis cs:-10000,-378.155934288066)
--(axis cs:0,-89.155934288066)
--(axis cs:10000,50.844065711934)
--(axis cs:20000,250.144065711934)
--(axis cs:30000,348.644065711934)
--(axis cs:40000,947.644065711934)
--(axis cs:50000,1208.64406571193)
--(axis cs:60000,1994.74406571193)
--(axis cs:70000,1536.84406571193)
--(axis cs:80000,2337.94406571193)
--(axis cs:90000,1971.84406571193)
--(axis cs:100000,2594.74406571193)
--(axis cs:110000,3040.44406571193)
--(axis cs:120000,3300.14406571193)
--(axis cs:130000,3746.84406571193)
--(axis cs:140000,3534.04406571193)
--(axis cs:150000,3767.94406571193)
--(axis cs:160000,3787.24406571193)
--(axis cs:170000,3992.64406571193)
--(axis cs:180000,3920.44406571193)
--(axis cs:190000,3967.94406571193)
--(axis cs:200000,4041.54406571193)
--(axis cs:210000,4012.94406571193)
--(axis cs:220000,4132.94406571193)
--(axis cs:230000,4059.34406571193)
--(axis cs:240000,4062.24406571193)
--(axis cs:250000,4092.94406571193)
--(axis cs:260000,4049.04406571193)
--(axis cs:270000,4088.34406571193)
--(axis cs:280000,4599.74406571193)
--(axis cs:290000,4263.64406571193)
--(axis cs:300000,4436.84406571193)
--(axis cs:310000,4242.94406571193)
--(axis cs:320000,4245.84406571193)
--(axis cs:330000,4477.24406571193)
--(axis cs:340000,4624.04406571193)
--(axis cs:350000,5506.14406571193)
--(axis cs:360000,5335.84406571193)
--(axis cs:370000,6065.44406571193)
--(axis cs:380000,6236.14406571193)
--(axis cs:390000,5738.64406571193)
--(axis cs:390000,6799.15593428807)
--(axis cs:390000,6799.15593428807)
--(axis cs:380000,7296.65593428807)
--(axis cs:370000,7125.95593428807)
--(axis cs:360000,6396.35593428807)
--(axis cs:350000,6566.65593428807)
--(axis cs:340000,5684.55593428807)
--(axis cs:330000,5537.75593428807)
--(axis cs:320000,5306.35593428807)
--(axis cs:310000,5303.45593428807)
--(axis cs:300000,5497.35593428807)
--(axis cs:290000,5324.15593428807)
--(axis cs:280000,5660.25593428807)
--(axis cs:270000,5148.85593428807)
--(axis cs:260000,5109.55593428807)
--(axis cs:250000,5153.45593428807)
--(axis cs:240000,5122.75593428807)
--(axis cs:230000,5119.85593428807)
--(axis cs:220000,5193.45593428807)
--(axis cs:210000,5073.45593428807)
--(axis cs:200000,5102.05593428807)
--(axis cs:190000,5028.45593428807)
--(axis cs:180000,4980.95593428807)
--(axis cs:170000,5053.15593428807)
--(axis cs:160000,4847.75593428807)
--(axis cs:150000,4828.45593428807)
--(axis cs:140000,4594.55593428807)
--(axis cs:130000,4807.35593428807)
--(axis cs:120000,4360.65593428807)
--(axis cs:110000,4100.95593428807)
--(axis cs:100000,3655.25593428807)
--(axis cs:90000,3032.35593428807)
--(axis cs:80000,3398.45593428807)
--(axis cs:70000,2597.35593428807)
--(axis cs:60000,3055.25593428807)
--(axis cs:50000,2269.15593428807)
--(axis cs:40000,2008.15593428807)
--(axis cs:30000,1409.15593428807)
--(axis cs:20000,1310.65593428807)
--(axis cs:10000,1111.35593428807)
--(axis cs:0,971.355934288066)
--(axis cs:-10000,682.355934288066)
--cycle;

\path [draw=peru20214597, fill=peru20214597, opacity=0.2]
(axis cs:-10000,649.852368427985)
--(axis cs:-10000,-334.052368427985)
--(axis cs:0,-243.752368427985)
--(axis cs:10000,-38.3523684279846)
--(axis cs:20000,-6.25236842798466)
--(axis cs:30000,126.947631572015)
--(axis cs:40000,161.647631572015)
--(axis cs:50000,603.747631572015)
--(axis cs:60000,457.647631572015)
--(axis cs:70000,814.847631572015)
--(axis cs:80000,1320.54763157202)
--(axis cs:90000,638.047631572015)
--(axis cs:100000,975.147631572015)
--(axis cs:110000,818.447631572015)
--(axis cs:120000,574.147631572015)
--(axis cs:130000,948.447631572015)
--(axis cs:140000,1337.64763157202)
--(axis cs:150000,2080.54763157202)
--(axis cs:160000,2175.54763157202)
--(axis cs:170000,2500.94763157202)
--(axis cs:180000,2926.64763157202)
--(axis cs:190000,2744.84763157202)
--(axis cs:200000,2358.44763157202)
--(axis cs:210000,2544.84763157202)
--(axis cs:220000,3172.34763157202)
--(axis cs:230000,2473.44763157202)
--(axis cs:240000,2863.74763157202)
--(axis cs:250000,2683.44763157202)
--(axis cs:260000,2892.34763157202)
--(axis cs:270000,3028.74763157202)
--(axis cs:280000,2995.54763157202)
--(axis cs:290000,3221.94763157202)
--(axis cs:300000,3423.04763157202)
--(axis cs:310000,3580.94763157202)
--(axis cs:320000,3517.64763157202)
--(axis cs:330000,3603.44763157202)
--(axis cs:340000,3937.64763157202)
--(axis cs:350000,3851.94763157201)
--(axis cs:360000,3942.34763157202)
--(axis cs:370000,4005.14763157202)
--(axis cs:380000,4031.24763157202)
--(axis cs:390000,4118.04763157202)
--(axis cs:390000,5101.95236842798)
--(axis cs:390000,5101.95236842798)
--(axis cs:380000,5015.15236842798)
--(axis cs:370000,4989.05236842799)
--(axis cs:360000,4926.25236842798)
--(axis cs:350000,4835.85236842798)
--(axis cs:340000,4921.55236842799)
--(axis cs:330000,4587.35236842798)
--(axis cs:320000,4501.55236842798)
--(axis cs:310000,4564.85236842798)
--(axis cs:300000,4406.95236842798)
--(axis cs:290000,4205.85236842798)
--(axis cs:280000,3979.45236842798)
--(axis cs:270000,4012.65236842798)
--(axis cs:260000,3876.25236842798)
--(axis cs:250000,3667.35236842798)
--(axis cs:240000,3847.65236842798)
--(axis cs:230000,3457.35236842798)
--(axis cs:220000,4156.25236842798)
--(axis cs:210000,3528.75236842798)
--(axis cs:200000,3342.35236842798)
--(axis cs:190000,3728.75236842798)
--(axis cs:180000,3910.55236842798)
--(axis cs:170000,3484.85236842798)
--(axis cs:160000,3159.45236842798)
--(axis cs:150000,3064.45236842798)
--(axis cs:140000,2321.55236842798)
--(axis cs:130000,1932.35236842798)
--(axis cs:120000,1558.05236842798)
--(axis cs:110000,1802.35236842798)
--(axis cs:100000,1959.05236842798)
--(axis cs:90000,1621.95236842798)
--(axis cs:80000,2304.45236842798)
--(axis cs:70000,1798.75236842798)
--(axis cs:60000,1441.55236842798)
--(axis cs:50000,1587.65236842798)
--(axis cs:40000,1145.55236842798)
--(axis cs:30000,1110.85236842798)
--(axis cs:20000,977.652368427985)
--(axis cs:10000,945.552368427985)
--(axis cs:0,740.152368427985)
--(axis cs:-10000,649.852368427985)
--cycle;

\addplot [semithick, darkcyan1115178, mark=square*, mark size=1.5, mark options={solid}]
table {%
-10000 149.3
0 507.9
10000 585.4
20000 756.1
30000 767.1
40000 1471.4
50000 1335.7
60000 2149.3
70000 2318.6
80000 2961.1
90000 2580
100000 2357.5
110000 2926.8
120000 2589.3
130000 2770.7
140000 3680.7
150000 3790
160000 4055.7
170000 3609.6
180000 3813.9
190000 3978.2
200000 3801.8
210000 4331.8
220000 4620.4
230000 4926.4
240000 4998.9
250000 4965
260000 5270.4
270000 5249.6
280000 5318.2
290000 5011.4
300000 5885
310000 5875.7
320000 6038.6
330000 6875.7
340000 6725.7
350000 5884.6
360000 6054.3
370000 6136.1
380000 6525.7
390000 6844.6
};
\addplot [semithick, darkorange2221435, mark=triangle*, mark size=1.5, mark options={solid}]
table {%
-10000 110.2
0 356.8
10000 589.5
20000 704.2
30000 783
40000 785
50000 754.2
60000 749.5
70000 1158.8
80000 1671.2
90000 1846.5
100000 1521.2
110000 2244.8
120000 2356
130000 2907.8
140000 2264
150000 2430.5
160000 2706.5
170000 3265
180000 2658.8
190000 3205
200000 2905.2
210000 3341.2
220000 3053.8
230000 3102.8
240000 3185.8
250000 3480.8
260000 3537.8
270000 3658
280000 3569.5
290000 3685.2
300000 3692.5
310000 3695
320000 3809.2
330000 3826.5
340000 3793.5
350000 3757
360000 3972.5
370000 3830.8
380000 3762.2
390000 3780.2
};
\addplot [semithick, orchid204120188, mark=+, mark size=1.5, mark options={solid}]
table {%
-10000 152.1
0 441.1
10000 581.1
20000 780.4
30000 878.9
40000 1477.9
50000 1738.9
60000 2525
70000 2067.1
80000 2868.2
90000 2502.1
100000 3125
110000 3570.7
120000 3830.4
130000 4277.1
140000 4064.3
150000 4298.2
160000 4317.5
170000 4522.9
180000 4450.7
190000 4498.2
200000 4571.8
210000 4543.2
220000 4663.2
230000 4589.6
240000 4592.5
250000 4623.2
260000 4579.3
270000 4618.6
280000 5130
290000 4793.9
300000 4967.1
310000 4773.2
320000 4776.1
330000 5007.5
340000 5154.3
350000 6036.4
360000 5866.1
370000 6595.7
380000 6766.4
390000 6268.9
};
\addplot [semithick, peru20214597, mark=diamond*, mark size=1.5, mark options={solid}]
table {%
-10000 157.9
0 248.2
10000 453.6
20000 485.7
30000 618.9
40000 653.6
50000 1095.7
60000 949.6
70000 1306.8
80000 1812.5
90000 1130
100000 1467.1
110000 1310.4
120000 1066.1
130000 1440.4
140000 1829.6
150000 2572.5
160000 2667.5
170000 2992.9
180000 3418.6
190000 3236.8
200000 2850.4
210000 3036.8
220000 3664.3
230000 2965.4
240000 3355.7
250000 3175.4
260000 3384.3
270000 3520.7
280000 3487.5
290000 3713.9
300000 3915
310000 4072.9
320000 4009.6
330000 4095.4
340000 4429.6
350000 4343.9
360000 4434.3
370000 4497.1
380000 4523.2
390000 4610
};
\end{axis}

\end{tikzpicture}

%% file: Image/road_runner.tex
\begin{tikzpicture}

\definecolor{darkcyan1115178}{RGB}{1,115,178}
\definecolor{darkorange2221435}{RGB}{222,143,5}
\definecolor{darkslategray38}{RGB}{38,38,38}
\definecolor{lavender234234242}{RGB}{234,234,242}
\definecolor{lightgray204}{RGB}{204,204,204}
\definecolor{orchid204120188}{RGB}{204,120,188}
\definecolor{peru20214597}{RGB}{202,145,97}

\begin{axis}[
axis background/.style={fill=lavender234234242},
axis line style={white},
legend cell align={left},
legend style={
  fill opacity=0.8,
  draw opacity=1,
  text opacity=1,
  at={(0.03,0.97)},
  anchor=north west,
  draw=lightgray204,
  fill=lavender234234242
},
tick align=outside,
x grid style={white},
xlabel=\textcolor{darkslategray38}{Time Step},
xmajorgrids,
xmajorticks=true,
xmin=-30000, xmax=410000,
xtick style={color=darkslategray38},
xtick={0,50000,100000,150000,200000,250000,300000,350000,400000},
xticklabels={0k,50k,100k,150k,200k,250k,300k,350k,400k},
y grid style={white},
ylabel=\textcolor{darkslategray38}{Performance},
ymajorgrids,
ymajorticks=true,
ymin=-16017.2904084005, ymax=77338.6904084005,
ytick style={color=darkslategray38}
]
\path [draw=darkcyan1115178, fill=darkcyan1115178, opacity=0.2]
(axis cs:-10000,11836.6367349095)
--(axis cs:-10000,-11773.8367349095)
--(axis cs:0,-11196.6367349095)
--(axis cs:10000,-10866.6367349095)
--(axis cs:20000,-10125.2367349095)
--(axis cs:30000,-8289.53673490954)
--(axis cs:40000,-7343.83673490954)
--(axis cs:50000,-6616.63673490954)
--(axis cs:60000,-4006.63673490954)
--(axis cs:70000,-2425.23673490954)
--(axis cs:80000,-1469.53673490954)
--(axis cs:90000,-396.636734909538)
--(axis cs:100000,-303.836734909539)
--(axis cs:110000,456.163265090461)
--(axis cs:120000,-663.836734909539)
--(axis cs:130000,91.8632650904619)
--(axis cs:140000,270.463265090462)
--(axis cs:150000,-1280.93673490954)
--(axis cs:160000,64.7632650904616)
--(axis cs:170000,537.663265090461)
--(axis cs:180000,-433.836734909539)
--(axis cs:190000,1711.86326509046)
--(axis cs:200000,1613.36326509046)
--(axis cs:210000,2543.36326509046)
--(axis cs:220000,2877.66326509046)
--(axis cs:230000,4190.46326509046)
--(axis cs:240000,5981.86326509046)
--(axis cs:250000,3189.06326509046)
--(axis cs:260000,4431.86326509046)
--(axis cs:270000,6493.36326509046)
--(axis cs:280000,14534.7632650905)
--(axis cs:290000,19651.8632650905)
--(axis cs:300000,9171.86326509046)
--(axis cs:310000,14680.4632650905)
--(axis cs:320000,33956.1632650905)
--(axis cs:330000,22664.7632650905)
--(axis cs:340000,47587.6632650905)
--(axis cs:350000,30491.8632650905)
--(axis cs:360000,49484.7632650905)
--(axis cs:370000,11776.1632650905)
--(axis cs:380000,38657.6632650905)
--(axis cs:390000,9683.36326509046)
--(axis cs:390000,33293.8367349095)
--(axis cs:390000,33293.8367349095)
--(axis cs:380000,62268.1367349095)
--(axis cs:370000,35386.6367349095)
--(axis cs:360000,73095.2367349095)
--(axis cs:350000,54102.3367349095)
--(axis cs:340000,71198.1367349095)
--(axis cs:330000,46275.2367349095)
--(axis cs:320000,57566.6367349095)
--(axis cs:310000,38290.9367349095)
--(axis cs:300000,32782.3367349095)
--(axis cs:290000,43262.3367349095)
--(axis cs:280000,38145.2367349095)
--(axis cs:270000,30103.8367349095)
--(axis cs:260000,28042.3367349095)
--(axis cs:250000,26799.5367349095)
--(axis cs:240000,29592.3367349095)
--(axis cs:230000,27800.9367349095)
--(axis cs:220000,26488.1367349095)
--(axis cs:210000,26153.8367349095)
--(axis cs:200000,25223.8367349095)
--(axis cs:190000,25322.3367349095)
--(axis cs:180000,23176.6367349095)
--(axis cs:170000,24148.1367349095)
--(axis cs:160000,23675.2367349095)
--(axis cs:150000,22329.5367349095)
--(axis cs:140000,23880.9367349095)
--(axis cs:130000,23702.3367349095)
--(axis cs:120000,22946.6367349095)
--(axis cs:110000,24066.6367349095)
--(axis cs:100000,23306.6367349095)
--(axis cs:90000,23213.8367349095)
--(axis cs:80000,22140.9367349095)
--(axis cs:70000,21185.2367349095)
--(axis cs:60000,19603.8367349095)
--(axis cs:50000,16993.8367349095)
--(axis cs:40000,16266.6367349095)
--(axis cs:30000,15320.9367349095)
--(axis cs:20000,13485.2367349095)
--(axis cs:10000,12743.8367349095)
--(axis cs:0,12413.8367349095)
--(axis cs:-10000,11836.6367349095)
--cycle;

\path [draw=darkorange2221435, fill=darkorange2221435, opacity=0.2]
(axis cs:-10000,1182.30998675555)
--(axis cs:-10000,-934.30998675555)
--(axis cs:0,1435.69001324445)
--(axis cs:10000,1874.69001324445)
--(axis cs:20000,1653.69001324445)
--(axis cs:30000,2478.69001324445)
--(axis cs:40000,2564.69001324445)
--(axis cs:50000,3102.69001324445)
--(axis cs:60000,3320.69001324445)
--(axis cs:70000,4491.69001324445)
--(axis cs:80000,4526.69001324445)
--(axis cs:90000,5128.69001324445)
--(axis cs:100000,5214.69001324445)
--(axis cs:110000,5534.69001324445)
--(axis cs:120000,5471.69001324445)
--(axis cs:130000,5882.69001324445)
--(axis cs:140000,5452.69001324445)
--(axis cs:150000,5755.69001324445)
--(axis cs:160000,5919.69001324445)
--(axis cs:170000,5393.69001324445)
--(axis cs:180000,5530.69001324445)
--(axis cs:190000,5608.69001324445)
--(axis cs:200000,5992.69001324445)
--(axis cs:210000,5546.69001324445)
--(axis cs:220000,5695.69001324445)
--(axis cs:230000,5961.69001324445)
--(axis cs:240000,6320.69001324445)
--(axis cs:250000,6109.69001324445)
--(axis cs:260000,6641.69001324445)
--(axis cs:270000,6567.69001324445)
--(axis cs:280000,6581.69001324445)
--(axis cs:290000,6304.69001324445)
--(axis cs:300000,6038.69001324445)
--(axis cs:310000,6151.69001324445)
--(axis cs:320000,6517.69001324445)
--(axis cs:330000,6113.69001324445)
--(axis cs:340000,6671.69001324445)
--(axis cs:350000,6491.69001324445)
--(axis cs:360000,6844.69001324445)
--(axis cs:370000,6697.69001324445)
--(axis cs:380000,6671.69001324445)
--(axis cs:390000,6768.69001324445)
--(axis cs:390000,8885.30998675555)
--(axis cs:390000,8885.30998675555)
--(axis cs:380000,8788.30998675555)
--(axis cs:370000,8814.30998675555)
--(axis cs:360000,8961.30998675555)
--(axis cs:350000,8608.30998675555)
--(axis cs:340000,8788.30998675555)
--(axis cs:330000,8230.30998675555)
--(axis cs:320000,8634.30998675555)
--(axis cs:310000,8268.30998675555)
--(axis cs:300000,8155.30998675555)
--(axis cs:290000,8421.30998675555)
--(axis cs:280000,8698.30998675555)
--(axis cs:270000,8684.30998675555)
--(axis cs:260000,8758.30998675555)
--(axis cs:250000,8226.30998675555)
--(axis cs:240000,8437.30998675555)
--(axis cs:230000,8078.30998675555)
--(axis cs:220000,7812.30998675555)
--(axis cs:210000,7663.30998675555)
--(axis cs:200000,8109.30998675555)
--(axis cs:190000,7725.30998675555)
--(axis cs:180000,7647.30998675555)
--(axis cs:170000,7510.30998675555)
--(axis cs:160000,8036.30998675555)
--(axis cs:150000,7872.30998675555)
--(axis cs:140000,7569.30998675555)
--(axis cs:130000,7999.30998675555)
--(axis cs:120000,7588.30998675555)
--(axis cs:110000,7651.30998675555)
--(axis cs:100000,7331.30998675555)
--(axis cs:90000,7245.30998675555)
--(axis cs:80000,6643.30998675555)
--(axis cs:70000,6608.30998675555)
--(axis cs:60000,5437.30998675555)
--(axis cs:50000,5219.30998675555)
--(axis cs:40000,4681.30998675555)
--(axis cs:30000,4595.30998675555)
--(axis cs:20000,3770.30998675555)
--(axis cs:10000,3991.30998675555)
--(axis cs:0,3552.30998675555)
--(axis cs:-10000,1182.30998675555)
--cycle;

\path [draw=orchid204120188, fill=orchid204120188, opacity=0.2]
(axis cs:-10000,1292.59869853864)
--(axis cs:-10000,-1232.59869853864)
--(axis cs:0,-635.498698538636)
--(axis cs:10000,187.401301461364)
--(axis cs:20000,168.801301461364)
--(axis cs:30000,151.701301461364)
--(axis cs:40000,1567.40130146136)
--(axis cs:50000,2411.70130146136)
--(axis cs:60000,2936.00130146136)
--(axis cs:70000,4114.50130146136)
--(axis cs:80000,4297.40130146136)
--(axis cs:90000,4667.40130146136)
--(axis cs:100000,4683.10130146136)
--(axis cs:110000,5878.80130146136)
--(axis cs:120000,6191.70130146136)
--(axis cs:130000,6280.30130146136)
--(axis cs:140000,6546.00130146136)
--(axis cs:150000,6411.70130146136)
--(axis cs:160000,6603.10130146136)
--(axis cs:170000,6880.30130146136)
--(axis cs:180000,6966.00130146136)
--(axis cs:190000,6998.80130146136)
--(axis cs:200000,6717.40130146136)
--(axis cs:210000,6471.70130146136)
--(axis cs:220000,7207.40130146136)
--(axis cs:230000,6926.00130146136)
--(axis cs:240000,6773.10130146136)
--(axis cs:250000,7006.00130146136)
--(axis cs:260000,6801.70130146136)
--(axis cs:270000,6674.50130146136)
--(axis cs:280000,6954.50130146136)
--(axis cs:290000,6788.80130146136)
--(axis cs:300000,7000.30130146136)
--(axis cs:310000,7481.70130146136)
--(axis cs:320000,7063.10130146136)
--(axis cs:330000,7467.40130146136)
--(axis cs:340000,6810.30130146136)
--(axis cs:350000,7054.50130146136)
--(axis cs:360000,6816.00130146136)
--(axis cs:370000,7193.10130146136)
--(axis cs:380000,6720.30130146136)
--(axis cs:390000,7258.80130146136)
--(axis cs:390000,9783.99869853863)
--(axis cs:390000,9783.99869853863)
--(axis cs:380000,9245.49869853863)
--(axis cs:370000,9718.29869853864)
--(axis cs:360000,9341.19869853864)
--(axis cs:350000,9579.69869853864)
--(axis cs:340000,9335.49869853863)
--(axis cs:330000,9992.59869853864)
--(axis cs:320000,9588.29869853864)
--(axis cs:310000,10006.8986985386)
--(axis cs:300000,9525.49869853863)
--(axis cs:290000,9313.99869853863)
--(axis cs:280000,9479.69869853864)
--(axis cs:270000,9199.69869853864)
--(axis cs:260000,9326.89869853864)
--(axis cs:250000,9531.19869853864)
--(axis cs:240000,9298.29869853864)
--(axis cs:230000,9451.19869853864)
--(axis cs:220000,9732.59869853864)
--(axis cs:210000,8996.89869853864)
--(axis cs:200000,9242.59869853864)
--(axis cs:190000,9523.99869853863)
--(axis cs:180000,9491.19869853864)
--(axis cs:170000,9405.49869853863)
--(axis cs:160000,9128.29869853864)
--(axis cs:150000,8936.89869853864)
--(axis cs:140000,9071.19869853864)
--(axis cs:130000,8805.49869853863)
--(axis cs:120000,8716.89869853864)
--(axis cs:110000,8403.99869853863)
--(axis cs:100000,7208.29869853864)
--(axis cs:90000,7192.59869853864)
--(axis cs:80000,6822.59869853864)
--(axis cs:70000,6639.69869853864)
--(axis cs:60000,5461.19869853864)
--(axis cs:50000,4936.89869853864)
--(axis cs:40000,4092.59869853864)
--(axis cs:30000,2676.89869853864)
--(axis cs:20000,2693.99869853864)
--(axis cs:10000,2712.59869853864)
--(axis cs:0,1889.69869853864)
--(axis cs:-10000,1292.59869853864)
--cycle;

\path [draw=peru20214597, fill=peru20214597, opacity=0.2]
(axis cs:-10000,1505.3379065978)
--(axis cs:-10000,-1448.13790659781)
--(axis cs:0,87.5620934021949)
--(axis cs:10000,424.662093402195)
--(axis cs:20000,-425.337906597805)
--(axis cs:30000,-328.137906597805)
--(axis cs:40000,-275.337906597805)
--(axis cs:50000,-342.437906597805)
--(axis cs:60000,1036.1620934022)
--(axis cs:70000,2811.8620934022)
--(axis cs:80000,2580.36209340219)
--(axis cs:90000,3437.5620934022)
--(axis cs:100000,3104.66209340219)
--(axis cs:110000,3684.66209340219)
--(axis cs:120000,3488.9620934022)
--(axis cs:130000,3753.2620934022)
--(axis cs:140000,3983.2620934022)
--(axis cs:150000,4121.8620934022)
--(axis cs:160000,4376.16209340219)
--(axis cs:170000,4894.66209340219)
--(axis cs:180000,4758.9620934022)
--(axis cs:190000,4468.9620934022)
--(axis cs:200000,5004.66209340219)
--(axis cs:210000,5448.9620934022)
--(axis cs:220000,5530.3620934022)
--(axis cs:230000,6040.3620934022)
--(axis cs:240000,7466.16209340219)
--(axis cs:250000,7407.56209340219)
--(axis cs:260000,6748.9620934022)
--(axis cs:270000,8488.9620934022)
--(axis cs:280000,7757.56209340219)
--(axis cs:290000,9024.66209340219)
--(axis cs:300000,8371.86209340219)
--(axis cs:310000,9414.66209340219)
--(axis cs:320000,8403.26209340219)
--(axis cs:330000,6934.66209340219)
--(axis cs:340000,7990.3620934022)
--(axis cs:350000,7806.16209340219)
--(axis cs:360000,8140.3620934022)
--(axis cs:370000,8281.86209340219)
--(axis cs:380000,8137.56209340219)
--(axis cs:390000,7864.66209340219)
--(axis cs:390000,10818.1379065978)
--(axis cs:390000,10818.1379065978)
--(axis cs:380000,11091.0379065978)
--(axis cs:370000,11235.3379065978)
--(axis cs:360000,11093.8379065978)
--(axis cs:350000,10759.6379065978)
--(axis cs:340000,10943.8379065978)
--(axis cs:330000,9888.13790659781)
--(axis cs:320000,11356.7379065978)
--(axis cs:310000,12368.1379065978)
--(axis cs:300000,11325.3379065978)
--(axis cs:290000,11978.1379065978)
--(axis cs:280000,10711.0379065978)
--(axis cs:270000,11442.4379065978)
--(axis cs:260000,9702.43790659781)
--(axis cs:250000,10361.0379065978)
--(axis cs:240000,10419.6379065978)
--(axis cs:230000,8993.83790659781)
--(axis cs:220000,8483.83790659781)
--(axis cs:210000,8402.4379065978)
--(axis cs:200000,7958.1379065978)
--(axis cs:190000,7422.4379065978)
--(axis cs:180000,7712.4379065978)
--(axis cs:170000,7848.1379065978)
--(axis cs:160000,7329.6379065978)
--(axis cs:150000,7075.33790659781)
--(axis cs:140000,6936.7379065978)
--(axis cs:130000,6706.7379065978)
--(axis cs:120000,6442.4379065978)
--(axis cs:110000,6638.1379065978)
--(axis cs:100000,6058.1379065978)
--(axis cs:90000,6391.0379065978)
--(axis cs:80000,5533.83790659781)
--(axis cs:70000,5765.33790659781)
--(axis cs:60000,3989.63790659781)
--(axis cs:50000,2611.0379065978)
--(axis cs:40000,2678.13790659781)
--(axis cs:30000,2625.33790659781)
--(axis cs:20000,2528.13790659781)
--(axis cs:10000,3378.13790659781)
--(axis cs:0,3041.0379065978)
--(axis cs:-10000,1505.3379065978)
--cycle;

\addplot [semithick, darkcyan1115178, mark=square*, mark size=1.5, mark options={solid}]
table {%
-10000 31.4
0 608.6
10000 938.6
20000 1680
30000 3515.7
40000 4461.4
50000 5188.6
60000 7798.6
70000 9380
80000 10335.7
90000 11408.6
100000 11501.4
110000 12261.4
120000 11141.4
130000 11897.1
140000 12075.7
150000 10524.3
160000 11870
170000 12342.9
180000 11371.4
190000 13517.1
200000 13418.6
210000 14348.6
220000 14682.9
230000 15995.7
240000 17787.1
250000 14994.3
260000 16237.1
270000 18298.6
280000 26340
290000 31457.1
300000 20977.1
310000 26485.7
320000 45761.4
330000 34470
340000 59392.9
350000 42297.1
360000 61290
370000 23581.4
380000 50462.9
390000 21488.6
};
\addplot [semithick, darkorange2221435, mark=triangle*, mark size=1.5, mark options={solid}]
table {%
-10000 124
0 2494
10000 2933
20000 2712
30000 3537
40000 3623
50000 4161
60000 4379
70000 5550
80000 5585
90000 6187
100000 6273
110000 6593
120000 6530
130000 6941
140000 6511
150000 6814
160000 6978
170000 6452
180000 6589
190000 6667
200000 7051
210000 6605
220000 6754
230000 7020
240000 7379
250000 7168
260000 7700
270000 7626
280000 7640
290000 7363
300000 7097
310000 7210
320000 7576
330000 7172
340000 7730
350000 7550
360000 7903
370000 7756
380000 7730
390000 7827
};
\addplot [semithick, orchid204120188, mark=+, mark size=1.5, mark options={solid}]
table {%
-10000 30
0 627.1
10000 1450
20000 1431.4
30000 1414.3
40000 2830
50000 3674.3
60000 4198.6
70000 5377.1
80000 5560
90000 5930
100000 5945.7
110000 7141.4
120000 7454.3
130000 7542.9
140000 7808.6
150000 7674.3
160000 7865.7
170000 8142.9
180000 8228.6
190000 8261.4
200000 7980
210000 7734.3
220000 8470
230000 8188.6
240000 8035.7
250000 8268.6
260000 8064.3
270000 7937.1
280000 8217.1
290000 8051.4
300000 8262.9
310000 8744.3
320000 8325.7
330000 8730
340000 8072.9
350000 8317.1
360000 8078.6
370000 8455.7
380000 7982.9
390000 8521.4
};
\addplot [semithick, peru20214597, mark=diamond*, mark size=1.5, mark options={solid}]
table {%
-10000 28.6
0 1564.3
10000 1901.4
20000 1051.4
30000 1148.6
40000 1201.4
50000 1134.3
60000 2512.9
70000 4288.6
80000 4057.1
90000 4914.3
100000 4581.4
110000 5161.4
120000 4965.7
130000 5230
140000 5460
150000 5598.6
160000 5852.9
170000 6371.4
180000 6235.7
190000 5945.7
200000 6481.4
210000 6925.7
220000 7007.1
230000 7517.1
240000 8942.9
250000 8884.3
260000 8225.7
270000 9965.7
280000 9234.3
290000 10501.4
300000 9848.6
310000 10891.4
320000 9880
330000 8411.4
340000 9467.1
350000 9282.9
360000 9617.1
370000 9758.6
380000 9614.3
390000 9341.4
};
\end{axis}

\end{tikzpicture}

%% file: Image/up_n_down.tex
\begin{tikzpicture}

\definecolor{darkcyan1115178}{RGB}{1,115,178}
\definecolor{darkorange2221435}{RGB}{222,143,5}
\definecolor{darkslategray38}{RGB}{38,38,38}
\definecolor{lavender234234242}{RGB}{234,234,242}
\definecolor{lightgray204}{RGB}{204,204,204}
\definecolor{orchid204120188}{RGB}{204,120,188}
\definecolor{peru20214597}{RGB}{202,145,97}

\begin{axis}[
axis background/.style={fill=lavender234234242},
axis line style={white},
legend cell align={left},
legend style={
  fill opacity=0.8,
  draw opacity=1,
  text opacity=1,
  draw=lightgray204,
  fill=lavender234234242
},
tick align=outside,
x grid style={white},
xlabel=\textcolor{darkslategray38}{Time Step},
xmajorgrids,
xmajorticks=true,
xmin=-30000, xmax=410000,
xtick style={color=darkslategray38},
xtick={0,50000,100000,150000,200000,250000,300000,350000,400000},
xticklabels={0k,50k,100k,150k,200k,250k,300k,350k,400k},
y grid style={white},
ylabel=\textcolor{darkslategray38}{Performance},
ymajorgrids,
ymajorticks=true,
ymin=-4406.43418641908, ymax=30833.0341864191,
ytick style={color=darkslategray38}
]
\path [draw=darkcyan1115178, fill=darkcyan1115178, opacity=0.2]
(axis cs:-10000,3321.34035492289)
--(axis cs:-10000,-2212.74035492289)
--(axis cs:0,-1724.74035492289)
--(axis cs:10000,850.059645077109)
--(axis cs:20000,1429.35964507711)
--(axis cs:30000,593.959645077109)
--(axis cs:40000,3660.35964507711)
--(axis cs:50000,2712.35964507711)
--(axis cs:60000,4361.35964507711)
--(axis cs:70000,1620.05964507711)
--(axis cs:80000,651.859645077109)
--(axis cs:90000,655.859645077109)
--(axis cs:100000,3692.25964507711)
--(axis cs:110000,5947.35964507711)
--(axis cs:120000,15410.5596450771)
--(axis cs:130000,7568.55964507711)
--(axis cs:140000,7824.85964507711)
--(axis cs:150000,5650.85964507711)
--(axis cs:160000,5203.95964507711)
--(axis cs:170000,3907.05964507711)
--(axis cs:180000,5980.25964507711)
--(axis cs:190000,5081.95964507711)
--(axis cs:200000,4310.85964507711)
--(axis cs:210000,6883.95964507711)
--(axis cs:220000,4596.85964507711)
--(axis cs:230000,9182.35964507711)
--(axis cs:240000,5978.55964507711)
--(axis cs:250000,14267.8596450771)
--(axis cs:260000,6588.85964507711)
--(axis cs:270000,4412.95964507711)
--(axis cs:280000,14060.9596450771)
--(axis cs:290000,6031.65964507711)
--(axis cs:300000,5718.65964507711)
--(axis cs:310000,3682.35964507711)
--(axis cs:320000,3713.95964507711)
--(axis cs:330000,8993.35964507711)
--(axis cs:340000,4706.65964507711)
--(axis cs:350000,3809.05964507711)
--(axis cs:360000,6497.95964507711)
--(axis cs:370000,3312.25964507711)
--(axis cs:380000,4791.05964507711)
--(axis cs:390000,4803.25964507711)
--(axis cs:390000,10337.3403549229)
--(axis cs:390000,10337.3403549229)
--(axis cs:380000,10325.1403549229)
--(axis cs:370000,8846.34035492289)
--(axis cs:360000,12032.0403549229)
--(axis cs:350000,9343.14035492289)
--(axis cs:340000,10240.7403549229)
--(axis cs:330000,14527.4403549229)
--(axis cs:320000,9248.04035492289)
--(axis cs:310000,9216.44035492289)
--(axis cs:300000,11252.7403549229)
--(axis cs:290000,11565.7403549229)
--(axis cs:280000,19595.0403549229)
--(axis cs:270000,9947.04035492289)
--(axis cs:260000,12122.9403549229)
--(axis cs:250000,19801.9403549229)
--(axis cs:240000,11512.6403549229)
--(axis cs:230000,14716.4403549229)
--(axis cs:220000,10130.9403549229)
--(axis cs:210000,12418.0403549229)
--(axis cs:200000,9844.94035492289)
--(axis cs:190000,10616.0403549229)
--(axis cs:180000,11514.3403549229)
--(axis cs:170000,9441.14035492289)
--(axis cs:160000,10738.0403549229)
--(axis cs:150000,11184.9403549229)
--(axis cs:140000,13358.9403549229)
--(axis cs:130000,13102.6403549229)
--(axis cs:120000,20944.6403549229)
--(axis cs:110000,11481.4403549229)
--(axis cs:100000,9226.34035492289)
--(axis cs:90000,6189.94035492289)
--(axis cs:80000,6185.94035492289)
--(axis cs:70000,7154.14035492289)
--(axis cs:60000,9895.44035492289)
--(axis cs:50000,8246.44035492289)
--(axis cs:40000,9194.44035492289)
--(axis cs:30000,6128.04035492289)
--(axis cs:20000,6963.44035492289)
--(axis cs:10000,6384.14035492289)
--(axis cs:0,3809.34035492289)
--(axis cs:-10000,3321.34035492289)
--cycle;

\path [draw=darkorange2221435, fill=darkorange2221435, opacity=0.2]
(axis cs:-10000,3963.84016947189)
--(axis cs:-10000,-2804.64016947189)
--(axis cs:0,-2471.14016947189)
--(axis cs:10000,-1274.34016947189)
--(axis cs:20000,2430.95983052811)
--(axis cs:30000,10177.8598305281)
--(axis cs:40000,20528.3598305281)
--(axis cs:50000,22462.7598305281)
--(axis cs:60000,16965.0598305281)
--(axis cs:70000,12984.7598305281)
--(axis cs:80000,6335.95983052811)
--(axis cs:90000,6863.65983052811)
--(axis cs:100000,2458.95983052811)
--(axis cs:110000,2605.55983052811)
--(axis cs:120000,3336.55983052811)
--(axis cs:130000,7175.85983052811)
--(axis cs:140000,9194.15983052811)
--(axis cs:150000,6156.55983052811)
--(axis cs:160000,10697.2598305281)
--(axis cs:170000,5397.45983052811)
--(axis cs:180000,5749.95983052811)
--(axis cs:190000,7210.75983052811)
--(axis cs:200000,7414.05983052811)
--(axis cs:210000,6457.85983052811)
--(axis cs:220000,11837.0598305281)
--(axis cs:230000,6951.75983052811)
--(axis cs:240000,8629.15983052811)
--(axis cs:250000,4016.75983052811)
--(axis cs:260000,7999.95983052811)
--(axis cs:270000,10041.3598305281)
--(axis cs:280000,8218.35983052811)
--(axis cs:290000,7381.65983052811)
--(axis cs:300000,9153.95983052811)
--(axis cs:310000,4804.75983052811)
--(axis cs:320000,7180.75983052811)
--(axis cs:330000,7891.95983052811)
--(axis cs:340000,6376.05983052811)
--(axis cs:350000,6797.65983052811)
--(axis cs:360000,4403.65983052811)
--(axis cs:370000,6150.85983052811)
--(axis cs:380000,4285.85983052811)
--(axis cs:390000,5238.05983052811)
--(axis cs:390000,12006.5401694719)
--(axis cs:390000,12006.5401694719)
--(axis cs:380000,11054.3401694719)
--(axis cs:370000,12919.3401694719)
--(axis cs:360000,11172.1401694719)
--(axis cs:350000,13566.1401694719)
--(axis cs:340000,13144.5401694719)
--(axis cs:330000,14660.4401694719)
--(axis cs:320000,13949.2401694719)
--(axis cs:310000,11573.2401694719)
--(axis cs:300000,15922.4401694719)
--(axis cs:290000,14150.1401694719)
--(axis cs:280000,14986.8401694719)
--(axis cs:270000,16809.8401694719)
--(axis cs:260000,14768.4401694719)
--(axis cs:250000,10785.2401694719)
--(axis cs:240000,15397.6401694719)
--(axis cs:230000,13720.2401694719)
--(axis cs:220000,18605.5401694719)
--(axis cs:210000,13226.3401694719)
--(axis cs:200000,14182.5401694719)
--(axis cs:190000,13979.2401694719)
--(axis cs:180000,12518.4401694719)
--(axis cs:170000,12165.9401694719)
--(axis cs:160000,17465.7401694719)
--(axis cs:150000,12925.0401694719)
--(axis cs:140000,15962.6401694719)
--(axis cs:130000,13944.3401694719)
--(axis cs:120000,10105.0401694719)
--(axis cs:110000,9374.04016947189)
--(axis cs:100000,9227.44016947189)
--(axis cs:90000,13632.1401694719)
--(axis cs:80000,13104.4401694719)
--(axis cs:70000,19753.2401694719)
--(axis cs:60000,23733.5401694719)
--(axis cs:50000,29231.2401694719)
--(axis cs:40000,27296.8401694719)
--(axis cs:30000,16946.3401694719)
--(axis cs:20000,9199.44016947189)
--(axis cs:10000,5494.14016947189)
--(axis cs:0,4297.34016947189)
--(axis cs:-10000,3963.84016947189)
--cycle;

\path [draw=orchid204120188, fill=orchid204120188, opacity=0.2]
(axis cs:-10000,4196.90940577167)
--(axis cs:-10000,-2594.10940577167)
--(axis cs:0,-1901.60940577167)
--(axis cs:10000,-858.90940577167)
--(axis cs:20000,-217.50940577167)
--(axis cs:30000,-359.80940577167)
--(axis cs:40000,81.8905942283304)
--(axis cs:50000,15.8905942283304)
--(axis cs:60000,567.59059422833)
--(axis cs:70000,12018.4905942283)
--(axis cs:80000,2791.49059422833)
--(axis cs:90000,7808.49059422833)
--(axis cs:100000,4902.79059422833)
--(axis cs:110000,9281.49059422833)
--(axis cs:120000,6910.79059422833)
--(axis cs:130000,1532.09059422833)
--(axis cs:140000,9762.89059422833)
--(axis cs:150000,10243.5905942283)
--(axis cs:160000,12761.0905942283)
--(axis cs:170000,12818.8905942283)
--(axis cs:180000,8955.09059422833)
--(axis cs:190000,10593.8905942283)
--(axis cs:200000,2798.59059422833)
--(axis cs:210000,8019.39059422833)
--(axis cs:220000,5673.09059422833)
--(axis cs:230000,1977.09059422833)
--(axis cs:240000,2412.19059422833)
--(axis cs:250000,2429.49059422833)
--(axis cs:260000,3578.89059422833)
--(axis cs:270000,2552.39059422833)
--(axis cs:280000,1172.49059422833)
--(axis cs:290000,2862.79059422833)
--(axis cs:300000,3902.09059422833)
--(axis cs:310000,2320.79059422833)
--(axis cs:320000,1272.89059422833)
--(axis cs:330000,932.790594228331)
--(axis cs:340000,3453.19059422833)
--(axis cs:350000,4773.09059422833)
--(axis cs:360000,1731.19059422833)
--(axis cs:370000,2535.79059422833)
--(axis cs:380000,2012.19059422833)
--(axis cs:390000,1293.19059422833)
--(axis cs:390000,8084.20940577167)
--(axis cs:390000,8084.20940577167)
--(axis cs:380000,8803.20940577167)
--(axis cs:370000,9326.80940577167)
--(axis cs:360000,8522.20940577167)
--(axis cs:350000,11564.1094057717)
--(axis cs:340000,10244.2094057717)
--(axis cs:330000,7723.80940577167)
--(axis cs:320000,8063.90940577167)
--(axis cs:310000,9111.80940577167)
--(axis cs:300000,10693.1094057717)
--(axis cs:290000,9653.80940577167)
--(axis cs:280000,7963.50940577167)
--(axis cs:270000,9343.40940577167)
--(axis cs:260000,10369.9094057717)
--(axis cs:250000,9220.50940577167)
--(axis cs:240000,9203.20940577167)
--(axis cs:230000,8768.10940577167)
--(axis cs:220000,12464.1094057717)
--(axis cs:210000,14810.4094057717)
--(axis cs:200000,9589.60940577167)
--(axis cs:190000,17384.9094057717)
--(axis cs:180000,15746.1094057717)
--(axis cs:170000,19609.9094057717)
--(axis cs:160000,19552.1094057717)
--(axis cs:150000,17034.6094057717)
--(axis cs:140000,16553.9094057717)
--(axis cs:130000,8323.10940577167)
--(axis cs:120000,13701.8094057717)
--(axis cs:110000,16072.5094057717)
--(axis cs:100000,11693.8094057717)
--(axis cs:90000,14599.5094057717)
--(axis cs:80000,9582.50940577167)
--(axis cs:70000,18809.5094057717)
--(axis cs:60000,7358.60940577167)
--(axis cs:50000,6806.90940577167)
--(axis cs:40000,6872.90940577167)
--(axis cs:30000,6431.20940577167)
--(axis cs:20000,6573.50940577167)
--(axis cs:10000,5932.10940577167)
--(axis cs:0,4889.40940577167)
--(axis cs:-10000,4196.90940577167)
--cycle;

\path [draw=peru20214597, fill=peru20214597, opacity=0.2]
(axis cs:-10000,4281.51018309083)
--(axis cs:-10000,-1948.71018309083)
--(axis cs:0,-1779.11018309083)
--(axis cs:10000,-598.010183090832)
--(axis cs:20000,1320.88981690917)
--(axis cs:30000,-843.810183090831)
--(axis cs:40000,-379.710183090831)
--(axis cs:50000,1369.78981690917)
--(axis cs:60000,620.789816909169)
--(axis cs:70000,-297.410183090832)
--(axis cs:80000,-289.710183090831)
--(axis cs:90000,877.989816909168)
--(axis cs:100000,-258.810183090831)
--(axis cs:110000,-43.2101830908314)
--(axis cs:120000,95.9898169091684)
--(axis cs:130000,55.5898169091683)
--(axis cs:140000,155.989816909168)
--(axis cs:150000,571.889816909169)
--(axis cs:160000,747.889816909169)
--(axis cs:170000,673.889816909169)
--(axis cs:180000,1638.18981690917)
--(axis cs:190000,1375.98981690917)
--(axis cs:200000,3140.98981690917)
--(axis cs:210000,1390.48981690917)
--(axis cs:220000,2821.98981690917)
--(axis cs:230000,4081.78981690917)
--(axis cs:240000,3010.98981690917)
--(axis cs:250000,4398.88981690917)
--(axis cs:260000,9786.18981690917)
--(axis cs:270000,7653.98981690917)
--(axis cs:280000,5072.78981690917)
--(axis cs:290000,10337.5898169092)
--(axis cs:300000,9049.58981690917)
--(axis cs:310000,4009.58981690917)
--(axis cs:320000,7282.48981690917)
--(axis cs:330000,8167.58981690917)
--(axis cs:340000,4040.28981690917)
--(axis cs:350000,5566.88981690917)
--(axis cs:360000,4851.58981690917)
--(axis cs:370000,4342.98981690917)
--(axis cs:380000,13098.4898169092)
--(axis cs:390000,15742.7898169092)
--(axis cs:390000,21973.0101830908)
--(axis cs:390000,21973.0101830908)
--(axis cs:380000,19328.7101830908)
--(axis cs:370000,10573.2101830908)
--(axis cs:360000,11081.8101830908)
--(axis cs:350000,11797.1101830908)
--(axis cs:340000,10270.5101830908)
--(axis cs:330000,14397.8101830908)
--(axis cs:320000,13512.7101830908)
--(axis cs:310000,10239.8101830908)
--(axis cs:300000,15279.8101830908)
--(axis cs:290000,16567.8101830908)
--(axis cs:280000,11303.0101830908)
--(axis cs:270000,13884.2101830908)
--(axis cs:260000,16016.4101830908)
--(axis cs:250000,10629.1101830908)
--(axis cs:240000,9241.21018309083)
--(axis cs:230000,10312.0101830908)
--(axis cs:220000,9052.21018309083)
--(axis cs:210000,7620.71018309083)
--(axis cs:200000,9371.21018309083)
--(axis cs:190000,7606.21018309083)
--(axis cs:180000,7868.41018309083)
--(axis cs:170000,6904.11018309083)
--(axis cs:160000,6978.11018309083)
--(axis cs:150000,6802.11018309083)
--(axis cs:140000,6386.21018309083)
--(axis cs:130000,6285.81018309083)
--(axis cs:120000,6326.21018309083)
--(axis cs:110000,6187.01018309083)
--(axis cs:100000,5971.41018309083)
--(axis cs:90000,7108.21018309083)
--(axis cs:80000,5940.51018309083)
--(axis cs:70000,5932.81018309083)
--(axis cs:60000,6851.01018309083)
--(axis cs:50000,7600.01018309083)
--(axis cs:40000,5850.51018309083)
--(axis cs:30000,5386.41018309083)
--(axis cs:20000,7551.11018309083)
--(axis cs:10000,5632.21018309083)
--(axis cs:0,4451.11018309083)
--(axis cs:-10000,4281.51018309083)
--cycle;

\addplot [semithick, darkcyan1115178, mark=square*, mark size=1.5, mark options={solid}]
table {%
-10000 554.3
0 1042.3
10000 3617.1
20000 4196.4
30000 3361
40000 6427.4
50000 5479.4
60000 7128.4
70000 4387.1
80000 3418.9
90000 3422.9
100000 6459.3
110000 8714.4
120000 18177.6
130000 10335.6
140000 10591.9
150000 8417.9
160000 7971
170000 6674.1
180000 8747.3
190000 7849
200000 7077.9
210000 9651
220000 7363.9
230000 11949.4
240000 8745.6
250000 17034.9
260000 9355.9
270000 7180
280000 16828
290000 8798.7
300000 8485.7
310000 6449.4
320000 6481
330000 11760.4
340000 7473.7
350000 6576.1
360000 9265
370000 6079.3
380000 7558.1
390000 7570.3
};
\addplot [semithick, darkorange2221435, mark=triangle*, mark size=1.5, mark options={solid}]
table {%
-10000 579.6
0 913.1
10000 2109.9
20000 5815.2
30000 13562.1
40000 23912.6
50000 25847
60000 20349.3
70000 16369
80000 9720.2
90000 10247.9
100000 5843.2
110000 5989.8
120000 6720.8
130000 10560.1
140000 12578.4
150000 9540.8
160000 14081.5
170000 8781.7
180000 9134.2
190000 10595
200000 10798.3
210000 9842.1
220000 15221.3
230000 10336
240000 12013.4
250000 7401
260000 11384.2
270000 13425.6
280000 11602.6
290000 10765.9
300000 12538.2
310000 8189
320000 10565
330000 11276.2
340000 9760.3
350000 10181.9
360000 7787.9
370000 9535.1
380000 7670.1
390000 8622.3
};
\addplot [semithick, orchid204120188, mark=+, mark size=1.5, mark options={solid}]
table {%
-10000 801.4
0 1493.9
10000 2536.6
20000 3178
30000 3035.7
40000 3477.4
50000 3411.4
60000 3963.1
70000 15414
80000 6187
90000 11204
100000 8298.3
110000 12677
120000 10306.3
130000 4927.6
140000 13158.4
150000 13639.1
160000 16156.6
170000 16214.4
180000 12350.6
190000 13989.4
200000 6194.1
210000 11414.9
220000 9068.6
230000 5372.6
240000 5807.7
250000 5825
260000 6974.4
270000 5947.9
280000 4568
290000 6258.3
300000 7297.6
310000 5716.3
320000 4668.4
330000 4328.3
340000 6848.7
350000 8168.6
360000 5126.7
370000 5931.3
380000 5407.7
390000 4688.7
};
\addplot [semithick, peru20214597, mark=diamond*, mark size=1.5, mark options={solid}]
table {%
-10000 1166.4
0 1336
10000 2517.1
20000 4436
30000 2271.3
40000 2735.4
50000 4484.9
60000 3735.9
70000 2817.7
80000 2825.4
90000 3993.1
100000 2856.3
110000 3071.9
120000 3211.1
130000 3170.7
140000 3271.1
150000 3687
160000 3863
170000 3789
180000 4753.3
190000 4491.1
200000 6256.1
210000 4505.6
220000 5937.1
230000 7196.9
240000 6126.1
250000 7514
260000 12901.3
270000 10769.1
280000 8187.9
290000 13452.7
300000 12164.7
310000 7124.7
320000 10397.6
330000 11282.7
340000 7155.4
350000 8682
360000 7966.7
370000 7458.1
380000 16213.6
390000 18857.9
};
\end{axis}

\end{tikzpicture}